\documentclass[letterpaper]{article} 
\usepackage{aaai2027}  
\renewcommand{\bibliographystyle}[1]{}

\nocopyright

\usepackage[hyphens]{url}
\usepackage{graphicx}
\usepackage{natbib}
\usepackage{caption}
\usepackage{booktabs}
\usepackage{longtable}
\usepackage{array}
\usepackage{amsmath}
\usepackage{amssymb}
\usepackage{algorithm}
\usepackage{algorithmic}
\usepackage{multicol}
\usepackage{needspace}
\raggedcolumns

\newcommand{\figref}[1]{Figure~\ref{#1}}
\newcommand{\figrefs}[2]{Figures~\ref{#1} and~\ref{#2}}
\newcommand{\figrangeref}[2]{Figures~\ref{#1}--\ref{#2}}
\newcommand{\tabref}[1]{Table~\ref{#1}}
\newcommand{\appref}[1]{Appendix~\ref{#1}}
\newcommand{\apprefs}[2]{Appendices~\ref{#1} and~\ref{#2}}
\newcolumntype{L}[1]{>{\raggedright\arraybackslash}p{#1}}
\newcolumntype{R}[1]{>{\raggedleft\arraybackslash}p{#1}}
\newcolumntype{C}[1]{>{\centering\arraybackslash}p{#1}}
\newcounter{appendixsection}
\newcounter{appendixsubsection}[appendixsection]
\newcounter{appendixdomain}[appendixsubsection]
\renewcommand{\theappendixsection}{\Alph{appendixsection}}
\renewcommand{\theappendixsubsection}{\theappendixsection.\arabic{appendixsubsection}}
\renewcommand{\theappendixdomain}{\theappendixsubsection.\arabic{appendixdomain}}
\newcommand{\appendixsection}[2]{%
  \Needspace{7\baselineskip}%
  \refstepcounter{appendixsection}%
  \section*{\theappendixsection. #1}%
  \label{#2}%
}
\newcommand{\appendixsubsection}[1]{%
  \refstepcounter{appendixsubsection}%
  \subsection*{\theappendixsubsection\ #1}%
}

\title{Solver-Integrated Adversarial Attacking and Training of Neural Operators}

\author{
  Yifei Sun\textsuperscript{\dag}
}
\affiliations{}

\newcommand{\pendingappendixfigureheading}{}
\newcommand{\nextappendixfigureheading}[1]{\gdef\pendingappendixfigureheading{#1}}
\newcommand{\useappendixfigureheading}{%
  \pendingappendixfigureheading\par
  \global\let\pendingappendixfigureheading\empty
}

\newcommand{\appendixdomainheading}[1]{%
  \refstepcounter{appendixdomain}%
  \par\medskip
  \noindent{\normalsize\bfseries \theappendixdomain\ #1}\par\smallskip
}

\begin{document}

\maketitle
\thispagestyle{plain}
\pagestyle{plain}
\begingroup
\renewcommand{\thefootnote}{\fnsymbol{footnote}}
\footnotetext[2]{\textsuperscript{\dag} Corresponding author. Email: \texttt{yifeisun@umich.edu}.}
\endgroup
\begin{abstract}
Neural operators are widely used as fast surrogates for numerical PDE
solvers, mapping input functions to solution functions. However, their
generalizability and robustness are not yet clearly defined in the
operator-learning setting, which differs from traditional adversarial robustness
definitions. This paper studies the generalizability and robustness of a
learned neural operator from a solver-integrated perspective, addressing the
challenge that the output of a learned operator and a numerical solver tends to
change in tandem under input perturbation. First, we formalize the definition
of generalization and robustness through a model-solver error operator,
identifying fixed-input model-solver loss as generalization metric, and
norm-bounded adversarial attack loss increase and Jacobian-error function norm as
robustness metric. Second, we
identify the solver-integrated adversarial attack as appropriate
for PDE operator learning and show why model-only or fixed-ground-truth attacks
can be insufficient when the solver output also changes with the input. Third, we
develop solver-integrated adversarial training methods for neural operators.
Experiments on representative PDE benchmarks show that this solver-integrated
adversarial training clearly improves both generalizability and robustness.
Deeper solver integration yields more effective attacks, more informative
samples, and more efficient training than less integrated alternatives. These
results provide a general
framework for robust operator training and automatic sample selection without
heavy manual intervention. More broadly, the formulation applies to
adversarial regression whenever a ground-truth oracle can evaluate, and ideally
differentiate, the true input--output map; PDE operator learning is one such
case.
\end{abstract}

\section{Introduction}

Neural operators, including Fourier Neural Operators and DeepONets, learn maps
between function spaces and are commonly utilized as fast data-driven
surrogates for numerical PDE solvers \citep{li2021fourier,lu2021deeponet}.
They map input functions to solution functions in one forward pass, making them
attractive when many related PDE solution queries are needed. Practical use, however,
requires the learned operator to generalize beyond the training dataset and
remain close to the solver when the input function is perturbed. Recent neural
operator studies have therefore begun to examine adversarial robustness and
robustness-aware training
\citep{adesoji2022evaluating,huang2026stablepdenet,roy2026beyond}.

Prior work has often not clearly separated the notions of generalization and
robustness.
Classical theory defines generalization through expected-versus-empirical loss
\citep{bousquet2002stability,xu2012robustness}. Several modern robustness
notions are defined by worst-case loss increase under norm-bounded
perturbations \citep{madry2018towards}, and global Lipschitz constants and
local spectral norms are also used to measure robustness
\citep{cisse2017parseval,huang2026stablepdenet}. We inherit these ideas and
specialize them to the operator learning setting. Generalization is a static
definition: the norm of the output of our error operator. Robustness is defined
in a dynamic way: how the output of our error operator amplifies under an input
perturbation. We further develop a Jacobian-error metric, which is cheaper to
compute and closely aligned with adversarial loss increase in the local
robustness setting.

This setting also differs from traditional adversarial robustness. Most
classical adversarial attacks were developed for classification models, where it
is reasonable to assume that a small input perturbation should not change the
ground-truth label \citep{goodfellow2015explaining,madry2018towards}. However,
the PDE operator learning problem is not only a regression problem, but a
regression problem with an existing numerical solver as a trusted oracle. When
the input function is perturbed, the correct solution should change accordingly
but stay close to the solver's output. Therefore, the model is not expected to
be robust by itself; rather, the model-solver discrepancy is expected to be
robust. We therefore define the error operator as the discrepancy between the
learned operator and the solver.
Classical white-box access already supplies the model derivative. The additional
information here is the oracle derivative and the oracle output recomputed at
each current input. Relabeling selected examples supplies output values; a
solver VJP or adjoint also supplies the input sensitivity used during attack
optimization.
\figrefs{fig:training-vs-attack-gradients}{fig:pde-solver-training-attack} in
\appref{app:diagram-flow} give a systematic comparison
between the usual neural-network attack/training flow and the PDE-operator
setting.

For the displayed Navier--Stokes input in \figref{fig:app-g-ns-steepest-add-final}, the bounded full-L3 attack increases the reported model--solver loss from $34.82$ to $263.9$, about $7.6$ times the initial loss. This illustrates the reliability failure quantified by the error operator: the perturbed prediction ceases to track the corresponding perturbed solver solution.

Building on this view, we develop solver-integrated adversarial training. The
method can be interpreted as a teacher-student solver distillation framework:
the numerical solver acts as the teacher, while the neural operator is the
student repeatedly trained on informative inputs discovered from the local
model-solver discrepancy, so that the neural operator becomes an increasingly
faithful behavioral twin of the solver. In this sense, the solver's behavior is
distilled into the neural operator through automatically generated samples
guided by model and solver gradients. This process reduces manual intervention
and avoids the bias of hand-selected training inputs.
Experiments on representative PDE benchmarks show that the proposed training
clearly improves generalization and robustness, and that deeper solver
integration yields more effective attacks, more informative samples, and more
efficient training. Moreover, the contribution is not tied to neural operators
alone: it applies to adversarial regression whenever a ground-truth oracle can
provide true outputs and their input gradients, as PDE solvers do here.

Our main contributions can be summarized as follows:
\begin{itemize}
  \item In the setting of operator learning, we propose a new robustness
  definition and metric.
  \item We identify the most accurate and appropriate target loss for
  neural-operator adversarial attack, which involves deep solver integration.
  \item We propose solver-integrated adversarial training for neural operators
  to improve both generalization and robustness.
\end{itemize}

\section{Related Work}

Most classical adversarial attacks were designed for classification. They
usually define the attack objective as a loss-maximization problem under
perturbation constraints, for example maximizing a classification loss
$\ell(f_\theta(x+\delta),y)$ over a norm-bounded perturbation $\delta$
\citep{goodfellow2015explaining,madry2018towards}. Different methods mainly
vary in optimization specifics rather than the target loss in the attack
process: FGSM and PGD use white-box gradients with different step rules; JSMA
uses saliency gradients for sparse feature changes; C\&W changes the
constrained optimization form; DeepFool searches for a nearby decision
boundary; Boundary Attack, HopSkipJump, and Square Attack are black-box
methods; and one-pixel or patch attacks impose sparse or structured
perturbations
\citep{carlini2017towards,papernot2016limitations,moosavi2016deepfool,brendel2017decision,chen2019hopskipjump,andriushchenko2019square,su2017onepixel,brown2017adversarialpatch,croce2020autoattack}.
Other variants may change the norm constraint, gradient scaling, step schedule,
or available information. Despite these optimization-specific differences, they
usually assume that the ground-truth label does not change.

Regression attacks have also been studied, including adversarial regression and
time-series forecasting attacks
\citep{nguyen2018regression,tong2018adversarialregression,ribeiro2023linearregression,mode2020mtsregression,liu2022forecastingattacks}.
These works use an observed and fixed ground truth, or the model's own output,
as ground truth. They do not put the true input-to-output oracle inside the
attack loop: the target loss does not use the ground-truth output associated
with the perturbed input, and the backward propagation does not use gradient
information from that ground-truth oracle. They do not recompute the true loss
with the true output for the perturbed input inside the attack loop. They
acknowledge that the true output, and therefore the true loss, should change
accordingly with input perturbations, but they lack the generating function in
that setting. Therefore, they assume that small perturbations do not
meaningfully change the observed label. However, in our operator-learning
setting, the numerical solver is exactly this generating function, so the true
output and the corresponding true loss can be recomputed for each perturbed
input inside the loop. This is the key difference from our solver-integrated
setting. In the notation used throughout the paper, the first two conventions
are $\mathcal L_1(\delta)=\|\mathcal F_\theta(x+\delta)-\mathcal F_\theta(x)\|$
and $\mathcal L_2(\delta)=\|\mathcal F_\theta(x+\delta)-\mathcal G(x)\|$.
We introduce the solver-integrated objective
$\mathcal L_3(\delta)=\|\mathcal F_\theta(x+\delta)-\mathcal G(x+\delta)\|$,
which recomputes the oracle output at the perturbed input.

Beyond generic adversarial-robustness papers, one study focuses on robustness
in operator learning and finds that neural-operator error can grow significantly
when the perturbation norm is increased \citep{adesoji2022evaluating}. Solver
outputs are used as references. Its Burgers/Darcy nearest-surrogate procedure
approximates the reference at a perturbed input by a precomputed neighboring
solver label; that label can change when a different neighbor is selected, but
the lookup does not supply the derivative of the true solver at that input.
Our dictionary baselines isolate this principle at three dictionary sizes.
The cited study describes its Navier--Stokes case separately using a direct
solver reference, so the dictionary characterization is specific to its
Burgers/Darcy procedure.

A related line studies adversarial training with PDE residual/physics loss.
WbAR combines adversarial training with PINNs by moving collocation points
toward larger PDE residual \citep{shi2025wbar}. StablePDENet applies
adversarial training to neural operators, but its loss is still a physics loss
computed from the predicted solution rather than supervised by the true solver
output \citep{huang2026stablepdenet}. RAMS systematizes this physics-loss
gradient moving-sample idea for PINNs and operator learning, but it also uses
physics loss to move samples \citep{ouyang2025rams}. RAMS can call a solver to
label selected samples in data-driven operator learning; its selection gradient
still comes from the PDE residual. Our Physics baseline compares this residual
selection mechanism under the same operator architecture and data. It does not
reproduce every collocation schedule or outer regularizer in those papers.
The StablePDENet comparison refers to the residual objective in the cited
original version; later residual-sensitivity regularization is a distinct
outer-training component. In our setting, physics loss is less direct than
solver supervision: a solver output should satisfy the physics loss, but a
small physics loss does not necessarily identify the solver output, especially
for time-dependent problems where physics-consistent trajectories may not match
the solver's time-stepping output. Second, many neural-operator setups output
only the target field or final state, rather than dense neighboring time frames,
so temporal derivatives cannot be computed by finite differences. Our
experiments compare the corresponding objectives with solver-integrated
training, with the differences in complete training pipelines kept explicit.

Other operator-learning works use heuristic, active, or generative sampling.
Neural-operator digital-twin studies use gradient-free sparse attacks and
attack-driven active learning with input denoising to find vulnerability
locations and add data \citep{roy2026vulnerabilities,roy2026beyond}.
Adversarial autoencoders, GANO, and turbulent-flow generative models use
generative sampling, Gaussian-random-field inputs, discriminator losses, or
conditional reconstruction to improve representations or generated fields
\citep{enyeart2024adversarial,rahman2022gano,oommen2025turbulent}. These
sampling strategies are useful, but they are heuristic or distributional: they
do not use the existing numerical solver as the teacher in sample selection,
nor use solver gradients to choose perturbation directions. In this sense, they
do not fully exploit the special structure of PDE operator learning, where a
numerical solver is already available as a supervision oracle.

\section{Preliminaries}
\subsection{Problem Formulation}

Let $\mathcal{X}$ and $\mathcal{Y}$ be Hilbert function spaces. An input is a
function $x\in\mathcal{X}$ and the output is a function $y\in\mathcal{Y}$; for
example, $x$ may be an initial condition or coefficient field, and $y$ may be a
final condition or solution function. We learn a neural operator as a surrogate
and use a numerical solver as the oracle:
\begin{align}
  \mathcal{F}_\theta &: \mathcal{X}\to\mathcal{Y},
  &x&\mapsto \mathcal{F}_\theta(x),
  \nonumber\\
  \mathcal{G} &: \mathcal{X}\to\mathcal{Y},
  &x&\mapsto \mathcal{G}(x).
\end{align}
Throughout the paper, model (surrogate, neural operator) refers to
$\mathcal{F}_\theta$, while solver (oracle, ground-truth input-to-output mapping)
refers to $\mathcal{G}$. Their difference defines the model-solver error operator
\begin{align}
  \mathcal{E}_\theta
  &:=
  \mathcal{F}_\theta-\mathcal{G}
  :
  \mathcal{X}\to\mathcal{Y},
  \nonumber\\
  e_\theta[x]
  &:=
  \mathcal{E}_\theta(x)
  =
  \mathcal{F}_\theta(x)-\mathcal{G}(x)\in\mathcal{Y}.
\end{align}
When the three operators are Frechet differentiable at $x$, their local
linearizations are bounded linear maps from input perturbation functions to
output perturbation functions:
\begin{align}
  D\mathcal{F}_\theta[x] &: \mathcal{X}\to\mathcal{Y},
  &
  D\mathcal{G}[x] &: \mathcal{X}\to\mathcal{Y},
  \nonumber\\
  D\mathcal{E}_\theta[x]
  &:=
  D\mathcal{F}_\theta[x]-D\mathcal{G}[x]
  :
  \mathcal{X}\to\mathcal{Y}.
\end{align}
This distinction is important for robustness.  If robustness is defined only by
the model's own local Lipschitz constant, such as the spectral norm
\(\|D\mathcal{F}_\theta[x]\|\), then a smaller value would mean that the neural
operator itself changes less under input perturbations.  StablePDENet reports
this type of model-Jacobian spectral norm as a stability diagnostic
\citep{huang2026stablepdenet}.  We do not assume that a PDE surrogate should be
robust in this model-only sense.  The exact solver \(\mathcal{G}\) can have
non-negligible input-output sensitivity, so changing the input should normally
change both \(\mathcal{G}(x)\) and \(\mathcal{F}_\theta(x)\).  The desired local
property is instead that the model and solver move together, i.e.,
\(D\mathcal{F}_\theta[x]\approx D\mathcal{G}[x]\), so the error derivative
\(D\mathcal{E}_\theta[x]\) is small; the norm
\(\|D\mathcal{F}_\theta[x]\|\) itself need not be small. This gives the first
refinement over model-Jacobian spectral diagnostics: use the error operator
\(\mathcal{E}_\theta\), not the model operator alone. The second refinement is
that even the spectral norm of \(D\mathcal{E}_\theta[x]\) need not measure the
fastest growth of the actual squared error loss; that loss grows in the
error-aligned adjoint direction \(D\mathcal{E}_\theta[x]^*e_\theta[x]\).

For an infinitesimal direction $h\in\mathcal{X}$,
\begin{align}
  \mathcal{F}_\theta(x+h)
  &=
  \mathcal{F}_\theta(x)+D\mathcal{F}_\theta[x](h)
  +o(\|h\|_{\mathcal{X}}),
  \nonumber\\
  \mathcal{G}(x+h)
  &=
  \mathcal{G}(x)+D\mathcal{G}[x](h)
  +o(\|h\|_{\mathcal{X}}),
  \nonumber\\
  e_\theta[x+h]
  &=
  e_\theta[x]+D\mathcal{E}_\theta[x](h)
  +o(\|h\|_{\mathcal{X}}).
\end{align}
Thus $h$ denotes a local perturbation direction, while later finite-budget
attacks use a finite perturbation $\delta$. We define the perturbation ball as
the admissible constraint set for perturbed inputs in the function space
$\mathcal{X}$:
\begin{equation}
  \mathcal{B}_{p,\varepsilon}(x)
  =
  \{x+\delta:\delta\in\mathcal{X},
  \|\delta\|_{\mathcal{X},p}\le\varepsilon\}.
  \label{eq:perturbation-ball}
\end{equation}
It is the radius-$\varepsilon$ ball around $x$ under the
$\mathcal{X}$-space norm.

All perturbation budgets and errors are function norms. When inner products,
adjoints, or first variations are used below, we take $p=q=2$ by default.
In implementation, finite-dimensional discretizations approximate these
infinite-dimensional functions and operators.
\tabref{tab:discrete-notation} gives the corresponding vector-space
equivalents.

\begin{table}[H]
\centering
\footnotesize
\setlength{\tabcolsep}{3pt}
\renewcommand{\arraystretch}{1.25}
\caption{Discretized notation.}
\label{tab:discrete-notation}
\begin{tabular*}{\columnwidth}{@{}p{0.39\columnwidth}p{0.55\columnwidth}@{}}
\toprule
Function-space object & Vector-space equivalent \\
\midrule
Functions \(x,y,e_\theta[x]\) &
vectors \(x,y,e_\theta(x)\) \\
Operators \(\mathcal{F}_\theta,\mathcal{G},\mathcal{E}_\theta\) &
maps \(\mathcal{F}_\theta,\mathcal{G},\mathcal{E}_\theta\) \\
Function norms &
vector norms \\
Local derivative \(D\mathcal{E}_\theta[x]\) &
Jacobian \(J_{\mathcal{E}}(x)\) \\
Adjoint action \(D\mathcal{E}_\theta[x]^*e_\theta[x]\) &
\(J_{\mathcal{E}}(x)^\top e_\theta(x)\) \\
Function inner product &
vector dot product \\
\begin{tabular}[t]{@{}l@{}}Operator spectral norm\\and singular functions\end{tabular} &
\begin{tabular}[t]{@{}l@{}}matrix spectral norm\\and singular vectors\end{tabular} \\
\bottomrule
\end{tabular*}
\end{table}

\subsection{Generalization and Robustness Metrics}

\par\smallskip
\noindent\makebox[1.3em][l]{\textbf{1.}}\textbf{Generalization.}\quad
Generalization is the static model-solver discrepancy at a fixed input
function:
\begin{equation}
  \mathcal{L}_{\mathrm{gen}}(x)
  =
  \|e_\theta[x]\|_{\mathcal{Y},q}
  =
  \|\mathcal{F}_\theta(x)-\mathcal{G}(x)\|_{\mathcal{Y},q}.
\end{equation}

\par\smallskip
\noindent\makebox[1.3em][l]{\textbf{2.}}\textbf{Robustness.}\quad
Robustness asks how the model-solver error changes when the input function is
perturbed; locally, it is the largest directional derivative of the scalar error
loss \(\mathcal{L}_{\mathrm{err}}=\tfrac12\mathcal{L}_{\mathrm{gen}}^2\) at
\(x\). We organize it into three local metrics.

For all three metrics below, assume $\mathcal{E}_\theta$ is Frechet
differentiable at $x$. We use $h$ for an infinitesimal direction and keep
$\delta$ for a finite adversarial perturbation. Along $h$,
\begin{equation}
  e_\theta[x+t h]
  =
  e_\theta[x]+tD\mathcal{E}_\theta[x](h)+o(t),
  \qquad t\to 0,
\end{equation}
where $D\mathcal{E}_\theta[x]:\mathcal{X}\to\mathcal{Y}$ is a bounded linear
operator. In the Hilbert norm used for first variations, define the scalar
error loss as one half of the squared generalization loss:
\begin{align}
  \mathcal{L}_{\mathrm{err}}(x)
  &:=
  \frac{1}{2}\mathcal{L}_{\mathrm{gen}}(x)^2
  =
  \frac{1}{2}\|e_\theta[x]\|_{\mathcal{Y}}^2
  =
  \frac{1}{2}\|\mathcal{E}_\theta(x)\|_{\mathcal{Y}}^2
  \nonumber\\
  &=
  \frac{1}{2}\|\mathcal{F}_\theta(x)-\mathcal{G}(x)\|_{\mathcal{Y}}^2 .
\end{align}
Before taking the first variation, define the operator-level Jacobian-error
function
\begin{equation}
  g_\theta(x):=D\mathcal{E}_\theta[x]^*e_\theta[x]\in\mathcal{X},
\end{equation}
where $D\mathcal{E}_\theta[x]^*:\mathcal{Y}\to\mathcal{X}$ is the adjoint
derivative. After discretization this becomes the vector
$J_{\mathcal{E}}(x)^\top e_\theta(x)$. The first variation is
\begin{align}
  D\mathcal{L}_{\mathrm{err}}[x](h)
  &:=
  \left.\frac{d}{dt}
  \frac{1}{2}
  \langle e_\theta[x+t h],e_\theta[x+t h]\rangle_{\mathcal{Y}}
  \right|_{t=0}
  \nonumber\\
  &=
  \langle e_\theta[x],D\mathcal{E}_\theta[x](h)\rangle_{\mathcal{Y}}
  =
  \langle D\mathcal{E}_\theta[x]^*e_\theta[x],h\rangle_{\mathcal{X}}
  \nonumber\\
  &=
  \langle g_\theta(x),h\rangle_{\mathcal{X}}.
\end{align}
This derivative is the first-order growth rate of the squared model-solver error
loss when $x$ is perturbed along $h$. It is the function-space inner product
between the Jacobian-error function $g_\theta(x)$ and the perturbation direction
$h$. Hence the local loss-growth robustness value is
\begin{align*}
  \max_{\|h\|_{\mathcal{X}}\le1}
  D\mathcal{L}_{\mathrm{err}}[x](h)
  &=
  \|g_\theta(x)\|_{\mathcal{X}},\\
  h^*=
  \frac{g_\theta(x)}{\|g_\theta(x)\|_{\mathcal{X}}}
  &\quad (g_\theta(x)\neq0).
\end{align*}
This maximizer follows the adjoint-error direction, not generally the leading
right singular direction of \(D\mathcal{E}_\theta[x]\).

\smallskip
\noindent\textbf{Jacobian operator norm.}
The first robustness metric is the induced norm of the derivative of the
error operator. In a Lipschitz view, a regional robustness bound uses the
largest derivative norm over an input region; at a fixed input $x$, this reduces
to the pointwise induced norm of the Frechet derivative. The Frechet derivative
is defined from perturbations $x+t h$ as $t\to0$; after this infinitesimal scale
is factored out, the operator norm maximizes over normalized input-space
directions $h$:
\begin{equation}
  \|D\mathcal{E}_\theta[x]\|_{p\to q}
  =
  \sup_{\substack{h\in\mathcal{X}\\ \|h\|_{\mathcal{X},p}\le 1}}
  \|D\mathcal{E}_\theta[x](h)\|_{\mathcal{Y},q}.
\end{equation}
For a small perturbation $t h$, the linearized change is
$tD\mathcal{E}_\theta[x](h)$. Thus the operator norm gives the first-order
growth rate of the nonlinear error operator at $x$. A growth direction is any
maximizing direction
\begin{equation}
  h_{\mathrm{op}}^*
  \in
  \operatorname*{arg\,max}_{\substack{h\in\mathcal{X}\\ \|h\|_{\mathcal{X},p}\le 1}}
  \|D\mathcal{E}_\theta[x](h)\|_{\mathcal{Y},q},
\end{equation}
when the maximum is attained. After discretization, with
$J=J_{\mathcal{E}}(x)$, common cases are
$2\to2:\sigma_{\max}(J)$ with the leading right singular vector as direction,
$1\to1:\max_j\sum_i|J_{ij}|$ with a largest-column coordinate direction, and
$\infty\to\infty:\max_i\sum_j|J_{ij}|$ with a sign vector aligned with a
largest row.

\smallskip
\noindent\textbf{Adversarial attack loss increase.}
For a fixed input function $x$, the attack optimizes over the perturbation
$\delta$; the model, solver, and clean input are fixed. This metric directly
measures the worst loss increase over a norm-bounded perturbation ball, rather
than a local Jacobian/operator norm. For one perturbation, define
\begin{align}
  \Delta\mathcal{L}_{\mathrm{err}}(\delta)
  &:=
  \mathcal{L}_{\mathrm{err}}(x+\delta)
  -
  \mathcal{L}_{\mathrm{err}}(x).
\end{align}
Since the clean error $\|e_\theta[x]\|_{\mathcal{Y}}^2$ is constant in
$\delta$, maximizing $\|e_\theta[x+\delta]\|_{\mathcal{Y}}^2$ and maximizing
$\Delta\mathcal{L}_{\mathrm{err}}(\delta)$ give the same perturbation. The
finite-budget attack and its reported growth amount are
\begin{align}
  \delta_\varepsilon^*
  &\in
  \operatorname*{arg\,max}_{\|\delta\|_{\mathcal{X}}\le\varepsilon}
  \|e_\theta[x+\delta]\|_{\mathcal{Y}}^2,
  \nonumber\\
  \Delta\mathcal{L}_{\mathrm{err}}^*(\varepsilon)
  &:=
  \Delta\mathcal{L}_{\mathrm{err}}(\delta_\varepsilon^*)
  =
  \max_{\|\delta\|_{\mathcal{X}}\le\varepsilon}
  \Delta\mathcal{L}_{\mathrm{err}}(\delta).
\end{align}

\smallskip
\noindent\textbf{Jacobian-error metric.}
The third robustness metric is the Jacobian-error metric, defined by the
input-space function
$g_\theta(x)=D\mathcal{E}_\theta[x]^*e_\theta[x]$. It combines the derivative
of the error operator with the current error field. Its scalar growth value is
$\|g_\theta(x)\|_{\mathcal{X}}$, and its growth direction is
$g_\theta(x)/\|g_\theta(x)\|_{\mathcal{X}}$ when $g_\theta(x)\neq0$.

\smallskip
\noindent\textbf{Comparison with finite adversarial attack.}
This comparison explains why the finite-budget attack reduces to the
Jacobian-error metric in the small-budget limit. Write
$\delta=\varepsilon h$ with $\|h\|_{\mathcal{X}}\le 1$. As
$\varepsilon\to0$,
\begin{align}
  \Delta\mathcal{L}_{\mathrm{err}}(\varepsilon h)
  &=
  \varepsilon D\mathcal{L}_{\mathrm{err}}[x](h)+o(\varepsilon)
  \nonumber\\
  &=
  \varepsilon\langle g_\theta(x),h\rangle_{\mathcal{X}}+o(\varepsilon).
\end{align}
Therefore
\begin{align}
  \Delta\mathcal{L}_{\mathrm{err}}^*(\varepsilon)
  &=
  \varepsilon
  \max_{\|h\|_{\mathcal{X}}\le 1}
  \langle g_\theta(x),h\rangle_{\mathcal{X}}
  +o(\varepsilon)
  \nonumber\\
  &=
  \varepsilon
  \left\langle
  g_\theta(x),
  \frac{g_\theta(x)}{\|g_\theta(x)\|_{\mathcal{X}}}
  \right\rangle_{\mathcal{X}}
  +o(\varepsilon)
  \nonumber\\
  &=
  \varepsilon\|g_\theta(x)\|_{\mathcal{X}}+o(\varepsilon).
  \nonumber\\
  \frac{\delta_\varepsilon^*}{\varepsilon}
  &\to
  \frac{g_\theta(x)}{\|g_\theta(x)\|_{\mathcal{X}}},
  \qquad g_\theta(x)\neq0 .
\end{align}
Thus, in the small-budget limit, the finite-budget adversarial attack has the
same first-order growth value and direction as the Jacobian-error metric.

\smallskip
\noindent\textbf{Comparison with Jacobian operator norm.}
The Jacobian-error metric differs from the Jacobian operator norm because the
two metrics optimize different local objectives. The
Jacobian-error metric measures the error-aligned first-order growth of the
final squared error:
\begin{align}
  \frac{1}{2}\|e_\theta[x+h]\|_{\mathcal{Y}}^2
  -
  \frac{1}{2}\|e_\theta[x]\|_{\mathcal{Y}}^2
  &=
  \bigl\langle e_\theta[x],
  D\mathcal{E}_\theta[x](h)\bigr\rangle_{\mathcal{Y}}
  \nonumber\\
  &\quad+
  O(\|h\|_{\mathcal{X}}^2).
\end{align}
This first-order term depends on both the current error $e_\theta[x]$ and the
local Jacobian through $D\mathcal{E}_\theta[x](h)$, so the local growth
direction is governed by $D\mathcal{E}_\theta[x]^*e_\theta[x]$. In contrast,
the operator norm measures the largest local displacement of the error field:
\begin{equation}
  \|e_\theta[x+h]-e_\theta[x]\|_{\mathcal{Y}}
  \approx
  \|D\mathcal{E}_\theta[x](h)\|_{\mathcal{Y}}.
\end{equation}
Thus, the operator norm maximizes the size of the local error displacement,
not the error-aligned loss increase. This error-vector geometry is illustrated
in \figref{fig:error-difference-vs-final-norm}.
In the finite-dimensional, or compact Hilbert $2$-norm, case, the
operator-norm input direction is the leading right singular direction of
$D\mathcal{E}_\theta[x]$, equivalently governed by
$D\mathcal{E}_\theta[x]^*D\mathcal{E}_\theta[x]$. The Jacobian-error and
small-budget attack directions are governed by
$D\mathcal{E}_\theta[x]^*e_\theta[x]$. According to
\appref{app:robustness-metric-diagnostics}, the two agree only when the current
error field has a dominant weighted projection onto the leading left singular
function; otherwise, their directions need not be close. Our experiments
confirm this geometry: the finite adversarial attack metric tracks the
Jacobian-error metric more closely than the operator-norm metric, especially
for small $\varepsilon$. Hence, in our
solver-consistent robustness setting, the usual spectral/operator-norm metric
is not the most suitable primary robustness metric. Empirically, the proposed
Jacobian-error metric is also fast to compute: for a discretized
$J_{\mathcal{E}}\in\mathbb{R}^{m\times n}$ and error vector
$e_\theta\in\mathbb{R}^{m}$, computing
$J_{\mathcal{E}}^\top e_\theta$ and its norm costs $O(mn)$, whereas computing
the leading singular value/vector for the operator norm costs
$O(mn\min\{m,n\})$ for an exact dense SVD or $O(kmn)$ for $k$ iterative
matrix-vector passes. Thus, the Jacobian-error metric is a practical
lightweight local proxy.

\begin{table*}[t]
\centering
\footnotesize
\setlength{\tabcolsep}{4pt}
\renewcommand{\arraystretch}{1.55}
\caption{Comparison of the three local robustness metrics. The first column
labels growth value and growth direction.}
\label{tab:local-robustness-metrics}
\begin{tabular*}{\textwidth}{@{\extracolsep{\fill}}p{0.115\textwidth}p{0.265\textwidth}p{0.265\textwidth}p{0.265\textwidth}@{}}
\toprule
Metric
&
Jacobian operator norm
&
Finite adversarial attack
&
Jacobian-error / $J^\top$ times error
\\
\midrule
\begin{tabular}[t]{@{}l@{}}
\textbf{Growth}\\
\textbf{value}
\end{tabular}
&
\(
\begin{aligned}
&\|D\mathcal{E}_\theta[x]\|_{p\to q}\\
&\text{linearized growth over }\varepsilon:\\
&\varepsilon\|D\mathcal{E}_\theta[x]\|_{p\to q}+o(\varepsilon)
\end{aligned}
\)
&
\(
\begin{aligned}
&\Delta\mathcal{L}_{\mathrm{err}}^*(\varepsilon)
=\max_{\|\delta\|_{\mathcal{X}}\le\varepsilon}
\Delta\mathcal{L}_{\mathrm{err}}(\delta)\\
&\Delta\mathcal{L}_{\mathrm{err}}^*(\varepsilon)
=\varepsilon\|g_\theta(x)\|_{\mathcal{X}}+o(\varepsilon)
\end{aligned}
\)
&
\(
\begin{aligned}
&g_\theta(x)=D\mathcal{E}_\theta[x]^*e_\theta[x]\\
&\text{discrete: }J_{\mathcal{E}}(x)^\top e_\theta(x)\\
&\text{unit-budget scalar: }\|g_\theta(x)\|_{\mathcal{X}}\\
&\text{budgeted scalar: }
\ \varepsilon\|g_\theta(x)\|_{\mathcal{X}}
\end{aligned}
\)
\\
\midrule
\begin{tabular}[t]{@{}l@{}}
\textbf{Growth}\\
\textbf{direction}
\end{tabular}
&
\begin{minipage}[t]{0.265\textwidth}
\raggedright
\(h_{\mathrm{op}}^*\): unit-direction maximizer of
\(\|D\mathcal{E}_\theta[x](h)\|_{\mathcal{Y},q}\)
over \(\|h\|_{\mathcal{X},p}\le1\).
\end{minipage}
&
\(
\begin{aligned}
&\delta_\varepsilon^*
\in
\operatorname*{arg\,max}_{\|\delta\|_{\mathcal{X}}\le\varepsilon}
\Delta\mathcal{L}_{\mathrm{err}}(\delta)\\
&\frac{\delta_\varepsilon^*}{\varepsilon}
\to
\frac{g_\theta(x)}{\|g_\theta(x)\|_{\mathcal{X}}}
\quad(\varepsilon\to0)
\end{aligned}
\)
&
\(
\begin{aligned}
&h_{\mathrm{JE}}^*
=
\frac{g_\theta(x)}{\|g_\theta(x)\|_{\mathcal{X}}}\\
&\text{discrete: }
\frac{J_{\mathcal{E}}(x)^\top e_\theta(x)}
{\|J_{\mathcal{E}}(x)^\top e_\theta(x)\|}
\end{aligned}
\)
\\
\bottomrule
\end{tabular*}
\end{table*}

\section{Proposed Framework}

\subsection{Solver-Integrated Adversarial Attack}
\label{subsec:solver-integrated-attack}

The notation above separates three finite attack objective functions. In this
subsection, $x$ is fixed and the optimization variable is the finite
perturbation $\delta$:
\begin{align}
  \mathcal{L}_1(\delta) &=
  \|\mathcal{F}_\theta(x+\delta)-\mathcal{F}_\theta(x)\|_{\mathcal{Y},q},
  \label{eq:l1-attack-loss}\\
  \mathcal{L}_2(\delta) &=
  \|\mathcal{F}_\theta(x+\delta)-\mathcal{G}(x)\|_{\mathcal{Y},q},\\
  \mathcal{L}_3(\delta) &=
  \|\mathcal{F}_\theta(x+\delta)-\mathcal{G}(x+\delta)\|_{\mathcal{Y},q}
  =\|e_\theta[x+\delta]\|_{\mathcal{Y},q}.
\end{align}
For $i\in\{1,2,3\}$, the corresponding attack optimization is
\begin{equation}
  \delta_{i,\varepsilon}^*
  \in
  \operatorname*{arg\,max}_{\|\delta\|_{\mathcal{X}}\le\varepsilon}
  \mathcal{L}_i(\delta),
\end{equation}
with $\delta$ as the optimized variable.
The evaluation reference is always the solver at the same perturbed input.
For a perfect surrogate $\mathcal F_\theta=\mathcal G$ throughout the admissible
neighborhood, $\mathcal L_3=0$, while $\mathcal L_1=\mathcal L_2$ may be large
because the true solution changes. This separates the objective used to find a
perturbation from the solver-referenced criterion used to assess its damage;
it applies regardless of the training objective.
These losses correspond to different attack conventions. Model-output-change
objectives, including attacks or regularizers that compare perturbed and clean
predictions, are closest to $\mathcal{L}_1$
\citep{nguyen2018regression,mode2020mtsregression,liu2022forecastingattacks}.
Fixed-label classification, fixed-response regression, specified-target
regression/forecasting, and neural-operator tests with precomputed solver
targets are closest to $\mathcal{L}_2$, because the reference is held fixed
rather than recomputed at $x+\delta$
\citep{goodfellow2015explaining,madry2018towards,tong2018adversarialregression,ribeiro2023linearregression,liu2022forecastingattacks,adesoji2022evaluating}.

The high-level losses above have implementation-level variants. Let
$x_k=x+\delta_k$ be the current perturbed input and let
$\operatorname{sg}(\cdot)$ denote stop-gradient. A fixed-target variant keeps the
clean solver reference, whereas an updated variant uses $x_k$:
\begin{align}
  \mathcal{L}_2^{\mathrm{fixed}}(\delta_k)
  &=
  \|\mathcal{F}_\theta(x_k)-\mathcal{G}(x)\|_{\mathcal{Y},q},
  \nonumber\\
  \mathcal{L}_3^{\mathrm{full}}(\delta_k)
  &=
  \|\mathcal{F}_\theta(x_k)-\mathcal{G}(x_k)\|_{\mathcal{Y},q}.
\end{align}
The fixed-target version uses the solver output only as a held reference label,
so the attack gradient does not pass through the solver. The updated
$\mathcal{L}_3$ version instead pairs the model output with the solver output at
the same perturbed input.
Dictionary attacks replace the online solver output at $x_k$ by
$y_{\mathrm{dict},N}(x_k)$, the precomputed solver output attached to the nearest
dictionary input to $x_k$. The loss is
\begin{equation*}
  \mathcal{L}_{2,N}^{\mathrm{dict}}(\delta_k)
  =
  \|\mathcal{F}_\theta(x_k)-y_{\mathrm{dict},N}(x_k)\|_{\mathcal{Y},q}.
\end{equation*}
This is the $\mathcal{L}_2$-like structure used by dictionary-based FNO
robustness evaluation \citep{adesoji2022evaluating}; it uses an external solver
target without differentiating through the solver. A stop-gradient solver
variant uses the current solver output in the target loss but blocks the
solver branch in backpropagation:
\begin{equation}
  \mathcal{L}_3^{\mathrm{sg}}(\delta_k)
  =
  \|\mathcal{F}_\theta(x_k)-\operatorname{sg}(\mathcal{G}(x_k))\|_{\mathcal{Y},q}.
\end{equation}
For the squared-loss version, the backpropagated first-order directions differ
as
\begin{align}
  \nabla_{\delta_k}\tfrac{1}{2}\mathcal{L}_3^{\mathrm{sg}}(\delta_k)^2
  &=
  D\mathcal{F}_\theta[x_k]^*
  (\mathcal{F}_\theta(x_k)-\mathcal{G}(x_k)),
  \nonumber\\
  \nabla_{\delta_k}\tfrac{1}{2}\mathcal{L}_3^{\mathrm{full}}(\delta_k)^2
  &=
  (D\mathcal{F}_\theta[x_k]-D\mathcal{G}[x_k])^*
  \nonumber\\
  &\quad
  (\mathcal{F}_\theta(x_k)-\mathcal{G}(x_k)).
\end{align}
Physics-residual attacks such as WbAR, StablePDENet, and RAMS replace
solver-output supervision by a residual surrogate
\citep{shi2025wbar,huang2026stablepdenet,ouyang2025rams}. This implementation
evaluates the predicted field $\mathcal{F}_\theta(x_k)$ in the PDE and boundary
residuals, and uses the perturbed input $x_k$ for the initial-condition term:
\begin{align}
  \mathcal{L}_{\mathrm{phys}}(\delta_k)
  &=
  \lambda_{\mathrm{PDE}}
  \|R_{\mathrm{PDE}}(\mathcal{F}_\theta(x_k))\|^2
  \nonumber\\
  &\quad+
  \lambda_{\mathrm{BC}}
  \|R_{\mathrm{BC}}(\mathcal{F}_\theta(x_k))\|^2
  +
  \lambda_{\mathrm{IC}}
  \|R_{\mathrm{IC}}(x_k)\|^2 .
\end{align}
This objective is self-supervised: it does not compare against the solver output
$\mathcal{G}(x_k)$, but only asks the predicted field to satisfy the governing
equation and boundary condition while the perturbed input satisfies the initial
condition. It has two limitations in our setting. First, a small residual is a
weaker signal than solver supervision; it can enforce local physical consistency
without ensuring that the predicted solution is close to the true solver
trajectory, especially when source terms or intermediate time evolution are not
fully supervised. In the Darcy Flow adversarial-training experiment below, the
physics-loss baseline is much less effective than direct solver supervision.
Second, many neural operators output only the target field or final state,
rather than dense neighboring time frames, so temporal derivatives needed by a
physics residual cannot be computed by finite differences. In the Burgers and
Navier--Stokes experiments, this means that model-side physics laws cannot be
computed by finite-difference temporal derivatives.
Overall, these attack variants increase solver integration in the order
$\mathcal{L}_{\mathrm{phys}}$, $\mathcal{L}_1$, $\mathcal{L}_2$,
$\mathcal{L}_{2,N}^{\mathrm{dict}}$, $\mathcal{L}_3^{\mathrm{sg}}$, and
$\mathcal{L}_3^{\mathrm{full}}$.
\figrefs{fig:attack-loss-variants-twocol}{fig:ns-recurrent-adversarial-attack}
summarize their forward and
backward paths. Our experiments below show that deeper solver integration
generally yields stronger attacks and more effective adversarial training.
Full and stop-gradient L3 have identical forward values at a fixed input;
their different search directions isolate the effect of solver differentiation.
The VJP/adjoint interface and the forward-only fallback are detailed in
\appref{app:numerical-solver-details}.

\subsection{Optimization Methods for Solver-Integrated Attacks}

The losses above specify what is attacked; the optimizer specifies how the
perturbation is updated. Let $\mathcal{A}(\delta)$ denote any chosen attack
objective and let
$g_k=\nabla_\delta\mathcal{A}(\delta_k)$ denote the input-space gradient at
attack step $k$. Raw updates use $g_k$ as the search vector. Let $p^\star$ be
the dual exponent, defined by
$1/p+1/p^\star=1$, with the endpoint convention
$1^\star=\infty$ and $\infty^\star=1$. Steepest updates use $s_k$, the
$p$-norm steepest ascent direction associated with the dual norm of $g_k$:
\begin{equation}
  s_k \in \arg\max_{\|v\|_{\mathcal{X},p}\le 1}
  \langle g_k,v\rangle .
\end{equation}
Equivalently,
$\max_{\|v\|_{\mathcal{X},p}\le 1}\langle g_k,v\rangle
=\|g_k\|_{\mathcal{X},p^\star}$, so $s_k$ is the primal direction that attains
the dual norm of the gradient.

\begin{table}[H]
\centering
\footnotesize
\caption{Four attack optimizers under one projected framework.}
\label{tab:attack-optimizers}
\setlength{\tabcolsep}{1pt}
\begin{tabular}{p{0.17\columnwidth}p{0.375\columnwidth}p{0.375\columnwidth}}
\toprule
 & \textbf{Add} & \textbf{Replace}\\
\midrule
\textbf{Raw}
&
\begin{tabular}[t]{@{}l@{}}
\textbf{raw add}\\
$\delta_{k+1}=$\\[-1pt]
$\Pi_{\mathcal{B}_\varepsilon}(\delta_k+\alpha g_k)$
\end{tabular}
&
\begin{tabular}[t]{@{}l@{}}
\textbf{raw replace}\\
$\delta_{k+1}=\varepsilon$\\[-1pt]
$g_k/\|g_k\|_{\mathcal X,p}$
\end{tabular}
\\
\midrule
\textbf{Steepest}
&
\begin{tabular}[t]{@{}l@{}}
\textbf{steepest add}\\
$\delta_{k+1}=$\\[-1pt]
$\Pi_{\mathcal{B}_\varepsilon}(\delta_k+\alpha s_k)$
\end{tabular}
&
\begin{tabular}[t]{@{}l@{}}
\textbf{steepest replace}\\
$\delta_{k+1}=\varepsilon s_k$
\end{tabular}
\\
\bottomrule
\end{tabular}
\end{table}

Add-style methods accumulate: the new direction is added to the previous
perturbation and then projected. Replace-style methods discard the previous
perturbation and directly replace it with a new boundary perturbation. Since
first-order norm-constrained attack models usually use the full perturbation
budget
\citep{goodfellow2015explaining,kurakin2016physical,madry2018towards}, add-style
updates can spend early steps moving toward the boundary, whereas replace-style
updates start on it.

The four cells in \tabref{tab:attack-optimizers} have simple optimization
interpretations: raw add is standard PGD; steepest add is $\ell_p$-steepest PGD;
raw replace is normalized-gradient boundary replacement; steepest replace is
steepest-direction boundary replacement. In the Euclidean case $p=2$, the two
replace updates coincide. For the local linearized $\mathcal{L}_1$ model at a
fixed input, with $q=2$ and $J=D\mathcal{F}_\theta[x]$, this replace update
reduces to normalized multiplication by the local matrix $J^\top J$:
\begin{equation}
  \delta_{k+1}
  =
  \varepsilon
  \frac{J^\top J\delta_k}{\|J^\top J\delta_k\|_2}.
\end{equation}
In this limited setting, the replace update is power-iteration-like; for
general \(p,q\), the steepest replace variant is analogous to a generalized
power iteration for the corresponding frozen linear operator-norm problem.
This is a frozen-Jacobian local statement, not a global linear description of
the finite nonlinear attack.

A small implementation caveat follows for $\mathcal{L}_1$: it must not be
started from the zero perturbation. Although \eqref{eq:l1-attack-loss} contains
no explicit Jacobian, Taylor expansion around $x$ gives the local Jacobian form
\begin{align*}
  \mathcal{F}_\theta(x+\delta)-\mathcal{F}_\theta(x)
  &= J\delta+O(\|\delta\|^2),\quad J=D\mathcal{F}_\theta[x],\\
  g_k=\nabla_\delta\tfrac{1}{2}\mathcal{L}_1(\delta_k)^2
  &\approx A\delta_k,\quad A=J^\top J,\\
  \delta_{k+1}^{\mathrm{add}}
  &\approx (I+\alpha A)\delta_k .
\end{align*}
Thus $\delta_k=0$ gives $g_k=0$, so additive updates cannot move and replacement
updates have no direction to normalize. The formulas above are only a local
linearization of a nonlinear attack objective, with $J$ frozen at the current
input. Locally, an add step applies $I+\alpha A$ and a replacement step would
normalize $A\delta_k$; both require a nonzero current direction. Thus a tiny
nonzero initial function is needed. This should not be read as a global
quadratic or eigenvalue characterization of the finite nonlinear attack.

For solver-integrated losses, any frozen-Jacobian view is only heuristic because
the squared local model contains offset terms rather than a pure homogeneous
quadratic. If $\delta$ is treated as infinitesimal and both linear and
quadratic terms are present, the linear terms dominate locally:
\begin{align}
  \mathcal{L}_1^2:\quad
  \|J\delta\|_2^2
  &=
  \delta^\top J^\top J\delta,
  \nonumber\\
  \mathcal{L}_2^2:\quad
  \|r_0+J\delta\|_2^2
  &=
  \delta^\top J^\top J\delta
  +2\langle J^\top r_0,\delta\rangle+\|r_0\|_2^2,
  \nonumber\\
  \mathcal{L}_3^2:\quad
  \|e_0+H\delta\|_2^2
  &=
  \delta^\top H^\top H\delta
  +2\langle H^\top e_0,\delta\rangle+\|e_0\|_2^2,
\end{align}
where $r_0=\mathcal{F}_\theta(x)-\mathcal{G}(x)$,
$e_0=e_\theta[x]$, and $H=D\mathcal{F}_\theta[x]-D\mathcal{G}[x]$.
Thus $\mathcal{L}_2$ and $\mathcal{L}_3$ are not homogeneous local quadratic
problems. Our experiments reflect this distinction: replace-style updates are
fast and often strong on Burgers and Darcy Flow, but can be too aggressive for
recurrent Navier--Stokes, where add-style updates are more stable.

\subsection{Solver-Integrated Adversarial Training}

The training stage wraps the attack objectives above in an online
attack-then-train loop. We start from a trained baseline neural operator. At
step $t$, the current model is frozen during attack generation, and an attacked
batch $x^{\mathrm{adv}}=x+\delta_N$ is obtained by maximizing one of
$\mathcal{L}_1$, $\mathcal{L}_2$, $\mathcal{L}_3$, or
$\mathcal{L}_{\mathrm{phys}}$ for $N$ projected steps. This attack step searches
for locally high-loss input functions where the current model has large
model-solver discrepancy or large physics residual. The trainable model is
then updated on the attacked batch, and the updated weights become the next
current model.

This process can be summarized as a function-space min-max risk.  Let
$\mu_{\mathcal X}$ be the data distribution over the input function space
$\mathcal X$, so that $x\sim\mu_{\mathcal X}$.  The usual expected risk is
\begin{equation*}
  \mathcal R(\theta)
  =
  \mathbb E_{x\sim\mu_{\mathcal X}}
  \left[
    \mathcal L_\theta(x)
  \right].
\end{equation*}
After the inner attack, define
\begin{align*}
  \delta_\theta(x)
  &\in
  \operatorname*{arg\,max}_{\|\delta\|_{\mathcal X,p}\le\varepsilon}
  \mathcal L_\theta(x;\delta),\\
  x^{\mathrm{adv}}
  &=T_\theta(x)=x+\delta_\theta(x),\\
  \mu_\theta^{\mathrm{adv}}
  &=(T_\theta)_\#\mu_{\mathcal X}.
\end{align*}
The transformation \(T_\theta:\mathcal X\to\mathcal X\) sends the input-function
distribution \(\mu_{\mathcal X}\) to another distribution on the same function
space by adding the perturbation \(\delta_\theta(x)\).  The resulting
attack-induced pushforward distribution is
\(\mu_\theta^{\mathrm{adv}}=(T_\theta)_\#\mu_{\mathcal X}\), which places more
probability near local high-loss regions of the current loss landscape.  The
corresponding local robust risk is
\begin{align*}
  \mathcal R_{\mathrm{rob}}(\theta)
  &=
  \mathbb E_{x\sim\mu_{\mathcal X}}
  \left[
    \max_{\|\delta\|_{\mathcal X,p}\le \varepsilon}
    \mathcal L_\theta(x;\delta)
  \right]\\
  &=
  \mathbb E_{x\sim\mu_{\mathcal X}}
  \left[
    \mathcal L_\theta(T_\theta(x))
  \right]
  =
  \mathbb E_{x^{\mathrm{adv}}\sim\mu_\theta^{\mathrm{adv}}}
  \left[
    \mathcal L_\theta(x^{\mathrm{adv}})
  \right].
\end{align*}
The adversarial training objective is therefore
\begin{align*}
  \min_\theta \mathcal R_{\mathrm{rob}}(\theta)
  &=
  \min_\theta
  \mathbb E_{x\sim\mu_{\mathcal X}}
  \left[
    \max_{\|\delta\|_{\mathcal X,p}\le \varepsilon}
    \mathcal L_\theta(x;\delta)
  \right]\\
  &=
  \min_\theta
  \mathbb E_{x\sim\mu_{\mathcal X}}
  \left[
    \mathcal L_\theta(T_\theta(x))
  \right]\\
  &=
  \min_\theta
  \mathbb E_{x^{\mathrm{adv}}\sim\mu_\theta^{\mathrm{adv}}}
  \left[
    \mathcal L_\theta(x^{\mathrm{adv}})
  \right].
\end{align*}
Equivalently, on a finite training batch
\(\mathcal B=\{x_i\}_{i=1}^N\) drawn from \(\mu_{\mathcal X}\), the same
quantities are empirical sums.  The samples are first drawn from the data
distribution, but the attack then moves each sample toward a high-loss region
on the current loss landscape before it enters the training loss.  The
low-dimensional visualization in
\figref{fig:loss-landscape-pgd-visualization} illustrates this
loss-guided motion of sampled input functions:
\begin{align*}
  \delta_{\theta,i}
  &\in
  \operatorname*{arg\,max}_{\|\delta_i\|_{\mathcal X,p}\le\varepsilon}
  \mathcal L_\theta(x_i;\delta_i),\\
  x_i^{\mathrm{adv}}
  &=T_\theta(x_i)=x_i+\delta_{\theta,i}.
\end{align*}
The empirical risk, empirical robust risk, and empirical adversarial objective
are then
\begin{align*}
  \widehat{\mathcal R}_N(\theta)
  &=
  \frac{1}{N}\sum_{i=1}^N
  \mathcal L_\theta(x_i),\\
  \widehat{\mathcal R}_{\mathrm{rob},N}(\theta)
  &=
  \frac{1}{N}\sum_{i=1}^N
  \max_{\|\delta_i\|_{\mathcal X,p}\le\varepsilon}
  \mathcal L_\theta(x_i;\delta_i)\\
  &=
  \frac{1}{N}\sum_{i=1}^N
  \mathcal L_\theta(x_i^{\mathrm{adv}}),\\
  \min_\theta\widehat{\mathcal R}_{\mathrm{rob},N}(\theta)
  &=
  \min_\theta
  \frac{1}{N}\sum_{i=1}^N
  \max_{\|\delta_i\|_{\mathcal X,p}\le\varepsilon}
  \mathcal L_\theta(x_i;\delta_i)\\
  &=
  \min_\theta
  \frac{1}{N}\sum_{i=1}^N
  \mathcal L_\theta(x_i^{\mathrm{adv}}).
\end{align*}
The full pipeline and both random perturbation variants are shown in
\figrangeref{fig:adversarial-training-pipeline}{fig:random-solver-training-pipeline}.

This is a PDE operator-learning embodiment of standard adversarial training:
for a chosen attack objective and corresponding training loss, the training
batches directly use attacked samples rather than clean samples.
Existing theory says that the inner attack quality matters: Wang et al. relate
inner-maximization convergence quality to robustness and give a min-max
convergence guarantee under their assumptions, while TRADES gives a
robustness-accuracy bound for adversarial training
\citep{goodfellow2015explaining,madry2018towards,wang2019convergence,zhang2019trades}.
Here the classification label is replaced by a solver output or by a physics
residual.

Experimentally, we also compare with two hand-designed random perturbation
variants. Unlike adversarial training, which adapts $\delta_N$ to the current
model-solver discrepancy, these variants draw perturbations from a fixed random
sampler; kernel choices and Burgers per-sample $\epsilon$ jitter are independent
of the current model. They differ only in target: \emph{random solver} uses
$y=G(x+\delta)$, while \emph{random clean} keeps $y=G(x)$.

\section{Experiments}

\subsection{Benchmark PDEs}

We evaluate on the three canonical PDE operator-learning benchmarks used in the
Fourier Neural Operator (FNO) paper \citep{li2021fourier}: 1D Burgers, 2D
Darcy Flow, and 2D Navier--Stokes. Their
equations, boundary conditions, input distributions, and solver details are
collected in \apprefs{app:benchmark-pdes}{app:numerical-solver-details}.

\subsection{Solver-Integrated Adversarial Attacks}
\label{subsec:exp-solver-integrated-attacks}

For Burgers and recurrent Navier--Stokes, we compare
$\mathcal{L}_1$, $\mathcal{L}_2$, and $\mathcal{L}_3$. For Darcy Flow, we also
include the physics-residual objective $\mathcal{L}_{\mathrm{phys}}$ as a fourth
attack objective. Auxiliary sweeps further include $\mathcal{L}_3^{\mathrm{sg}}$,
$\mathcal{L}_{2,N}^{\mathrm{dict}}$ with $N\in\{200,2000,20000\}$, and
$\mathcal{L}_2^{\mathrm{fixed}}$. See the attack-loss diagram in
\figref{fig:attack-loss-variants-twocol} and the attack figure set in
\appref{app:si-attack-training-figures}.

Unless stated otherwise, these attack experiments use FNO models for Burgers,
Darcy Flow, and recurrent Navier--Stokes. We additionally include Burgers
DeepONet attacks for the easier $\nu=10^{-2}$ setting; for Burgers at
$\nu=10^{-3}$, we report only FNO because the DeepONet baseline was difficult
to train to a sufficiently accurate clean model. Across these settings, the
attack-loss increase grows approximately as a power law in the attack budget,
$\Delta\mathcal{L}\approx C b^a$ (with $b=\epsilon$ for norm attacks and
$b=K$ for Darcy Flow flips), as summarized in
\figref{fig:app-g-cross-benchmark-epsilon-loss}.

\begin{table*}[t]
\centering
\footnotesize
\setlength{\tabcolsep}{2pt}
\caption{\textbf{Attack Objective Comparison.} Evaluation metric for every
column: final true $\mathcal{L}_3=\|\mathcal{F}_\theta(x_{\mathrm{adv}})
-\mathcal{G}(x_{\mathrm{adv}})\|$, reported as mean $\pm$ standard deviation.
Superscript \(^{*}\) on the Loss 3 entry marks a one-sided test against the
second-highest objective in that row after BH correction, \(q<0.05\), using
the per-sample final-\(\mathcal L_3\) distribution for that attack setting.
Fixed settings and sample sizes are: Burgers uses $\nu=0.001$, an $L_2$
attack, $100$ steps, and $N=50$ samples per row; Darcy Flow uses a binary
coefficient-flip budget, $\alpha_{\mathrm{flips}}=5$, $100$ steps, and
$N=20$ samples per row; NS2D uses $p=q=2$, an $L_2/L_2$ attack, $50$ steps,
and $N=20$ samples per row.}
\label{tab:attack_objective_true_l3_summary}
\begin{tabular}{@{}lllccccc@{}}
\toprule
System & Budget & Alpha & loss1 & loss2 & \textbf{loss3} & loss4 physics & L3 win \\
\midrule
Burgers & $\epsilon=0.125$ & $\alpha=0.0125$
& 0.1027 \(\pm\) 0.0620 & 0.1021 \(\pm\) 0.0624 & \textbf{0.1306 \(\pm\) 0.0862}\(^{*}\) & -- & 50/50 \\
Burgers & $\epsilon=0.25$ & $\alpha=0.025$
& 0.1008 \(\pm\) 0.0568 & 0.0999 \(\pm\) 0.0578 & \textbf{0.1664 \(\pm\) 0.1136}\(^{*}\) & -- & 50/50 \\
Burgers & $\epsilon=0.5$ & $\alpha=0.05$
& 0.0982 \(\pm\) 0.0486 & 0.0963 \(\pm\) 0.0494 & \textbf{0.2793 \(\pm\) 0.2118}\(^{*}\) & -- & 50/50 \\
Burgers & $\epsilon=0.75$ & $\alpha=0.075$
& 0.0976 \(\pm\) 0.0448 & 0.0934 \(\pm\) 0.0428 & \textbf{0.4612 \(\pm\) 0.3825}\(^{*}\) & -- & 50/50 \\
Burgers & $\epsilon=1.0$ & $\alpha=0.1$
& 0.0988 \(\pm\) 0.0500 & 0.0913 \(\pm\) 0.0398 & \textbf{0.7262 \(\pm\) 0.6240}\(^{*}\) & -- & 50/50 \\
\midrule
Darcy Flow & $K=100$ & $\alpha_{\mathrm{flips}}=5$
& 0.0254 \(\pm\) 0.0074 & 0.0256 \(\pm\) 0.0071 & \textbf{0.0307 \(\pm\) 0.0104}\(^{*}\) & 0.0245 \(\pm\) 0.0071 & 14/20 \\
Darcy Flow & $K=250$ & $\alpha_{\mathrm{flips}}=5$
& 0.0266 \(\pm\) 0.0081 & 0.0270 \(\pm\) 0.0069 & \textbf{0.0403 \(\pm\) 0.0142}\(^{*}\) & 0.0245 \(\pm\) 0.0071 & 17/20 \\
Darcy Flow & $K=437$ & $\alpha_{\mathrm{flips}}=5$
& 0.0276 \(\pm\) 0.0084 & 0.0284 \(\pm\) 0.0068 & \textbf{0.0500 \(\pm\) 0.0179}\(^{*}\) & 0.0244 \(\pm\) 0.0071 & 17/20 \\
Darcy Flow & $K=875$ & $\alpha_{\mathrm{flips}}=5$
& 0.0288 \(\pm\) 0.0086 & 0.0309 \(\pm\) 0.0084 & \textbf{0.0771 \(\pm\) 0.0173}\(^{*}\) & 0.0245 \(\pm\) 0.0070 & 20/20 \\
\midrule
NS2D & $\epsilon=8$ & $\alpha=0.25$
& 64.979 \(\pm\) 28.525 & 59.567 \(\pm\) 28.883 & \textbf{126.832 \(\pm\) 32.477}\(^{*}\) & -- & 20/20 \\
NS2D & $\epsilon=16$ & $\alpha=0.5$
& 68.913 \(\pm\) 26.433 & 61.469 \(\pm\) 30.800 & \textbf{186.683 \(\pm\) 35.291}\(^{*}\) & -- & 20/20 \\
NS2D & $\epsilon=32$ & $\alpha=1$
& 118.588 \(\pm\) 46.057 & 69.505 \(\pm\) 30.926 & \textbf{276.760 \(\pm\) 44.681}\(^{*}\) & -- & 20/20 \\
NS2D & $\epsilon=64$ & $\alpha=2$
& 190.810 \(\pm\) 90.681 & 88.554 \(\pm\) 25.322 & \textbf{324.618 \(\pm\) 61.349}\(^{*}\) & -- & 18/20 \\
\bottomrule
\end{tabular}%
\vspace{0.65em}

\footnotesize
\setlength{\tabcolsep}{3pt}
\renewcommand{\arraystretch}{1.08}
\caption{Adversarial-training robustness summary.  Values are mean \(\pm\)
standard deviation, and smaller is better.  Boldface marks the best method
within each PDE and diagnostic.  Burgers statistics use the same \(25\)
mixed diagnostic records for each method.  Darcy Flow attack \(\Delta\)loss
uses the \(50\) shifted datasets with \(50\) inputs each (\(n=2500\) per
method), while its two local Jacobian diagnostics use \(25\) fixed records
(\(2\) train, \(2\) test, and \(21\) shifted).  Superscripts test the bold
entry against the second-smallest entry in the same PDE and diagnostic using
the corresponding per-record distribution and a one-sided paired test after
BH correction: \(^{*}\) \(q<0.05\), and \(^{\mathrm{ns}}\) not significant.}
\label{tab:adversarial_training_robustness_summary}
\begin{tabular*}{\textwidth}{@{\extracolsep{\fill}}llccc@{}}
\toprule
PDE & Method & Attack \(\Delta\)loss &
\(\|J_{\mathcal E}^{\top}e_\theta\|_2\) & \(\sigma_1(J_{\mathcal E})\) \\
\midrule
Burgers & Baseline & 0.00916 \(\pm\) 0.00743 & 0.4630 \(\pm\) 0.4381 & 2.367 \(\pm\) 1.378 \\
 & Loss 1 & 0.00584 \(\pm\) 0.00367 & 0.3159 \(\pm\) 0.3246 & 1.702 \(\pm\) 1.194 \\
 & Loss 2 & 0.00562 \(\pm\) 0.00346 & 0.3433 \(\pm\) 0.3473 & 1.855 \(\pm\) 1.227 \\
 & Loss 3 & \textbf{0.00306 \(\pm\) 0.00260}\(^{*}\) & \textbf{0.1119 \(\pm\) 0.1285}\(^{*}\) & \textbf{1.272 \(\pm\) 0.8203}\(^{\mathrm{ns}}\) \\
 & Random clean & 0.0428 \(\pm\) 0.0126 & 4.785 \(\pm\) 2.888 & 4.131 \(\pm\) 0.9373 \\
 & Random solver & 0.00800 \(\pm\) 0.00557 & 0.3712 \(\pm\) 0.4211 & 1.553 \(\pm\) 1.221 \\
\midrule
Darcy Flow & Baseline & 9.681e-6 \(\pm\) 1.364e-6 & 9.293e-5 \(\pm\) 4.646e-5 & \textbf{1.272e-3 \(\pm\) 1.350e-4}\(^{*}\) \\
 & Loss 1 & 3.927e-6 \(\pm\) 9.965e-7 & 9.922e-5 \(\pm\) 5.783e-5 & 1.821e-3 \(\pm\) 2.651e-4 \\
 & Loss 2 & 3.565e-6 \(\pm\) 9.173e-7 & 9.225e-5 \(\pm\) 5.206e-5 & 1.751e-3 \(\pm\) 2.545e-4 \\
 & Loss 3 & \textbf{1.699e-6 \(\pm\) 8.868e-7}\(^{*}\) & \textbf{6.157e-5 \(\pm\) 3.486e-5}\(^{*}\) & 2.354e-3 \(\pm\) 3.994e-4 \\
 & Physics loss & 3.974e-6 \(\pm\) 9.114e-7 & 1.034e-4 \(\pm\) 5.895e-5 & 1.856e-3 \(\pm\) 2.779e-4 \\
 & Random clean & 3.691e-6 \(\pm\) 1.047e-6 & 9.112e-5 \(\pm\) 5.351e-5 & 1.725e-3 \(\pm\) 2.604e-4 \\
 & Random solver & 4.824e-6 \(\pm\) 7.415e-7 & 1.236e-4 \(\pm\) 6.885e-5 & 1.898e-3 \(\pm\) 2.869e-4 \\
\bottomrule
\end{tabular*}
\end{table*}

The main result is simple: to make the model output differ strongly from the
solver output, the attack should optimize the full $\mathcal{L}_3$ objective
directly. \tabref{tab:attack_objective_true_l3_summary} shows this pattern
across Burgers, Darcy Flow, and recurrent Navier--Stokes. Attacks based on
$\mathcal{L}_1$, $\mathcal{L}_2$, physics residuals, fixed or dictionary
targets, or $\mathcal{L}_3$ with the solver gradient removed usually increase
the true model--solver loss more slowly and reach a smaller final loss. Thus the
solver is useful both as the forward target $\mathcal{G}(x+\delta)$ and as a
backward path through $D\mathcal{G}$.
Expanded dictionary, stop-gradient, and recurrent gradient-path comparisons
appear in \tabref{tab:revision-burgers-expanded-attacks},
\tabref{tab:revision-darcy-expanded-attacks}, and
\tabref{tab:revision-ns-expanded-attacks}. These retain their own normalization
and attack settings. The largest NS budget includes a partial-gradient variant
that exceeds Full L3 in mean final loss, so stronger full integration is an
empirical trend rather than a universal optimizer guarantee.

A second pattern is spectral. In Burgers and Navier--Stokes, perturbations
generated by full $\mathcal{L}_3$ are visibly higher-frequency than those from
the other objectives; see
\figrefs{fig:app-g-burgers-steepest-add-final}{fig:app-g-ns-steepest-add-final}.
Local SVD diagnostics in \appref{app:local-svd-diagnostics} give a mechanism:
the neural operator and solver can share dominant local response directions,
whereas the residual Jacobian exposes different and often more oscillatory
input directions.

\subsection{Projected Attack Optimization Methods}
\label{subsec:exp-projected-optimization}

The Burgers optimizer ablation gives a somewhat surprising result.  A
replace-style update is closest to a frozen-linear boundary iteration: it
discards the old perturbation and places the next iterate directly on the
constraint boundary in the current first-order direction.  This is naturally
motivated for local linear or quadratic models, but the Loss 3 attack is a
finite-radius nonlinear model--solver objective, not a fixed quadratic problem.
Empirically, replacement nevertheless reaches a strong Burgers attack very
quickly.  The boundary-ratio and delta-angle diagnostics in
\figrefs{fig:app-g-burgers-qmean-boundary-markers}{fig:app-g-burgers-delta-angle-grid}
show what happens: add-style methods spend many early steps moving outward,
whereas replacement methods reach the perturbation boundary in only a few steps
and then continue changing direction on that boundary.  The reason appears to
be that Burgers has a broad high-loss region in these experiments.  Many nearby
boundary directions already produce large model--solver discrepancy, so raw add
or steepest add can eventually approach, and in a few settings slightly exceed,
the replacement loss.  Thus for Burgers the main difference is speed of entry
into a high-loss region, not access to a unique optimum.  We observe a similar
practical pattern in Darcy Flow: the replace-style binary flip update also
reaches high true Loss 3 quickly and performs well.

Navier--Stokes shows the opposite tendency.  In the recurrent NS2D optimizer
ablation, steepest add reaches the largest mean Loss 3 in
\figref{fig:app-g-cross-benchmark-optimizer-ablation}, while the replacement
curves remain much lower.  This supports a different interpretation: the
high-loss region for the NS attack appears much narrower and sharper than in
Burgers.  Moving even moderately away from a good perturbation direction can
lead to a much smaller attack loss, so an aggressive boundary replacement is no
longer reliable.  The optimizer conclusion is therefore geometry-dependent:
replacement is fast and often strong when the high-loss region is broad, as in
Burgers and Darcy Flow, whereas recurrent Navier--Stokes favors the more gradual
additive steepest update.  \appref{app:loss-landscape-mechanism-diagnostics}
defines the loss-landscape diagnostics behind this interpretation and collects
their cross-system evidence.

\subsection{Adversarial Training}
\label{subsec:exp-adversarial-training}

We perform adversarial training by generating each training batch through an
online attack against the current model and then using that adversarial batch
directly for model updates. Because the losses and random perturbation variants
have different costs, the added controlled studies report both matched epochs
and matched measured training time. The historical robustness entries in
\tabref{tab:adversarial_training_robustness_summary} use their original
checkpoints; they are not relabeled as strictly time-matched measurements.
The corrected endpoint selections and ID/OOD outcomes are reported in
\appref{app:generalization-random-perturbation}. We measure generalization to shifted inputs
and robustness under subsequent attacks. Burgers and Darcy Flow support broad
comparisons across losses and random perturbation variants because solver calls
are feasible.

\paragraph{Generalization.}
The Burgers figure sequence in \appref{app:generalization-random-perturbation} presents measured-time views first (\figref{fig:revision-burgers-common-time-mean-original}--\ref{fig:revision-burgers-common-time-rmse-26-50}), followed by epoch-axis views. Hours and epoch count are separate horizontal coordinates; they are not interchangeable budget labels.
On Burgers, Loss 3 gives the lowest 50-dataset generalization relative \(L_2\).
Random solver is often the
strongest non-Loss3 training variant early in wall-clock time; however, its
perturbations come from a fixed hand-designed distribution, so the sampled
directions do not adapt to the current model's weak points and the method
overfits later. Loss 3 instead regenerates adversarial samples from the current
model--solver error. Darcy Flow shows the same pattern
(\tabref{tab:appendix-darcy-52-relative-l2}), with Loss 3 again giving
the lowest shifted-set relative \(L_2\). On the original Darcy train/test
references, Loss 3 is not the clean-error winner because the run is trained
adv-only on solver-relabelled finite-radius binary-flip batches; the recorded
alignment and replay probes showed that such Loss 3 samples can trade away
clean in-distribution fit while improving the shifted datasets.

For 2D Navier--Stokes, memory and wall-clock constraints limited training to six
successive Full-L3 rounds. In the retained 402-dataset GRF subset, 364 have lower
dataset-mean final-frame RMSE after round six. The full 482-dataset GRF pool has
366 improvements; the broader 1,442-dataset pool has 1,015. The subset rules,
unfavorable cases, and RMSE definition are given in
\tabref{tab:revision-ns-quality-scope} and
\figref{fig:app-g-ns-training-delta-rmse}. These are comparisons with
the original baseline, not a completed multi-method NS training study.

\paragraph{Robustness.}
\tabref{tab:adversarial_training_robustness_summary} reports the finite-budget
attack loss increase, the Jacobian-error norm
\(\|J_{\mathcal E}^{\top}e_\theta\|_2\), and the local spectral norm
\(\sigma_1(J_{\mathcal E})\), where smaller values indicate stronger
robustness.  Loss 3 gives the smallest attack loss increase for both Burgers
and Darcy Flow and the smallest Jacobian-error norm, so it is the most robust
method in this comparison.  The appendix diagnostics in
\figrefs{fig:app-g-burgers-training-loss-gain}{fig:app-g-burgers-training-frequency}
show the same trend dynamically: Loss 3 training steadily shrinks the
self-attack loss increase and shifts perturbations toward higher frequencies,
unlike Loss 1 and Loss 2.  They also show that
\(\|J_{\mathcal E}^{\top}e_\theta\|_2\) tracks the self-attack loss increase
more closely than \(\sigma_1(J_{\mathcal E})\); see
\appref{app:robustness-metric-diagnostics}.
Independent Burgers cross-attack records in \tabref{tab:revision-burgers-id-cross-attacks}--\ref{tab:revision-burgers-cross-attack-summary} distinguish ID and OOD: Random solver achieves the lowest worst-case ID MSE, whereas Full L3 achieves the lowest worst-case OOD MSE. This limits any claim of a universal cross-method robustness ranking.
The NS matched-input study is reported explicitly in
\figref{fig:revision-ns-jacobian-scatter}, with Pearson/Spearman
coefficients in \tabref{tab:revision-ns-scatter-correlations} and the
20-input, four-budget data in \tabref{tab:revision-ns-scatter-input-data}.
For Darcy,
\tabref{tab:revision-darcy-cross-attacks} evaluates each Darcy training method
under eight attack objectives using the correct perturbed solver output.

\section{Conclusion}

This paper studies the generalizability and robustness of neural operators from
a solver-integrated perspective.  We first define these quantities by the
model--solver discrepancy under distribution shift and adversarial
perturbation, and explain why model-only stability indicators such as a
Jacobian spectral norm are not sufficient for PDE operator learning.  We then propose
a solver-integrated adversarial attack whose forward loss and backward gradient
both use the PDE solver, so the attack directly maximizes the relevant
model--solver error.  From this attack, we derive an adversarial-training
procedure that generates solver-relabelled adversarial batches online.
Experiments on Burgers and Darcy Flow support improvements in shifted-input
accuracy and robustness under the tested solver-consistent attacks, with an
ID/OOD accuracy tradeoff. The six-round Navier--Stokes study provides limited
baseline-versus-L3 evidence. These experiments keep the PDE and boundary
conditions fixed and do not test transfer to a new governing equation.
More broadly, any regression model with a
ground-truth oracle that tracks how true outputs change under perturbation can
use the same formulation; PDE operator learning supplies such an oracle through
a solver.

The main limitation is computational cost.  Solver-integrated attacks and
training require expensive solver forward passes and gradients through the
solver.  For time-dependent PDEs, the solver backward pass must keep the
intermediate states and stage variables from many time steps in GPU memory so
that gradients can be propagated back through the entire rollout.  This creates
a large memory footprint and a long wall-clock cost, especially for
Navier--Stokes.  Future work can study how to reduce the time and memory cost
of solver propagation.  One possible direction is a more hierarchical approach
in which part of the explicit solver rollout is replaced by an implicit
neural-network rollout, forming a hierarchy from the PDE solver, to an implicit
neural rollout, and finally to the neural operator. Coarse-grid solvers,
reduced-order models, approximate adjoints, and learned gradient surrogates
could select candidate inputs before periodic high-fidelity verification.
Their gradient bias and cost--quality tradeoffs require separate validation;
the present experiments do not establish performance on 3D PDEs.

\FloatBarrier
\clearpage
\onecolumn
\begingroup
\section*{References}
\vspace{-0.4em}
\makeatletter
\renewcommand{\bibsection}{}%
\renewcommand{\bibfont}{\fontsize{9.4pt}{10.45pt}\selectfont}%
\let\paperorigthebibliography\thebibliography
\renewcommand{\thebibliography}[1]{%
  \paperorigthebibliography{#1}%
  \setlength{\bibsep}{0pt}%
  \setlength{\itemsep}{0pt}%
  \setlength{\parsep}{0pt}%
  \setlength{\parskip}{0pt}%
  \setlength{\topsep}{0pt}%
  \setlength{\partopsep}{0pt}%
}
\makeatother
\begin{multicols}{2}
\FloatBarrier

\end{multicols}
\endgroup

\appendix
\raggedbottom
\appendixsection{Benchmark PDEs}{app:benchmark-pdes}
\vspace{0.4em}
\begin{multicols}{2}

We evaluate on three canonical PDE operator-learning benchmarks used in the
original Fourier Neural Operator paper \citep{li2021fourier}: 1D Burgers,
2D Darcy Flow, and 2D Navier--Stokes. They represent two different operator
types. Burgers and Navier--Stokes are time-dependent evolution problems, where
the input is an initial field and the target is a later-time solution field.
Darcy Flow is a steady elliptic problem, where the input is a coefficient field
and the target is the steady solution. For time-dependent PDEs, the solver can
return a full stored rollout, but the supervised neural-operator loss used here
is evaluated on the final frame.

\paragraph{1D Burgers equation.}
On a periodic 1D domain $\mathbb{T}_L=[0,L)$, we solve
\begin{align}
  u_t + u u_x
  &=
  \nu_{\mathrm{B}} u_{xx},
  \nonumber\\
  u(x,0)&=u_0(x),
  \nonumber\\
  u(x+L,t)&=u(x,t).
\end{align}
At the solver level, Burgers is a rollout map from the initial velocity field to
the stored trajectory, while the final-time FNO task selects the final frame as
the supervised target:
\begin{equation}
  \mathcal{G}_{\mathrm{B}}^{\mathrm{roll}}:\ u_0
  \mapsto
  \{u(\cdot,t_j)\}_{j=0}^{T_{\mathrm{B}}},
  \qquad
  \mathcal{G}_{\mathrm{B}}^{\mathrm{final}}:\ u_0
  \mapsto u(\cdot,T_{\mathrm{B}}).
\end{equation}
The learned Burgers FNO has the corresponding terminal-frame map
$\mathcal F_{\theta,\mathrm{B}}:\ u_0\mapsto
\widehat u(\cdot,T_{\mathrm{B}})$.
The initial condition is a periodic Gaussian random field. In the Burgers data
generator used by our experiments, this is implemented by drawing white noise
$\xi$ on the grid, applying a periodic Gaussian smoothing kernel
$K^{\mathrm{per}}_\ell$, and normalizing each sample:
\begin{equation}
  u_0
  =
  \operatorname{Norm}_{[0,1]}
  \!\left(K^{\mathrm{per}}_\ell * \xi\right),
  \qquad
  \xi_i\overset{\mathrm{i.i.d.}}{\sim}\mathcal{N}(0,1).
\end{equation}
The Burgers experiments use $N=1024$, $T_{\mathrm{B}}=1$,
$\nu_{\mathrm{B}}\in\{10^{-3},10^{-2}\}$, periodic boundary conditions, and
correlation length $\ell=0.03$.

\paragraph{2D Darcy Flow.}
Darcy Flow is a steady coefficient-to-solution problem on
$D=(0,1)^2$. Given a permeability coefficient $a(x)$, the solver computes
$u(x)$ from
\begin{align}
  -\nabla\cdot\big(a(x)\nabla u(x)\big) &= 1,
  \qquad x\in D,
  \nonumber\\
  u(x)&=0,
  \qquad x\in\partial D .
\end{align}
The operator is therefore
\begin{equation}
  \mathcal{G}_{\mathrm{D}}:\ a \mapsto u .
\end{equation}
There is no temporal rollout and no initial condition. The Fourier neural
operator for Darcy Flow uses the same coefficient-to-solution task as the solver:
$\mathcal F_{\theta,\mathrm{D}}:\ a\mapsto \widehat u$.
The input coefficient is obtained by thresholding a latent Gaussian random field
sampled from the
Neumann-Laplacian covariance used in the FNO Darcy Flow benchmark:
\begin{equation}
  z \sim
  \mathcal{N}
  \!\left(
    0,\,
    (-\Delta_{\mathrm{N}}+\tau^2 I)^{-\alpha}
  \right),
  \qquad
  \alpha=2,\ \tau=3.
\end{equation}
The binary permeability is
\begin{equation}
  a(x)=
  \begin{cases}
    12, & z(x)\ge 0,\\
    3, & z(x)<0.
  \end{cases}
\end{equation}
Thus, in Darcy Flow experiments, an adversarial perturbation changes the coefficient
field, while the PDE itself remains a steady zero-Dirichlet elliptic solve.  The
perturbed coefficient is still constrained to the binary input class, so the
attack changes which grid locations take values \(3\) or \(12\) rather than
introducing continuous permeability values.

\paragraph{2D Navier--Stokes equation.}
The incompressible Navier--Stokes equation is usually written in
velocity--pressure form in three space dimensions,
\begin{align}
  \partial_t u+(u\cdot\nabla)u
  &=
  -\nabla p+\nu_{\mathrm{NS}}\Delta u+F,
  \nonumber\\
  \nabla\cdot u&=0 .
\end{align}
For the 2D benchmark, the flow is planar and periodic:
$u=(v_1,v_2,0)$ and all fields are independent of the third coordinate. Define
the vorticity vector $\Omega=\nabla\times u$. In this planar setting,
\begin{equation}
  \Omega=(0,0,\omega),
  \qquad
  \omega=\partial_x v_2-\partial_y v_1 .
\end{equation}
Taking the curl of the momentum equation removes the pressure term and gives
the vorticity equation
\begin{equation}
  \partial_t\Omega+(u\cdot\nabla)\Omega
  =
  (\Omega\cdot\nabla)u+\nu_{\mathrm{NS}}\Delta\Omega+\nabla\times F .
\end{equation}
For planar 2D flow, $(\Omega\cdot\nabla)u=0$. Taking the third component gives
\begin{equation}
  \partial_t\omega+v\cdot\nabla\omega
  =
  \nu_{\mathrm{NS}}\Delta\omega+g,
  \qquad
  g=(\nabla\times F)_3 .
\end{equation}
We use the vorticity-streamfunction formulation on the periodic torus
$\mathbb{T}^2=[0,1)^2$:
\begin{align}
  \omega_t + v\cdot\nabla\omega
  &=
  \nu_{\mathrm{NS}}\Delta\omega + g,
  \nonumber\\
  -\Delta\psi &= \omega,
  \nonumber\\
  v&=(\partial_y\psi,-\partial_x\psi),
  \nonumber\\
  \omega(x,0)&=\omega_0(x),
  \nonumber\\
  \omega(x+k,t)&=\omega(x,t),
  \qquad k\in\mathbb{Z}^2 .
\end{align}
The default external forcing is
\begin{equation}
  g(x,y)=0.1
  \left[
    \sin(2\pi(x+y))+\cos(2\pi(x+y))
  \right].
\end{equation}
The input is an initial vorticity field $\omega_0$ sampled from the periodic
Gaussian random-field distribution of the FNO Navier--Stokes benchmark,
commonly written as
\begin{equation}
  \omega_0
  \sim
  \mathcal{N}
  \!\left(
    0,\,
    7^{3/2}(-\Delta+49I)^{-5/2}
  \right).
\end{equation}
The corresponding solver rollout maps the initial vorticity to the stored
later-time trajectory,
\begin{equation}
  \mathcal{G}_{\mathrm{NS}}^{\mathrm{roll}}:\ \omega_0
  \mapsto
  \{\omega(\cdot,t_j)\}_{j=0}^{T_{\mathrm{NS}}}.
\end{equation}
In our recurrent neural-operator setting, the model receives
$\omega_0,\ldots,\omega_9$ and predicts
$\omega_{10},\ldots,\omega_{19}$:
\begin{equation}
  \mathcal F_{\theta,\mathrm{NS}}^{\mathrm{rec}}:
  (\omega_0,\ldots,\omega_9)
  \mapsto
  (\widehat\omega_{10},\ldots,\widehat\omega_{19}).
\end{equation}
For the losses reported here, we use only the final predicted frame
$\widehat\omega_{19}$ as the supervised target; the loss is not accumulated
over the whole predicted window.
In our recurrent Navier--Stokes experiments, the generated rollouts use
$\nu_{\mathrm{NS}}=10^{-5}$, $T_{\mathrm{NS}}=20$, and periodic Fourier
time-stepping, with stored integer-time vorticity frames.

\end{multicols}
\vspace{0.5em}
\appendixsection{Numerical Solver Details}{app:numerical-solver-details}
\vspace{0.4em}
\begin{multicols}{2}

\subsection*{1D Burgers}

For the Burgers benchmark, the solver advances the viscous periodic Burgers
equation
\begin{align}
  u_t + c\,u u_x &= \nu u_{xx},
  \nonumber\\
  x&\in[0,L),\qquad u(x+L,t)=u(x,t),
\end{align}
from an initial field $u_0$. With $N$ equispaced grid points, the code uses a
real Fourier representation
\begin{equation}
  u(x,t)=\sum_k \hat u_k(t)\exp(i k x),
  \qquad k = \frac{2\pi m}{L}.
\end{equation}
In Fourier variables the semi-discrete equation has the form
\begin{equation}
  \frac{d\hat u_k}{dt}
  =
  -c\,\widehat{u u_x}_k
  -\nu k^2 \hat u_k .
\end{equation}
Spatial derivatives are spectral multiplications: $D_k=i k$ and
$\Delta_k=-k^2$. The nonlinear product is evaluated pseudospectrally: the
dealiased Fourier coefficients are transformed to physical space, $u_x$ is
formed by inverse transforming $D_k\hat u_k$, the pointwise product $u u_x$ is
computed, and the result is transformed back. Equivalently, the nonlinear term
can be written as $u u_x=\partial_x(u^2/2)$.
For the quadratic nonlinear term, modes outside the default $2/3$ de-aliasing
fraction are masked.

Time integration uses Cox--Matthews ETDRK4. The typical differentiable Burgers
rollout uses $N_x=1024$, spatial period $L=2$, $T=1$,
$\Delta t=10^{-3}$, $N_t=1000$, $\Delta x=L/N_x$, $\nu=10^{-3}$,
and a $2/3$ de-aliasing mask. This solver supplies the Burgers reference
rollout and solver-integrated attack evaluations. The solver loop is:

\begingroup
\footnotesize
\begin{algorithm}[H]
\caption{1D Burgers Fourier pseudospectral solver}
\begin{algorithmic}[1]
\STATE \textbf{Input:} $u_0$, $\nu$, $c$, $N_x$, spatial period $L$,
$\Delta t$, $N_t$, de-aliasing mask $M$.
\STATE Set $k=2\pi m/L$, $L_k=-\nu k^2$, $h=\Delta t$,
$E=\exp(hL_k)$, and $E_2=\exp(hL_k/2)$.
\STATE Precompute ETDRK4 coefficients $Q,f_1,f_2,f_3$ from $hL_k$.
\STATE $\hat u\leftarrow \mathrm{FFT}(u_0)$.
\FOR{$n=0,\ldots,N_t-1$}
  \STATE For any stage state $\hat z$, evaluate
  $N(\hat z)=-c\,M\odot\mathrm{FFT}(z\,z_x)$, where
  $z=\mathrm{IFFT}(\hat z)$ and $z_x=\mathrm{IFFT}(ik\hat z)$.
  \STATE $a\leftarrow E_2\hat u+QN(\hat u)$.
  \STATE $b\leftarrow E_2\hat u+QN(a)$.
  \STATE $d\leftarrow E_2a+Q\{2N(b)-N(\hat u)\}$.
  \STATE $\hat u\leftarrow E\hat u+f_1N(\hat u)
  +2f_2\{N(a)+N(b)\}+f_3N(d)$.
\ENDFOR
\STATE \textbf{Output:} $u(T)=\mathrm{IFFT}(\hat u)$.
\end{algorithmic}
\end{algorithm}
\endgroup

\subsection*{2D Darcy Flow}

For the Darcy Flow benchmark, the operator maps a coefficient field $a(x)$ to the
solution $u(x)$ of the steady elliptic problem
\begin{equation}
  -\nabla\cdot(a(x)\nabla u(x)) = f(x),
  \qquad x\in(0,1)^2,\qquad u|_{\partial\Omega}=0.
\end{equation}
We use the standard FNO Darcy setting with $f(x)=1$.
The PDE solve uses a second-order finite-difference discretization on the
interior grid. Let $h=1/(n-1)$ and let $u_{ij}$ denote an interior unknown.
The resulting finite-difference operator uses arithmetic averages of $a$ on
grid faces:
\begin{align}
  (A_h u)_{ij}
  =
  \frac{1}{h^2}
  \big[
    &\frac{a_{ij}+a_{i+1,j}}{2}(u_{ij}-u_{i+1,j})
    \nonumber\\
    &+\frac{a_{ij}+a_{i-1,j}}{2}(u_{ij}-u_{i-1,j})
    \nonumber\\
    &+\frac{a_{ij}+a_{i,j+1}}{2}(u_{ij}-u_{i,j+1})
    \nonumber\\
    &+\frac{a_{ij}+a_{i,j-1}}{2}(u_{ij}-u_{i,j-1})
  \big].
\end{align}
The interior linear system $A_h u=\mathbf{1}$ is solved by JAX conjugate
gradient and the zero Dirichlet boundary is padded back afterward. Gradients
through the Darcy Flow solver are obtained by implicit differentiation through
the linear solve. This solve supplies the Darcy reference solution and
solver-integrated attack evaluations.
There is no temporal rollout for this benchmark. The default solver grid in the
current differentiable experiments is $n=85$, with $h=1/(n-1)$; the data
generation scripts also use high-resolution solve/downsample variants. The
conjugate-gradient stopping tolerance is $10^{-5}$.

\begingroup
\footnotesize
\begin{algorithm}[H]
\caption{2D Darcy Flow finite-difference CG solver}
\begin{algorithmic}[1]
\STATE \textbf{Input:} coefficient field $a$, grid size $n$, forcing $f(x)=1$,
and stopping tolerance $\tau_{\mathrm{cg}}$.
\STATE Set $h=1/(n-1)$ and initialize the interior unknown $u\leftarrow0$.
\STATE Define $A_h z$ by the finite-difference stencil above, using arithmetic
averages of $a$ on grid faces and zero Dirichlet boundary values.
\STATE $r\leftarrow \mathbf 1-A_hu$, $p\leftarrow r$.
\FOR{CG iterations $m=0,1,\ldots$ until $\|r\|<\tau_{\mathrm{cg}}$}
  \STATE $q\leftarrow A_hp$.
  \STATE $\alpha\leftarrow (r^\top r)/(p^\top q)$.
  \STATE $u\leftarrow u+\alpha p$; \quad $r_{\mathrm{new}}\leftarrow r-\alpha q$.
  \STATE $\beta\leftarrow (r_{\mathrm{new}}^\top r_{\mathrm{new}})/(r^\top r)$.
  \STATE $p\leftarrow r_{\mathrm{new}}+\beta p$; \quad $r\leftarrow r_{\mathrm{new}}$.
\ENDFOR
\STATE Pad zero boundary values around the interior solution.
\STATE \textbf{Output:} Darcy Flow solution field $u$.
\end{algorithmic}
\end{algorithm}
\endgroup

Algorithm 2 is written at a lower level than Algorithms 1 and 3.
It expands the CG iteration used to solve the steady Darcy linear system; this
is comparable to spelling out one low-level iterative solve within a time step,
while Algorithms 1 and 3 describe the overall time-marching procedures for
Burgers and Navier--Stokes. This also explains why the Darcy Flow solver is
much faster: it is a steady solve, whereas Burgers and Navier--Stokes repeat
many small time-discretization steps.

\subsection*{2D Navier--Stokes}
\paragraph{Availability of the physics-residual baseline.}
For steady Darcy Flow, with $\hat u=\mathcal F_\theta(a_{\rm adv})$, the
residual is
\begin{equation}
 R_D(s)=-\nabla_s\!\cdot\!\bigl(a_{\rm adv}(s)\nabla_s\hat u(s)\bigr)-1.
\end{equation}
The Physics objective combines the spatial residual and the homogeneous
Dirichlet boundary penalty. One predicted field and the matching coefficient
field suffice to evaluate it. Our Burgers FNO instead maps the initial field
to a single terminal frame. The usual residual
$R_B=\partial_t\hat u+\hat u\,\partial_s\hat u-\nu\partial_{ss}\hat u$
requires a local time derivative that this architecture does not expose.
The secant $(\hat u(T)-u_0)/T$ averages the entire time interval and does not
determine $\partial_t\hat u(T)$. A time-conditioned or trajectory-output model
could support that residual, but would change the present comparison.

The solver starts from the vorticity-streamfunction system:
\begin{align}
  \omega_t + {\bf v}\cdot\nabla\omega
  &=
  \nu\Delta\omega + g,
  \nonumber\\
  -\Delta\psi &= \omega,
  \nonumber\\
  {\bf v}=(\partial_y\psi,-\partial_x\psi),
\end{align}
On an $N\times N$ periodic grid, write the scalar vorticity in a complex
Fourier basis
\begin{equation}
  \omega(x,t)=\sum_{\kappa}\hat\omega_{\kappa}(t)
  \exp(i\kappa\cdot x),
  \qquad
  \kappa=\frac{2\pi}{L}(m,n).
\end{equation}
For each nonzero mode, the streamfunction and velocity are recovered by
\begin{equation}
  \begin{aligned}
  \hat\psi_{\kappa}
  &=
  \frac{\hat\omega_{\kappa}}{|\kappa|^2},
  \qquad \kappa\ne0,
  \qquad
  \hat\psi_{0}=0,
  \\
  \hat v_{1,\kappa}
  &=
  i\kappa_y\hat\psi_{\kappa},
  \qquad
  \hat v_{2,\kappa}
  =
  -i\kappa_x\hat\psi_{\kappa}.
  \end{aligned}
\end{equation}
Thus the Fourier-space evolution has the semi-discrete form
\begin{equation}
  \frac{d\hat\omega_{\kappa}}{dt}
  =
  -\nu|\kappa|^2\hat\omega_{\kappa}
  +\hat g_{\kappa}
  -
  \widehat{
    v_1\,\partial_x\omega+v_2\,\partial_y\omega
  }_{\kappa}.
\end{equation}
The nonlinear advection term is evaluated pseudospectrally: transform
$v_1,v_2,\partial_x\omega,\partial_y\omega$ back to physical space, form the
pointwise product $v_1\partial_x\omega+v_2\partial_y\omega$, and transform that
field back to Fourier space. After time stepping, the scalar vorticity is
recovered by inverse FFT, the vorticity vector is
$\Omega=(0,0,\omega)$, and the velocity field is recovered from the inverse FFT
of the Fourier coefficients above.
The solver then uses a real 2D FFT and Cox--Matthews ETDRK4.
The typical differentiable rollout uses
$N_x=N_y=256$, spatial period $L=1$, $\Delta t=0.005$, $200$ internal steps
per unit time, $\nu=10^{-5}$, $t_{\mathrm{final}}=20$, $N_t=4000$, and the
default $2/3$ de-aliasing mask. This solver supplies the Navier--Stokes
reference trajectory and solver-integrated attack evaluations. The solver loop is:

\begingroup
\footnotesize
\begin{algorithm}[H]
\caption{2D Navier--Stokes vorticity pseudospectral solver}
\begin{algorithmic}[1]
\STATE \textbf{Input:} initial vorticity $\omega_0$, viscosity $\nu$,
vorticity forcing $g$, grid size $N$, spatial period $L$, $\Delta t$, $N_t$,
save interval, and de-aliasing mask $M$.
\STATE Set angular wavenumbers $k_x,k_y$, $K^2=k_x^2+k_y^2$, and
$L_{k_x,k_y}=-\nu K^2$.
\STATE Precompute ETDRK4 coefficients from $\Delta t\,L_{k_x,k_y}$.
\STATE $\hat\omega\leftarrow\mathrm{FFT}(\omega_0)$; save the initial frame.
\FOR{$n=0,\ldots,N_t-1$}
  \STATE For any stage state $\hat z$, set
  $\hat\psi=\hat z/K^2$ for $K^2>0$ and $\hat\psi_{0,0}=0$.
  \STATE Form velocities and derivatives spectrally:
  $\hat v_1=ik_y\hat\psi$, $\hat v_2=-ik_x\hat\psi$,
  $\widehat{z_x}=ik_x\hat z$, and $\widehat{z_y}=ik_y\hat z$.
  \STATE Transform $v_1,v_2,z_x,z_y$ to physical space and compute
  $\widehat{\mathrm{adv}}=\mathrm{FFT}(v_1z_x+v_2z_y)$.
  \STATE Define $N(\hat z)=M\odot(\widehat g-\widehat{\mathrm{adv}})$.
  \STATE Update $\hat\omega$ by one Cox--Matthews ETDRK4 step using
  $L_{k_x,k_y}$ and $N(\cdot)$.
  \STATE Save $\mathrm{IFFT}(\hat\omega)$ at each requested integer-time frame.
\ENDFOR
\STATE \textbf{Output:} stored vorticity frames.
\end{algorithmic}
\end{algorithm}
\endgroup

\subsection*{Batch Differentiation}

Let $\mathcal X$ and $\mathcal Y$ be Hilbert spaces for input and output
functions. For a batch $a=(a_1,\ldots,a_B)\in\mathcal X^B$, the batched neural
operator and solver act componentwise,
\begin{align}
  \mathcal{F}_\theta^{B}(a)
  &=
  \bigl(\mathcal{F}_\theta(a_1),\ldots,\mathcal{F}_\theta(a_B)\bigr),
  \\
  \mathcal{G}^{B}(a)
  &=
  \bigl(\mathcal{G}(a_1),\ldots,\mathcal{G}(a_B)\bigr).
\end{align}
Given a scalar sample loss
$\ell:\mathcal Y\times\mathcal Y\rightarrow\mathbb R$, the batch attack loss is
the sum, or mean, of paired model--solver losses:
\begin{align}
  \mathcal L_{\mathrm{sum}}(a)
  &=
  \sum_{i=1}^{B}
  \ell\!\left(\mathcal{F}_\theta(a_i),\mathcal{G}(a_i)\right),
  \\
  \mathcal L_{\mathrm{avg}}(a)
  &=
  \frac{1}{B}\mathcal L_{\mathrm{sum}}(a).
\end{align}
In an attack step, $a_i=x_i+\delta_i$; sample $i$ is paired only with its own
model and solver outputs, with no cross-sample loss terms.

Assume $\ell$, $\mathcal{F}_\theta$, and $\mathcal{G}$ are Frechet
differentiable at the relevant points. For a batch attack, write
$a_i=x_i+\delta_i$ and
\(\delta=(\delta_1,\ldots,\delta_B)\in\mathcal X^B\). This is the same as
concatenating the \(B\) perturbations into one large optimization variable,
while still applying the model, solver, and loss sample by sample. With the
clean batch \(x=(x_1,\ldots,x_B)\) fixed, the summed loss as a function of the
batch perturbation is
\begin{align}
  \mathcal L_{\mathrm{sum}}(\delta;x)
  &=
  \sum_{i=1}^{B}
  \ell\!\left(
    \mathcal{F}_\theta(x_i+\delta_i),
    \mathcal{G}(x_i+\delta_i)
  \right)
  \nonumber\\
  &=
  \sum_{i=1}^{B}\mathcal L_i(\delta_i;x_i).
\end{align}
Because the loss is additive over samples, term \(i\) depends only on its own
perturbation \(\delta_i\). Thus all terms with \(i\ne j\) are constant when
differentiating with respect to \(\delta_j\), and
\begin{align}
  \nabla_{\delta_j}\mathcal L_{\mathrm{sum}}(\delta;x)
  &=
  \nabla_{\delta_j}
  \ell\!\left(
    \mathcal{F}_\theta(x_j+\delta_j),
    \mathcal{G}(x_j+\delta_j)
  \right)
  \nonumber\\
  &=
  \nabla_{\delta_j}\mathcal L_j(\delta_j;x_j).
\end{align}
For the averaged loss,
\(\nabla_{\delta_j}\mathcal L_{\mathrm{avg}}(\delta;x)
=\frac{1}{B}\nabla_{\delta_j}\mathcal L_{\mathrm{sum}}(\delta;x)\).
Thus attacking the whole batch at once gives the same gradient block for
\(\delta_j\) as attacking sample \(j\) alone, up to the constant $1/B$ scaling
when the mean loss is used. It amounts to running \(B\) independent
single-sample attacks in one vectorized computation: the backward pass returns
one concatenated gradient whose \(j\)-th block is the single-sample gradient for
\(\delta_j\). The benefit is execution-level amortization of model, solver, FFT,
and linear-solve overhead; the cost is larger memory and longer wall time per
batch.

\subsection*{Differentiating Through the Solvers}

\paragraph{Backend choice and tensor transfer.}
For the Burgers timing run, the best backend path was JAX-solver/PyTorch-model:
$87.4$ s over $100$ attack steps with $669$ MiB peak global GPU memory. The
other backend paths were slower and used more memory: JAX-solver/JAX-model took
$95.5$ s and $747$ MiB, PyTorch-solver/PyTorch-model took $210.2$ s and $703$
MiB, and PyTorch-solver/JAX-model took $224.9$ s and $807$ MiB over the same
$100$ attack steps. DLPack transfers totaled $0.1261$ s over the same $100$
steps, far smaller than the $87.4$ s full attack loop, so the transfer time is
negligible. These measurements belong to the backend timing matrix in
\tabref{tab:backend_bridge_timing}.

\paragraph{Backward-mode memory and time.}
The PDE solvers do not introduce learned parameters, but differentiating
through a time-dependent solver introduces many latent solver states. In our
experiments, these time-dependent numerical solvers require substantially more
GPU memory and time during backpropagation than steady solves. For each
sample, write the state after time step \(n\) as \(z_n^i\). In Algorithm 3 for
Navier--Stokes, this state is the current Fourier vorticity
\(\hat\omega_n^i\), and one application of \(\Phi_{\Delta t}\) is one
Cox--Matthews ETDRK4 loop update using \(L_{k_x,k_y}\) and \(N(\cdot)\).
Abstractly, the fixed-step solver is
\begin{equation}
  z^{i}_{n+1}=\Phi_{\Delta t}(z^{i}_n),\qquad n=0,\ldots,N_t-1 .
\end{equation}
Reverse-mode differentiation initializes
$\bar z^i_{N_t}=\partial\mathcal L_{\mathrm{batch}}/\partial z^i_{N_t}$
and propagates adjoints backward by
\begin{equation}
  \bar z^i_n
  =
  \left(D\Phi_{\Delta t}(z^i_n)\right)^*\bar z^i_{n+1}.
\end{equation}
The memory gap comes from what each pass must retain. A forward-only solve is a
streaming computation: after forming $z_{n+1}$ from $z_n$, it can discard
$z_n$ and most temporary arrays. Reverse mode starts at the final loss and
moves backward, so applying $(D\Phi_{\Delta t}(z_n))^*$ requires the state $z_n$ and the
intermediate values created inside that step. A differentiable time-dependent
solver must therefore store or reconstruct the forward trajectory: time levels,
spatial grid values, ETDRK4 stage states, FFT and inverse-FFT variables,
spectral derivatives, nonlinear products, and masks. For a batch, even the
lowest-order storage scales like $B N_t$ times the number of spatial degrees of
freedom before those stage and Fourier work arrays are counted.

There is also a serial time-depth cost. Spatial operations, FFTs, pointwise
products, and the batch dimension can be parallelized, but the standard time
rollout cannot evaluate all frames simultaneously: $z_{n+1}$ depends on
$z_n$. The solver must finish one time step before the next one is available.
Thus thousands of stability-constrained small time steps in Burgers and
Navier--Stokes translate directly into wall time and into a long reverse-mode
tape.

\paragraph{Checkpointing and rematerialization.}
In reverse-mode AD, differentiating a time-stepping solver creates a solver
tape: the saved primal states and temporary arrays needed to propagate
gradients back through the solver steps. Without checkpointing, this tape spans
the full internal rollout. Checkpointing stores only selected boundary states,
such as every $m$ microsteps or every output frame; rematerialization recomputes
the missing states inside each segment during backward. Thus memory drops from
all internal time-step intermediates to checkpoint boundaries plus one local
segment, while wall time rises because parts of the forward solve are rerun.
\paragraph{Idealized checkpoint count and measured deviations.}
For a rollout of $N$ time steps, an idealized equal-weight checkpoint model
stores $K$ boundary states and rematerializes segments of length approximately
$N/K$. Its combined storage/recomputation proxy is
\begin{equation}
  C(K)=K+\frac{N}{K}.
  \label{eq:ideal-checkpoint-proxy}
\end{equation}
The derivative $C'(K)=1-N/K^2$ (equivalently, the AM--GM inequality) gives the
continuous optimum
\begin{equation}
  K^*=\sqrt{N},
  \label{eq:ideal-checkpoint-count}
\end{equation}
with the nearest feasible integer used in practice. This is a useful rule of
thumb, not a universal wall-time law. Real solver steps contain FFT stages and
nonuniform temporary arrays; GPU kernels, allocator behavior, compilation
caches, batch size, fixed model cost, and discrete chunk boundaries change the
constants and can shift the measured optimum. The Navier--Stokes sweep in
\tabref{tab:revision-ns-remat-capacity} denotes the segment length by $K$, so
its approximate checkpoint count is $N/K$. The proxy is symmetric in these two
quantities and also suggests a segment length near $\sqrt{N}$. The fastest
passing measured segment is $K=20$, whereas $\sqrt{4000}\simeq63.25$; this
chunk size is therefore selected empirically.
For the time-dependent rows in Tables~\ref{tab:remat_time_memory_tradeoff} and
\ref{tab:adv_training_epoch_timing}, the checkpoint/rematerialization labels
use this convention: ``none'' means no explicit rematerialization; ``micro'' or
``per-step'' rematerializes individual solver updates; ``chunk'' rematerializes
blocks of microsteps; ``second'' stores coarse rollout-frame boundaries such as
integer-time Navier--Stokes frames; and ``all'' stores the listed rollout states
without chunking them. Darcy Flow throughput rows are not rematerialization
sweeps, because the recorded steady elliptic solve has no rollout-tape
checkpointing setting.

\paragraph{Solver VJPs and adjoint access.}
For $e=\mathcal F_\theta(x)-\mathcal G(x)$, the full squared-error gradient is
$D\mathcal F_\theta[x]^*e-D\mathcal G[x]^*e$. Thus an implementation needs
the solver VJP against an output sensitivity, rather than a dense solver
Jacobian. An existing adjoint can supply the second term even when the solver
code cannot be differentiated by JAX or PyTorch. At one attack iterate, hold
$\hat u=\mathcal F_\theta(x)$ fixed inside the solver-only scalar objective
\begin{equation}
 Q(z;\hat u)=\tfrac12\|\mathcal G(z)-\hat u\|_2^2.
 \label{eq:revision-adjoint-objective}
\end{equation}
Then $\nabla_zQ|_{z=x}=-D\mathcal G[x]^*e$; adding the model VJP recovers
the full gradient. The adjoint must support the actual attacked input and this
objective. Design-parameter sensitivities alone do not establish access to an
arbitrary initial-field gradient. In recurrent NS, the model also consumes
solver-generated intermediate frames: $\mathcal F_\theta$ is the composite
map, and solver sensitivities through those input frames must be retained or
explicitly identified as detached. Stopping the final reference branch does
not automatically stop those other solver paths.

If a forward solve is available without a usable sensitivity, L3 Stop still
recomputes $\mathcal G(x+\delta)$ at every iterate and uses consistent labels
for training, while omitting the detached solver VJP in attack search. It is
different from keeping $\mathcal G(x)$ fixed. Coordinate finite differences
could approximate a missing gradient, but a central difference of a scalar
solver objective costs $2d$ solves for $d$ input coordinates per iterate; at
$256^2$ resolution, this is 131,072 solves. Random-direction approximations
can reduce calls at the expense of variance or bias. These approximations and
commercial-solver integration are not evaluated here. If no online solver or
validated reference surrogate is available, the fixed/dictionary-reference
setting applies instead.

The forward-only fallback is also tested in the matched-time Burgers study.
At the 7.427462-h budget, L3 Stop obtains OOD50 mean relative $L_2$ error
0.040487, compared with 0.043012 for L1 and 0.023001 for Full L3
(\tabref{tab:revision-burgers-alignment}). Thus consistent online solver
labels remain useful without the solver derivative, while the full-gradient
method has a further advantage in this comparison.

\paragraph{Higher dimensions and other surrogates.}
For $n$ points along each of $d$ spatial dimensions, $N$ time steps, and batch
$B$, the retained-state contribution scales as $O(BNn^d)$ before stage and FFT
workspaces. A $256^3$ field has 256 times as many spatial points as a $256^2$
field; 4,001 float32 scalar states alone occupy about 250 GiB versus 1 GiB per
sample. These are illustrative storage counts, not measured 3D GPU footprints.
Checkpointing reduces retained states but leaves the serial solver work.

The formulation also admits finite-dimensional, many-to-one regressors. For
example, a molecular-energy network $\hat E_\theta(R)$ approximates a reference
electronic-structure energy $E_{\rm ref}(R)$ for atomic coordinates $R$.
Uncertainty-driven molecular adversarial sampling has been demonstrated by
\citet{schwalbe2021differentiable}. A proposed solver-error variant instead uses
\begin{equation}
 \nabla_R\tfrac12(\hat E_\theta-E_{\rm ref})^2
 =(\hat E_\theta-E_{\rm ref})
   (\nabla_R\hat E_\theta-\nabla_RE_{\rm ref}),
\end{equation}
where the reference energy gradient is minus the reference force. Applying our
full-gradient objective in this domain is a proposed extension, not a completed
experiment; repeated reference calculations remain costly.

\paragraph{Complex-valued JAX gradients.}
One JAX-specific implementation detail is important for FNO models. The
spectral weights are naturally complex, but a real loss on $z=x+\mathrm{i}y$
should update the real coordinates $x$ and $y$ by ordinary real-coordinate
gradients. In Wirtinger calculus, this descent direction is the conjugate
Wirtinger derivative
$\partial L/\partial \bar z=\frac12(\partial L/\partial x+
\mathrm{i}\partial L/\partial y)$, up to the conventional factor of two. In our
JAX FNO experiments, directly optimizing native complex Fourier weights with
raw JAX complex gradients followed the opposite conjugation convention from the
PyTorch conjugate-Wirtinger descent direction for the imaginary part. We
therefore use real/imaginary split spectral weights in the JAX neural operator:
the real and imaginary parts are stored as two real parameter tensors, and the
complex multiplication is reconstructed only in the forward pass. Thus the
neural operator has no native complex trainable parameters, and autodiff returns
ordinary real gradients for both parts.

\subsection*{Wall-Clock Time and GPU Memory Consumption}

The solver-tape discussion above explains the large cost gap between the steady
Darcy Flow solver and the time-dependent Burgers/Navier--Stokes solvers. Darcy
Flow is a steady elliptic solve rather than a temporal rollout; operationally,
it is closer to one stationary solve than to thousands of small
stability-constrained time increments. It therefore does not create a long
time-stepping tape. For time-dependent solvers, the first storage lower bound is
simply the number of time steps times the number of spatial grid points. Burgers
gives about $1001\cdot1024$ stored physical-state values. Navier--Stokes is much
larger: $4001\cdot256\cdot256$ values, which is about $1$ GB per sample for one
float32 physical field alone. The actual differentiable tape is larger because
each time step also stores Fourier/physical transforms, velocity and gradient
fields, nonlinear terms, and ETDRK4 stage states.

The batch-one GPU phase probes in the codebase record the same pattern at the
solver level. A Darcy Flow CG-solver probe at $85\times85$ takes $0.0095$ s for
the solver forward phase, $0.0198$ s for the solver VJP/backward phase, and
$0.0293$ s total for solver forward plus backward. This Darcy number does not
include a neural-operator model forward pass. In the comparable Burgers
full-solver row below, the $1024$-point, $1000$-step solver path takes $0.1593$ s
forward, $0.2870$ s backward, and $0.4463$ s total. Thus Burgers is about
$17\times$ slower in solver forward time, $15\times$ slower in solver backward
time, and $15\times$ slower in total solver-update time.

Checkpointing and rematerialization trade memory for recomputation. Instead of
saving every solver state and stage value in the reverse-mode tape, they store
checkpoint boundaries and recompute local segments during backward. This can
admit larger feasible batches by omitting many saved internal states, but it is
not a multi-fold solver speedup.

Tables~6--8 give batch-one forward/backward phase timings, compare backend
choices, and separate model-only from solver-only component costs. The solver
dominates the time-dependent rows and tensor-transfer overhead is negligible.
Tables~9--18 then report attack throughput, checkpointing tradeoffs, completed
method-level costs, and full-epoch estimates.

The completed tables retain the measurement convention of each source probe:
framework-peak sums are conservative cross-framework upper bounds, PyTorch
peaks are allocator peaks, and process memory is the observed whole-process
allocation. These quantities should be compared within a table rather than
mixed across measurement conventions. ``Checkpoint'' denotes the recorded
chunk-rematerialized path; a dash means that the method has no solver backward
path to rematerialize. The production-budget table supersedes the earlier
short-probe estimate that treated nine batches as a complete Navier--Stokes
epoch; the quality study itself contains six L3 rounds, not seven long training
runs.

\end{multicols}
\vspace{0.5em}
\begin{center}
\footnotesize
\captionof{table}{Low-level attack-path phase probes. Rows record batch-size-one
forward/backward components; Burgers probes use Tesla V100-SXM2-32GB, and the
NS probe uses an NVIDIA A100-SXM4-80GB record.}
\label{tab:attack_step_timing_summary}
\setlength{\tabcolsep}{4pt}
\begin{tabular*}{\textwidth}{@{\extracolsep{\fill}}llrrrrr@{}}
\toprule
PDE & Path & Forward (s) & Model (s) & Solver (s) & Backward (s) & Total (s) \\
\midrule
Burgers & Full solver & $0.1593$ & $0.0030$ & $0.1557$ & $0.2870$ & $0.4463$ \\
Burgers & Detached solver & $0.1592$ & $0.0028$ & $0.1559$ & $0.0032$ & $0.1624$ \\
Burgers & Approx. target & $0.0024$ & $0.0022$ & $0.0001$ & $0.0023$ & $0.0047$ \\
NS & Full solver & $1.8599$ & $0.0192$ & $1.8407$ & $3.8295$ & $5.6932$ \\
\bottomrule
\end{tabular*}
\end{center}
\vspace{0.5em}

\begin{center}
\footnotesize
\captionof{table}{PGD backend matrix with phase-level time and GPU memory consumption. Burgers uses
$n_x=1024$, $\Delta t=0.001$, $T=1$, $\nu=0.001$. NS uses $256^2$ fields,
$\Delta t=0.005$, $\nu=10^{-5}$. Each row attacks one sample (batch size $1$).
Memory is reported as whole-GPU peak memory.
Hardware: this mixed-backend matrix was recorded on NVIDIA A100-SXM4-80GB.}
\label{tab:backend_bridge_timing}
\setlength{\tabcolsep}{2.3pt}
\begin{tabular}{llrrrrr}
\toprule
PDE & Solver/model & Step (s) & Fwd (s) & Solver (s) & Backward (s) & Peak GiB \\
\midrule
Burgers & JAX solver/Torch model & $0.895$ & $0.164$ & $0.156$ & $0.321$ & $0.65$ \\
Burgers & JAX solver/JAX model & $0.952$ & $0.169$ & $0.153$ & $0.329$ & $0.73$ \\
Burgers & Torch solver/Torch model & $2.114$ & $0.711$ & $0.706$ & $0.949$ & $0.69$ \\
Burgers & Torch solver/JAX model & $2.199$ & $0.747$ & $0.727$ & $0.989$ & $0.79$ \\
NS & JAX solver/Torch model & $4.194$ & $1.163$ & $1.095$ & $2.194$ & $34.30$ \\
NS & JAX solver/JAX model & $4.732$ & $1.297$ & $1.065$ & $2.563$ & $32.80$ \\
NS & Torch solver/Torch model & $9.882$ & $5.616$ & $5.564$ & $3.638$ & $25.73$ \\
NS & Torch solver/JAX model & $10.783$ & $5.914$ & $5.517$ & $4.283$ & $26.40$ \\
\bottomrule
\end{tabular}
\end{center}
\vspace{0.5em}

\begin{center}
\footnotesize
\captionof{table}{Standalone component time and GPU memory consumption. These rows isolate
model-only and solver-only baselines within each PDE. ``Plain fwd'' is a
prediction or PDE solve without a backward graph. ``Graph fwd'' is the loss
evaluation with the graph needed for differentiation. ``Forward/Infer peak''
records forward-only prediction or solve memory. Burgers solver rows use
a 1D grid with $n_x=1024$ and $T=1$; NS solver rows use a $256\times256$
spatial grid and $T=20$. Hardware: Burgers rows are from Tesla V100-SXM2-32GB
records; NS rows are from NVIDIA A100-SXM4-80GB records.}
\label{tab:standalone_component_timing_memory}
\setlength{\tabcolsep}{2.1pt}
\begin{tabular}{lllcrrrcc}
\toprule
PDE & Component & Backend & Batch & Plain fwd (s) & Graph fwd (s) & Backward (s) & Forward/Infer peak & Backward/Train peak \\
\midrule
Burgers & Model & PyTorch & $64$ & $0.0036$ & $0.0027$ & $0.0039$ & $140.2$ MiB & $505.8$ MiB \\
Burgers & Model & JAX & $64$ & $0.0020$ & $0.0020$ & $0.0034$ & $717$ MiB & $2733$ MiB \\
Burgers & Solver & PyTorch & $1$ & $0.6394$ & $0.8940$ & $1.3840$ & $0.45$ MiB & $53.5$ MiB \\
Burgers & Solver & JAX & $1$ & $0.2711$ & $0.1370$ & $0.1420$ & $20$ MiB & $46$ MiB \\
NS & Model & PyTorch & $8$ & $0.0633$ & $0.0588$ & $0.1085$ & $534.7$ MiB & $11.45$ GiB \\
NS & Model & JAX & $8$ & $0.0540$ & $0.0612$ & $0.0805$ & $1533$ MiB & $29.88$ GiB \\
NS & Solver & PyTorch & $1$ & $5.5949$ & $7.9290$ & $11.1270$ & $33.8$ MiB & $21.05$ GiB \\
NS & Solver & JAX & $1$ & $3.6890$ & $1.8360$ & $1.6650$ & $45.6$ MiB & $32.04$ GiB \\
\bottomrule
\end{tabular}
\end{center}
\vspace{0.2em}

\begin{table}[H]
\centering
\footnotesize
\caption{Checkpointing/rematerialization tradeoff for three-step attack probes.
Rows report pass/OOM status, throughput, and source-record GPU peak memory on a
single Tesla V100-SXM2-32GB path.}
\label{tab:remat_time_memory_tradeoff}
\setlength{\tabcolsep}{3.0pt}
\begin{tabular}{llllrrrr}
\toprule
PDE & Remat & Rollout recomp. \% & Batch & Status & Batch (s) & Samples/s & Peak GiB \\
\midrule
NS & chunk-20 & $95.0$ & $5$ & pass & $13.844$ & $0.361$ & $25.18$ \\
NS & chunk-20 & $95.0$ & $6$ & pass & $14.522$ & $0.413$ & $28.98$ \\
NS & chunk-20 & $95.0$ & $7$ & OOM & -- & -- & $31.64$ \\
NS & micro & n/a & $4$ & pass & $11.377$ & $0.352$ & $24.64$ \\
NS & micro & n/a & $5$ & pass & $13.111$ & $0.381$ & $31.38$ \\
NS & second-level & $99.5$ & $5$ & pass & $13.890$ & $0.360$ & $28.18$ \\
NS & none & $0.0$ & $2$ & OOM & -- & -- & $10.89$ \\
Burgers & chunk-50 & $98.0$ & $768$ & pass & $17.941$ & $42.807$ & $6.10$ \\
Burgers & none & $0.0$ & $384$ & pass & $11.626$ & $33.029$ & $22.56$ \\
Burgers & chunk-50 & $98.0$ & $512$ & pass & $17.124$ & $29.899$ & $4.07$ \\
Burgers & all & $\approx99.9$ & $512$ & pass & $18.478$ & $27.709$ & $25.49$ \\
Burgers & chunk-50 & $98.0$ & $384$ & pass & $16.956$ & $22.646$ & $3.06$ \\
Burgers & none & $0.0$ & $256$ & pass & $15.056$ & $17.003$ & $14.72$ \\
Burgers & none & $0.0$ & $512$ & OOM & -- & -- & -- \\
\bottomrule
\end{tabular}
\end{table}

\begin{table}[H]
\centering
\footnotesize
\caption{10-step attack batch capacity after selecting practical memory-saving
settings. Rows report pass/OOM status, per-batch and per-sample time, and
source-record GPU peak memory on a single Tesla V100-SXM2-32GB path.}
\label{tab:corrected_attack_throughput}
\setlength{\tabcolsep}{2.8pt}
\begin{tabular}{lllrrrr}
\toprule
PDE & Randomization & Batch & Status & Batch (s) & Sample (s) & Peak GiB \\
\midrule
NS & fixed $\epsilon,\alpha$ & $6$ & pass & $431.943$ & $71.9905$ & $26.96$ \\
NS & fixed $\epsilon,\alpha$ & $8$ & OOM & -- & -- & -- \\
Burgers & fixed $\epsilon$ & $1350$ & pass & $53.075$ & $0.0393$ & $10.73$ \\
Burgers & fixed $\epsilon$ & $1280$ & pass & $50.349$ & $0.0393$ & $10.17$ \\
Burgers & fixed $\epsilon$ & $1024$ & pass & $49.301$ & $0.0481$ & $8.13$ \\
Burgers & fixed $\epsilon$ & $768$ & pass & $50.119$ & $0.0653$ & $6.10$ \\
Darcy Flow & fixed top-$k$ flips & $448$ & pass & $14.522$ & $0.0324$ & $24.81$ \\
Darcy Flow & fixed top-$k$ flips & $480$ & pass & $15.649$ & $0.0326$ & $26.48$ \\
Darcy Flow & fixed top-$k$ flips & $384$ & pass & $12.773$ & $0.0333$ & $21.48$ \\
Darcy Flow & fixed top-$k$ flips & $256$ & pass & $9.486$ & $0.0371$ & $14.82$ \\
Darcy Flow & fixed top-$k$ flips & $496$ & OOM & -- & -- & -- \\
\bottomrule
\end{tabular}
\end{table}

\begin{table}[H]
\centering
\footnotesize
\caption{Historical short-probe adversarial-training estimates from measured
one-batch timings under the probe-specific epoch sizes. The attack policy is
the Loss 3 online self-attack used to generate adversarial batches. Hardware:
single Tesla V100-SXM2-32GB path. The Navier--Stokes row uses nine probe batches
and is not the production 1,150-sample epoch budget; the corrected prewarmed
production estimate is reported in
\tabref{tab:revision-ns-500epoch-budget}.}
\label{tab:adv_training_epoch_timing}
\setlength{\tabcolsep}{2.4pt}
\begin{tabular}{llllrrrr}
\toprule
Task & Loss 3 attack policy & Batch & Batches/epoch & Attack (s) & Train (s) & Step (s) & Epoch (s) \\
\midrule
Burgers & $10$-step steepest replace, chunk $50$ & $1350$ & $1$ & $53.075$ & $1.418$ & $54.498$ & $54.498$ \\
Darcy Flow & $10$-step binary steepest replace & $448$ & $3$ & $14.522$ & $2.060$ & $16.594$ & $49.781$ \\
NS & $10$-step steepest add, chunk $20$ & $6$ & $9$ & $431.943$ & $2.537$ & $434.494$ & $3910.445$ \\
\bottomrule
\end{tabular}
\end{table}

\begin{center}
\footnotesize
\captionof{table}{Adversarial-training and random clean/random solver wall-clock time and GPU memory per epoch.
Hardware: Tesla V100-SXM2-32GB CUDA training records.}
\label{tab:training_epoch_time_comparison}
\setlength{\tabcolsep}{5pt}
\renewcommand{\arraystretch}{1.14}
\begin{tabular*}{\textwidth}{@{\extracolsep{\fill}}llcccccc@{}}
\toprule
PDE & Metric & Loss 1 adv & Loss 2 adv & Loss 3 adv & Physics adv & Rand clean & Rand solver \\
\midrule
Burgers & Time (s/epoch) & $7.3405$ & $19.7521$ & $28.1989$ & -- & $2.0767$ & $4.8479$ \\
& Time / Loss 3 & $0.26{\times}$ & $0.70{\times}$ & $1.00{\times}$ & -- & $0.07{\times}$ & $0.17{\times}$ \\
& Peak GiB & $10.65$ & $10.65$ & $10.75$ & -- & $2.71$ & $2.72$ \\
\midrule
Darcy Flow & Time (s/epoch) & $4.9267$ & $4.3757$ & $4.6249$ & $4.0338$ & $2.8917$ & $3.0107$ \\
 & Time / Loss 3 & $1.07{\times}$ & $0.95{\times}$ & $1.00{\times}$ & $0.87{\times}$ & $0.63{\times}$ & $0.65{\times}$ \\
 & Peak GiB & $7.97$ & $7.45$ & $7.45$ & $7.45$ & $5.12$ & $5.12$ \\
\bottomrule
\end{tabular*}
\end{center}
\vspace{0.7em}

\subsection*{Completed Attack, Training, and Rematerialization Costs}

\begin{table}[H]
\centering
\footnotesize
\setlength{\tabcolsep}{4pt}
\caption{Burgers attack-core cost for batch 50. Times are seconds per attack
step (mean $\pm$ standard deviation); memory is the framework-peak-sum upper
bound in GiB. L2-dict reports three dictionary sizes.}
\label{tab:revision-burgers-attack-cost}
\begin{tabular*}{\textwidth}{@{\extracolsep{\fill}}lrrrr@{}}
\toprule
Attack objective & No-remat time & No-remat GiB & Checkpoint time & Checkpoint GiB\\
\midrule
L1 & $0.008282\pm0.000055$ & 0.3481 & -- & --\\
L2 fixed target & $0.008792\pm0.000544$ & 0.3481 & -- & --\\
L2 dictionary, 200 & $0.008492\pm0.000133$ & 0.3497 & -- & --\\
L2 dictionary, 2,000 & $0.008777\pm0.000245$ & 0.3634 & -- & --\\
L2 dictionary, 20,000 & $0.009473\pm0.001093$ & 0.5547 & -- & --\\
L3 Stop & $0.166627\pm0.000055$ & 0.3485 & -- & --\\
L3 Full & $0.629719\pm0.007720$ & 3.4002 & $0.785095\pm0.001764$ & 0.4685\\
\bottomrule
\end{tabular*}
\end{table}

\begin{table}[H]
\centering
\footnotesize
\setlength{\tabcolsep}{3.5pt}
\caption{Burgers training cost for batch 128. Time is seconds per complete
epoch and memory is PyTorch peak GiB. The L3 Full rows expose the
time--memory tradeoff across checkpoint chunks; the no-remat reference is
repeated to make each comparison explicit.}
\label{tab:revision-burgers-training-cost}
\begin{tabular*}{\textwidth}{@{\extracolsep{\fill}}lrrrr@{}}
\toprule
Training method & No-remat time & No-remat GiB & Checkpoint time & Checkpoint GiB\\
\midrule
L1 & 12.094 & 0.8908 & -- & --\\
L2 & 59.296 & 0.8908 & -- & --\\
L3 Stop & 58.803 & 0.8908 & -- & --\\
Random clean & 1.083 & 0.3256 & -- & --\\
Random solver & 10.527 & 0.3266 & -- & --\\
L3 Full, no remat & 177.819 & 7.2404 & -- & --\\
L3 Full, chunk 16 & 177.819 & 7.2404 & 299.369 & 0.9211\\
L3 Full, chunk 32 & 177.819 & 7.2404 & 281.480 & 0.9060\\
L3 Full, chunk 50 & 177.819 & 7.2404 & 271.777 & 0.9001\\
L3 Full, chunk 64 & 177.819 & 7.2404 & 270.559 & 0.8981\\
\bottomrule
\end{tabular*}
\end{table}

\begin{table}[H]
\centering\footnotesize
\setlength{\tabcolsep}{3pt}
\renewcommand{\arraystretch}{1.12}
\caption{Burgers checkpoint numerical-equivalence probe: batch 2, 1,000 solver steps, float32, tolerance $10^{-7}$. Differences are against the uncheckpointed run; all measured forward, gradient, and loss differences are zero. The common loss is $2.4184571\times10^{-13}$. Torch allocation is a separate probe and must not be substituted for the batch-128 training-memory table.}\label{tab:revision-checkpoint-equivalence}
\begin{tabular*}{\textwidth}{@{\extracolsep{\fill}}lrrrrl@{}}
\toprule
Mode & $|\Delta$loss$|$ & Max forward diff. & Max gradient diff. & Torch MiB & Status\\
\midrule
None & 0 & 0 & 0 & 104.144 & PASS\\
Chunk 16 & 0 & 0 & 0 & 2.297 & PASS\\
Chunk 32 & 0 & 0 & 0 & 3.719 & PASS\\
Chunk 50 & 0 & 0 & 0 & 5.505 & PASS\\
Chunk 64 & 0 & 0 & 0 & 6.922 & PASS\\
\bottomrule
\end{tabular*}
\end{table}

\begingroup\footnotesize
\setlength{\tabcolsep}{3pt}
\renewcommand{\arraystretch}{1.18}
\setlength{\LTcapwidth}{\textwidth}
\begin{longtable}{@{}L{0.33\textwidth}C{0.19\textwidth}R{0.08\textwidth}C{0.19\textwidth}R{0.08\textwidth}@{}}
\caption{All 17 Navier--Stokes attack-core configurations at batch one. Times are prewarmed seconds per attack update (mean $\pm$ sample standard deviation of two independent process measurements), excluding final relabeling and model training. Checkpointing uses chunk 20. Memory is the sum of separate PyTorch and JAX framework peaks, a conservative upper bound excluding idle caches; it is not a synchronized whole-process peak. W retains a solver path, D detaches it, and A substitutes an approximated input frame.}\label{tab:revision-ns-attack-cost}\\
\toprule
Attack path & No-remat s/step & No-remat GiB & Ckpt. s/step & Ckpt. GiB\\
\midrule
\endfirsthead
\toprule
Attack path & No-remat s/step & No-remat GiB & Ckpt. s/step & Ckpt. GiB\\
\midrule
\endhead
\midrule
\multicolumn{5}{r}{continued on next page}\\
\endfoot
\bottomrule
\endlastfoot
L3 approximate inputs; target gradient retained & $7.372962\pm0.050077$ & 20.7129 & $8.526814\pm0.002480$ & 5.3249\\
L3 all-input/target Stop (D1--9,19) & $1.944396\pm0.019797$ & 5.0962 & $1.919743\pm0.016663$ & 5.0962\\
L3 D1--9 & $7.155460\pm0.013226$ & 20.7129 & $8.377990\pm0.083881$ & 5.3249\\
L3 D1--8 & $7.164214\pm0.024458$ & 20.7129 & $8.358188\pm0.046168$ & 5.3249\\
L3 D1--7 & $7.171153\pm0.001647$ & 20.7129 & $8.391740\pm0.074922$ & 5.3249\\
L3 D1--6 & $7.170723\pm0.060463$ & 20.7129 & $8.388739\pm0.066036$ & 5.3249\\
L3 D1--5 & $7.151467\pm0.039892$ & 20.7129 & $8.390130\pm0.060111$ & 5.3249\\
L3 D1,3,5,7,9 & $7.176548\pm0.007697$ & 20.7129 & $8.369445\pm0.018625$ & 5.3249\\
L3 D2,4,6,8 & $7.146156\pm0.043775$ & 20.7129 & $8.396116\pm0.046464$ & 5.3249\\
L3 D5--9 & $7.178800\pm0.009098$ & 20.7129 & $8.351778\pm0.020798$ & 5.3249\\
L3 Full (all W) & $7.196903\pm0.064727$ & 20.7129 & $8.348192\pm0.046342$ & 5.3249\\
L1 & $3.274370\pm0.117762$ & 12.8880 & $4.209594\pm0.081393$ & 5.1899\\
L2 approximate inputs; fixed clean target & $0.418659\pm0.004576$ & 5.0642 & $0.423071\pm0.013272$ & 5.0642\\
L3 W1--5, D6--9 & $7.153686\pm0.023474$ & 20.7129 & $8.365060\pm0.001144$ & 5.3249\\
L3 A1--5, D6--9 & $7.407104\pm0.024174$ & 20.7129 & $8.455341\pm0.040983$ & 5.3249\\
L3 target Stop (original D19) & $7.193827\pm0.039058$ & 20.7080 & $8.582307\pm0.285208$ & 5.3201\\
L3 target Stop (prefix backward) & $4.090714\pm0.022607$ & 12.8931 & $5.015836\pm0.028954$ & 5.1951\\
\end{longtable}
\endgroup
In this recurrent benchmark the solver-generated input frames are 1--9 and the target is frame 19. Detaching only the target leaves solver paths through the model inputs. The all-input/target Stop row removes both. The approximate-input row still retains the target solver gradient, and is therefore different from a fixed-target L2 dictionary attack. These attack measurements use a JAX final-frame target; the short training-batch table uses native PyTorch multi-frame MSE and a different timing window.

\begin{table}[H]
\centering
\footnotesize
\setlength{\tabcolsep}{3.2pt}
\caption{Navier--Stokes short training-batch cost for batch one. Time is
seconds per training batch (mean $\pm$ standard deviation); memory is PyTorch
peak GiB. These are completed short-batch probes, not seven long training
runs.}
\label{tab:revision-ns-training-cost}
\begin{tabular*}{\textwidth}{@{\extracolsep{\fill}}lrrrr@{}}
\toprule
Training path & No-remat time & No-remat GiB & Checkpoint time & Checkpoint GiB\\
\midrule
L1 & $104.694\pm8.123$ & 14.7125 & $161.619\pm2.460$ & 5.8080\\
L2, approximate inputs/fixed clean target & $18.019\pm2.886$ & 5.8078 & $15.966\pm0.371$ & 5.8078\\
L3 Full & $201.604\pm3.103$ & 26.2792 & $344.521\pm1.313$ & 5.8031\\
L3 target Stop & $202.059\pm9.109$ & 26.2822 & $352.371\pm31.143$ & 5.8082\\
L3 all-solver-path Stop & $69.406\pm3.131$ & 5.8080 & $76.100\pm18.039$ & 5.8080\\
L3 input-frame Stop, target gradient retained & $218.232\pm22.593$ & 26.2871 & $372.455\pm48.705$ & 5.8082\\
L3 approximate input frames & $216.135\pm0.808$ & 26.2871 & $377.586\pm34.951$ & 5.8082\\
\bottomrule
\end{tabular*}
\end{table}

\begin{table}[H]
\centering
\footnotesize
\setlength{\tabcolsep}{4pt}
\caption{Navier--Stokes rematerialization chunk sweep at batch one and attack
batch-capacity sweep at chunk 20. Core-step time is measured after warm-up.
Process memory is whole-process GiB. The no-remat row reached OOM under the
standard allocation; chunk 20 is the fastest passing chunk in this sweep.}
\label{tab:revision-ns-remat-capacity}
\begin{minipage}[t]{0.47\textwidth}
\centering
\begin{tabular}{lrrrr}
\toprule
Mode & $K$ & Core s/step & Process GiB & Status\\
\midrule
None & -- & -- & 5.596 & OOM\\
Chunk & 20 & 8.350 & 5.848 & pass\\
Chunk & 32 & 8.859 & 5.850 & pass\\
Chunk & 64 & 9.076 & 6.100 & pass\\
Chunk & 96 & 8.883 & 6.100 & pass\\
\bottomrule
\end{tabular}
\end{minipage}\hfill
\begin{minipage}[t]{0.50\textwidth}
\centering
\begin{tabular}{rrrrr}
\toprule
Batch & 50-step attack (s) & Samples/s & Process GiB & Status\\
\midrule
1 & 418.086 & 0.120 & 5.848 & pass\\
2 & 508.181 & 0.197 & 10.342 & pass\\
4 & 591.300 & 0.338 & 19.387 & pass\\
5 & 687.102 & 0.364 & 23.531 & pass\\
6 & 717.634 & 0.418 & 27.750 & pass\\
7 & -- & -- & 31.092 & OOM\\
\bottomrule
\end{tabular}
\end{minipage}
\end{table}

\begin{table}[H]
\centering
\footnotesize
\setlength{\tabcolsep}{4pt}
\caption{Prewarmed Navier--Stokes L3 Full production-budget estimates for
1,150 training samples on one V100. The values extrapolate measured steady
batch throughput; they are not claims that 500 epochs were actually run, and
they exclude validation and checkpoint-I/O overhead.}
\label{tab:revision-ns-500epoch-budget}
\begin{tabular*}{\textwidth}{@{\extracolsep{\fill}}lrrrrrr@{}}
\toprule
Batch / checkpoint & Full batches/epoch & s/full batch & h/epoch & days/500 epochs & Active GiB & Evidence\\
\midrule
5 / chunk 20 & 230 & 335.346 & 21.42 & 446.35 & 25.5522 & prewarmed measured throughput\\
1 / none & 1,150 & 200.487 & 64.04 & 1,334.26 & 27.8705 & prewarmed measured throughput\\
\bottomrule
\end{tabular*}
\end{table}

The production-budget table supersedes the earlier short-probe estimate that
treated nine batches as a complete Navier--Stokes epoch.  It also explains why
the long-training quality result is limited to six L3 rounds: the cost tables
compare mechanisms and feasible batches, but they do not imply that seven
alternative methods were each trained for hundreds of Navier--Stokes epochs.

\subsection*{Steady Darcy Flow Cost Controls}
Darcy uses a steady elliptic CG solve with implicit differentiation. Its
records therefore have no temporal checkpoint/rematerialization toggle.
The paired hot-epoch probe below excludes evaluation; its process and PyTorch
memory counters are reported separately.

\begin{table}[H]
\centering\footnotesize
\setlength{\tabcolsep}{3pt}
\caption{Darcy Full/Stop training microbenchmark, batch 64 and one batch per
epoch. All times are synchronized seconds; evaluation is excluded.}
\label{tab:revision-darcy-training-cost}
\begin{tabular*}{\textwidth}{@{\extracolsep{\fill}}lrrrrrr@{}}
\toprule
Method & s/epoch & Process GiB & Torch GiB & Attack s & Relabel s & Update s\\
\midrule
L3 Full & 0.484185 & 8.441406 & 7.452587 & 0.183024 & 0.022486 & 0.278675\\
L3 Stop & 0.438895 & 8.433594 & 7.450473 & 0.137058 & 0.022598 & 0.279239\\
\bottomrule
\end{tabular*}
\end{table}

\begin{table}[H]
\centering\footnotesize
\setlength{\tabcolsep}{4pt}
\caption{Darcy attack-core costs on the same batch of 20 ID held-out inputs.
The final true-L3 evaluation is excluded. Solver calls count 100 attack
iterations; memory counters are not interchangeable with framework-peak sums
in the time-dependent probes.}
\label{tab:revision-darcy-attack-cost}
\begin{tabular*}{\textwidth}{@{\extracolsep{\fill}}lrrrr@{}}
\toprule
Attack & s/step & Process GiB & Torch GiB & Solver forward/backward\\
\midrule
L1 & 0.043337 & 3.851562 & 3.069851 & 0/0\\
L2 fixed & 0.042741 & 3.851562 & 3.069851 & 0/0\\
Dictionary 200 & 0.044634 & 3.851562 & 3.069852 & 0/0\\
Dictionary 2,000 & 0.043822 & 3.851562 & 3.069852 & 0/0\\
Dictionary 20,000 & 0.044325 & 3.851562 & 3.069852 & 0/0\\
L3 Stop & 0.055449 & 3.851562 & 3.069851 & 100/0\\
L3 Full & 0.077847 & 3.851562 & 3.069119 & 100/100\\
Physics & 0.042544 & 3.855469 & 3.069119 & 0/0 (100 residuals)\\
\bottomrule
\end{tabular*}
\end{table}

\clearpage
\onecolumn
\appendixsection{Adversarial Attack Optimization Methods}{app:adversarial-attack-optimization-methods}
\label{app:loss-landscape-mechanism-diagnostics}
\label{app:diagram-flow}
\label{app:si-attack-training-figures}
\vspace{0.4em}
\begin{multicols}{2}

\raggedbottom

These attacks update the perturbation $\delta$ by projected gradient-ascent
steps on the attack loss $\ell(x_0+\delta)$. In adversarial-attack terminology,
this procedure is commonly called PGD even though the sign is ascent rather
than descent. Let
$\mathcal B_p(\varepsilon)=\{\delta:\|\delta\|_{\mathcal X,p}\le\varepsilon\}$.
Under an $L_\infty$ budget, the standard Madry-style update
\citep{madry2018towards} takes a sign-gradient step and clips the perturbation
elementwise to $[-\varepsilon,\varepsilon]$.

\begin{algorithm}[H]
\caption{PGD attack in an $L_\infty$ ball}
\begin{algorithmic}[1]
\STATE \textbf{Input:} clean input $x_0=a$, radius $\varepsilon$, step size
\(\alpha\), steps \(T\).
\STATE Initialize $\delta^{(0)}\leftarrow0$.
\FOR{$t=0,1,\ldots,T-1$}
  \STATE $g^{(t)}\leftarrow\nabla_x \ell(x_0+\delta^{(t)})$.
  \STATE $\delta^{(t+1)}\leftarrow
  \operatorname{clip}_{[-\varepsilon,\varepsilon]}
  \!\left(\delta^{(t)}+\alpha\,\operatorname{sign}(g^{(t)})\right)$.
\ENDFOR
\STATE \textbf{Output:} attacked input $x_0+\delta^{(T)}$.
\end{algorithmic}
\end{algorithm}

For an $L_2$ budget, the perturbation step is normalized and then projected
back to $\mathcal B_2(\varepsilon)$; hence $\alpha$ is an approximate $L_2$
step length.

\begin{algorithm}[H]
\caption{PGD attack in an $L_2$ ball}
\begin{algorithmic}[1]
\STATE \textbf{Input:} clean input $x_0=a$, radius $\varepsilon$, step size
$\alpha$, steps $T$, stability constant $\delta_{\mathrm{stab}}$.
\STATE Initialize $\delta^{(0)}\leftarrow0$.
\FOR{$t=0,\ldots,T-1$}
  \STATE $g^{(t)}\leftarrow\nabla_x \ell(x_0+\delta^{(t)})$.
  \STATE $\delta^{(t+1)}\leftarrow
  \Pi_{\mathcal B_2(\varepsilon)}
  \!\left(\delta^{(t)}+
  \alpha\,g^{(t)}/(\|g^{(t)}\|_2+\delta_{\mathrm{stab}})\right)$.
\ENDFOR
\STATE \textbf{Output:} attacked input $x_0+\delta^{(T)}$.
\end{algorithmic}
\end{algorithm}

For the raw and steepest variants, use the current gradient $g^{(t)}$ and define
$\widehat g_p^{(t)}=g^{(t)}/\|g^{(t)}\|_p$ for $g^{(t)}\ne0$, and $0$
otherwise. The four rules are
\begin{align*}
\text{raw add:}\quad
&\delta^{(t+1)}=
\Pi_{\mathcal B_p(\varepsilon)}(\delta^{(t)}+\alpha g^{(t)}),\\
\text{raw replace:}\quad
&\delta^{(t+1)}=\varepsilon\widehat g_p^{(t)},\\
\text{steepest add:}\quad
&\delta^{(t+1)}=
\Pi_{\mathcal B_p(\varepsilon)}(\delta^{(t)}+\alpha s_p(g^{(t)})),\\
\text{steepest replace:}\quad
&\delta^{(t+1)}=\varepsilon s_p(g^{(t)}).
\end{align*}

The steepest rules use \(s_p(g)\); the following calculation gives this
direction. The gradient comes from the scalar loss, not from the constraint.
Write the current perturbed input as \(z=x_0+\delta^{(t)}\), the residual/error
field as \(e(z)\), and a powered loss as
\begin{equation*}
  \ell(z)=\Phi_q(e(z)),\qquad
  \Phi_q(e)=\frac{1}{q}\|e\|_q^q .
\end{equation*}
For \(q>1\),
\begin{equation*}
  \nabla_e\Phi_q(e)
  =
  \operatorname{sign}(e)\odot |e|^{q-1},
\end{equation*}
By the chain rule,
\begin{equation*}
  g^{(t)}
  =
  \nabla_z\ell(z)
  =
  De[z]^*\,\nabla_e\Phi_q(e(z)).
\end{equation*}
Thus a small step \(\eta=\alpha v\) changes the loss by
\(\ell(z+\eta)=\ell(z)+\alpha\langle g^{(t)},v\rangle+o(\alpha)\). Therefore
the \(L_p\)-unit steepest-ascent direction is
\begin{equation*}
  s_p(g^{(t)})\in
  \operatorname*{arg\,max}_{\|v\|_p\le1}\langle g^{(t)},v\rangle .
\end{equation*}
Raw updates instead use \(g^{(t)}\) itself and rely on projection or rescaling.
Let \(p^\star\) be the Holder conjugate of \(p\). Equality in Holder's
inequality gives the interior case
\begin{equation*}
  \bigl(s_p(g)\bigr)_i=
  \frac{\operatorname{sign}(g_i)|g_i|^{p^\star-1}}
       {\|g\|_{p^\star}^{p^\star-1}} .
\end{equation*}
The endpoint limits are $s_2(g)=g/\|g\|_2$,
$s_\infty(g)=\operatorname{sign}(g)$, and $s_1(g)$ is concentrated on an index
attaining $\max_i|g_i|$. Thus $p=2$ makes raw and steepest replace identical,
while $p=\infty$ recovers sign-PGD.

\paragraph{Darcy Flow binary-flip attack.}
Darcy Flow uses a binary coefficient field
($a_{\mathrm{lo}}=3$, $a_{\mathrm{hi}}=12$), so the attack variable is a flip set
$F$ under a Hamming budget $|F|\le K$, not a continuous perturbation. Gradients
only rank candidate flips. The solver always receives a binary coefficient; in
training, the target is recomputed by solving the PDE at the attacked
coefficient rather than reusing the clean target.

\begin{algorithm}[H]
\caption{Darcy Flow binary coefficient-flip attack}
\label{alg:darcy-binary-flip}
\begin{algorithmic}[1]
\STATE \textbf{Input:} clean coefficient
$a_0\in\{a_{\mathrm{lo}},a_{\mathrm{hi}}\}^m$, valid pixels $V$, Hamming
budget $K$, and steps $T$.
\STATE Define $a_{\mathrm{opp}}(a_{\mathrm{lo}})=a_{\mathrm{hi}}$ and
$a_{\mathrm{opp}}(a_{\mathrm{hi}})=a_{\mathrm{lo}}$.
\STATE For a flip set $F$, set $a(F)_i=a_{\mathrm{opp}}(a_{0,i})$ if
$i\in F$, and $a(F)_i=a_{0,i}$ otherwise.
\STATE Initialize $F^{(0)}\leftarrow\emptyset$ and accumulated score
$R^{(0)}\leftarrow0$.
\STATE For each $i\in V$, precompute the flip displacement
$\Delta_i\leftarrow a_{\mathrm{opp}}(a_{0,i})-a_{0,i}$.
\FOR{$t=0,\ldots,T-1$}
  \STATE Build the current binary coefficient
  $a^{(t)}\leftarrow a(F^{(t)})$.
  \STATE Compute the attack gradient
  $g^{(t)}\leftarrow\nabla_a\ell(a^{(t)})$.
  \STATE For each $i\in V$, estimate the gain from flipping pixel $i$:
  $\gamma_i^{(t)}\leftarrow g^{(t)}_i\Delta_i$.
  \STATE Accumulate scores:
  $R_i^{(t+1)}\leftarrow R_i^{(t)}+\gamma_i^{(t)}$ for $i\in V$.
  \STATE Set $F^{(t+1)}$ to the at most $K$ pixels in $V$ with largest
  positive $R_i^{(t+1)}$.
\ENDFOR
\STATE \textbf{Output:} binary attacked coefficient $a(F^{(T)})$.
\end{algorithmic}
\end{algorithm}

Algorithm~\ref{alg:darcy-binary-flip} accumulates first-order flip gains in
\(R\) and projects to a top-\(K\) Hamming mask. The score
\(\gamma_i^{(t)}=g_i^{(t)}\Delta_i\) is the Taylor-predicted loss change from a
single binary flip: \(\Delta_i\) is the actual change in coefficient value, and
\(g_i^{(t)}\) tells how the loss changes with that coefficient. Positive scores
are beneficial candidate flips. Other
optimizers keep the same loss, gradient, and binary map, changing only the
score update and mask selection. Let \(T_r(C;w)\) be the \(r\) pixels in \(C\)
with largest weight \(w_i\), \(C_t^+=\{i\in V:\gamma_i^{(t)}>0\}\),
\(C_{t,\mathrm{new}}^+=C_t^+\setminus F^{(t)}\), and
\(r_t=\min\{\alpha_f,K-|F^{(t)}|\}\). The four binary analogues are
\begin{align*}
F_{\mathrm{raw\ add}}^{(t+1)}
  &=F^{(t)}\cup T_{r_t}(C_{t,\mathrm{new}}^+;|g_i^{(t)}|),\\
F_{\mathrm{raw\ replace}}^{(t+1)}
  &=T_K(C_t^+;|g_i^{(t)}|),\\
F_{\mathrm{steepest\ add}}^{(t+1)}
  &=F^{(t)}\cup T_{r_t}(C_{t,\mathrm{new}}^+;\gamma_i^{(t)}),\\
F_{\mathrm{steepest\ replace}}^{(t+1)}
  &=T_K(C_t^+;\gamma_i^{(t)}).
\end{align*}
Pixels outside the selected positive-score set stay at their clean value.
Add-style updates preserve \(F^{(t)}\) and add at most \(\alpha_f\) new flips;
replace-style updates redraw the mask and can remove stale flips.

The equivalences are important. For continuous \(p=2\) attacks, as in our
Burgers \(L_2\) runs, raw replace and steepest replace are the same normalized
boundary update. For \(L_\infty\) PGD, \(s_\infty(g)=\operatorname{sign}(g)\),
so sign-PGD is the steepest-add update. For Darcy Flow, each valid flip has
\(\Delta_i\in\{-9,9\}\). On \(C_t^+\),
\[
  \gamma_i^{(t)}=g_i^{(t)}\Delta_i=9|g_i^{(t)}|,
\]
so \(|g_i^{(t)}|\) and \(\gamma_i^{(t)}\) rank candidates identically. Hence
raw add and steepest add coincide, raw replace and steepest replace coincide,
and the binary case has only two distinct behaviors: add versus replace.

In several differentiable-solver attacks we update the perturbation with an
Adam-adapted direction before the $L_2$ projection. The gradient is evaluated at
the attacked input $x_0+\delta^{(t)}$, while the projected variable remains
$\delta$. In our experiments, this adaptation makes the attack loss increase
more smoothly and reduces the magnitude of oscillations, especially when solver
gradients are noisy across time steps. We mainly use $L_2$ attacks because their
smoother perturbations are more plausible PDE initial conditions.

\begin{algorithm}[H]
\caption{Adam-adapted PGD update for the perturbation under an $L_2$ constraint}
\begin{algorithmic}[1]
\STATE \textbf{Input:} $x_0$, $\varepsilon$, $\alpha$, steps $T$,
$\beta_1,\beta_2$, $\epsilon_{\mathrm{adam}}$, $\delta_{\mathrm{stab}}$.
\STATE Initialize $\delta^{(0)}\leftarrow0$, $m^{(0)}\leftarrow0$, and $v^{(0)}\leftarrow0$.
\FOR{$t=0,\ldots,T-1$}
  \STATE $g^{(t)}\leftarrow\nabla_x\ell(x_0+\delta^{(t)})$.
  \STATE $m^{(t+1)}\leftarrow\beta_1m^{(t)}+(1-\beta_1)g^{(t)}$.
  \STATE $v^{(t+1)}\leftarrow\beta_2v^{(t)}+(1-\beta_2)(g^{(t)}\odot g^{(t)})$.
  \STATE $\hat m^{(t+1)}\leftarrow m^{(t+1)}/(1-\beta_1^{t+1})$.
  \STATE $\hat v^{(t+1)}\leftarrow v^{(t+1)}/(1-\beta_2^{t+1})$.
  \STATE $d^{(t+1)}\leftarrow
  \hat m^{(t+1)}/(\sqrt{\hat v^{(t+1)}}+\epsilon_{\mathrm{adam}})$.
  \STATE $u^{(t+1)}\leftarrow d^{(t+1)}/(\|d^{(t+1)}\|_2+\delta_{\mathrm{stab}})$.
  \STATE $\delta^{(t+1)}\leftarrow
  \Pi_{\mathcal B_2(\varepsilon)}(\delta^{(t)}+\alpha u^{(t+1)})$.
\ENDFOR
\STATE \textbf{Output:} attacked input $x_0+\delta^{(T)}$.
\end{algorithmic}
\end{algorithm}

\end{multicols}
\subsection*{Expanded Objective and Gradient-Path Comparisons}
The following controls separate attack quality from training quality. Each table uses the correct solver reference at the attacked input, while retaining its source-specific normalization and protocol. Larger final discrepancy indicates a stronger attack within a table.
\begin{table}[H]
\centering\footnotesize
\setlength{\tabcolsep}{3pt}
\renewcommand{\arraystretch}{1.1}
\caption{Expanded Burgers attack objectives. Entries are final true MSE in units of $10^{-4}$, mean $\pm$ sample standard deviation on the same 50 inputs; 100 attack steps and $\alpha=\epsilon/10$. The true MSE divides source squared-error sums by 1024.}\label{tab:revision-burgers-expanded-attacks}
\begin{tabular*}{\textwidth}{@{\extracolsep{\fill}}lrrrrr@{}}
\toprule
Attack & $\epsilon=0.125$ & $\epsilon=0.25$ & $\epsilon=0.5$ & $\epsilon=0.75$ & $\epsilon=1$\\
\midrule
L1 & \shortstack{1.003\\$\pm$ 0.605} & \shortstack{0.984\\$\pm$ 0.555} & \shortstack{0.959\\$\pm$ 0.475} & \shortstack{0.953\\$\pm$ 0.438} & \shortstack{0.964\\$\pm$ 0.488}\\
L2 fixed & \shortstack{0.997\\$\pm$ 0.610} & \shortstack{0.976\\$\pm$ 0.564} & \shortstack{0.941\\$\pm$ 0.483} & \shortstack{0.912\\$\pm$ 0.418} & \shortstack{0.891\\$\pm$ 0.389}\\
Dictionary 200 & \shortstack{1.041\\$\pm$ 0.671} & \shortstack{1.065\\$\pm$ 0.698} & \shortstack{1.138\\$\pm$ 0.823} & \shortstack{1.298\\$\pm$ 1.132} & \shortstack{1.617\\$\pm$ 1.807}\\
Dictionary 2,000 & \shortstack{1.049\\$\pm$ 0.681} & \shortstack{1.089\\$\pm$ 0.750} & \shortstack{1.207\\$\pm$ 0.958} & \shortstack{1.402\\$\pm$ 1.259} & \shortstack{1.646\\$\pm$ 1.577}\\
Dictionary 20,000 & \shortstack{1.058\\$\pm$ 0.706} & \shortstack{1.102\\$\pm$ 0.776} & \shortstack{1.244\\$\pm$ 1.005} & \shortstack{1.457\\$\pm$ 1.324} & \shortstack{1.804\\$\pm$ 1.893}\\
L3 Stop & \shortstack{0.993\\$\pm$ 0.608} & \shortstack{0.966\\$\pm$ 0.560} & \shortstack{0.920\\$\pm$ 0.467} & \shortstack{0.883\\$\pm$ 0.392} & \shortstack{0.852\\$\pm$ 0.347}\\
L3 Full & \shortstack{1.275\\$\pm$ 0.842} & \shortstack{1.625\\$\pm$ 1.109} & \shortstack{2.727\\$\pm$ 2.069} & \shortstack{4.504\\$\pm$ 3.735} & \shortstack{7.092\\$\pm$ 6.094}\\
\bottomrule
\end{tabular*}
\end{table}

\begin{table}[H]
\centering\footnotesize
\setlength{\tabcolsep}{3pt}
\renewcommand{\arraystretch}{1.1}
\caption{Expanded Darcy attack objectives on 20 ID held-out inputs, 100 steps. Entries retain the recorded final true-loss normalization (mean $\pm$ standard deviation). The four $(K,\alpha)$ pairs are $(100,5)$, $(250,13)$, $(437,22)$, and $(875,44)$, distinct from the fixed-alpha historical main table.}\label{tab:revision-darcy-expanded-attacks}
\begin{tabular*}{\textwidth}{@{\extracolsep{\fill}}lrrrr@{}}
\toprule
Attack & $K=100$ & $K=250$ & $K=437$ & $K=875$\\
\midrule
L1 & \shortstack{0.02243\\$\pm$ 0.00588} & \shortstack{0.02315\\$\pm$ 0.00513} & \shortstack{0.02438\\$\pm$ 0.00417} & \shortstack{0.03118\\$\pm$ 0.00470}\\
L2 fixed & \shortstack{0.01902\\$\pm$ 0.00312} & \shortstack{0.01884\\$\pm$ 0.00325} & \shortstack{0.02058\\$\pm$ 0.00514} & \shortstack{0.02702\\$\pm$ 0.00963}\\
Dictionary 200 & \shortstack{0.02239\\$\pm$ 0.00649} & \shortstack{0.02414\\$\pm$ 0.00667} & \shortstack{0.02538\\$\pm$ 0.00665} & \shortstack{0.03471\\$\pm$ 0.01300}\\
Dictionary 2,000 & \shortstack{0.02319\\$\pm$ 0.00831} & \shortstack{0.02527\\$\pm$ 0.00879} & \shortstack{0.02541\\$\pm$ 0.00562} & \shortstack{0.03087\\$\pm$ 0.00751}\\
Dictionary 20,000 & \shortstack{0.02218\\$\pm$ 0.00574} & \shortstack{0.02408\\$\pm$ 0.00662} & \shortstack{0.02531\\$\pm$ 0.00528} & \shortstack{0.03196\\$\pm$ 0.00816}\\
L3 Stop & \shortstack{0.01862\\$\pm$ 0.00375} & \shortstack{0.01837\\$\pm$ 0.00299} & \shortstack{0.01898\\$\pm$ 0.00291} & \shortstack{0.02279\\$\pm$ 0.00633}\\
L3 Full & \shortstack{0.04815\\$\pm$ 0.00764} & \shortstack{0.06592\\$\pm$ 0.01150} & \shortstack{0.07723\\$\pm$ 0.01840} & \shortstack{0.08740\\$\pm$ 0.02550}\\
Physics & \shortstack{0.02158\\$\pm$ 0.00494} & \shortstack{0.02141\\$\pm$ 0.00517} & \shortstack{0.02143\\$\pm$ 0.00473} & \shortstack{0.02601\\$\pm$ 0.00744}\\
\bottomrule
\end{tabular*}
\end{table}

\begin{table}[H]
\centering\footnotesize
\setlength{\tabcolsep}{3pt}
\renewcommand{\arraystretch}{1.1}
\caption{Expanded NS attack-objective and gradient-path comparison on the same 20 inputs, 50 steps. Entries are final true L3 (mean $\pm$ standard deviation), not $\Delta$MSE. Budgets $8,16,32,64$ use stepsizes $0.25,0.5,1,2$. W retains solver differentiation, D detaches a solver path, and A substitutes an input frame. Full L3 wins 20/20, 19/20, 15/20, and 3/20 individual comparisons; D1--5/W6--9 has a higher mean at budget 64.}\label{tab:revision-ns-expanded-attacks}
\begin{tabular*}{\textwidth}{@{\extracolsep{\fill}}lrrrr@{}}
\toprule
Attack/path & $\epsilon=8$ & $\epsilon=16$ & $\epsilon=32$ & $\epsilon=64$\\
\midrule
L1 & \shortstack{59.83\\$\pm$ 26.09} & \shortstack{69.06\\$\pm$ 37.82} & \shortstack{114.17\\$\pm$ 53.52} & \shortstack{167.81\\$\pm$ 63.82}\\
L2 fixed / approximate inputs & \shortstack{59.63\\$\pm$ 28.85} & \shortstack{60.61\\$\pm$ 31.21} & \shortstack{67.58\\$\pm$ 30.68} & \shortstack{86.29\\$\pm$ 26.14}\\
L3 Full (all W) & \shortstack{126.83\\$\pm$ 32.48} & \shortstack{186.68\\$\pm$ 35.29} & \shortstack{276.76\\$\pm$ 44.68} & \shortstack{324.62\\$\pm$ 61.35}\\
L3 all D to W & \shortstack{73.79\\$\pm$ 34.33} & \shortstack{95.11\\$\pm$ 43.53} & \shortstack{150.66\\$\pm$ 80.33} & \shortstack{282.88\\$\pm$ 143.40}\\
L3 W1--5 / D6--9 & \shortstack{74.29\\$\pm$ 34.25} & \shortstack{95.76\\$\pm$ 44.96} & \shortstack{157.70\\$\pm$ 84.92} & \shortstack{305.45\\$\pm$ 142.24}\\
L3 D1--5 / W6--9 & \shortstack{120.46\\$\pm$ 33.20} & \shortstack{175.85\\$\pm$ 35.71} & \shortstack{263.03\\$\pm$ 43.63} & \shortstack{329.90\\$\pm$ 69.34}\\
L3 A1--5 / D6--9 & \shortstack{69.96\\$\pm$ 32.63} & \shortstack{88.68\\$\pm$ 34.44} & \shortstack{139.28\\$\pm$ 52.07} & \shortstack{262.26\\$\pm$ 84.87}\\
\bottomrule
\end{tabular*}
\end{table}

\begin{table}[H]
\centering\footnotesize
\setlength{\tabcolsep}{3pt}
\renewcommand{\arraystretch}{1.1}
\caption{Darcy cross-attack final MSE, first four attack columns, in units of $10^{-7}$. All eight trained models use the same 20 ID held-out inputs at $K=437$. Smaller is better across trained models within an attack column. Accepted-checkpoint ablation records are kept separate from common-budget generalization tables.}\label{tab:revision-darcy-cross-attacks}
\begin{tabular*}{\textwidth}{@{\extracolsep{\fill}}lrrrr@{}}
\toprule
Trained model & L1 & L2 fixed & Dict 200 & Dict 2,000\\
\midrule
Base & 0.838 & 0.672 & 1.045 & 0.986\\
L1 & 0.247 & 0.308 & 0.379 & 0.409\\
L2 & 0.327 & 0.309 & 0.383 & 0.397\\
L3 Stop & 0.286 & 0.286 & 0.365 & 0.380\\
L3 Full & 1.212 & 1.406 & 1.507 & 1.462\\
Physics & 0.283 & 0.308 & 0.413 & 0.458\\
Random clean & 0.325 & 0.360 & 0.485 & 0.490\\
Random solver & 0.262 & 0.307 & 0.397 & 0.440\\
\bottomrule
\end{tabular*}
\end{table}

\begin{table}[H]
\centering\footnotesize
\setlength{\tabcolsep}{3pt}
\renewcommand{\arraystretch}{1.1}
\caption{Darcy cross-attack final MSE, remaining attack columns and defense summary, in units of $10^{-7}$. Worst denotes the maximum across all eight tested attacks. Full L3 minimizes the worst measured attack increase, but does not minimize every attack column or the clean error.}\label{tab:revision-darcy-cross-attacks-rest}
\begin{tabular*}{\textwidth}{@{\extracolsep{\fill}}lrrrrrrr@{}}
\toprule
Trained model & Dict 20,000 & L3 Stop & L3 Full & Physics & Clean & Worst final & Worst $\Delta$\\
\midrule
Base & 0.976 & 0.544 & 7.381 & 0.933 & 0.625 & 7.381 & 6.756\\
L1 & 0.369 & 0.181 & 3.341 & 0.549 & 0.258 & 3.341 & 3.083\\
L2 & 0.348 & 0.184 & 3.066 & 0.389 & 0.270 & 3.066 & 2.796\\
L3 Stop & 0.343 & 0.201 & 3.173 & 0.468 & 0.258 & 3.173 & 2.915\\
L3 Full & 1.537 & 1.179 & 2.741 & 0.739 & 0.639 & 2.741 & 2.102\\
Physics & 0.434 & 0.220 & 3.140 & 0.447 & 0.242 & 3.140 & 2.898\\
Random clean & 0.445 & 0.231 & 4.002 & 0.324 & 0.314 & 4.002 & 3.688\\
Random solver & 0.410 & 0.204 & 3.187 & 0.435 & 0.230 & 3.187 & 2.957\\
\bottomrule
\end{tabular*}
\end{table}

\subsection*{Burgers ID and OOD Cross-Attack Controls}
These independent accepted-checkpoint records evaluate every retained model against seven attacks using the correct perturbed solver reference. They are distinct from both the historical robustness table and the maximum-common-budget 52-dataset generalization tables. Each attack cohort contains 50 evaluated inputs; the OOD attack cohort is not the 50-dataset mean used elsewhere. Random clean is excluded from this retained presentation.

\begin{table}[H]
\centering\footnotesize
\setlength{\tabcolsep}{3pt}
\renewcommand{\arraystretch}{1.12}
\caption{Burgers ID cross-attack final true MSE in units of $10^{-6}$. Rows are retained models and columns are evaluation attacks; bold marks the smallest final MSE in each attack column. The original-distribution held-out cohort uses global $L_2$ budget $\epsilon=0.5$, $\alpha=0.05$, 100 steps, and $n=50$.}\label{tab:revision-burgers-id-cross-attacks}
\begin{tabular*}{\textwidth}{@{\extracolsep{\fill}}lrrrrrrr@{}}
\toprule
Trained model & L1 & L2 fixed & Dict. 200 & Dict. 2,000 & Dict. 20,000 & L3 Stop & L3 Full\\
\midrule
Baseline & 95.870 & 94.064 & 113.846 & 120.742 & 124.441 & 91.992 & 272.726\\
L1 & 1.054 & 1.117 & 1.207 & 1.200 & 1.140 & 1.130 & 20.598\\
L2 & 3.343 & 3.298 & 3.537 & 3.613 & 3.482 & 3.220 & 30.397\\
L3 Full & 6.401 & 6.477 & 6.934 & 6.856 & 6.909 & 6.557 & 13.798\\
L3 Stop & 2.743 & 2.388 & 2.805 & 2.763 & 2.902 & 2.319 & 28.737\\
Random solver & \textbf{0.668} & \textbf{0.637} & \textbf{0.739} & \textbf{0.726} & \textbf{0.751} & \textbf{0.628} & \textbf{6.801}\\
\bottomrule
\end{tabular*}
\end{table}

\begin{table}[H]
\centering\footnotesize
\setlength{\tabcolsep}{3pt}
\renewcommand{\arraystretch}{1.12}
\caption{Burgers OOD cross-attack final true MSE in units of $10^{-6}$. Rows are retained models and columns are evaluation attacks; bold marks the smallest final MSE in each attack column. The supplied shifted-input attack cohort has $n=50$. Full L3 has the lowest final MSE in all seven columns for this cohort.}\label{tab:revision-burgers-ood-cross-attacks}
\begin{tabular*}{\textwidth}{@{\extracolsep{\fill}}lrrrrrrr@{}}
\toprule
Trained model & L1 & L2 fixed & Dict. 200 & Dict. 2,000 & Dict. 20,000 & L3 Stop & L3 Full\\
\midrule
Baseline & 902.411 & 897.870 & 1028.100 & 1026.757 & 1034.966 & 890.624 & 1457.609\\
L1 & 503.241 & 552.168 & 506.489 & 518.385 & 528.229 & 573.065 & 872.893\\
L2 & 537.777 & 614.021 & 557.602 & 568.368 & 577.287 & 642.988 & 934.593\\
L3 Full & \textbf{138.966} & \textbf{147.943} & \textbf{141.892} & \textbf{145.328} & \textbf{144.362} & \textbf{156.361} & \textbf{272.252}\\
L3 Stop & 451.466 & 496.852 & 462.004 & 477.592 & 481.087 & 516.157 & 787.921\\
Random solver & 394.942 & 464.391 & 395.485 & 415.701 & 428.401 & 509.642 & 769.781\\
\bottomrule
\end{tabular*}
\end{table}

\begin{table}[H]
\centering\footnotesize
\setlength{\tabcolsep}{3pt}
\renewcommand{\arraystretch}{1.12}
\caption{Burgers clean error and worst observed cross-attack outcome. MSE columns use $10^{-6}$ units. The worst attack is Full L3 for every row. Bold marks the minimum of each MSE column within its cohort; relative growth is $100\Delta\mathrm{MSE}/\mathrm{MSE}_{\rm clean}$ and is not the final-error ranking. Random solver has the lowest worst-case ID error, while Full L3 has the lowest worst-case OOD error. These results do not establish universal dominance on ID inputs.}\label{tab:revision-burgers-cross-attack-summary}
\begin{tabular*}{\textwidth}{@{\extracolsep{\fill}}llrrrrr@{}}
\toprule
Cohort & Model & Clean MSE & Worst final & $\Delta$MSE & Growth (\%) & $n$\\
\midrule
ID & Baseline & 102.387 & 272.726 & 170.339 & 166.37 & 50\\
ID & L1 & 1.069 & 20.598 & 19.529 & 1826.58 & 50\\
ID & L2 & 3.392 & 30.397 & 27.005 & 796.06 & 50\\
ID & L3 Full & 6.472 & 13.798 & 7.326 & 113.21 & 50\\
ID & L3 Stop & 2.582 & 28.737 & 26.155 & 1013.06 & 50\\
ID & Random solver & \textbf{0.639} & \textbf{6.801} & \textbf{6.162} & 964.50 & 50\\
OOD & Baseline & 893.953 & 1457.609 & 563.655 & 63.05 & 50\\
OOD & L1 & 506.481 & 872.893 & 366.411 & 72.34 & 50\\
OOD & L2 & 553.798 & 934.593 & 380.796 & 68.76 & 50\\
OOD & L3 Full & \textbf{137.595} & \textbf{272.252} & \textbf{134.657} & 97.86 & 50\\
OOD & L3 Stop & 462.785 & 787.921 & 325.136 & 70.26 & 50\\
OOD & Random solver & 397.699 & 769.781 & 372.082 & 93.56 & 50\\
\bottomrule
\end{tabular*}
\end{table}

\begin{multicols}{2}
\subsection*{Loss-Landscape Diagnostics for Projected Attack Optimization}
The following two mechanism probes directly support the
landscape interpretation behind the projected attack optimizers.  The claim is
not that we exhaustively scan the full nonconvex loss landscape.  Instead, we
reuse saved attack endpoints and traces, evaluate the true solver-integrated
Loss 3 at diagnostic points, and ask whether the optimizer-relevant high-loss
region is broad and connected or narrow and path-dependent.

For a fixed clean input \(x\), define the true attack objective
\[
  L(\delta)
  =
  \|\mathcal{F}_{\theta}(x+\delta)-\mathcal{G}(x+\delta)\|_q ,
  \qquad
  \|\delta\|_p\le \epsilon .
\]
The add-style and replace-style updates can be viewed schematically as
\[
  \delta^{\mathrm{add}}_{k+1}
  =
  \Pi_{\|\delta\|\le\epsilon}(\delta_k+\alpha d_k),
  \qquad
  \delta^{\mathrm{rep}}_{k+1}
  =
  \Pi_{\|\delta\|\le\epsilon}(\epsilon d_k),
\]
where \(d_k\) is the chosen ascent direction and \(\Pi\) denotes projection onto
the feasible perturbation set.  Thus add-style methods follow a path toward the
boundary, while replace-style methods repeatedly choose a new full-budget
boundary point.  The cleanest Burgers--Navier--Stokes contrast comes from the
two probes in \tabref{tab:loss-landscape-diagnostic-definitions}, with the
measured values summarized in \tabref{tab:loss-landscape-mechanism-results}.
\end{multicols}

\begin{table}[H]
\centering
\footnotesize
\setlength{\tabcolsep}{3pt}
\renewcommand{\arraystretch}{1.15}
\caption{The two loss-landscape diagnostics retained as mechanism evidence.}
\label{tab:loss-landscape-diagnostic-definitions}
\begin{tabular*}{\textwidth}{@{\extracolsep{\fill}}L{0.20\textwidth}L{0.34\textwidth}L{0.36\textwidth}@{}}
\toprule
Diagnostic & Computed quantity & Mechanism meaning \\
\midrule
Boundary arc &
For two final \(L_2\)-boundary perturbations \(\delta_a,\delta_b\), evaluate
\(\delta(s)=\epsilon ((1-s)\delta_a+s\delta_b)/
\|(1-s)\delta_a+s\delta_b\|_2\), \(s\in[0,1]\), and measure true Loss 3 along
the arc. &
This directly asks whether the add and replace endpoints lie on a shared
high-loss ridge.  If the arc stays high, the landscape near those endpoints is
broad and connected; if the arc drops, the endpoints are in different basins. \\
\midrule
Boundary/endpoint continuation &
After an optimizer reaches the feasible boundary, measure whether continued
optimization increases or decreases true Loss 3.  Around a replace endpoint,
also measure whether short continuation runs return to the high-loss region. &
This tests whether reaching the boundary is enough.  A forgiving landscape lets
replace land inside a basin that remains high; a narrow landscape requires the
path to keep correcting along the boundary. \\
\bottomrule
\end{tabular*}
\end{table}

\begin{table}[H]
\centering
\footnotesize
\setlength{\tabcolsep}{3pt}
\renewcommand{\arraystretch}{1.18}
\caption{The main Burgers--Navier--Stokes loss-landscape evidence.  Values are
computed by evaluating true Loss 3 at diagnostic points from saved attack runs.}
\label{tab:loss-landscape-mechanism-results}
\begin{tabular*}{\textwidth}{@{\extracolsep{\fill}}L{0.18\textwidth}L{0.32\textwidth}L{0.32\textwidth}L{0.12\textwidth}@{}}
\toprule
Probe & Burgers result & Navier--Stokes result & Reading \\
\midrule
Boundary arc between replace and add endpoints &
Steepest replace endpoint \(7.069\), steepest add endpoint \(6.339\).  The
minimum sampled arc loss is \(6.172\), or about \(97.4\%\) of the weaker
endpoint; all \(17/17\) sampled arc points exceed \(0.95\) of the weaker
endpoint. &
In the \(N=3\) exact probe, steepest replace is \(85.09\), steepest add is
\(255.48\), and only \(1/9\) arc points reach \(95\%\) of the add endpoint.  In
the \(N=2\) top-up probe, the corresponding values are \(114.54\), \(337.49\),
and \(2/9\). &
Burgers broad; NS narrow. \\
\midrule
Boundary/endpoint continuation &
Near the steepest-replace endpoint, endpoint-cap probes have
\(p(L\ge0.95L_{\max})\approx0.4\)--\(0.6\).  Continuing optimization from the
replace endpoint gives final mean \(L/L_{\max}=1.044\), with \(p_{0.95}=0.800\). &
After first reaching the boundary, steepest replace has mean gain \(-51.05\) in
the \(N=3\) exact probe and \(-59.43\) in the \(N=2\) top-up, while steepest add
has \(+43.43\) and \(+52.80\).  Continuing from the replace endpoint reaches
only mean \(L/L_{\max}=0.700\), with \(p_{0.95}=0\). &
Burgers forgiving; NS path-dependent. \\
\bottomrule
\end{tabular*}
\end{table}

\begin{multicols}{2}
These two probes are the most direct evidence for the mechanism.  In Burgers,
the add and replace endpoints sit on a connected high-loss ridge, and a replace
endpoint remains inside a basin that can be continued to high loss.  In
recurrent Navier--Stokes, the replace endpoint is far below the add endpoint,
the connecting arc mostly stays below the add-level loss, and reaching the
boundary is not enough: add improves along the boundary while replace often
loses value.  Thus the projected-optimizer difference is best interpreted as a
difference in optimizer-relevant landscape geometry: broad and forgiving for
Burgers, narrow and path-dependent for recurrent Navier--Stokes.  Darcy Flow is
discrete rather than continuous, and its main role is auxiliary: its binary
flip-set behavior is consistent with a forgiving high-loss region, but the
cleanest landscape contrast is the Burgers--Navier--Stokes comparison above.
\end{multicols}

\begingroup
\captionsetup[figure]{font=small,skip=2pt}
\setlength{\intextsep}{5pt}
\setlength{\textfloatsep}{5pt}
\setlength{\floatsep}{5pt}
\begin{figure}[H]
  \centering
  \includegraphics[width=0.98\textwidth]{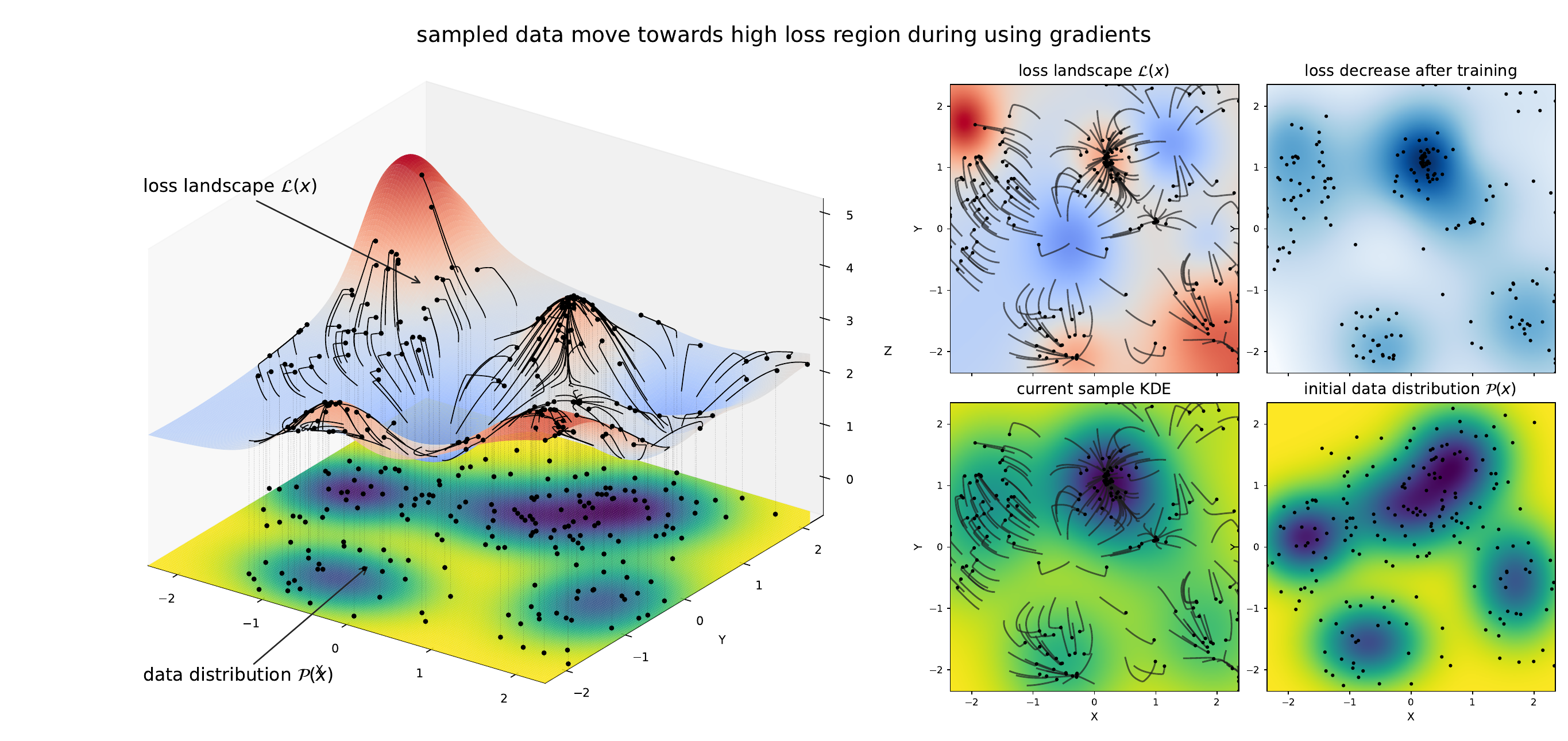}
  \caption{Low-dimensional PGD visualization of high-dimensional
  function-space sampling. Samples drawn from the dataset distribution
  \(\mathcal{P}(x)\) move along gradients of the loss landscape
  \(\mathcal{L}(x)\), blending data density with loss information and
  concentrating samples in high-loss regions.}
  \label{fig:loss-landscape-pgd-visualization}
\end{figure}
\endgroup

\begingroup
\captionsetup[figure]{font=small,skip=2pt}
\setlength{\intextsep}{5pt}
\setlength{\textfloatsep}{5pt}
\setlength{\floatsep}{5pt}
\subsection*{General Attack and Training}

\begin{figure}[H]
  \centering
  \includegraphics[width=0.94\textwidth]{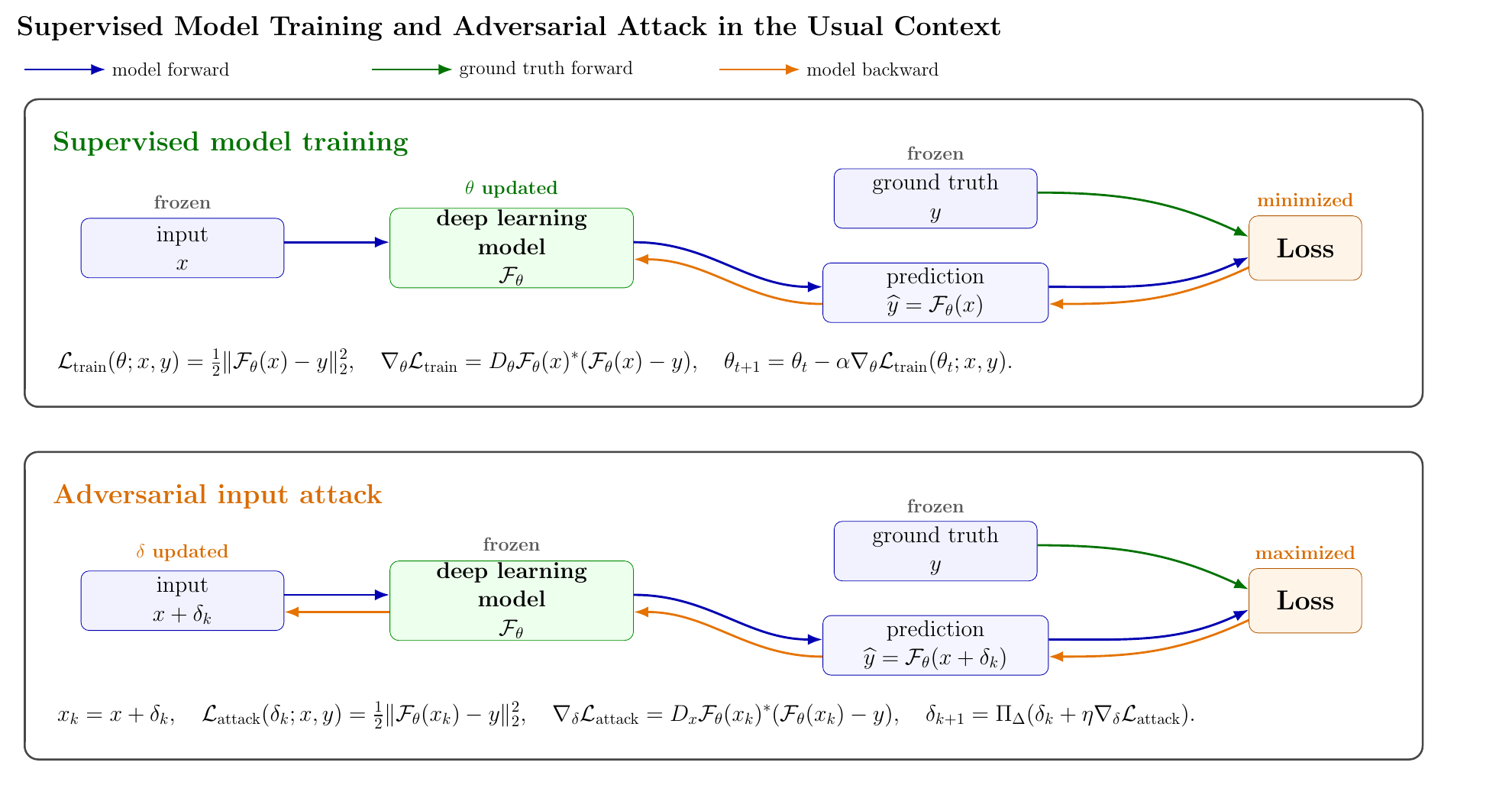}
  \caption{Training versus attack gradients. Training minimizes the loss by
  updating model parameters, whereas an adversarial attack maximizes the loss by
  updating the input perturbation.}
  \label{fig:training-vs-attack-gradients}
\end{figure}

\subsection*{PDE Operator Learning Attack and Training}

\begin{figure}[H]
  \centering
  \includegraphics[width=0.94\textwidth]{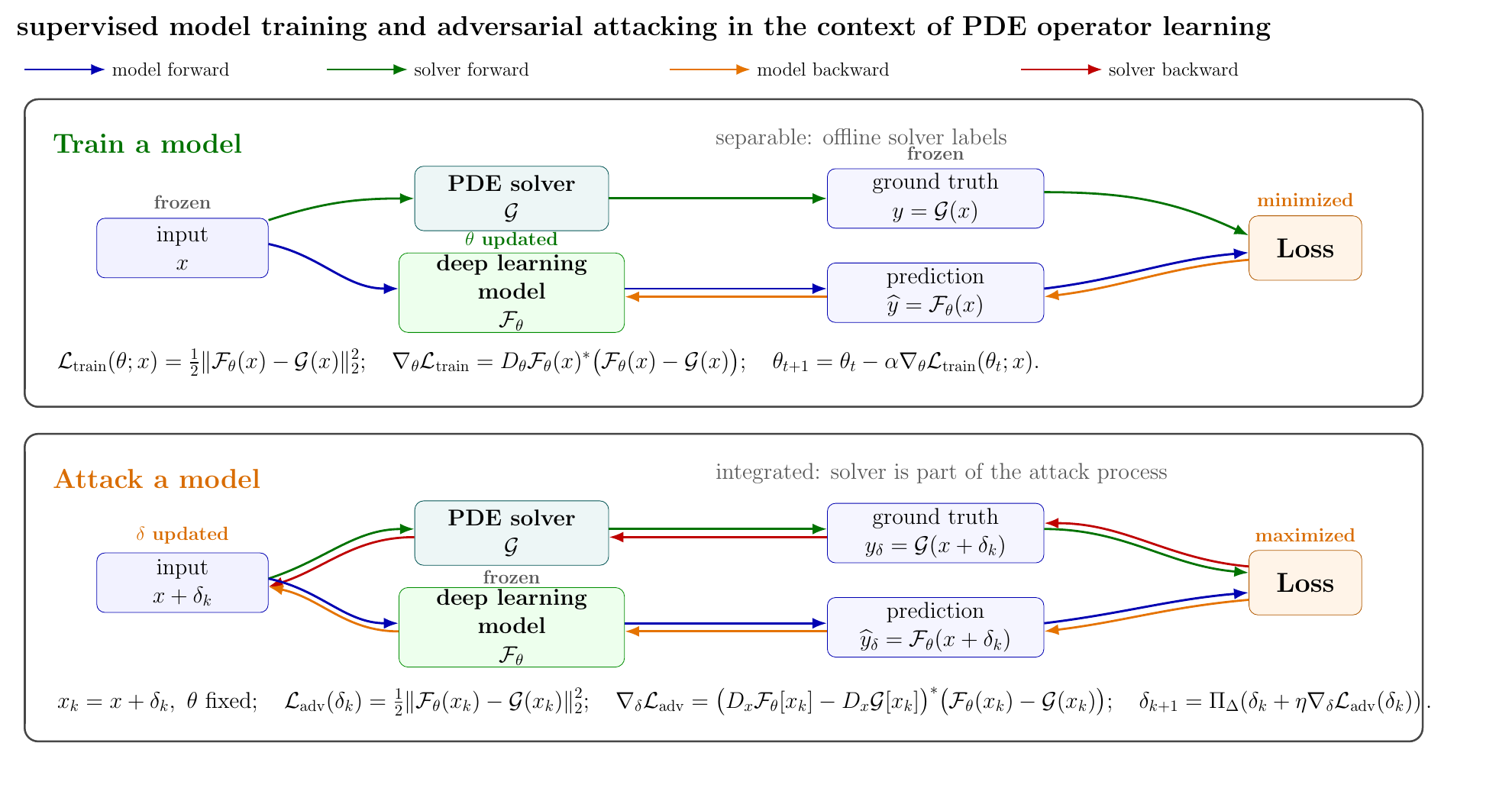}
  \caption{Supervised model training and adversarial attacking in PDE operator
  learning. The numerical solver maps the same input used by the model to the
  corresponding solver output, so it plays the role of a ground-truth oracle.}
  \label{fig:pde-solver-training-attack}
\end{figure}

\begin{figure}[H]
  \centering
  \includegraphics[width=0.94\textwidth]{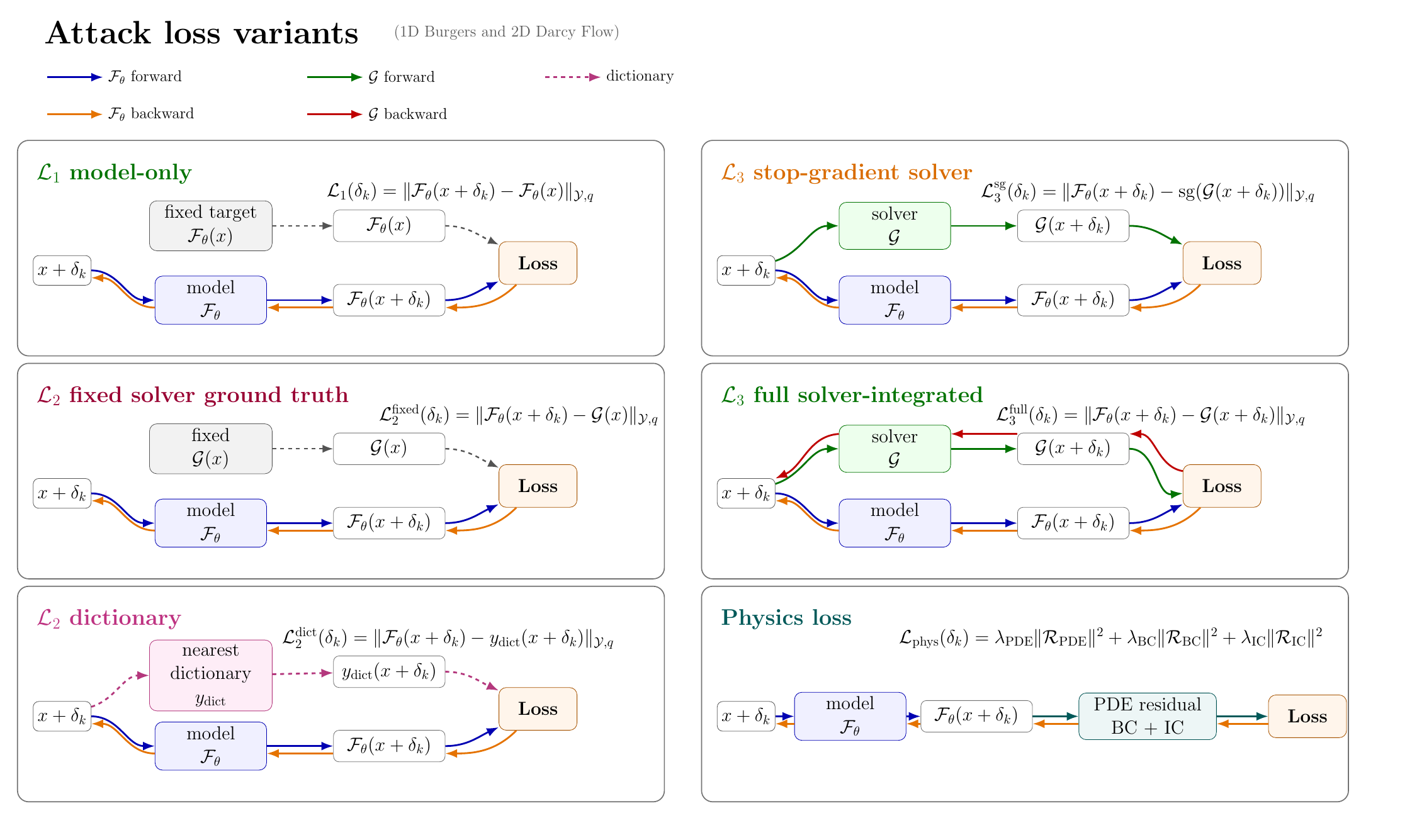}
  \caption{Attack-loss variants arranged by solver integration. The variants
  range from model-only objectives, through fixed or dictionary targets, to
  stop-gradient and fully solver-integrated losses.}
  \label{fig:attack-loss-variants-twocol}
\end{figure}

\begin{figure}[H]
  \centering
  \includegraphics[width=0.94\textwidth]{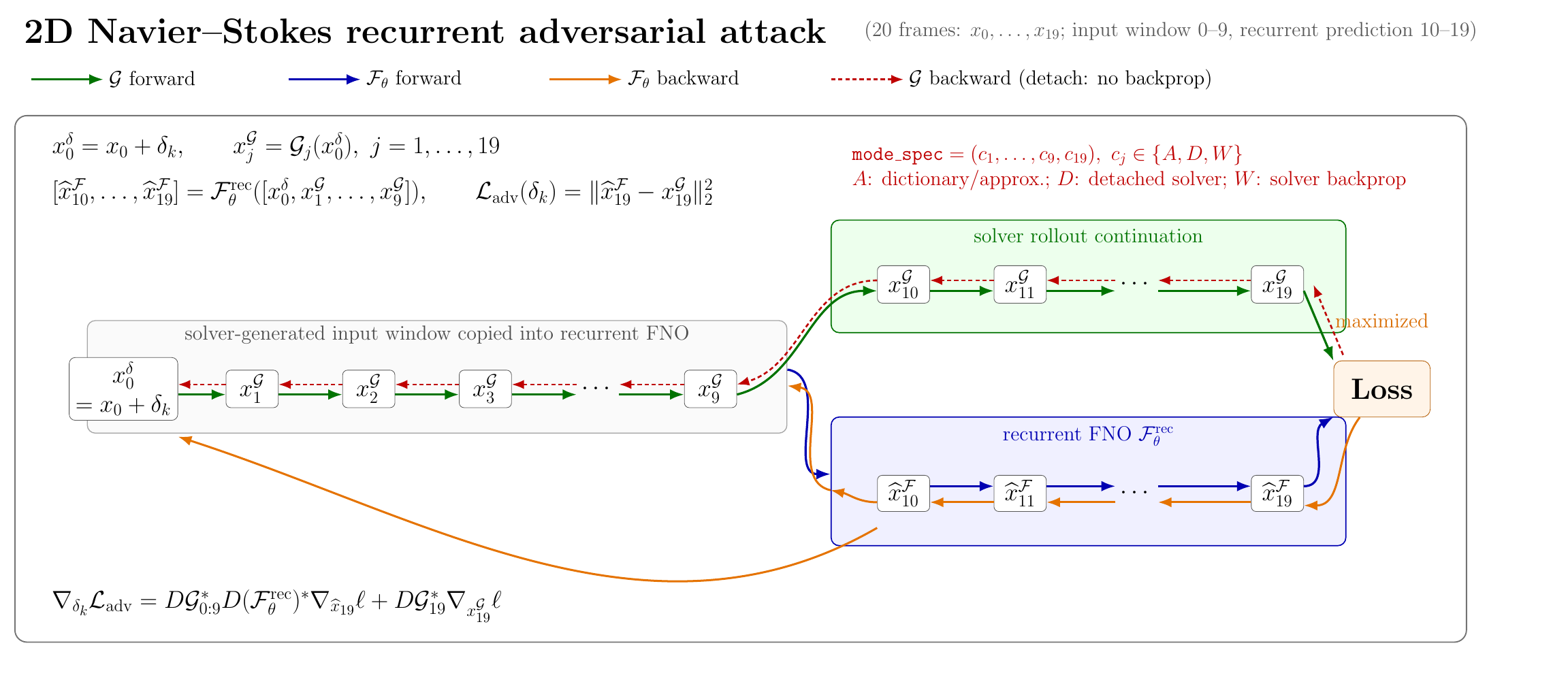}
  \caption{Recurrent Navier--Stokes adversarial attack flow. The attack updates
  the initial vorticity input while the recurrent solver rollout supplies the
  target trajectory used in the loss.}
  \label{fig:ns-recurrent-adversarial-attack}
\end{figure}
\endgroup

\begingroup
\captionsetup[figure]{font=small,skip=2pt}
\setlength{\intextsep}{5pt}
\setlength{\textfloatsep}{5pt}
\setlength{\floatsep}{5pt}
\nextappendixfigureheading{

\appendixsubsection{Solver-Integrated Adversarial Attack}

\appendixdomainheading{Burgers}

\noindent\textbf{Burgers: FNO final prediction and perturbation fields.}
}

\begin{figure}[H]
  \useappendixfigureheading
  \centering
  \includegraphics[width=1.02\textwidth,height=0.455\textheight,keepaspectratio]{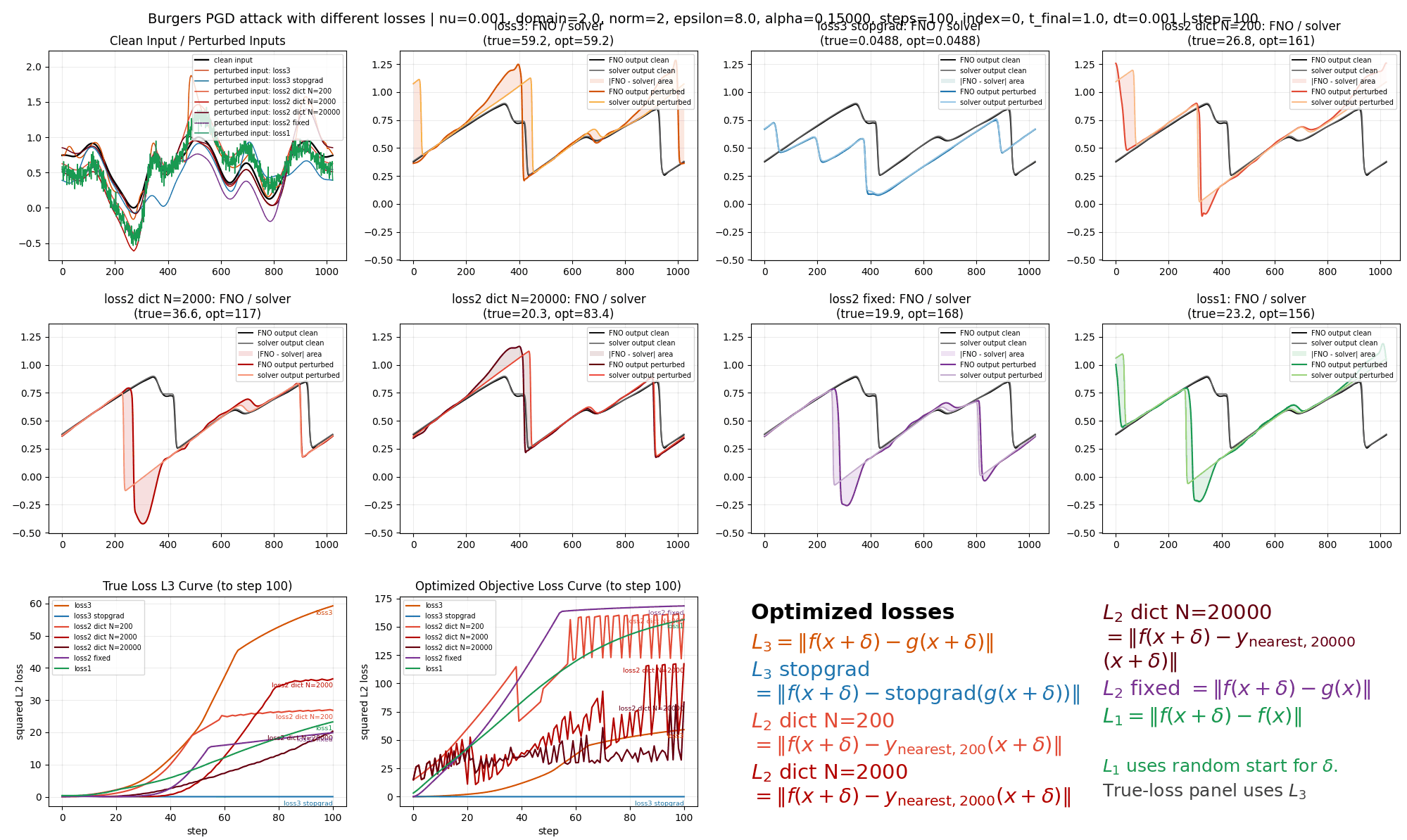}
  \caption{Burgers FNO $L_2$ PGD attack variants, including final perturbed inputs,
  model/solver predictions, residuals, and loss curves. Only optimizing Loss 3
  maximizes the true model--solver loss; optimizing the other losses often does
  not maximize the true discrepancy between the solver and the model.}
  \label{fig:app-g-burgers-fno-l2-final}
\end{figure}

\begin{figure}[H]
  \useappendixfigureheading
  \centering
  \includegraphics[width=1.02\textwidth,height=0.455\textheight,keepaspectratio]{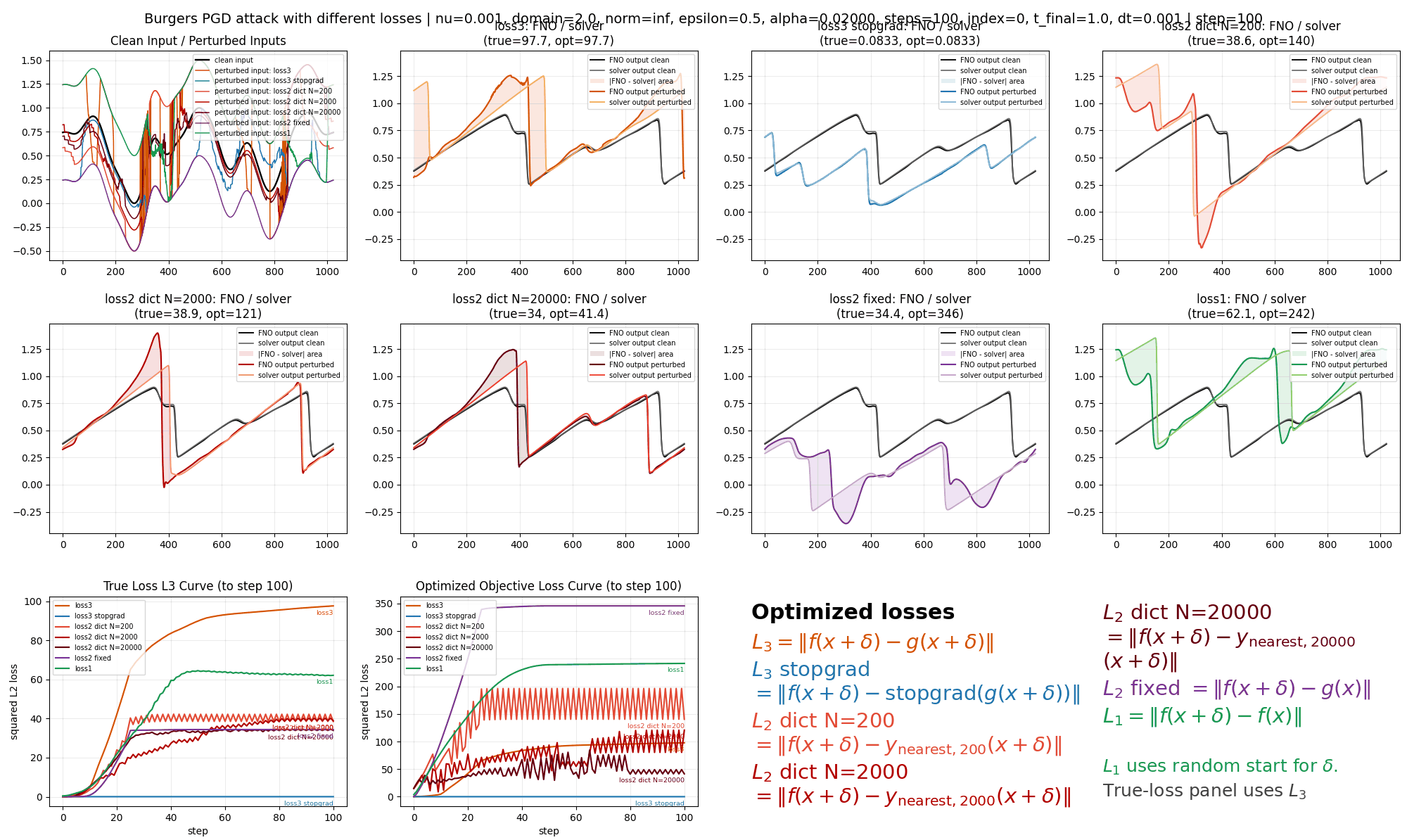}
  \caption{Burgers FNO $L_\infty$ PGD attack variants, including final perturbed
  inputs, model/solver predictions, residuals, and loss curves.}
  \label{fig:app-g-burgers-fno-linf-final}
\end{figure}

\nextappendixfigureheading{
\noindent\textbf{Burgers: FNO loss progression.}
}

\begin{figure}[H]
  \useappendixfigureheading
  \centering
  \includegraphics[width=0.96\textwidth,height=0.82\textheight,keepaspectratio]{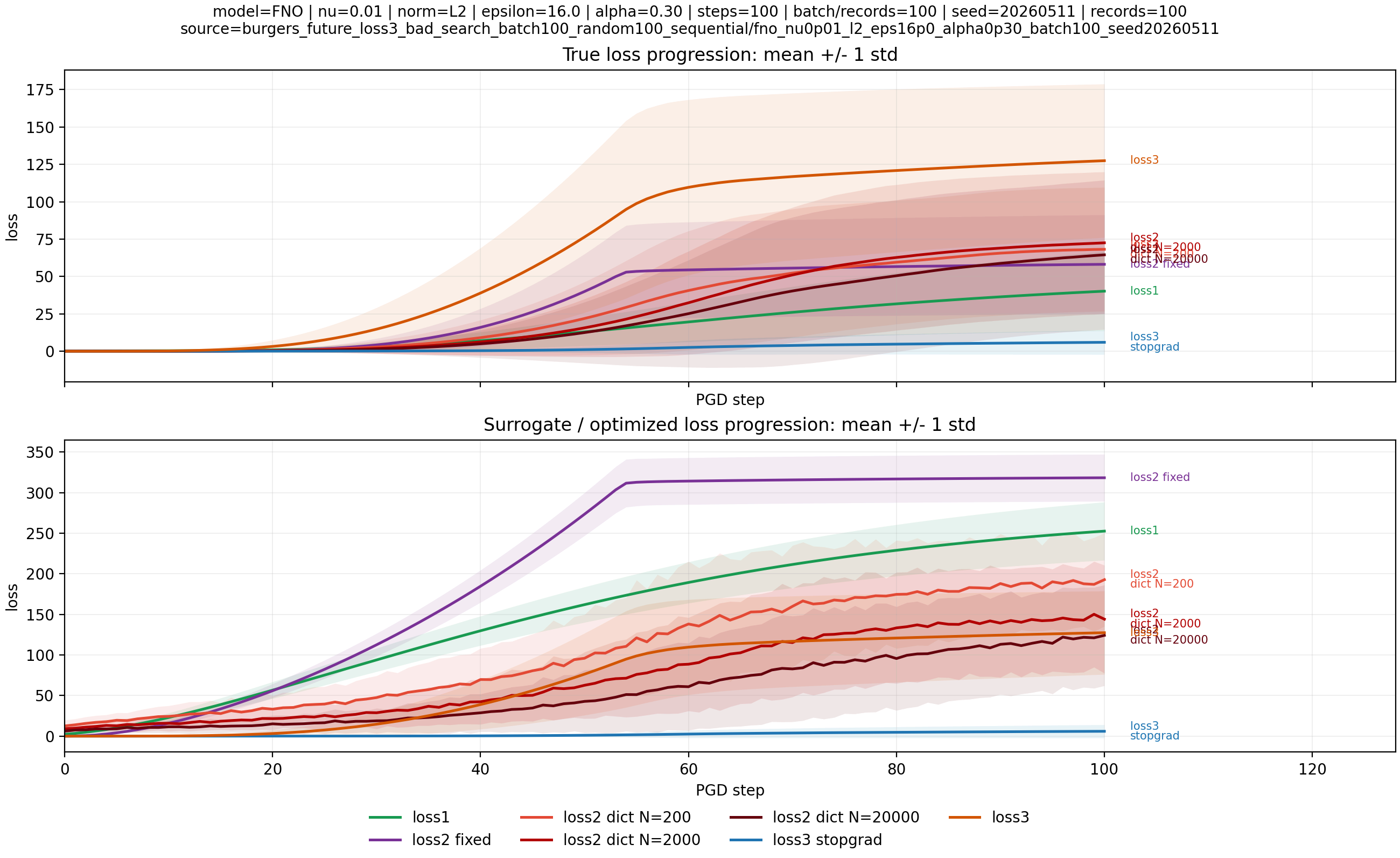}
  \caption{Burgers FNO $L_2$ attack loss progression with mean and standard-deviation
  bands across PGD steps.}
  \label{fig:app-g-burgers-fno-l2-loss}
\end{figure}

\begin{figure}[H]
  \useappendixfigureheading
  \centering
  \includegraphics[width=0.96\textwidth,height=0.82\textheight,keepaspectratio]{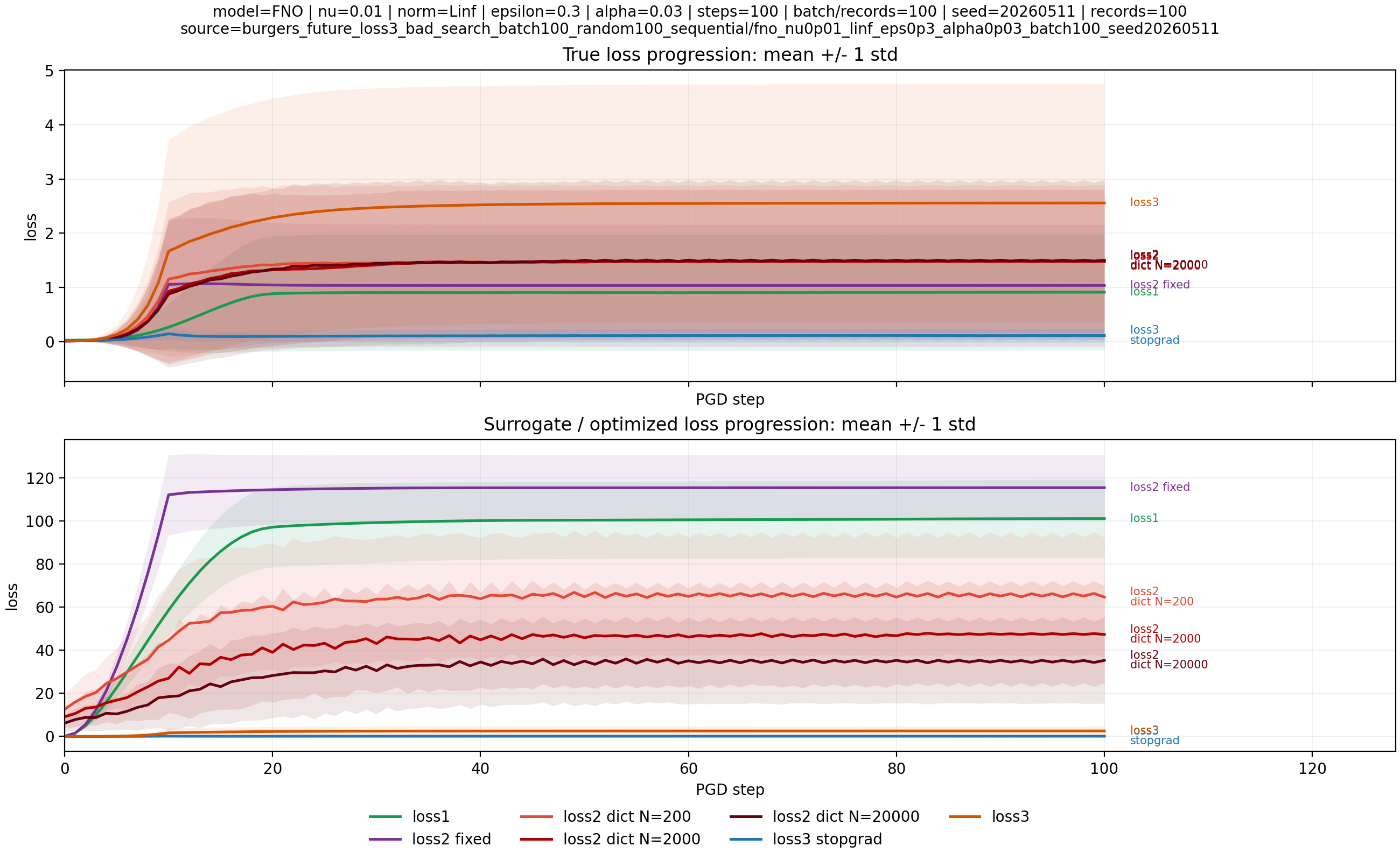}
  \caption{Burgers FNO $L_\infty$ attack loss progression with mean and
  standard-deviation bands at $\epsilon=0.3$ and $\alpha=0.03$.}
  \label{fig:app-g-burgers-fno-linf-eps03-loss}
\end{figure}

\FloatBarrier
\nextappendixfigureheading{
\noindent\textbf{Burgers: DeepONet final prediction and perturbation fields.}
}

\begin{figure}[H]
  \useappendixfigureheading
  \centering
  \includegraphics[width=0.96\textwidth,height=0.82\textheight,keepaspectratio]{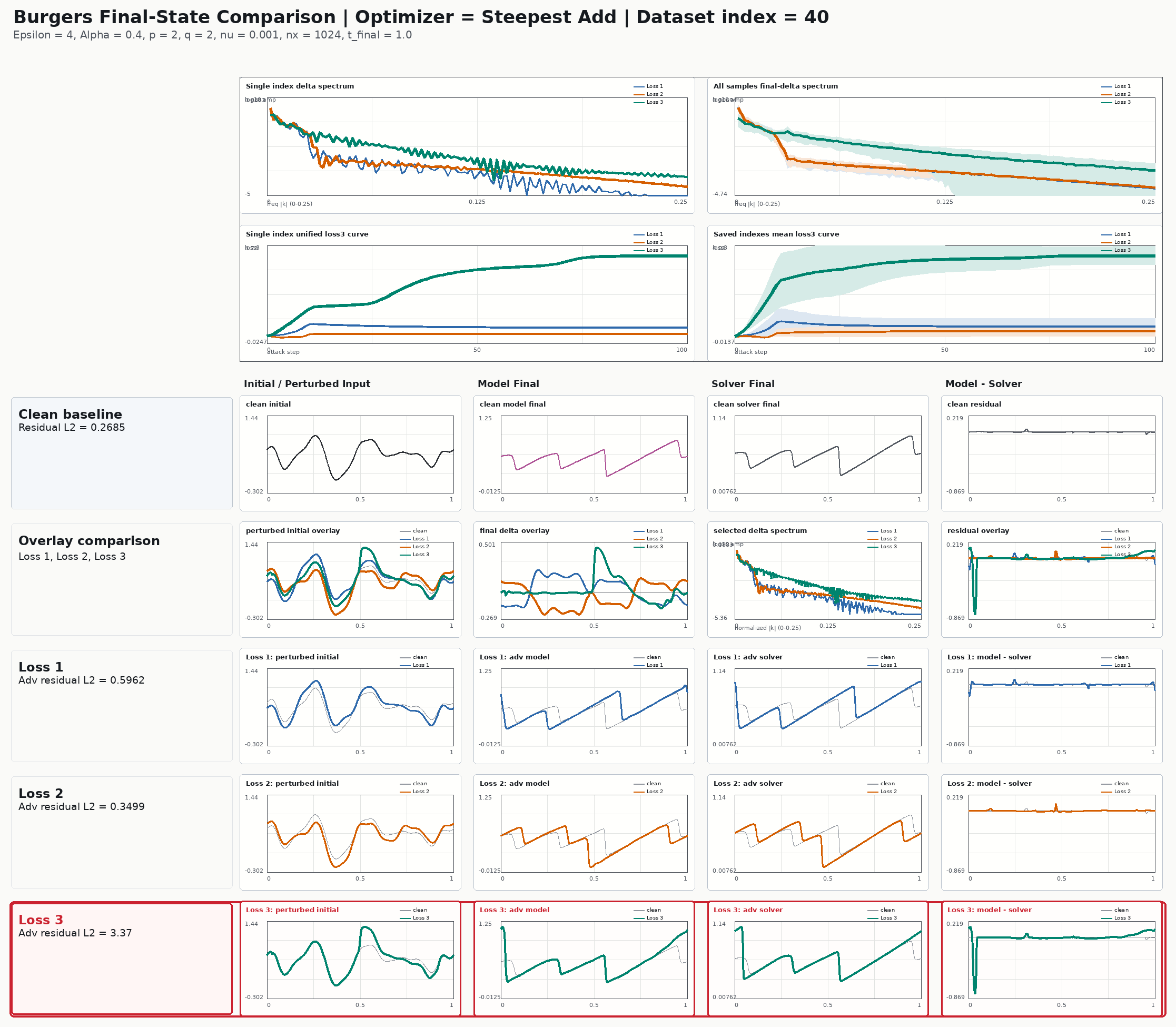}
  \caption{Burgers final-state comparison for the Steepest Add optimizer, showing the
  clean baseline, attacked inputs, final predictions, solver outputs, residuals,
  and loss-specific panels. In Burgers, the Delta perturbations optimized by
  loss 3 are often visibly higher-frequency.}
  \label{fig:app-g-burgers-steepest-add-final}
\end{figure}

\begin{figure}[H]
  \useappendixfigureheading
  \centering
  \includegraphics[width=1.02\textwidth,height=0.455\textheight,keepaspectratio]{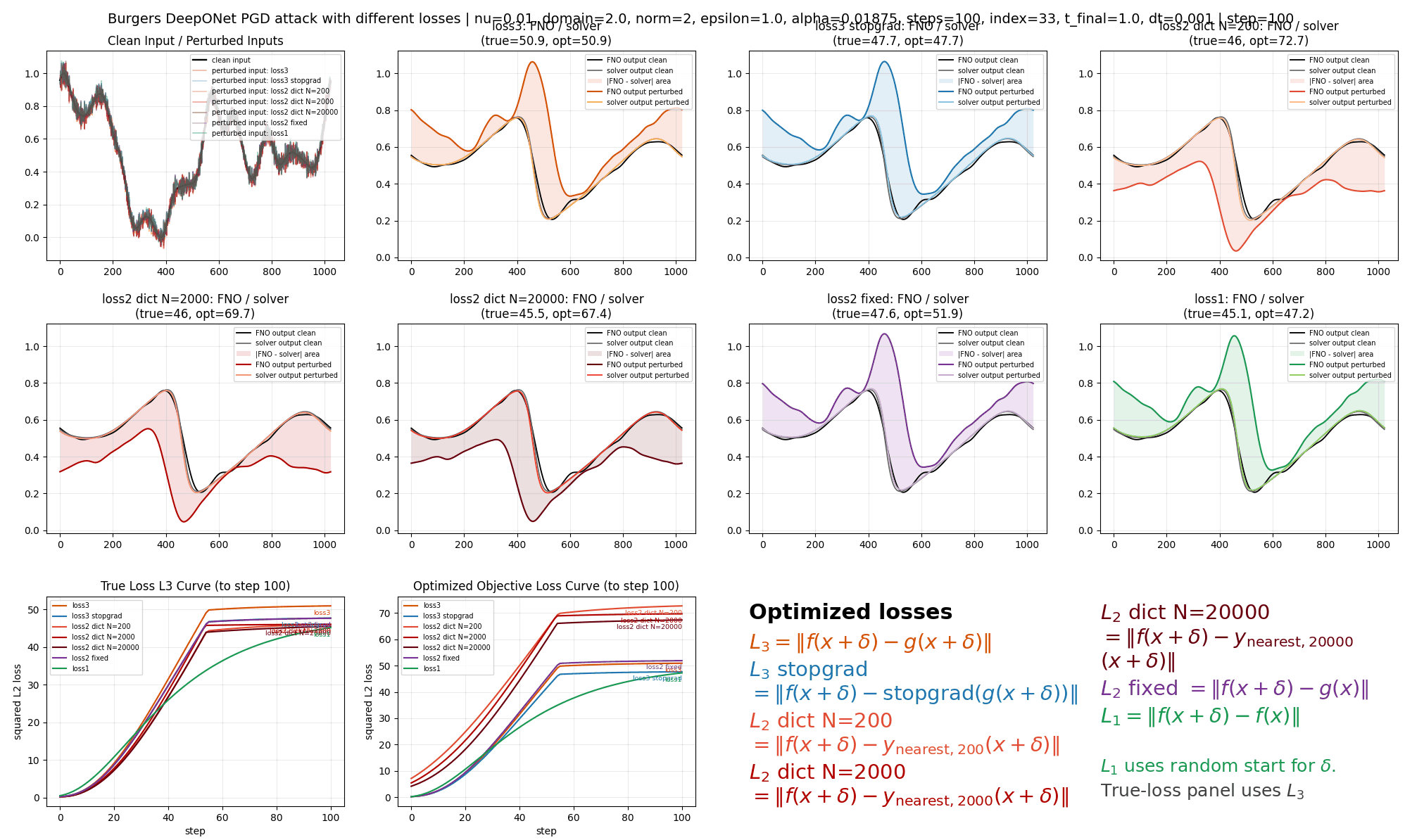}
  \caption{Burgers DeepONet $L_2$ PGD attack variants with final prediction and
  perturbation comparisons. Only optimizing Loss 3 maximizes the true
  model--solver loss; optimizing the other losses often does not maximize the
  true discrepancy between the solver and the model.}
  \label{fig:app-g-burgers-deeponet-l2-final}
\end{figure}

\begin{figure}[H]
  \useappendixfigureheading
  \centering
  \includegraphics[width=1.02\textwidth,height=0.455\textheight,keepaspectratio]{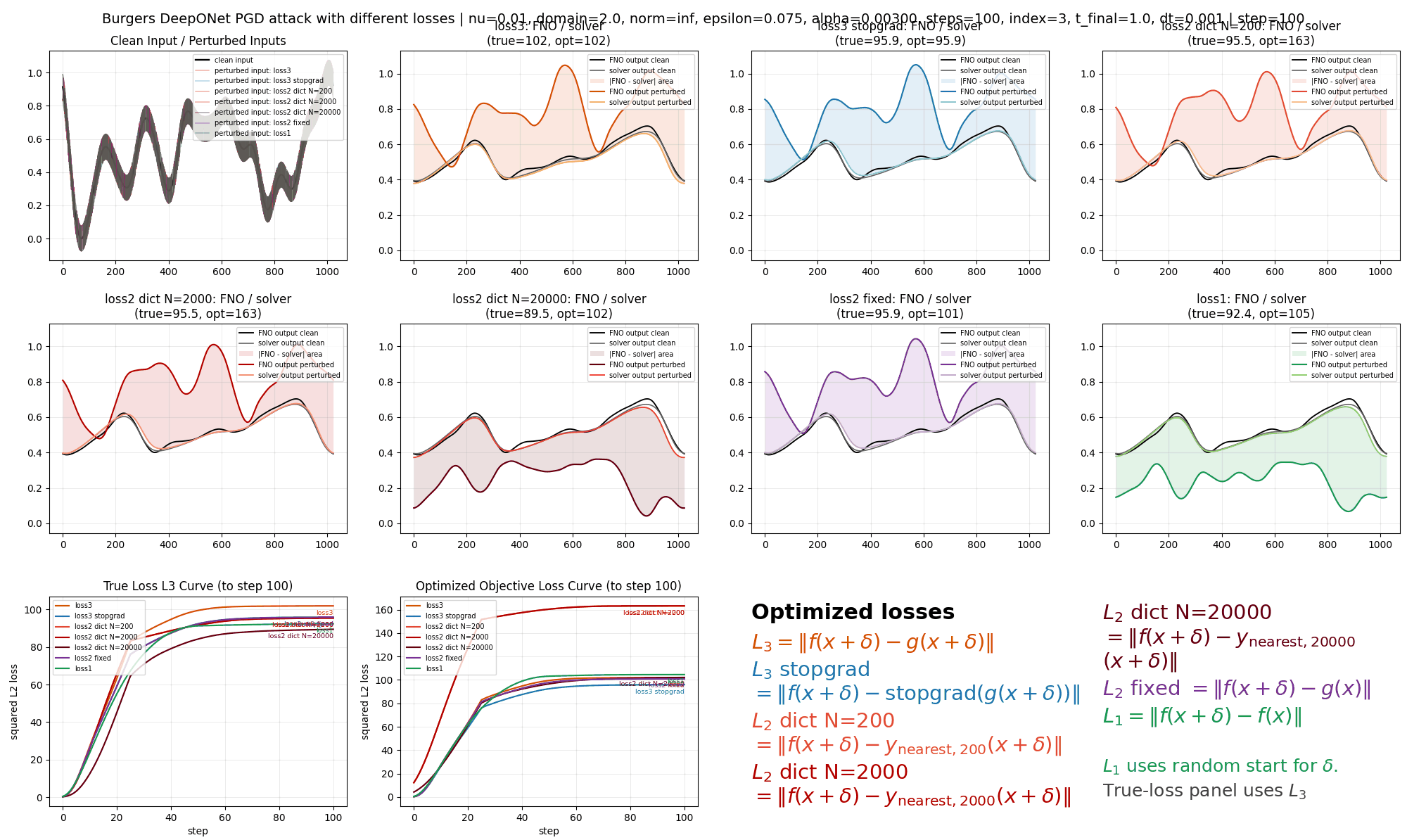}
  \caption{Burgers DeepONet $L_\infty$ PGD attack variants with final prediction and
  perturbation comparisons.}
  \label{fig:app-g-burgers-deeponet-linf-final}
\end{figure}

\nextappendixfigureheading{
\noindent\textbf{Burgers: DeepONet loss progression.}
}

\begin{figure}[H]
  \useappendixfigureheading
  \centering
  \includegraphics[width=0.96\textwidth,height=0.82\textheight,keepaspectratio]{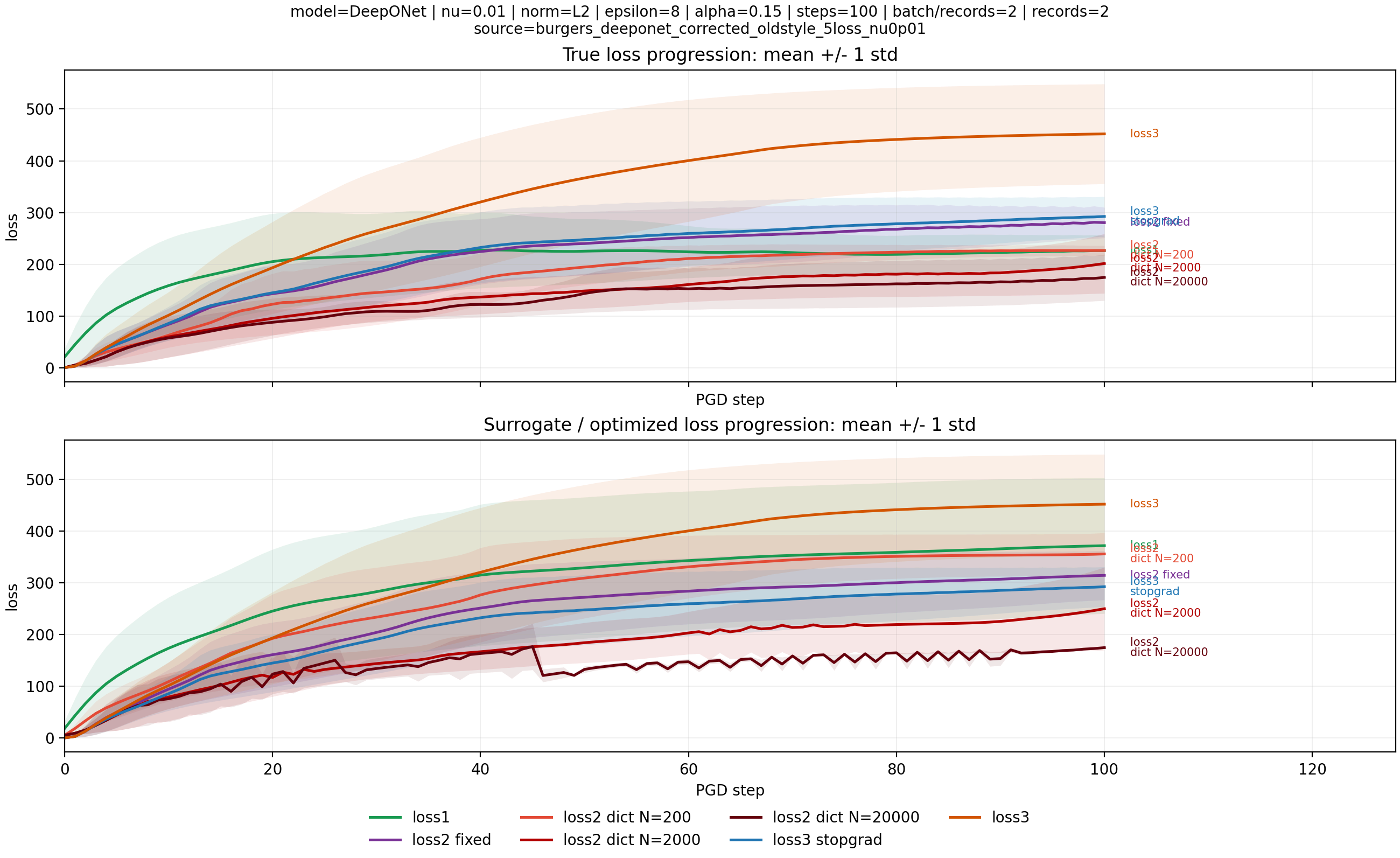}
  \caption{Burgers DeepONet $L_2$ attack loss progression with mean and
  standard-deviation bands.}
  \label{fig:app-g-burgers-deeponet-l2-loss}
\end{figure}

\begin{figure}[H]
  \useappendixfigureheading
  \centering
  \includegraphics[width=0.96\textwidth,height=0.82\textheight,keepaspectratio]{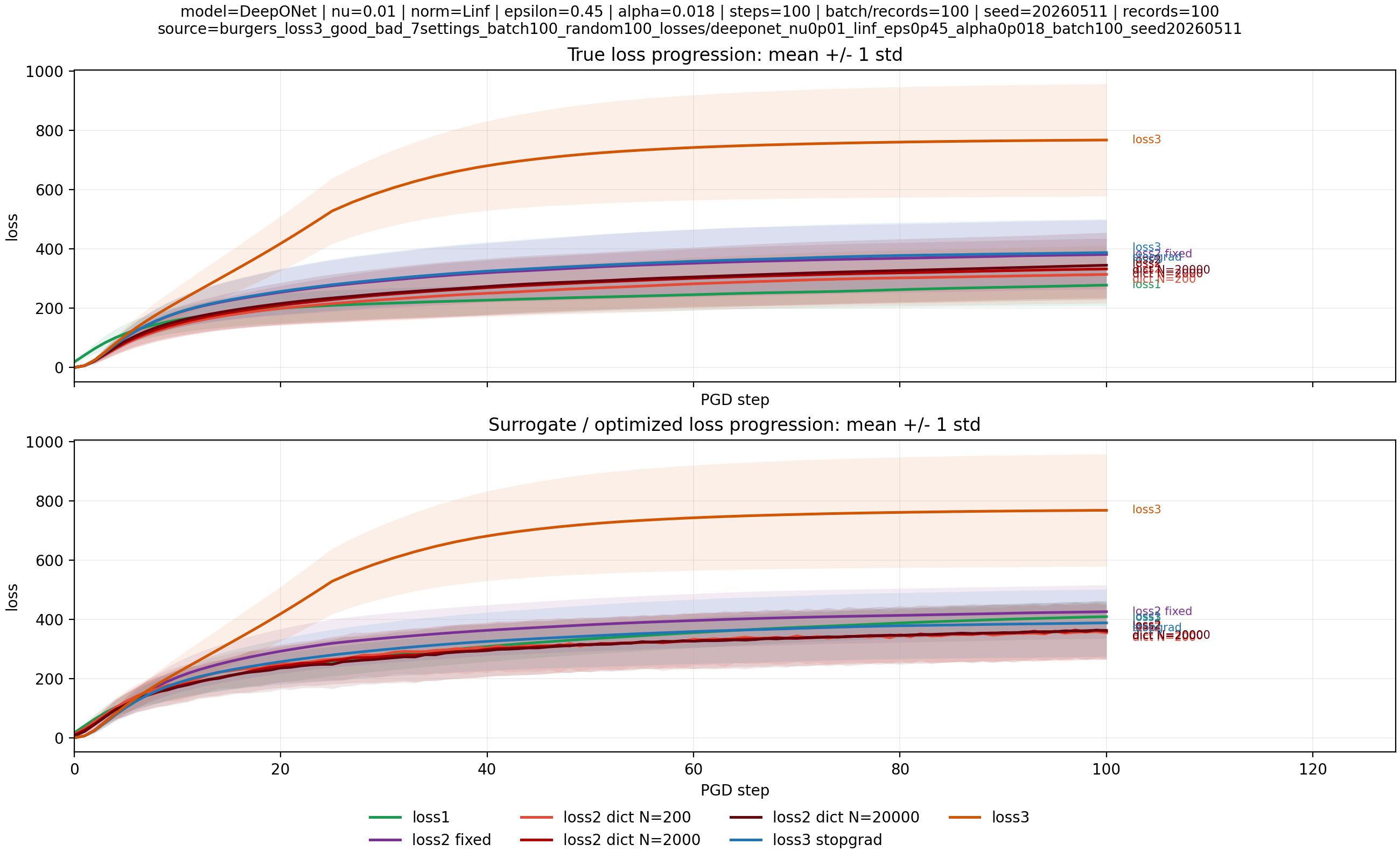}
  \caption{Burgers DeepONet $L_\infty$ attack loss progression with mean and
  standard-deviation bands at $\epsilon=0.45$.}
  \label{fig:app-g-burgers-deeponet-linf-loss}
\end{figure}

\nextappendixfigureheading{
\noindent\textbf{Burgers: optimizer and local SVD diagnostics.}

\noindent\textbf{Boundary and update diagnostics.}
}

\begin{figure}[H]
  \useappendixfigureheading
  \centering
  \includegraphics[width=0.96\textwidth,height=0.82\textheight,keepaspectratio]{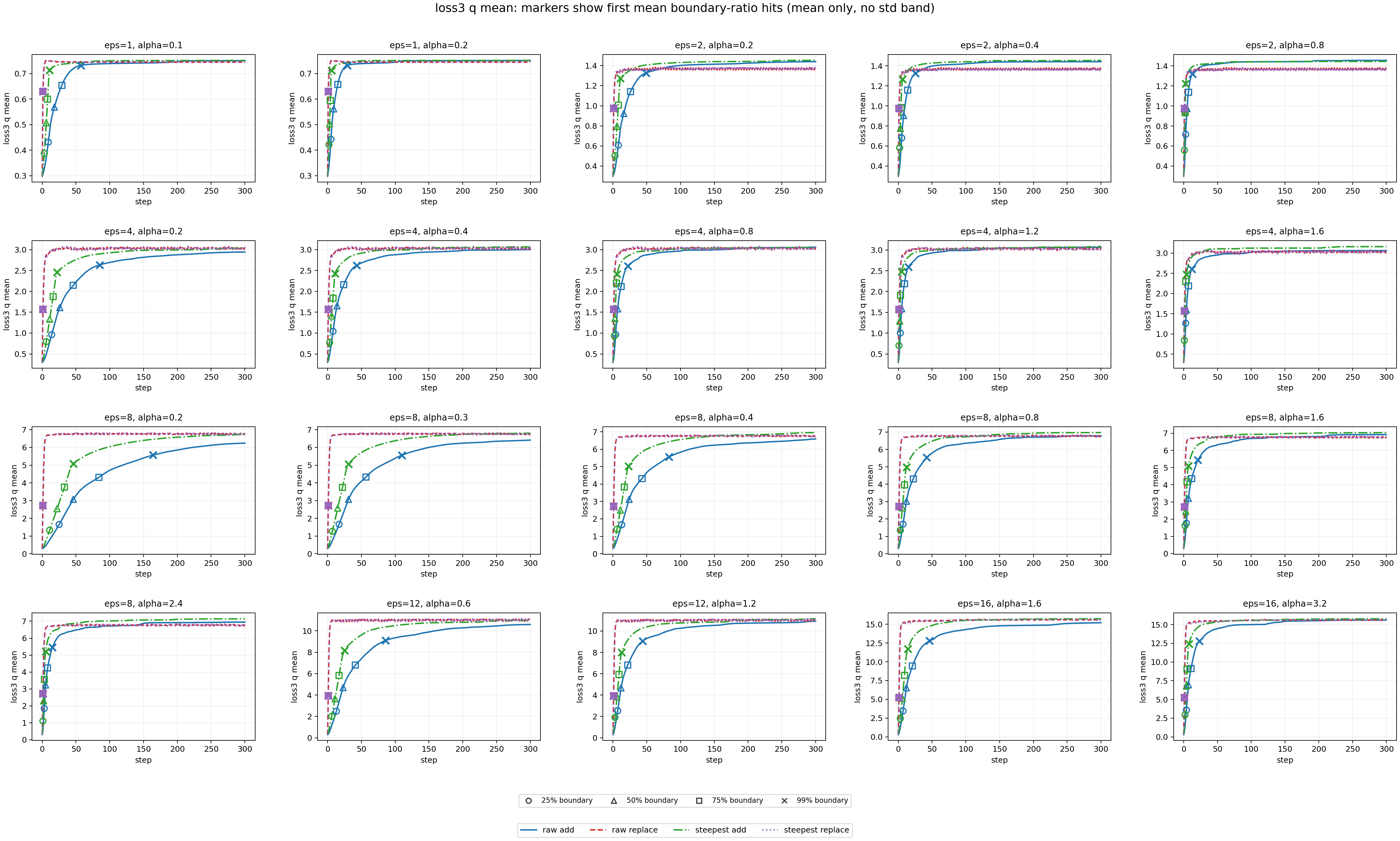}
  \caption{Burgers Loss 3 mean curves for the $p=q=2$ attack setting, with first mean
  boundary-ratio hit markers, shown without standard-deviation bands. The
  plotted quantity is the sample mean of the Loss 3 objective under output
  norm $q=2$; it is not an additional metric. In this $p=q=2$ geometry, raw
  replace and steepest replace coincide with the corresponding power-replace
  update and reach the
  perturbation boundary very quickly, whereas raw add and steepest add move
  toward the boundary step by step. In a few settings, such as
  $(\epsilon,\alpha)=(8,2.4)$, $(2,0.2)$, $(2,0.4)$, and $(2,0.8)$, raw add or
  steepest add can obtain a slightly larger final loss than the replacement
  update, but they usually do so more slowly. With small $\alpha$, the stepwise
  add methods are especially slow to reach the boundary, while the replacement
  update reaches it early, continues optimizing on the boundary for a few steps,
  and quickly reaches a strong loss.}
  \label{fig:app-g-burgers-qmean-boundary-markers}
\end{figure}

\begin{figure}[H]
  \useappendixfigureheading
  \centering
  \includegraphics[width=0.96\textwidth,height=0.82\textheight,keepaspectratio]{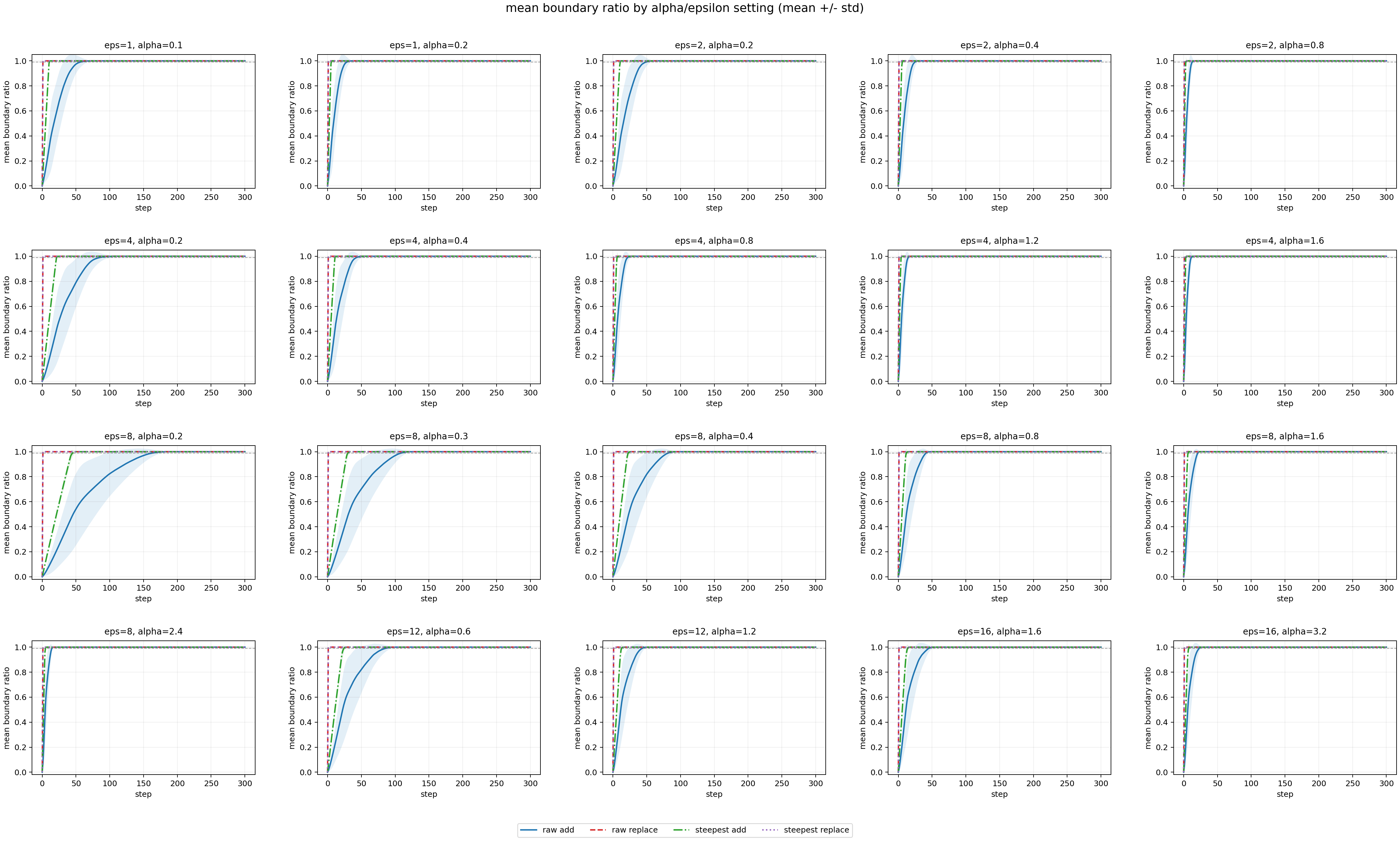}
  \caption{Burgers mean boundary-ratio trajectories by $\alpha$ and $\epsilon$ setting,
  with mean and standard-deviation bands.}
  \label{fig:app-g-burgers-boundary-ratio-grid}
\end{figure}

\begin{figure}[H]
  \useappendixfigureheading
  \centering
  \includegraphics[width=0.96\textwidth,height=0.82\textheight,keepaspectratio]{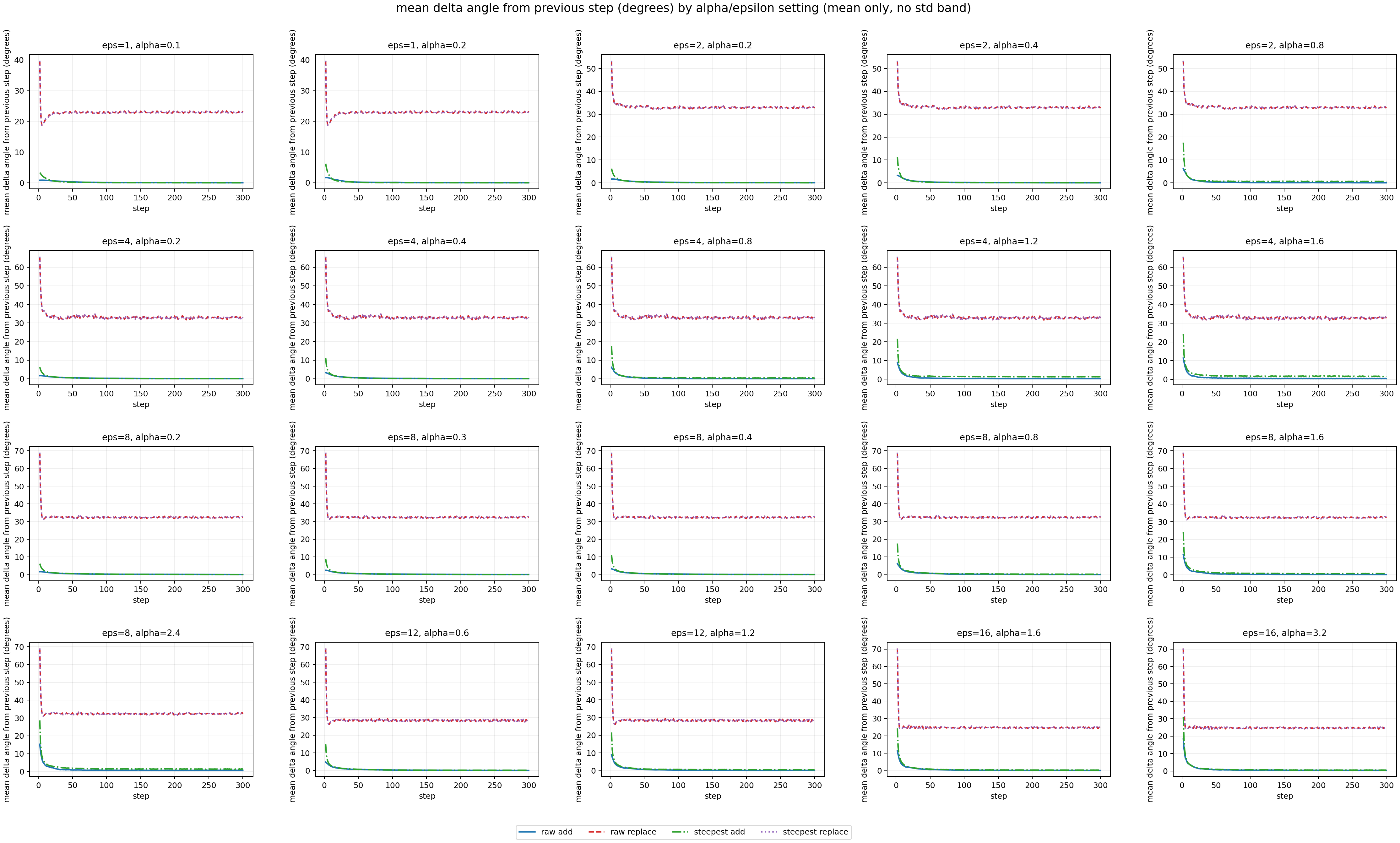}
  \caption{Burgers mean delta-angle trajectories from the previous step by $\alpha$ and
  $\epsilon$ setting. Each curve measures the angular change between the
  optimized perturbation vector $\delta$ after one attack step and the
  perturbation from the previous step. The two replace methods show much larger
  angular changes because they replace the perturbation at each step instead of
  gradually accumulating $\alpha$-scaled gradient increments; this may explain
  why replace-style updates are more effective in this Burgers setting.}
  \label{fig:app-g-burgers-delta-angle-grid}
\end{figure}

\nextappendixfigureheading{
\noindent\textbf{Local SVD singular-vector diagnostic.}
}

\begin{figure}[H]
  \useappendixfigureheading
  \centering
  \includegraphics[width=0.96\textwidth,height=0.82\textheight,keepaspectratio]{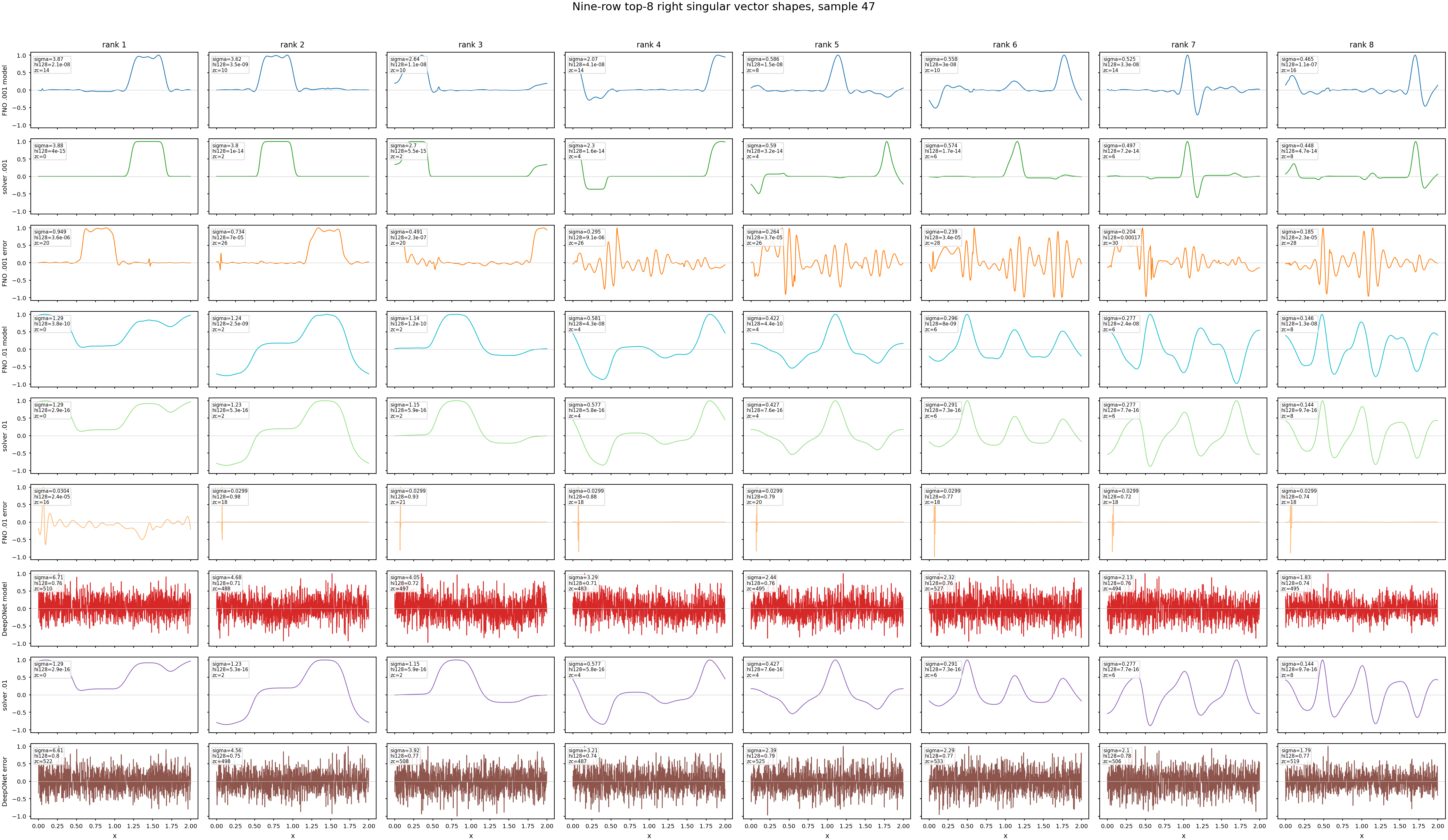}
  \caption{Local Burgers SVD diagnostic at sample 47, comparing the top eight right
  singular-vector shapes of trained models and the solver. FNO right singular
  vectors closely track the solver, while DeepONet right singular vectors remain
  much less aligned, indicating a mismatch in local model--solver behavior.}
  \label{fig:app-g-burgers-local-svd-right-vectors}
\end{figure}

\FloatBarrier
\nextappendixfigureheading{
\appendixdomainheading{Darcy Flow}

\noindent\textbf{Final prediction and perturbation fields.}
}

\begin{figure}[H]
  \useappendixfigureheading
  \centering
  \includegraphics[width=0.96\textwidth,height=0.82\textheight,keepaspectratio]{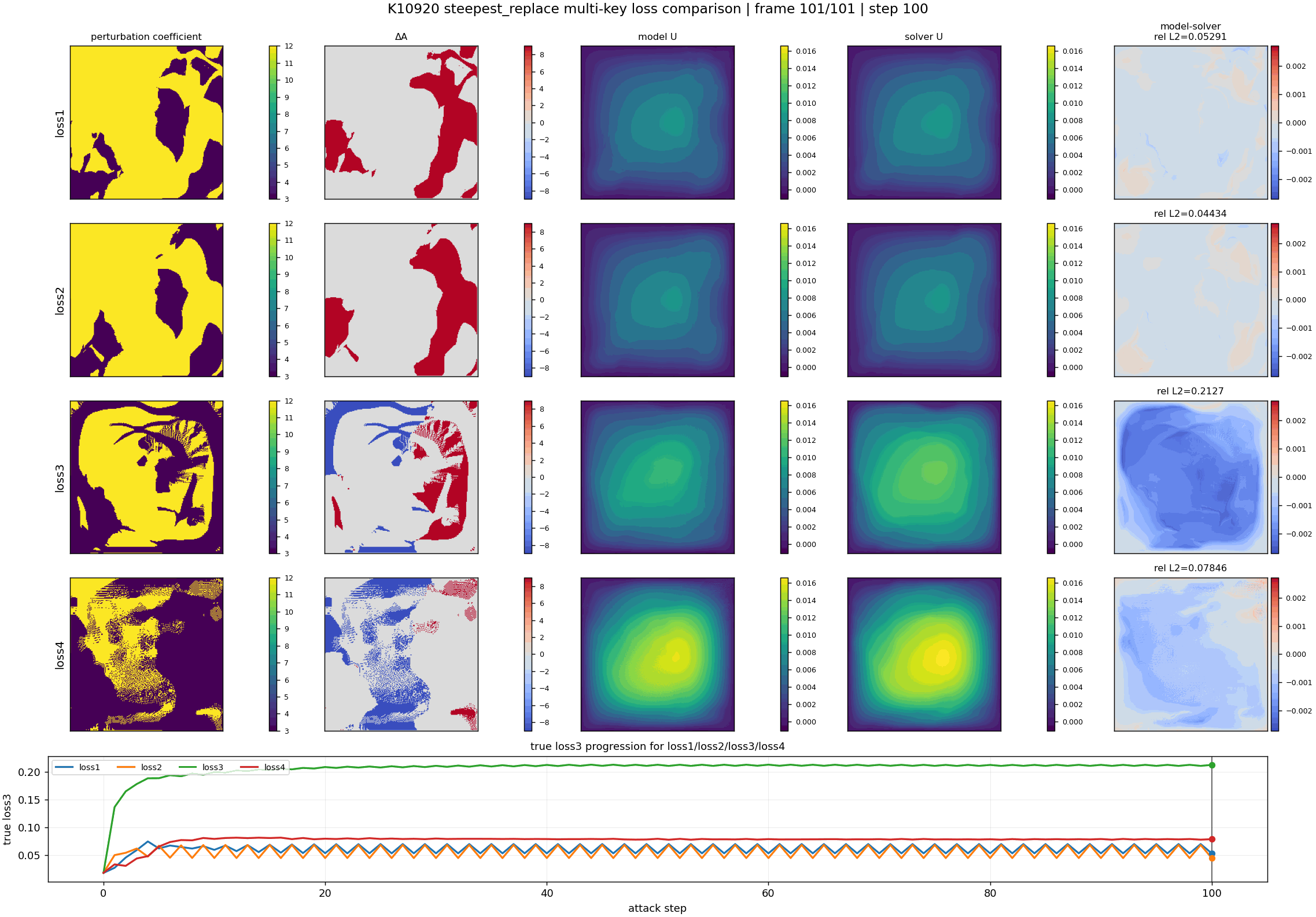}
  \caption{Darcy Flow multi-key loss comparison with coefficient perturbations, model
  predictions, solver solutions, residuals, and loss progression. Optimizing
  Loss 3 produces a larger model--solver output discrepancy than optimizing the
  other losses. Here Loss 4 denotes the physics-residual loss,
  $\mathcal{L}_{\mathrm{phys}}$.}
  \label{fig:app-g-darcy-multikey-final}
\end{figure}

\nextappendixfigureheading{
\noindent\textbf{Loss progression and optimizer comparison.}
}

\begin{figure}[H]
  \useappendixfigureheading
  \centering
  \captionsetup{font=small,skip=1pt}
  \includegraphics[width=0.99\textwidth,height=0.38\textheight,keepaspectratio]{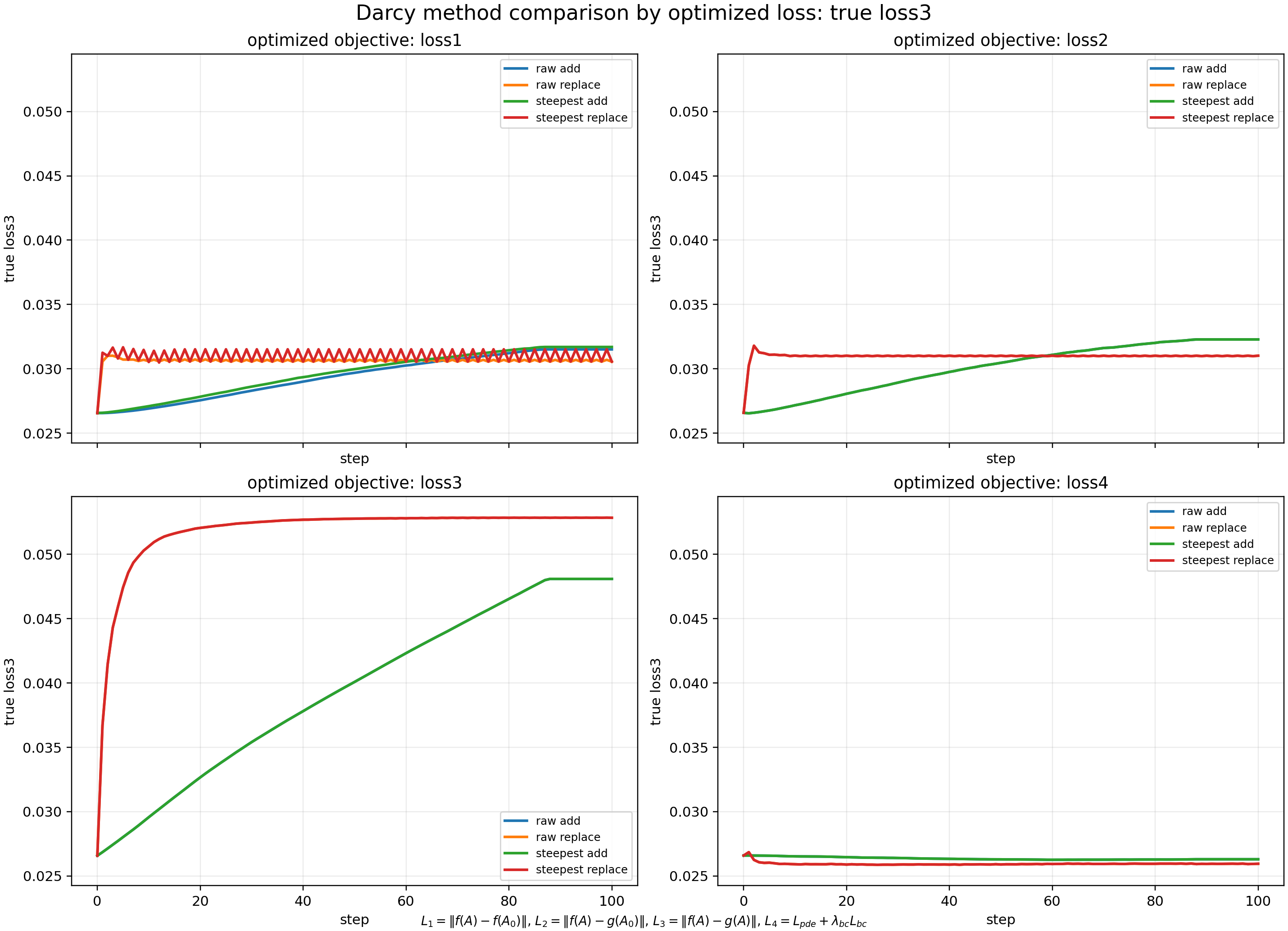}
  \caption{Darcy Flow optimizer-method comparison by optimized objective, evaluated
  with the true loss-3 metric. The replace-style optimizer converges very
  quickly and shows a practical advantage in this Darcy Flow benchmark. Here
  loss 4 denotes the physics-residual loss, $\mathcal{L}_{\mathrm{phys}}$.}
  \label{fig:app-g-darcy-optimizer-loss}
\end{figure}

\FloatBarrier
\nextappendixfigureheading{
\appendixdomainheading{Navier--Stokes}

\noindent\textbf{Periodic-boundary evidence and recurrent attack diagnostics.}
}

\begin{figure}[H]
  \useappendixfigureheading
  \centering
  \captionsetup{font=small,skip=1pt}
  \includegraphics[width=0.99\textwidth,height=0.38\textheight,keepaspectratio]{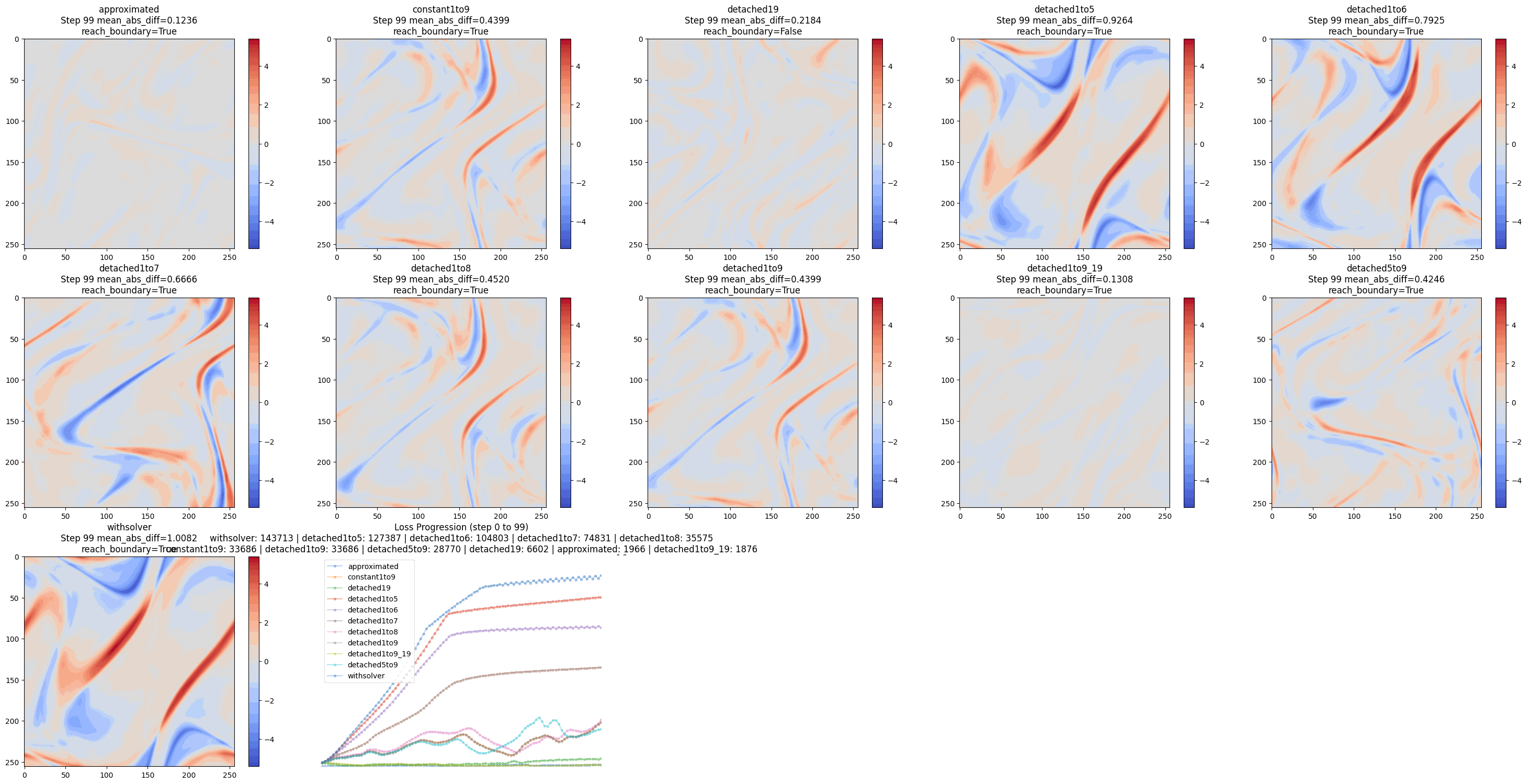}
  \caption{Navier--Stokes recurrent model-check heatmaps and loss progression. This
  diagnostic tests variants with reduced solver integration, implemented by
  detaching or approximating key intermediate frames in the recurrent solver
  process. Reducing solver integration often prevents the optimized attack from
  reaching the maximum achievable loss.}
  \label{fig:app-g-ns-recurrent-heatmaps}
\end{figure}

\begin{figure}[H]
  \useappendixfigureheading
  \centering
  \captionsetup{font=footnotesize,skip=1pt}
  \includegraphics[width=1.03\textwidth,height=0.88\textheight,keepaspectratio]{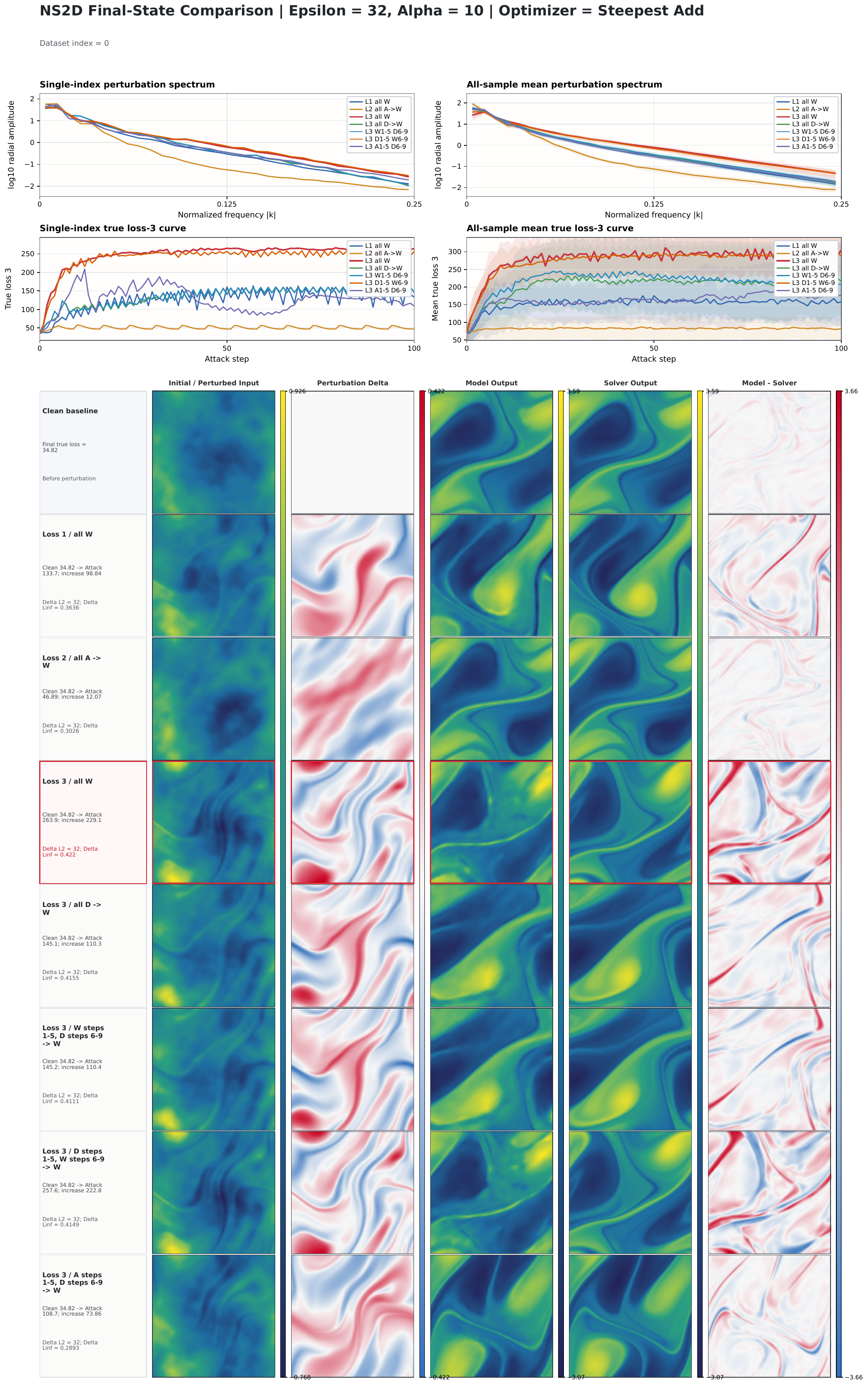}
  \caption{Navier--Stokes final-state comparison for the Steepest Add attack at
  $\epsilon=32$ and $\alpha=10$. The rows compare loss 1, loss 2, loss 3, and
  several loss-3 backpropagation variants, including detached or substituted
  gradient paths. We clearly observe that loss 3 generates relatively
  higher-frequency perturbations, while the perturbations generated by loss 2
  remain much lower-frequency.}
  \label{fig:app-g-ns-steepest-add-final}
\end{figure}

\begin{figure}[H]
  \useappendixfigureheading
  \centering
  \captionsetup{font=footnotesize,skip=1pt}
  \includegraphics[width=0.99\textwidth,height=0.82\textheight,keepaspectratio]{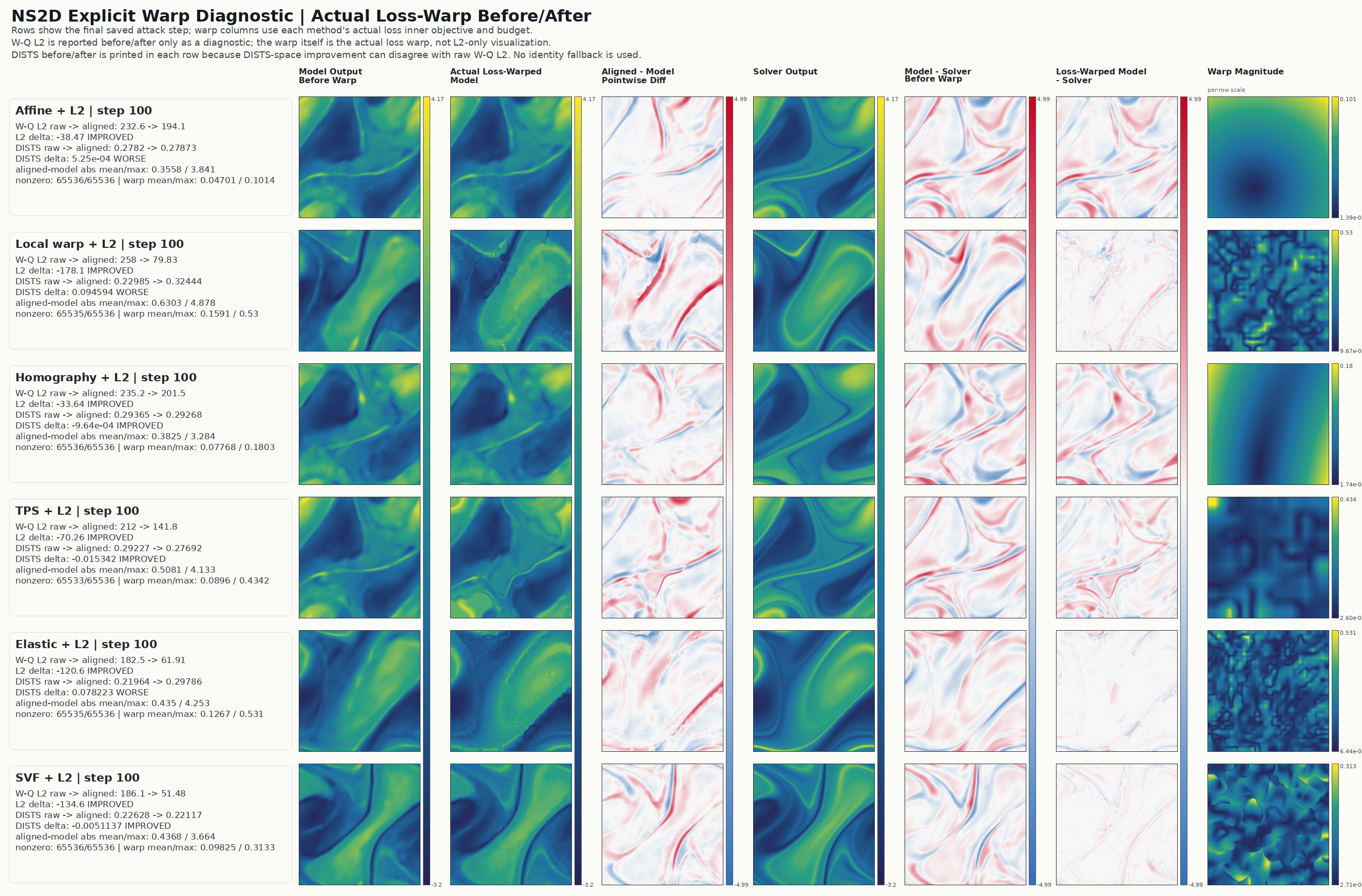}
  \caption{Navier--Stokes explicit-warp diagnostic, comparing the attack-loss behavior when
  the model and solver outputs are evaluated with and without the warp.}
  \label{fig:app-g-ns-explicit-warp}
\end{figure}

\begin{figure}[H]
  \useappendixfigureheading
  \centering
  \captionsetup{font=footnotesize,skip=1pt}
  \includegraphics[width=0.99\textwidth,height=0.82\textheight,keepaspectratio]{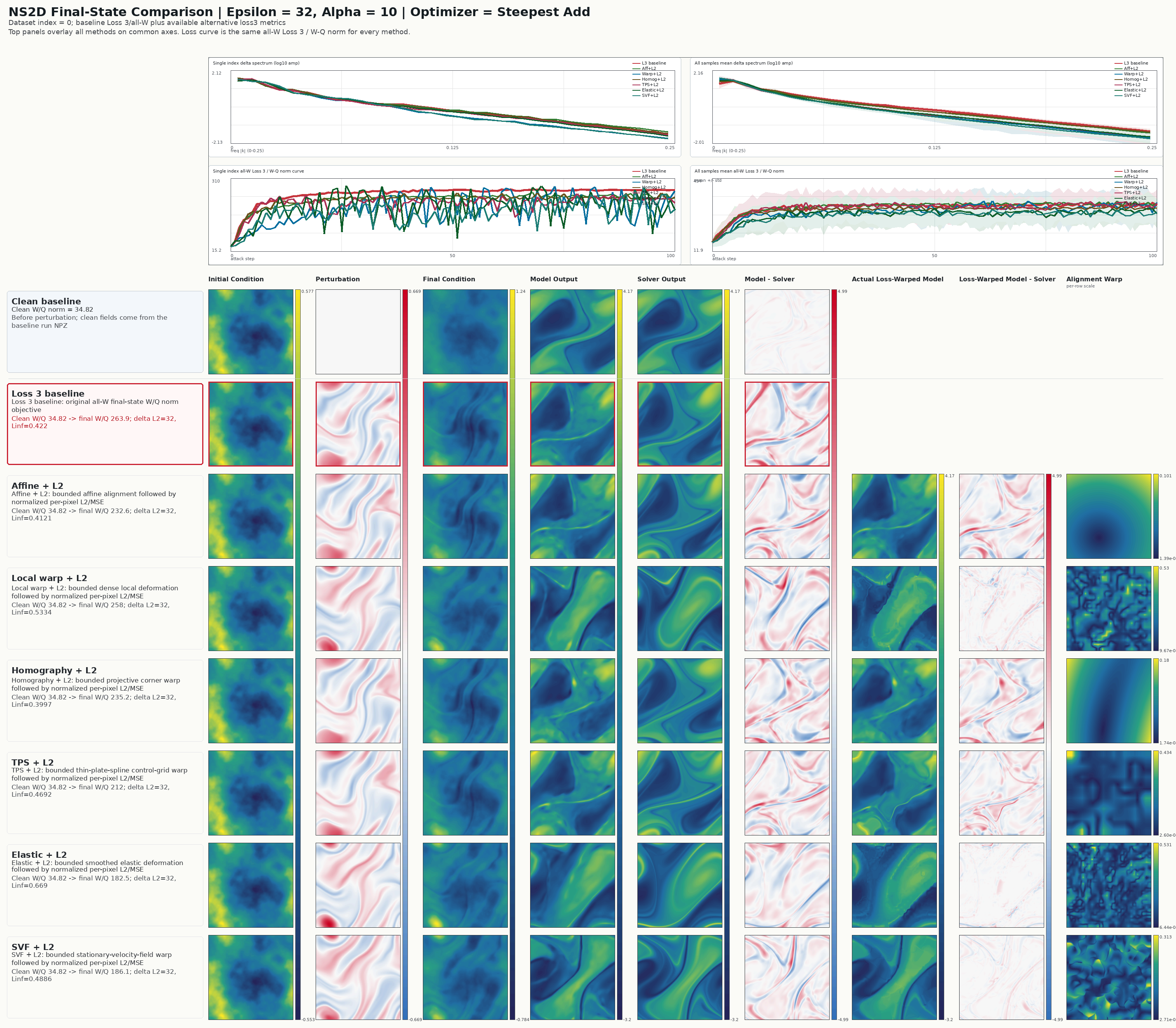}
  \caption{Navier--Stokes final-state comparison across warp methods at $\epsilon=32$
  and $\alpha=10$, showing how warp choices change the model--solver output
  mismatch after the attack. We tested several warp variants that deform the
  model and solver outputs toward each other, motivated by the possibility that
  their textures may be similar but shifted or locally warped. These variants
  did not produce a distinctive outcome, and the final perturbations remain
  qualitatively similar.}
  \label{fig:app-g-ns-warp-methods}
\end{figure}

\begin{figure}[H]
  \useappendixfigureheading
  \centering
  \includegraphics[width=0.96\textwidth,height=0.82\textheight,keepaspectratio]{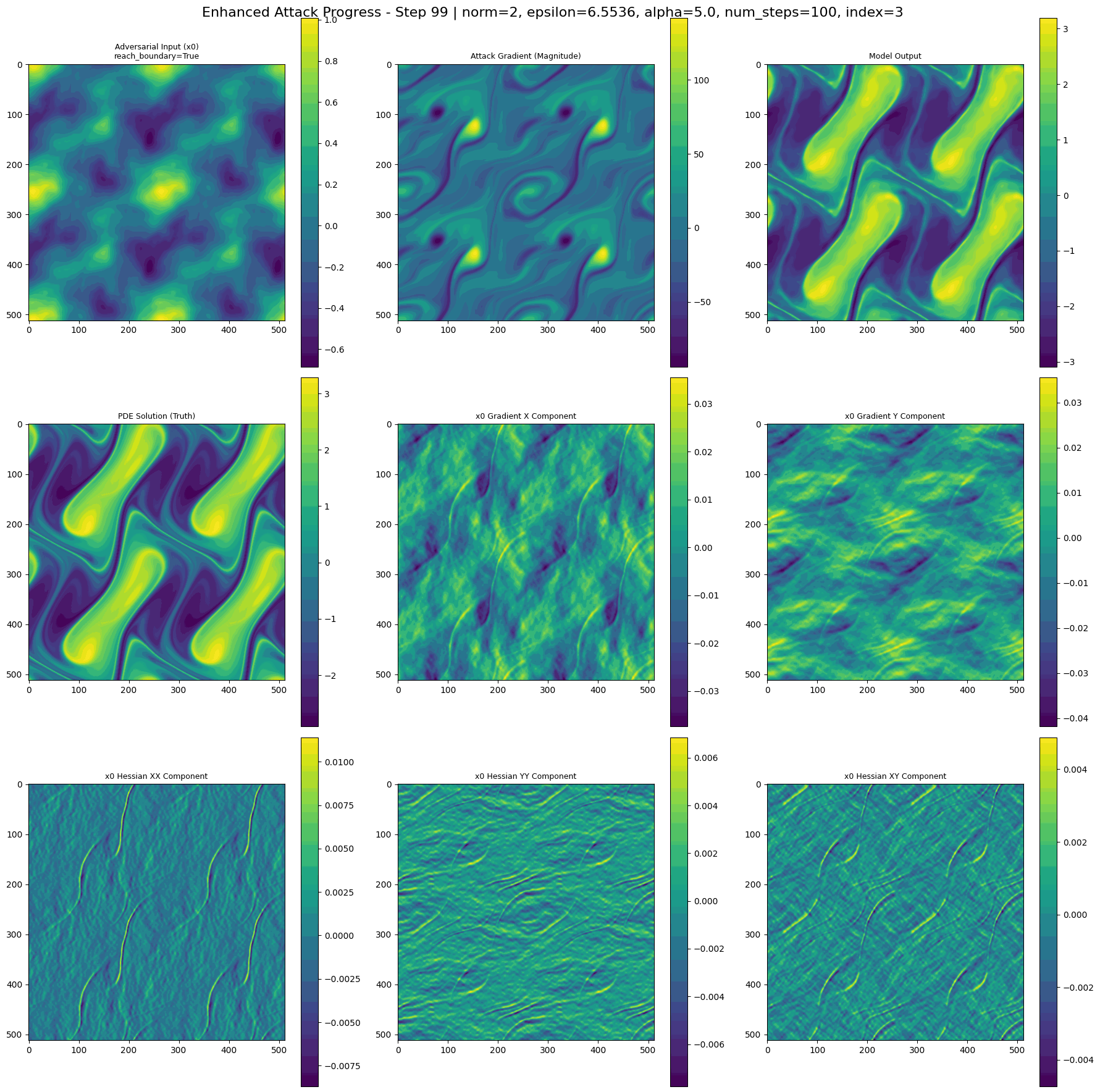}
  \caption{Navier--Stokes attack-progress and periodic-boundary diagnostic. The
  perturbation, gradients, model/PDE outputs, and Hessian-related components
  show that the added perturbation remains compatible with the periodic boundary
  structure of the PDE input. This figure is built by combining the attack-time
  gradient fields with the final generated perturbation; the tiled boundary
  panels show that the edges match periodically, confirming that the generated
  attack respects the periodic boundary condition.}
  \label{fig:app-g-ns-gradient-hessian}
\end{figure}

\nextappendixfigureheading{
\noindent\textbf{Final prediction, delta fields, and forcing perturbations.}
}

\begin{figure}[H]
  \useappendixfigureheading
  \centering
  \includegraphics[width=0.96\textwidth,height=0.82\textheight,keepaspectratio]{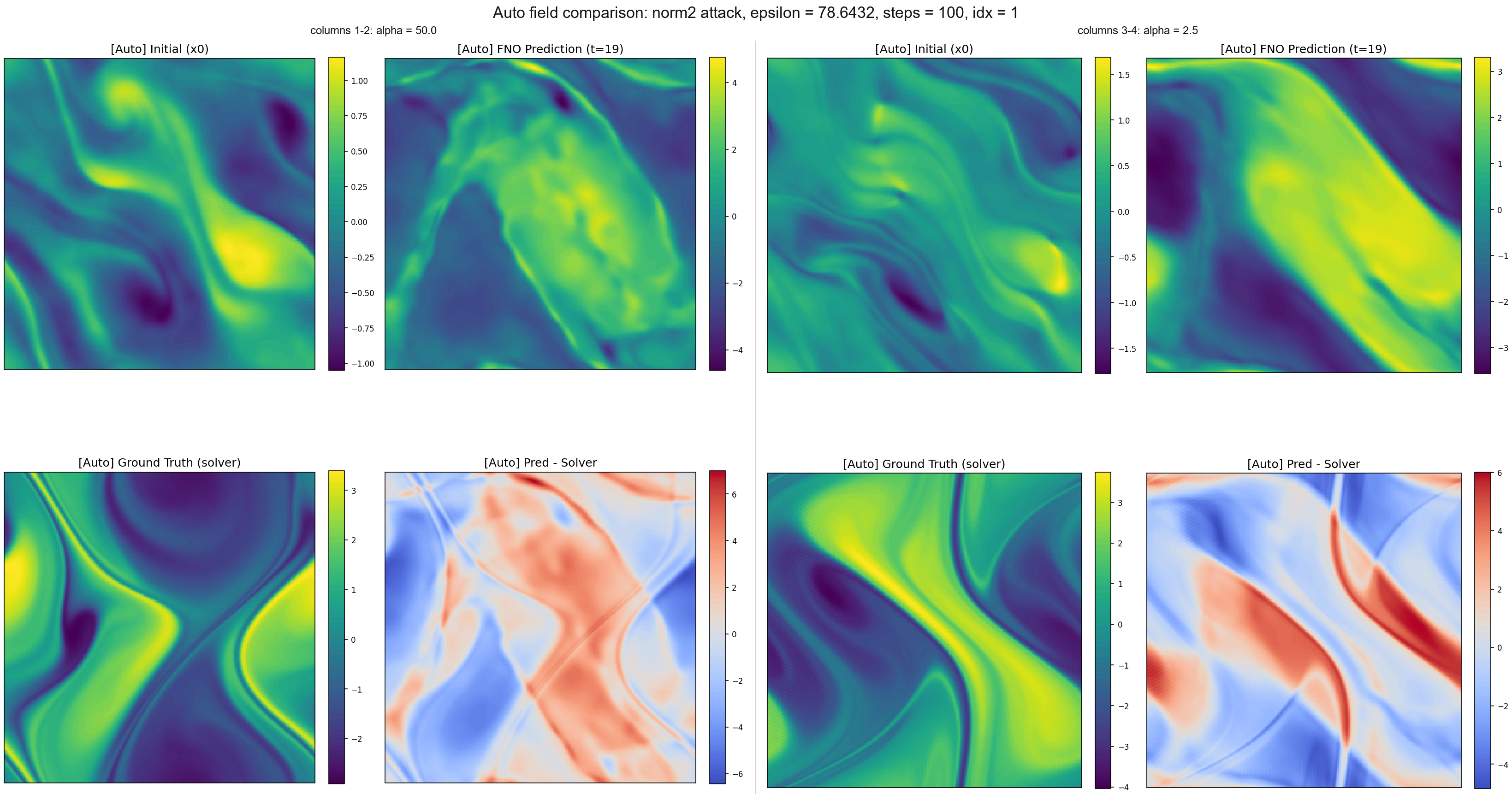}
  \caption{Navier--Stokes auto-field comparison after the norm-$2$ attack at
  $\epsilon=78.6432$, 100 attack steps, and sample index 1. Columns 1--2 use
  $\alpha=50.0$, while columns 3--4 use $\alpha=2.5$. In each two-column block,
  the first row shows the perturbed initial field $x_0$ and the FNO model
  prediction at $t=19$; the second row shows the software solver output and the
  model-minus-solver residual. The residual panels show that the attacked
  perturbation can make the FNO prediction differ sharply from the solver
  solution.}
  \label{fig:app-g-ns-auto-field-alpha-comparison}
\end{figure}

\begin{figure}[H]
  \useappendixfigureheading
  \centering
  \includegraphics[width=0.96\textwidth,height=0.82\textheight,keepaspectratio]{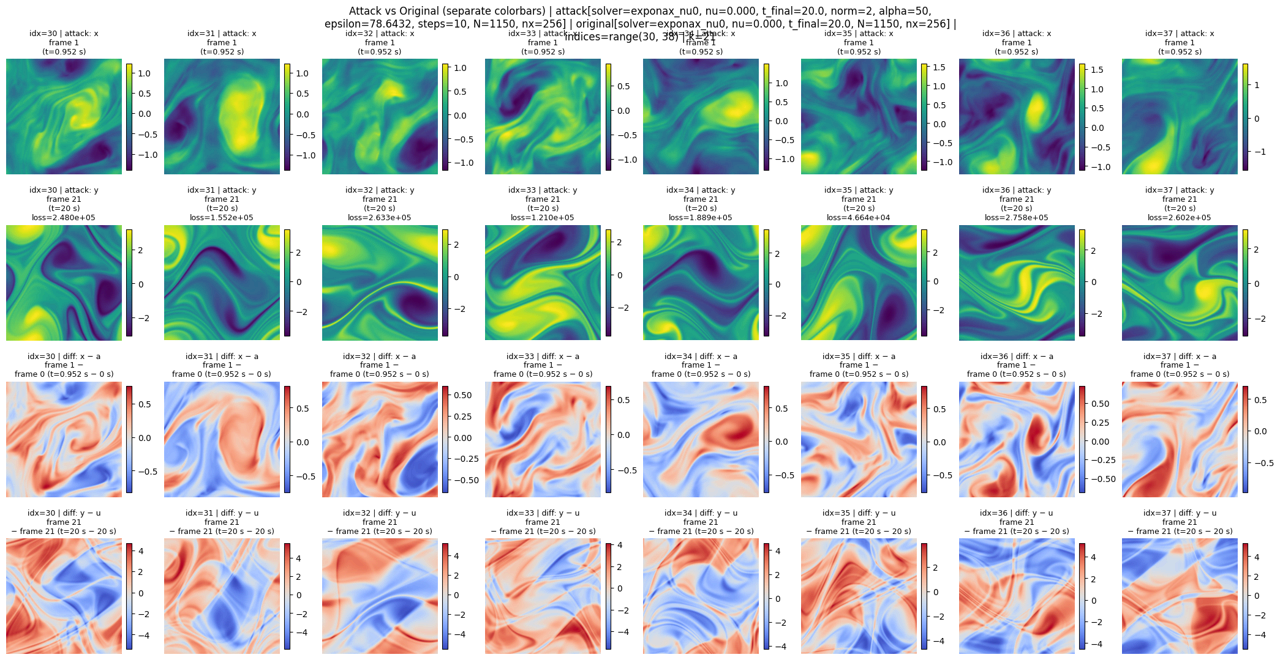}
  \caption{Navier--Stokes attack-versus-original multi-sample grid with $\alpha=50$.}
  \label{fig:app-g-ns-alpha50-grid}
\end{figure}

\begin{figure}[H]
  \useappendixfigureheading
  \centering
  \includegraphics[width=0.96\textwidth,height=0.82\textheight,keepaspectratio]{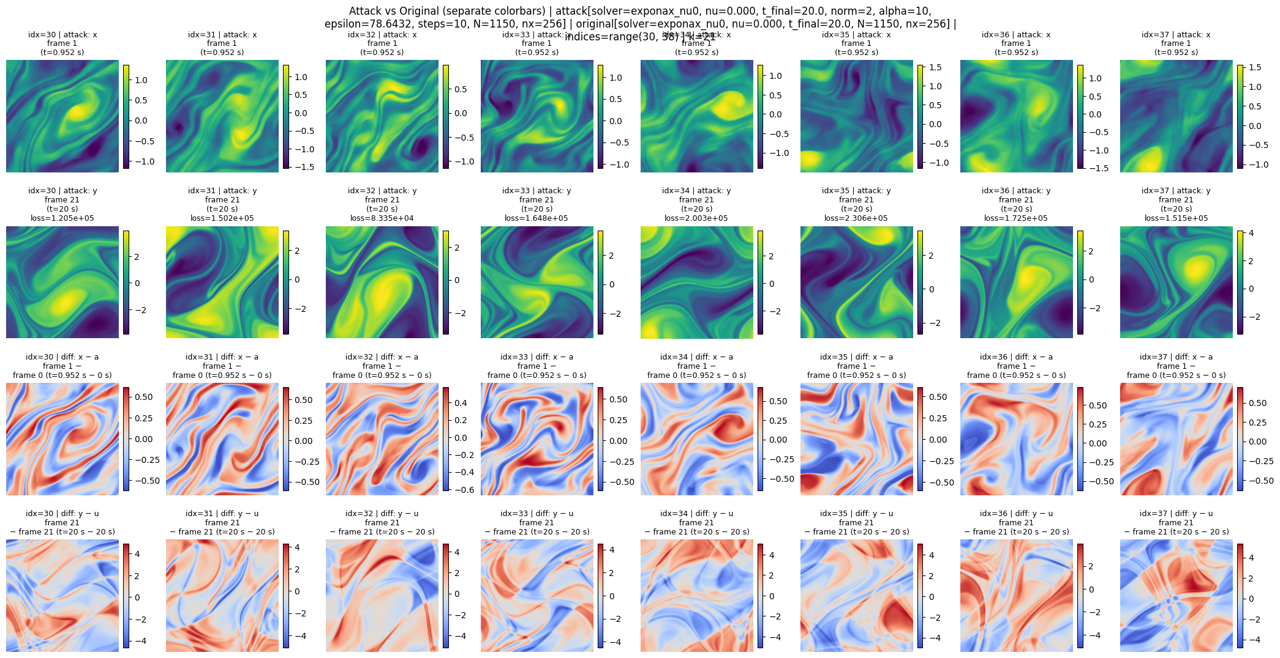}
  \caption{Navier--Stokes attack-versus-original multi-sample grid with $\alpha=10$.}
  \label{fig:app-g-ns-alpha10-grid}
\end{figure}

\begin{figure}[H]
  \useappendixfigureheading
  \centering
  \captionsetup{font=footnotesize,skip=1pt}
  \includegraphics[width=\textwidth,height=0.88\textheight,keepaspectratio]{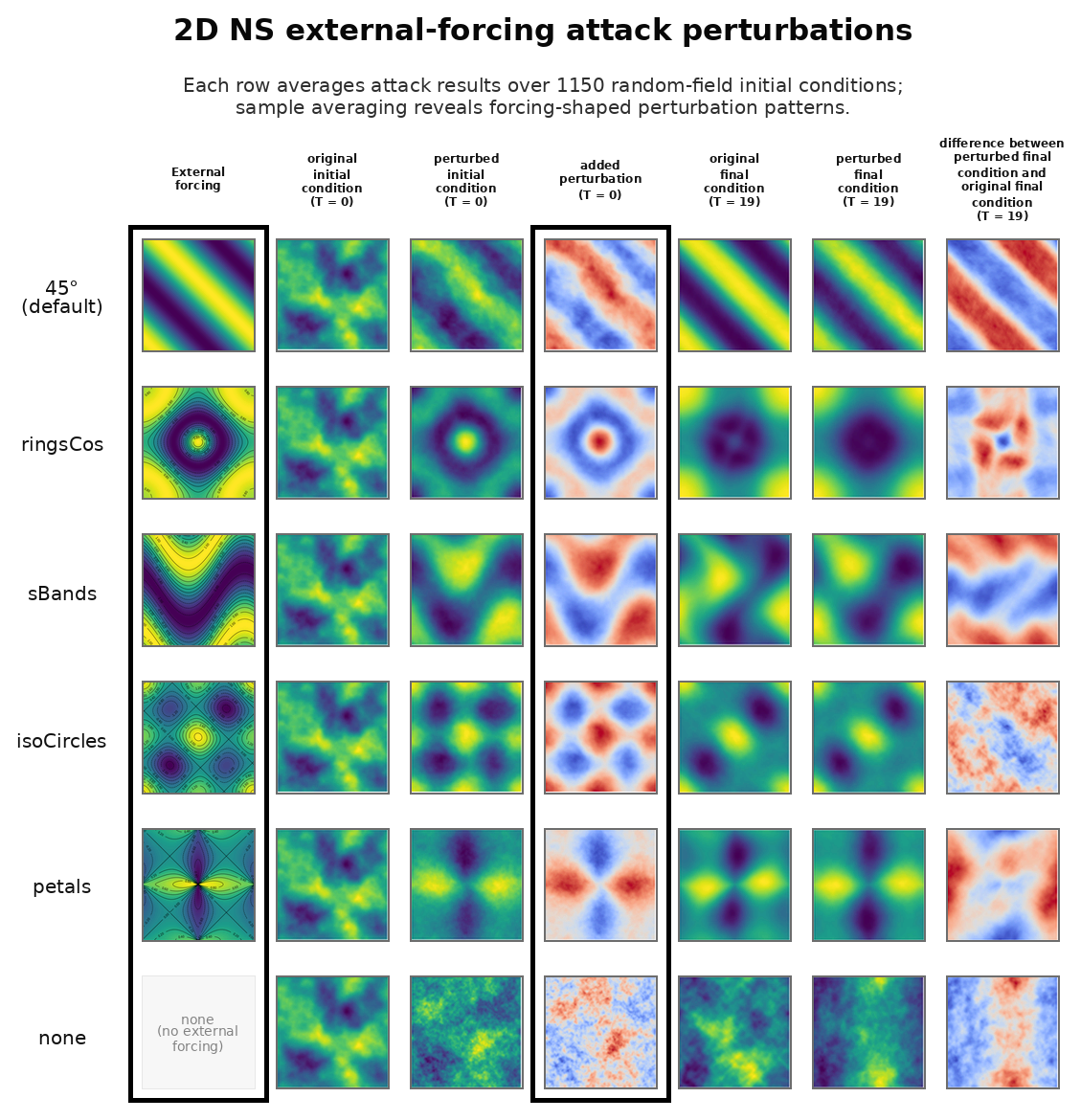}
  \caption{Navier--Stokes external-forcing attack perturbations across forcing
  families. In sample average, the gradient-generated perturbations become
  strongly aligned with the imposed forcing patterns, showing that the attack
  perturbation is not arbitrary noise but is correlated with the forcing
  structure of the Navier--Stokes system.}
  \label{fig:app-g-ns-external-forcing}
\end{figure}

\nextappendixfigureheading{
\noindent\textbf{Loss progression.}
}

\begin{figure}[H]
  \useappendixfigureheading
  \centering
  \includegraphics[width=0.96\textwidth,height=0.82\textheight,keepaspectratio]{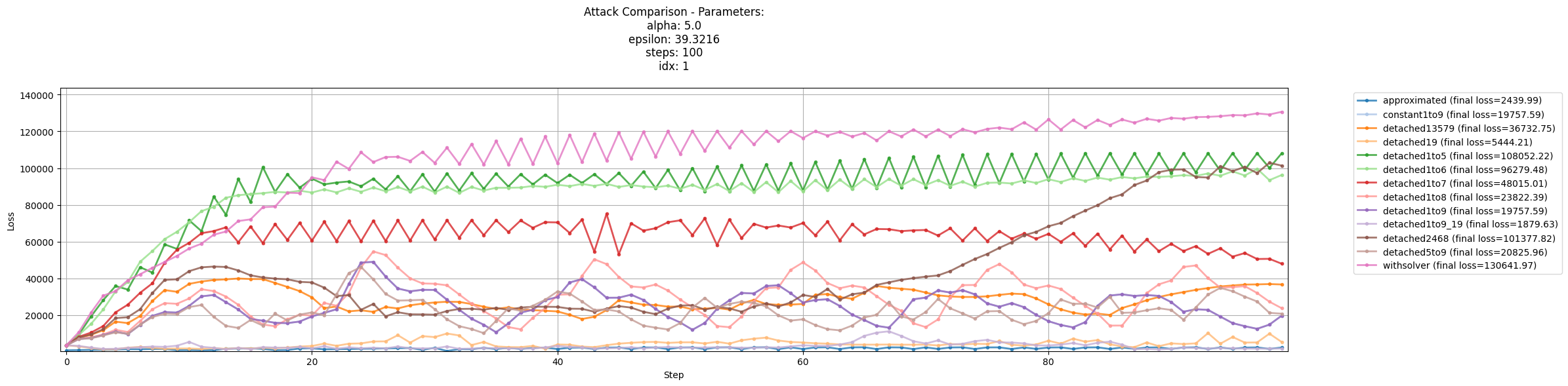}
  \caption{Navier--Stokes attack loss trace comparison. The gradient-backpropagated
  attack objectives and loss variants drive the attack loss upward quickly, and
  several methods reach much larger final losses, demonstrating that the
  proposed loss/backpropagation choices can produce strong loss growth during
  the attack.}
  \label{fig:app-g-ns-loss-trace}
\end{figure}

\FloatBarrier
\nextappendixfigureheading{
\appendixdomainheading{Cross-Benchmark Attack Diagnostics}

\noindent\textbf{Epsilon sweeps.}
}

\begin{figure}[H]
  \useappendixfigureheading
  \centering
  \includegraphics[width=0.96\textwidth,height=0.82\textheight,keepaspectratio]{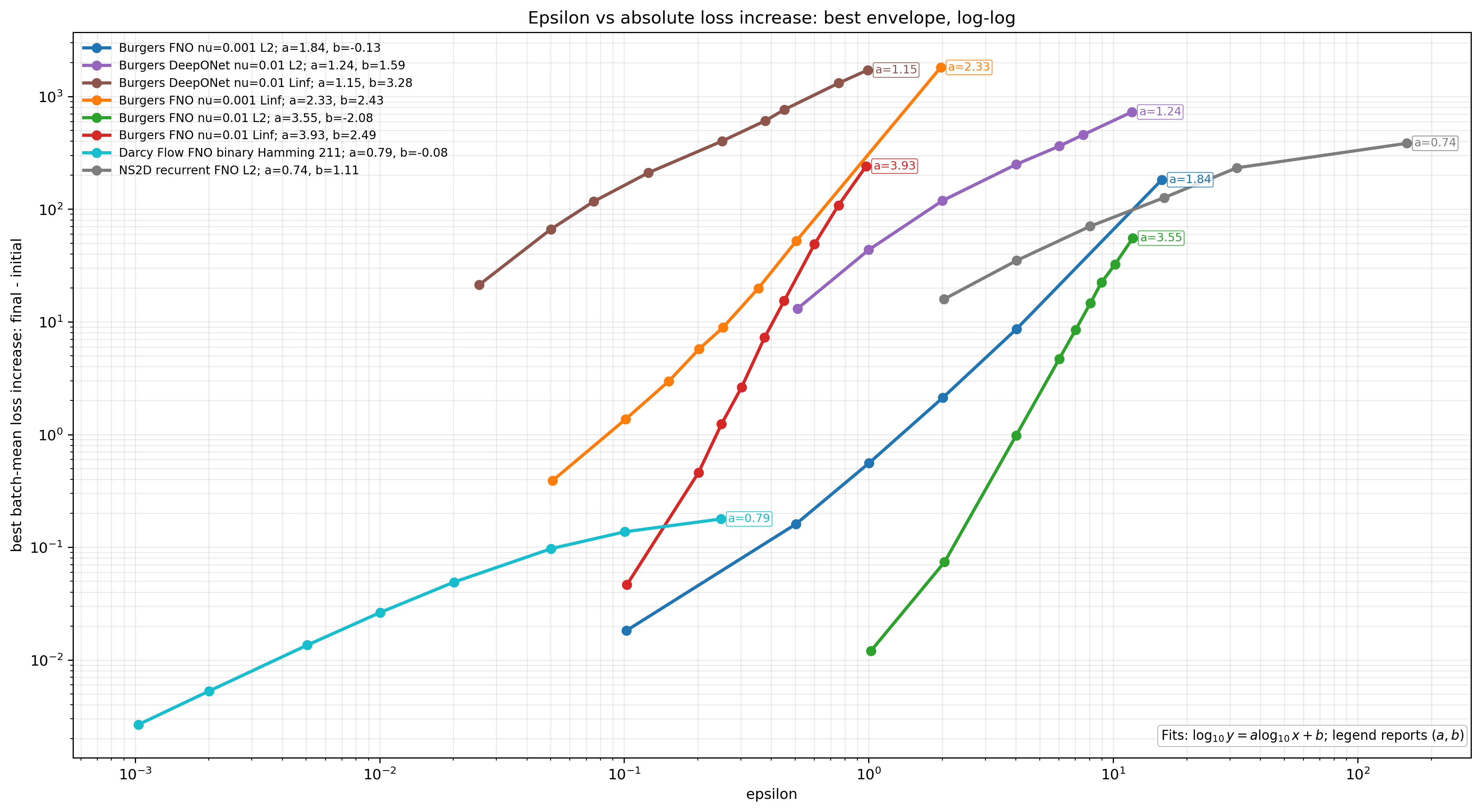}
  \caption{Cross-benchmark attack epsilon sweep showing the best-envelope absolute
  loss increase for Burgers, Darcy Flow, and recurrent Navier--Stokes settings
  on log-log axes. The attacks are applied to trained models before
  adversarial training, across different adversarial-attack budgets. The
  approximately linear trends on the log-log plot indicate an approximate
  power-law relation between attack budget and loss increase: since this plot
  uses base-10 logarithms, if $\log_{10} y = a\log_{10} x + b$, then
  $y = 10^b x^a$, so the slope is the power-law exponent. The legend reports
  the fitted $(a,b)$ values for each curve.}
  \label{fig:app-g-cross-benchmark-epsilon-loss}
\end{figure}

\begin{figure}[H]
  \useappendixfigureheading
  \centering
  \includegraphics[width=0.96\textwidth,height=0.82\textheight,keepaspectratio]{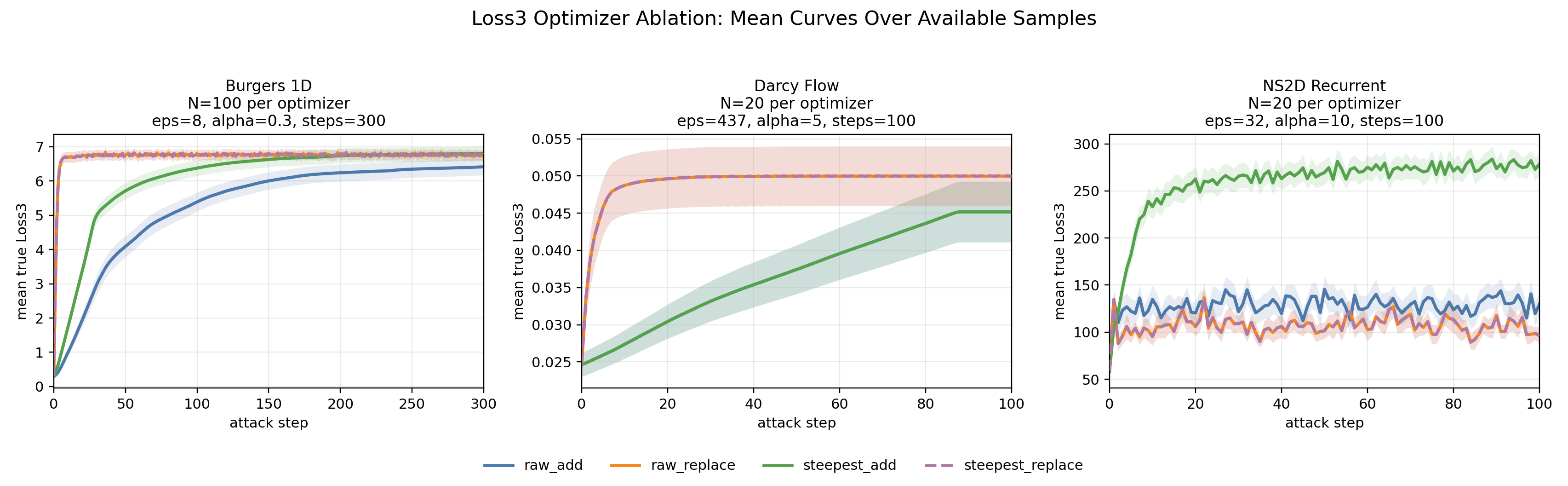}
  \caption{Cross-benchmark Loss 3 optimizer ablation for Burgers, Darcy Flow, and
  recurrent Navier--Stokes. Burgers and recurrent Navier--Stokes use continuous
  \(p=q=2\) attacks. Since \(s_2(g)=g/\|g\|_2\),
  raw-replace and steepest-replace coincide:
  \(\delta_{k+1}=\epsilon g/\|g\|_2\); add rules remain different,
  \(\Pi(\delta_k+\alpha g)\) versus
  \(\Pi(\delta_k+\alpha g/\|g\|_2)\), so NS shows three visible curves. Darcy
  Flow uses a binary Hamming flip budget. With \(a_i\in\{3,12\}\),
  the positive flip score satisfies
  \(\gamma_i=g_i(a_{\mathrm{opp}}-a_i)=9|g_i|\), so raw and steepest top-\(K\)
  rankings coincide for both add and replace. Thus Darcy shows two visible
  trajectories, not missing optimizer runs.}
  \label{fig:app-g-cross-benchmark-optimizer-ablation}
\end{figure}

\FloatBarrier
\endgroup

\clearpage
\begin{multicols}{2}
\appendixsection{Spectral Synthesis of Gaussian Random Fields}{app:spectral-synthesis}
\vspace{0.4em}
\noindent
This appendix records how the Gaussian random fields used in the benchmark
inputs and random perturbation generators are generated. A Gaussian random field
(GRF) is a random function whose values on any finite grid form a multivariate
Gaussian vector. Therefore, on the grid used by the code, generating a GRF
amounts to choosing a mean vector and a covariance matrix, then sampling a
Gaussian vector with that covariance. The spectral method does this efficiently
by diagonalizing the covariance matrix in a Fourier or cosine basis.
\vspace{0.75em}

\subsection*{Definitions and Random-Process Identities}

Let $\xi(x)$ be a second-order random field with mean
$m(x)=\mathbb E[\xi(x)]$. Its autocovariance, or autocorrelation function (ACF)
in the zero-mean normalized setting, is
\begin{equation}
  R(x,x')
  =
  \mathbb E\!\left[
    (\xi(x)-m(x))\overline{(\xi(x')-m(x'))}
  \right].
\end{equation}
For a stationary field this depends only on the lag
$r=x'-x$, so we write $K(r)=R(x,x+r)$. This kernel tells us how strongly two
locations of the random field are correlated as a function of their separation.

The power spectral density (PSD) is the same second-order information written
in frequency space. \textbf{Wiener--Khinchin theorem.} For a stationary field,
\begin{equation}
  S(\omega)=\mathcal F K(\omega),
  \qquad
  K(r)=\mathcal F^{-1}S(r),
\end{equation}
up to the chosen Fourier normalization. Thus the ACF/covariance kernel and the
PSD are Fourier-transform pairs. In particular, the PSD describes how much
power or energy the random field has at each frequency.

\textbf{Bochner theorem.} This theorem gives the validity condition behind the
spectral construction: $K$ is a valid stationary covariance kernel exactly when
it is positive definite, equivalently when it is the Fourier transform of a
nonnegative spectral measure,
\begin{equation}
  K(r)=\int e^{\mathrm i\omega\cdot r}\,d\mu(\omega),
  \qquad
  \mu\ge0 .
\end{equation}
When $d\mu(\omega)=S(\omega)d\omega$, $S$ is the PSD. Thus choosing a
nonnegative spectrum gives a valid stationary covariance kernel, which can be
used as the covariance of a stationary Gaussian random field in physical space.
In short, the ACF/covariance kernel is valid if and only if its spectral
measure is nonnegative.

\textbf{Operator diagonalization.} The same statement can be written as an
eigenfunction expansion. A covariance kernel defines a self-adjoint positive
covariance operator
\begin{equation}
  (Cf)(x)=\int K(x,y)f(y)\,dy .
\end{equation}
By the Mercer/Karhunen--Loeve expansion, if
$C\phi_j=\lambda_j\phi_j$, then a zero-mean Gaussian field can be sampled as
\begin{equation}
  \xi(x)
  =
  \sum_j \sqrt{\lambda_j}\,\eta_j\,\phi_j(x),
  \qquad
  \eta_j\sim\mathcal N(0,1).
\end{equation}
That is the mathematical reason the code samples independent Gaussian
coefficients in a diagonal basis and multiplies them by square roots of
eigenvalues.

\textbf{Fourier operator and DFT diagonalization.} The stationary periodic case
used here is a special case of the operator diagonalization above: the
covariance operator is a convolution operator, and its eigenfunctions are
Fourier modes. For such kernels, the covariance is first an operator on
\(L^2(\mathbb T^d)\), not only a finite covariance matrix. Define the
convolution operator
\(\mathcal C_K:L^2(\mathbb T^d)\to L^2(\mathbb T^d)\) by
\begin{equation*}
  (\mathcal C_K f)(x)
  =
  \int_{\mathbb T^d}K(x-y)f(y)\,dy
  =
  (K*f)(x).
\end{equation*}
Fourier modes $\varphi_k(x)=e^{\mathrm i k\cdot x}$ are eigenfunctions of this
operator:
\begin{align*}
  \mathcal C_K\varphi_k&=S(k)\varphi_k,\\
  \mathcal F(\mathcal C_K f)(k)&=S(k)\widehat f(k).
\end{align*}
Equivalently, the Fourier transform conjugates the covariance operator into a
diagonal multiplication operator:
\begin{equation}
  \begin{aligned}
  \mathcal F\mathcal C_K\mathcal F^{-1}&=M_S,\\
  \mathcal C_K&=\mathcal F^{-1}M_S\mathcal F,\\
  \mathcal C_K^{1/2}&=\mathcal F^{-1}M_{\sqrt S}\mathcal F .
  \end{aligned}
\end{equation}
Here \(M_S\) multiplies each Fourier coefficient by the PSD value \(S(k)\).
On a finite periodic grid, this operator identity becomes the matrix fact that
circulant covariance matrices in one dimension, and block-circulant covariance
matrices in multiple dimensions, are diagonalized by the discrete Fourier
transform (DFT):
\begin{equation}
  C_N=F_N^*\Lambda_NF_N,
  \qquad
  (\Lambda_N)_{kk}=S(k).
\end{equation}
With the usual normalization, the DFT matrix is unitary; after splitting real
and imaginary parts it is the finite-dimensional orthogonal change of basis used
by the sampler. FFT is only the fast algorithm for applying the same DFT. For
Neumann-boundary covariance generators, as in the Darcy Flow coefficient sampler, the
analogous diagonal basis is a discrete cosine transform (DCT).

\subsection*{Frequency-Domain Sampling Construction}

Let $\xi_q$ be a zero-mean random field on a periodic grid, where $q$ denotes
the high-level distribution parameters: kernel family, length scale,
smoothness, spectral decay, amplitude range, and post-processing transform. A
frequency-domain sampler proceeds as follows. First choose a nonnegative
discrete PSD $S_q(k)\ge0$. On a periodic grid this means
\begin{equation}
  \mathbb E|\widehat{\xi_q}(k)|^2=S_q(k),
\end{equation}
so $S_q(k)$ is the variance assigned to Fourier mode $k$. Equivalently, the
finite-grid covariance matrix has the DFT diagonalization
\begin{equation}
  C_q
  =
  F^*
  \operatorname{diag}\!\left(S_q(k)\right)
  F .
\end{equation}
The concrete sampling loop is:
\begin{enumerate}
  \item choose the high-level kernel parameters $q$ and form a nonnegative PSD
  $S_q(k)$;
  \item convert PSD to amplitude by $A_q(k)=S_q(k)^{1/2}$;
  \item draw independent Gaussian Fourier coefficients $\eta_k$;
  \item impose Hermitian symmetry $\eta_{-k}=\overline{\eta_k}$ so the inverse
  transform is real;
  \item set $\widehat{\xi_q}(k)=A_q(k)\eta_k$ and apply an inverse FFT.
\end{enumerate}
In formula form,
\begin{align}
  \widehat{\xi_q}(k)
  &=
  A_q(k)\eta_k,
  \\
  \xi_q
  &=
  F^*\widehat{\xi_q}.
\end{align}
Since $\xi_q$ is a linear transform of Gaussian coefficients, it is Gaussian.
Its covariance is
\begin{equation}
  \mathbb E[\xi_q\xi_q^*]
  =
  F^*\operatorname{diag}\!\left(A_q(k)^2\right)F
  =
  F^*\operatorname{diag}\!\left(S_q(k)\right)F
  =
  C_q .
\end{equation}
Thus the sampled physical-space field has exactly the covariance, ACF, and PSD
specified by $S_q$.

\textbf{Kernel/filter view.} The same construction can be written as filtering
white noise. Under a unitary Fourier transform, white Gaussian noise remains
white Gaussian noise in frequency space. If $\eta$ is white noise and
$\xi=h*\eta$, then the convolution theorem gives
\begin{equation}
  \widehat \xi(\omega)=\widehat h(\omega)\widehat\eta(\omega),
  \qquad
  S_\xi(\omega)=|\widehat h(\omega)|^2 .
\end{equation}
Thus choosing a smoothing filter $\widehat h=A=\sqrt S$ and transforming back
to real space is equivalent to multiplying independent Fourier Gaussian modes
by $\sqrt S$; this is the square-root covariance operator
$\mathcal F^{-1}M_{\sqrt S}\mathcal F$ applied to white noise.

\subsection*{Kernel and High-Level Parameters}

The random-field family is controlled by the shape of $S_q$ or $A_q$. An
RBF/Gaussian kernel has a Gaussian frequency envelope,
\begin{equation}
  S_{\mathrm{RBF}}(\omega)
  \propto
  \exp\!\left(-\frac{\ell^2\|\omega\|^2}{2}\right),
\end{equation}
so larger $\ell$ suppresses high frequencies more strongly and produces
smoother samples. A Matern kernel has polynomial spectral decay,
\begin{equation}
  S_{\mathrm{Mat}}(\omega)
  \propto
  \left(\frac{2\nu}{\ell^2}+\|\omega\|^2\right)^{-(\nu+d/2)},
\end{equation}
where $\nu$ controls roughness. The Neumann-Laplacian and periodic
power-law fields used below are discrete versions of the same spectral
construction, with amplitude filters of the form
\begin{equation}
  A_{\alpha,\tau}(k)
  \propto
  \tau^{\alpha-1}
  \left(c\,\|k\|^2+\tau^2\right)^{-\alpha/2}.
\end{equation}
Larger $\alpha$ or larger effective length scale moves energy toward low
frequencies; smaller values produce rougher fields. Periodic kernels or sine
mixtures concentrate energy at selected harmonics. Range transforms, sign
maps, log-absolute maps, square-centered maps, and sawtooth additions are
post-processing operations: they change amplitude, contrast, or morphology
after the base field is sampled, and need not remain Gaussian.

\subsection*{Benchmark Instantiations}

\paragraph{Burgers.}
The Burgers reference input is a periodic 1D Gaussian random field on $1024$
points with correlation length $0.03$. Matern variants use the amplitude
\begin{equation}
  A_{\nu,\ell}(k)
  =
  \left(
    (2\pi)^2/\ell_{\mathrm{grid}}^2
    +
    (2\pi k)^2
  \right)^{-(\nu+1/2)/2},
\end{equation}
and each generated sample is then normalized samplewise, usually to $[0,1]$
before any later range transform. Shifted Burgers sets also include power-law
Fourier fields, sine mixtures, piecewise-linear fields, sawtooth patterns, and
square waves.

\paragraph{Darcy Flow.}
Darcy Flow uses a binary permeability coefficient rather than a temporal
initial condition. We first form a real pre-threshold field $z(x)$ from
Gaussian coefficients in a cosine basis. With zero mode set to zero, the
coefficients are scaled by
\begin{equation}
  A_{\alpha,\tau}(k_1,k_2)
  =
  \tau^{\alpha-1}
  \left(
    \pi^2(k_1^2+k_2^2)+\tau^2
  \right)^{-\alpha/2},
\end{equation}
and transformed back by an inverse discrete cosine transform,
\begin{equation}
  z
  =
  \operatorname{IDCT}
  \left[
    A_{\alpha,\tau}(k_1,k_2)\eta_{k_1,k_2}
  \right].
\end{equation}
The binary coefficient is then
\begin{equation}
  a(x)=
  \begin{cases}
    12, & z(x)+b\ge 0,\\
    3, & z(x)+b<0,
  \end{cases}
\end{equation}
with training values $\alpha=2$, $\tau=3$, and $b=0$. Shifted variants change
$\alpha$, $\tau$, threshold bias, binary contrast, or the transform applied to
$z$ before thresholding. The PDE is still the zero-Dirichlet elliptic problem in
\appref{app:benchmark-pdes}.

\paragraph{Navier--Stokes.}
For 2D Navier--Stokes, the input is the initial vorticity field on
a periodic $256\times256$ grid. The training reference distribution is
\begin{equation}
  \omega_0
  \sim
  \mathcal N
  \left(
    0,\,
    7^{3/2}(-\Delta+49I)^{-5/2}
  \right).
\end{equation}
This is a periodic spectral power-law Gaussian field. The generalization generator uses
complex Fourier Gaussian coefficients with amplitude
\begin{align}
  A_{\alpha,\tau}(k_x,k_y)
  &=
  N^2\sqrt{2}\,
  \tau^{\alpha-1}
  \left(
    4\pi^2(k_x^2+k_y^2)+\tau^2
  \right)^{-\alpha/2},
  \nonumber\\
  A_{\alpha,\tau}(0,0)
  &=0 .
\end{align}
The shifted sets closest to this training distribution change only $\alpha$ and
$\tau$; larger shifts also apply range and morphology transforms such as scaling,
shifts, sign maps, square-centered maps, log-absolute maps, and sawtooth
additions.

\end{multicols}
\vspace{0.5em}
\appendixsection{Generalization Data and Random Perturbations}{app:generalization-random-perturbation}
\begin{multicols}{2}

\subsection*{Generalization Dataset Construction}
\label{subsec:ood-generalization-scope}

Here OOD generalization means prediction on unseen input-function
distributions for the same PDE, boundary conditions, and solution operator.
This usage includes changed random-field statistics, as in OOD DeepONet
prediction and operator extrapolation studies
\citep{wang2021physicsdeeponet,zhu2023reliable}. Burgers and Navier--Stokes
shift initial conditions; Darcy shifts coefficient fields. Every new input is
relabeled by the same numerical solver. Cross-PDE or changed-boundary-condition
transfer is an additional evaluation setting not tested here.

Across the shifted datasets, the solver, grid, and target-generation
pipeline are fixed while the input distribution changes.  A sampled input $x$ is
passed through the same numerical solver to obtain the label $y=\mathcal G(x)$;
therefore the shift is in the source distribution, not in the PDE definition.
For the generalization datasets and random perturbation generators, a high-level parameter
selection can be summarized as
\begin{align}
  q
  &=
  (\text{family},\ell,\nu,\alpha,\tau,\text{range},T),
  \nonumber\\
  x
  &=
  T(\xi_q),
  \qquad
  y=\mathcal G(x).
\end{align}
The random clean/random solver variants choose $q$ by hand and keep the
perturbation distribution fixed, while adversarial training chooses
$x+\delta$ from the current model--solver discrepancy.

\paragraph{Burgers.}
The selected record contains 12
Gaussian GRF sets with correlation lengths $0.009$ or $0.012$, 10 Matern GRF
sets with $(c,\nu)\in\{(0.03,1.5),(0.04,2.5),(0.055,4),(0.08,2.5)\}$, 14
power-law Fourier sets with $\alpha\in[1.2,4]$ and $k_0\in[8,36]$, 12 sine-mixture
sets, one sawtooth set, and one square-wave set.  The samplewise amplitude
range is also changed, covering intervals from $[-0.65,1.35]$ to $[0.15,1.25]$.

\paragraph{Darcy Flow.}
The 50 shifted sets keep the binary values 3 and 12 and use eight families:
Matern smooth, Matern fine, high-pass GRF, band-pass GRF, wave mix, blocky
tiles, rectangles, and cellular blobs.  The
target high-permeability fraction is drawn from $\{0.12,0.16,0.20,0.24\}$;
$\alpha$ ranges from 1.052 to 4.674, $\tau$ from 1.274 to 13.359, and the
average edge density ranges from 0.006 to 0.349.  The transforms include
identity, log-signed, exponential high-bias, tanh-plateau, cubic-skew, and
square-extreme maps before binary thresholding.

\paragraph{Navier--Stokes.}
The training reference uses the Zongyi real-initial-condition split: each
stored initial vorticity field is spectrally upsampled to $256\times256$ and
rolled out by the Exponax vorticity solver with $\nu=10^{-5}$, $T=20$, and 21
integer-time frames; the retained train/test files contain 1150/50 samples.
The generated NS2D sets use periodic-GRF parameter shifts with identity
transform for the near cases, using $\alpha\in[1.5,4.5]$ and $\tau\in[4,12]$.  The mid sets use
wider spectrum shifts with $\alpha\in[0.8,6]$ and $\tau\in[2,20]$.  The far sets
start from the $\alpha=2.5$, $\tau=7$ GRF family and apply scale,
positive/negative shift, scale-shift, sign, square-centered, log-abs-centered,
and sawtooth-add transforms.

\columnbreak

\subsection*{Random Perturbation Training Variants}

\paragraph{Burgers random clean and random solver.}
For each training batch, a fresh random field $\delta$ is sampled in Fourier
space.  The code randomly chooses a Gaussian/RBF or Matern kernel for each
sample.  Both kernels use correlation-length choices
$\{0.015,0.03,0.06,0.12,0.24\}$, and the Matern branch additionally draws
$\nu\in\{1.2,2.2,3.2,4.2,5.2\}$.  The zero-frequency mode is removed, the field
is mean-centered, and it is normalized so that the per-sample RMS radius equals
$\epsilon$.  In the reported runs, $\epsilon=(\max x-\min x)\epsilon_{\rm frac}u$
with uniform per-sample jitter $u$, so perturbation magnitudes vary.  Random clean trains on
$(x+\delta,y_{\rm clean})$; random solver trains on
$(x+\delta,\mathcal G(x+\delta))$.

\paragraph{Darcy Flow random clean and random solver.}
For each Darcy batch, the code first snaps the input to its two observed binary
levels, then samples a random spatial score field.  The score-field kernel is
sampled from Gaussian, Matern, high-pass, band-pass, and mixed kernels;
$\alpha\in\{1.2,2.2,3.2,4.2,5.2\}$ is chosen at random, and the length scale is
sampled uniformly from $[0.035,0.30]$.  The flip fraction is sampled uniformly
from $[0.005,0.05]$; the pixels with the largest absolute random scores are
flipped to the opposite binary value.  Random clean keeps the original solver target;
random solver recomputes the Darcy solution for the flipped coefficient field.
Neither random clean nor random solver uses an adversarial gradient.

\subsection*{Relative $L_2$ Loss Tables}

Each PDE has two complete 52-dataset comparisons: maximum common training
time first, then maximum common epoch. For the participating trained methods,
the common time is the smallest recorded final training-core time; each column
uses its last evaluated checkpoint at or before that limit. The common epoch
is the largest evaluated epoch shared by every participating method. Thus the
time-matched columns generally use different epochs, while the epoch-matched
columns generally have different training times. The static baseline does not
constrain either limit. Both checkpoint epoch and recorded core time are listed
in each table header; no interpolation or extension of a shorter run is used.

Burgers uses 7.427462 h for five retained training methods and epoch 1000 for
all six trained methods, additionally retaining Random clean. Including the
shorter failed Random-clean trajectory would instead restrict the common
time to 2.902409 h. The participating methods are therefore stated explicitly
in each comparison.
Darcy uses 1468.919765 s (24.481996 min) and epoch 3000 for all seven training
methods, including Physics and L3 Stop. Each table starts with the ID train/test
rows and then lists all 50 OOD rows. The separate accepted Darcy attack/ablation
records retain their own checkpoints and are not mixed into these comparisons.

\end{multicols}
\begingroup\footnotesize
\setlength{\tabcolsep}{2pt}
\renewcommand{\arraystretch}{1.06}
\setlength{\LTcapwidth}{\textwidth}
\begin{longtable}{@{}C{0.035\textwidth} L{0.39\textwidth} *{6}{R{0.078\textwidth}}@{}}
\caption{Burgers relative $L_2$ at the maximum common training-core time 7.427462 h. The 52 rows are two ID train/test references followed by 50 OOD datasets. The baseline is static; bold marks the smallest trained-method value in each row. The five retained training methods define this time support; Random clean has a shorter failed trajectory and limits the seven-model time range to 2.902409 h. It is retained in the common-epoch table and companion figures. Each method uses its last evaluated checkpoint at or before the time limit; the selected epochs differ, as recorded in the header.}\label{tab:appendix-burgers-52-relative-l2-time}\\
\toprule
No. & Dataset & Base & L1 & L2 & L3 Full & L3 Stop & Rand solver\\
\multicolumn{2}{l}{Selected checkpoint epoch} & 0 & 2756 & 1499 & 1000 & 2562 & 7859\\
\multicolumn{2}{l}{Recorded core time (h)} & 0.00000 & 7.42484 & 7.42670 & 7.42746 & 7.42686 & 7.42691\\
\midrule
\endfirsthead
\toprule
No. & Dataset & Base & L1 & L2 & L3 Full & L3 Stop & Rand solver\\
\multicolumn{2}{l}{Selected checkpoint epoch} & 0 & 2756 & 1499 & 1000 & 2562 & 7859\\
\multicolumn{2}{l}{Recorded core time (h)} & 0.00000 & 7.42484 & 7.42670 & 7.42746 & 7.42686 & 7.42691\\
\midrule
\endhead
\midrule
\multicolumn{8}{r}{continued on next page}\\
\endfoot
\bottomrule
\endlastfoot
1 & Train; GRF corr=0.03; range [0,1] & 0.0169 & 0.0027 & 0.0029 & 0.0045 & 0.0023 & \textbf{0.0015}\\
2 & Test; GRF corr=0.03; range [0,1] & 0.0178 & 0.0028 & 0.0031 & 0.0046 & 0.0024 & \textbf{0.0016}\\
3 & Gaussian GRF corr=0.012; range [-0.2,1.2] & 0.0409 & 0.0331 & 0.0345 & \textbf{0.0165} & 0.0309 & 0.0250\\
4 & Gaussian GRF corr=0.012; range [-0.3,1.3] & 0.0552 & 0.0437 & 0.0455 & \textbf{0.0225} & 0.0408 & 0.0356\\
5 & Gaussian GRF corr=0.012; range [0.15,1.25] & 0.0390 & 0.0184 & 0.0189 & \textbf{0.0100} & 0.0174 & 0.0132\\
6 & Gaussian GRF corr=0.012; range [0,1.2] & 0.0326 & 0.0213 & 0.0220 & \textbf{0.0110} & 0.0203 & 0.0148\\
7 & Gaussian GRF corr=0.009; range [-0.3,1.3] & 0.0506 & 0.0445 & 0.0462 & \textbf{0.0254} & 0.0420 & 0.0413\\
8 & Gaussian GRF corr=0.009; range [0.15,1.25] & 0.0339 & 0.0193 & 0.0197 & \textbf{0.0109} & 0.0186 & 0.0145\\
9 & Gaussian GRF corr=0.012; range [-0.1,1.1] & 0.0325 & 0.0240 & 0.0250 & \textbf{0.0119} & 0.0227 & 0.0164\\
10 & Gaussian GRF corr=0.009; range [0,1.2] & 0.0321 & 0.0238 & 0.0244 & \textbf{0.0129} & 0.0229 & 0.0184\\
11 & Gaussian GRF corr=0.012; range [-0.5,1.5] & 0.1002 & 0.0742 & 0.0754 & \textbf{0.0435} & 0.0692 & 0.0678\\
12 & Gaussian GRF corr=0.009; range [-0.5,1.5] & 0.0816 & 0.0693 & 0.0712 & \textbf{0.0417} & 0.0651 & 0.0682\\
13 & Gaussian GRF corr=0.009; range [-0.2,1.2] & 0.0410 & 0.0338 & 0.0353 & \textbf{0.0194} & 0.0322 & 0.0308\\
14 & Gaussian GRF corr=0.012; range [0.05,1.05] & 0.0255 & 0.0166 & 0.0172 & \textbf{0.0083} & 0.0161 & 0.0094\\
15 & Matern GRF c=0.055, nu=4; range [-0.3,1.3] & 0.0658 & 0.0361 & 0.0380 & \textbf{0.0188} & 0.0332 & 0.0296\\
16 & Matern GRF c=0.04, nu=2.5; range [-0.3,1.3] & 0.0485 & 0.0390 & 0.0402 & \textbf{0.0226} & 0.0370 & 0.0369\\
17 & Matern GRF c=0.04, nu=2.5; range [-0.5,1.5] & 0.0808 & 0.0626 & 0.0638 & \textbf{0.0384} & 0.0587 & 0.0635\\
18 & Matern GRF c=0.04, nu=2.5; range [0.15,1.25] & 0.0339 & 0.0174 & 0.0177 & \textbf{0.0099} & 0.0169 & 0.0126\\
19 & Matern GRF c=0.08, nu=2.5; range [-0.3,1.3] & 0.0693 & 0.0325 & 0.0351 & \textbf{0.0169} & 0.0298 & 0.0266\\
20 & Matern GRF c=0.04, nu=2.5; range [-0.65,1.35] & 0.1489 & 0.0941 & 0.0944 & \textbf{0.0612} & 0.0882 & 0.0982\\
21 & Matern GRF c=0.04, nu=2.5; range [0,1.5] & 0.0811 & 0.0374 & 0.0387 & \textbf{0.0223} & 0.0339 & 0.0406\\
22 & Matern GRF c=0.08, nu=2.5; range [-0.5,1.5] & 0.1329 & 0.0735 & 0.0766 & \textbf{0.0441} & 0.0671 & 0.0773\\
23 & Matern GRF c=0.055, nu=4; range [-0.5,1.5] & 0.1234 & 0.0723 & 0.0750 & \textbf{0.0428} & 0.0671 & 0.0718\\
24 & Matern GRF c=0.03, nu=1.5; range [0,1.5] & 0.0449 & 0.0236 & 0.0241 & \textbf{0.0157} & 0.0232 & 0.0239\\
25 & Power-law Fourier alpha=2.5, k0=18; range [-0.2,1.2] & 0.0397 & 0.0315 & 0.0324 & \textbf{0.0166} & 0.0299 & 0.0258\\
26 & Power-law Fourier alpha=1.5, k0=10; range [-0.5,1.5] & 0.0800 & 0.0536 & 0.0558 & \textbf{0.0304} & 0.0509 & 0.0500\\
27 & Power-law Fourier alpha=3, k0=24; range [-0.2,1.2] & 0.0394 & 0.0329 & 0.0338 & \textbf{0.0182} & 0.0313 & 0.0307\\
28 & Power-law Fourier alpha=2.5, k0=18; range [-0.5,1.5] & 0.0860 & 0.0655 & 0.0671 & \textbf{0.0378} & 0.0613 & 0.0610\\
29 & Power-law Fourier alpha=2.5, k0=18; range [-0.1,1.1] & 0.0329 & 0.0237 & 0.0246 & \textbf{0.0128} & 0.0228 & 0.0182\\
30 & Power-law Fourier alpha=1.2, k0=8; range [-0.5,1.5] & 0.0706 & 0.0443 & 0.0457 & \textbf{0.0255} & 0.0423 & 0.0428\\
31 & Power-law Fourier alpha=2.2, k0=16; range [-0.2,1.2] & 0.0371 & 0.0286 & 0.0297 & \textbf{0.0160} & 0.0272 & 0.0243\\
32 & Power-law Fourier alpha=3.5, k0=28; range [0.15,1.25] & 0.0331 & 0.0184 & 0.0189 & \textbf{0.0101} & 0.0180 & 0.0131\\
33 & Power-law Fourier alpha=3.5, k0=28; range [-0.5,1.5] & 0.0786 & 0.0629 & 0.0644 & \textbf{0.0382} & 0.0590 & 0.0623\\
34 & Power-law Fourier alpha=4, k0=36; range [-0.5,1.5] & 0.0739 & 0.0653 & 0.0666 & \textbf{0.0407} & 0.0616 & 0.0678\\
35 & Power-law Fourier alpha=4, k0=36; range [0.15,1.25] & 0.0299 & 0.0190 & 0.0193 & \textbf{0.0109} & 0.0186 & 0.0137\\
36 & Power-law Fourier alpha=3, k0=24; range [-0.1,1.1] & 0.0328 & 0.0248 & 0.0257 & \textbf{0.0135} & 0.0239 & 0.0209\\
37 & Power-law Fourier alpha=1.8, k0=12; range [-0.5,1.5] & 0.0846 & 0.0559 & 0.0578 & \textbf{0.0336} & 0.0522 & 0.0548\\
38 & Power-law Fourier alpha=3, k0=24; range [-0.5,1.5] & 0.0765 & 0.0650 & 0.0662 & \textbf{0.0398} & 0.0615 & 0.0655\\
39 & Sine mix f=[5, 9, 15, 23], decay=0.25; range [0,1.5] & 0.0512 & 0.0335 & 0.0330 & \textbf{0.0157} & 0.0307 & 0.0309\\
40 & Sine mix f=[7, 19, 43, 89], decay=0.15; range [0,1.5] & 0.0371 & 0.0274 & 0.0282 & \textbf{0.0096} & 0.0280 & 0.0130\\
41 & Sine mix f=[5, 9, 15, 23], decay=0.25; range [-0.3,1.3] & 0.0453 & 0.0440 & 0.0456 & \textbf{0.0199} & 0.0382 & 0.0305\\
42 & Sine mix f=[7, 19, 43, 89], decay=0.15; range [-0.3,1.3] & 0.0458 & 0.0430 & 0.0443 & \textbf{0.0146} & 0.0426 & 0.0251\\
43 & Sine mix f=[7, 19, 43, 89], decay=0.15; range [-0.2,1.2] & 0.0377 & 0.0348 & 0.0359 & \textbf{0.0118} & 0.0345 & 0.0203\\
44 & Sine mix f=[5, 9, 15, 23], decay=0.25; range [-0.65,1.35] & 0.0959 & 0.1000 & 0.1000 & \textbf{0.0521} & 0.0914 & 0.0830\\
45 & Sine mix f=[5, 9, 15, 23], decay=0.25; range [-0.5,1.5] & 0.0635 & 0.0670 & 0.0696 & \textbf{0.0305} & 0.0584 & 0.0506\\
46 & Sine mix f=[5, 9, 15, 23], decay=0.25; range [-0.2,1.2] & 0.0389 & 0.0338 & 0.0351 & \textbf{0.0161} & 0.0296 & 0.0228\\
47 & Sine mix f=[7, 19, 43, 89], decay=0.15; range [-0.5,1.5] & 0.0610 & 0.0584 & 0.0601 & \textbf{0.0213} & 0.0579 & 0.0363\\
48 & Sine mix f=[5, 10, 20, 40, 80], decay=0.9; range [-0.1,1.1] & 0.0230 & 0.0180 & 0.0183 & \textbf{0.0058} & 0.0179 & 0.0060\\
49 & Sine mix f=[4, 7, 11], decay=0.35; range [-0.3,1.3] & 0.0605 & 0.0375 & 0.0404 & \textbf{0.0164} & 0.0310 & 0.0260\\
50 & Sine mix f=[7, 19, 43, 89], decay=0.15; range [-0.65,1.35] & 0.0843 & 0.0774 & 0.0774 & \textbf{0.0418} & 0.0782 & 0.0538\\
51 & Sawtooth freq=2; range [0.15,1.25] & 0.0223 & 0.0166 & 0.0164 & \textbf{0.0101} & 0.0126 & 0.0346\\
52 & Square wave freq=7, duty=0.35; range [0,1.2] & 0.0802 & 0.0577 & 0.0578 & \textbf{0.0134} & 0.0592 & 0.0310\\
\end{longtable}
\endgroup

\begingroup\footnotesize
\setlength{\tabcolsep}{2pt}
\renewcommand{\arraystretch}{1.06}
\setlength{\LTcapwidth}{\textwidth}
\begin{longtable}{@{}C{0.035\textwidth} L{0.35\textwidth} *{7}{R{0.075\textwidth}}@{}}
\caption{Burgers relative $L_2$ at the maximum common epoch 1000. The 52 rows are two ID train/test references followed by 50 OOD datasets. The baseline is static; bold marks the smallest trained-method value in each row. All six trained methods, including Random clean, are retained at the maximum common epoch; their training times need not agree. A dash denotes Random-clean runtime metadata not supplied in the retained final checkpoint-selection file; its epoch-1000 errors are recorded.}\label{tab:appendix-burgers-52-relative-l2}\\
\toprule
No. & Dataset & Base & L1 & L2 & L3 Full & L3 Stop & Rand clean & Rand solver\\
\multicolumn{2}{l}{Selected checkpoint epoch} & 0 & 1000 & 1000 & 1000 & 1000 & 1000 & 1000\\
\multicolumn{2}{l}{Recorded core time (h)} & 0.00000 & 2.69636 & 4.94251 & 7.42746 & 2.95029 & -- & 0.93973\\
\midrule
\endfirsthead
\toprule
No. & Dataset & Base & L1 & L2 & L3 Full & L3 Stop & Rand clean & Rand solver\\
\multicolumn{2}{l}{Selected checkpoint epoch} & 0 & 1000 & 1000 & 1000 & 1000 & 1000 & 1000\\
\multicolumn{2}{l}{Recorded core time (h)} & 0.00000 & 2.69636 & 4.94251 & 7.42746 & 2.95029 & -- & 0.93973\\
\midrule
\endhead
\midrule
\multicolumn{9}{r}{continued on next page}\\
\endfoot
\bottomrule
\endlastfoot
1 & Train; GRF corr=0.03; range [0,1] & 0.0169 & \textbf{0.0032} & 0.0053 & 0.0045 & 0.0055 & 0.0217 & 0.0035\\
2 & Test; GRF corr=0.03; range [0,1] & 0.0178 & \textbf{0.0034} & 0.0055 & 0.0046 & 0.0056 & 0.1285 & 0.0038\\
3 & Gaussian GRF corr=0.012; range [-0.2,1.2] & 0.0409 & 0.0352 & 0.0357 & \textbf{0.0165} & 0.0330 & 0.1561 & 0.0328\\
4 & Gaussian GRF corr=0.012; range [-0.3,1.3] & 0.0552 & 0.0464 & 0.0469 & \textbf{0.0225} & 0.0432 & 0.1868 & 0.0443\\
5 & Gaussian GRF corr=0.012; range [0.15,1.25] & 0.0390 & 0.0192 & 0.0194 & \textbf{0.0100} & 0.0180 & 0.1055 & 0.0201\\
6 & Gaussian GRF corr=0.012; range [0,1.2] & 0.0326 & 0.0225 & 0.0226 & \textbf{0.0110} & 0.0214 & 0.1191 & 0.0215\\
7 & Gaussian GRF corr=0.009; range [-0.3,1.3] & 0.0506 & 0.0471 & 0.0474 & \textbf{0.0254} & 0.0443 & 0.1596 & 0.0442\\
8 & Gaussian GRF corr=0.009; range [0.15,1.25] & 0.0339 & 0.0199 & 0.0203 & \textbf{0.0109} & 0.0194 & 0.0944 & 0.0201\\
9 & Gaussian GRF corr=0.012; range [-0.1,1.1] & 0.0325 & 0.0256 & 0.0258 & \textbf{0.0119} & 0.0243 & 0.1290 & 0.0235\\
10 & Gaussian GRF corr=0.009; range [0,1.2] & 0.0321 & 0.0247 & 0.0250 & \textbf{0.0129} & 0.0241 & 0.1046 & 0.0233\\
11 & Gaussian GRF corr=0.012; range [-0.5,1.5] & 0.1002 & 0.0766 & 0.0772 & \textbf{0.0435} & 0.0716 & 0.2503 & 0.0775\\
12 & Gaussian GRF corr=0.009; range [-0.5,1.5] & 0.0816 & 0.0724 & 0.0726 & \textbf{0.0417} & 0.0679 & 0.2111 & 0.0716\\
13 & Gaussian GRF corr=0.009; range [-0.2,1.2] & 0.0410 & 0.0359 & 0.0363 & \textbf{0.0194} & 0.0342 & 0.1369 & 0.0333\\
14 & Gaussian GRF corr=0.012; range [0.05,1.05] & 0.0255 & 0.0176 & 0.0177 & \textbf{0.0083} & 0.0173 & 0.1083 & 0.0162\\
15 & Matern GRF c=0.055, nu=4; range [-0.3,1.3] & 0.0658 & 0.0390 & 0.0400 & \textbf{0.0188} & 0.0358 & 0.2279 & 0.0397\\
16 & Matern GRF c=0.04, nu=2.5; range [-0.3,1.3] & 0.0485 & 0.0409 & 0.0413 & \textbf{0.0226} & 0.0390 & 0.1587 & 0.0387\\
17 & Matern GRF c=0.04, nu=2.5; range [-0.5,1.5] & 0.0808 & 0.0646 & 0.0652 & \textbf{0.0384} & 0.0609 & 0.2113 & 0.0640\\
18 & Matern GRF c=0.04, nu=2.5; range [0.15,1.25] & 0.0339 & 0.0180 & 0.0183 & \textbf{0.0099} & 0.0178 & 0.0935 & 0.0184\\
19 & Matern GRF c=0.08, nu=2.5; range [-0.3,1.3] & 0.0693 & 0.0360 & 0.0373 & \textbf{0.0169} & 0.0335 & 0.2204 & 0.0378\\
20 & Matern GRF c=0.04, nu=2.5; range [-0.65,1.35] & 0.1489 & 0.0967 & 0.0966 & \textbf{0.0612} & 0.0912 & 0.2981 & 0.1075\\
21 & Matern GRF c=0.04, nu=2.5; range [0,1.5] & 0.0811 & 0.0386 & 0.0388 & \textbf{0.0223} & 0.0359 & 0.1360 & 0.0469\\
22 & Matern GRF c=0.08, nu=2.5; range [-0.5,1.5] & 0.1329 & 0.0771 & 0.0785 & \textbf{0.0441} & 0.0715 & 0.2961 & 0.0875\\
23 & Matern GRF c=0.055, nu=4; range [-0.5,1.5] & 0.1234 & 0.0765 & 0.0776 & \textbf{0.0428} & 0.0712 & 0.2937 & 0.0840\\
24 & Matern GRF c=0.03, nu=1.5; range [0,1.5] & 0.0449 & 0.0246 & 0.0250 & \textbf{0.0157} & 0.0243 & 0.0918 & 0.0269\\
25 & Power-law Fourier alpha=2.5, k0=18; range [-0.2,1.2] & 0.0397 & 0.0329 & 0.0333 & \textbf{0.0166} & 0.0316 & 0.1395 & 0.0307\\
26 & Power-law Fourier alpha=1.5, k0=10; range [-0.5,1.5] & 0.0800 & 0.0566 & 0.0580 & \textbf{0.0304} & 0.0536 & 0.2260 & 0.0575\\
27 & Power-law Fourier alpha=3, k0=24; range [-0.2,1.2] & 0.0394 & 0.0343 & 0.0347 & \textbf{0.0182} & 0.0331 & 0.1335 & 0.0320\\
28 & Power-law Fourier alpha=2.5, k0=18; range [-0.5,1.5] & 0.0860 & 0.0682 & 0.0688 & \textbf{0.0378} & 0.0641 & 0.2231 & 0.0681\\
29 & Power-law Fourier alpha=2.5, k0=18; range [-0.1,1.1] & 0.0329 & 0.0250 & 0.0253 & \textbf{0.0128} & 0.0244 & 0.1214 & 0.0230\\
30 & Power-law Fourier alpha=1.2, k0=8; range [-0.5,1.5] & 0.0706 & 0.0460 & 0.0470 & \textbf{0.0255} & 0.0443 & 0.2033 & 0.0477\\
31 & Power-law Fourier alpha=2.2, k0=16; range [-0.2,1.2] & 0.0371 & 0.0302 & 0.0308 & \textbf{0.0160} & 0.0291 & 0.1390 & 0.0283\\
32 & Power-law Fourier alpha=3.5, k0=28; range [0.15,1.25] & 0.0331 & 0.0191 & 0.0194 & \textbf{0.0101} & 0.0188 & 0.0922 & 0.0192\\
33 & Power-law Fourier alpha=3.5, k0=28; range [-0.5,1.5] & 0.0786 & 0.0653 & 0.0656 & \textbf{0.0382} & 0.0614 & 0.2101 & 0.0643\\
34 & Power-law Fourier alpha=4, k0=36; range [-0.5,1.5] & 0.0739 & 0.0670 & 0.0676 & \textbf{0.0407} & 0.0636 & 0.1942 & 0.0654\\
35 & Power-law Fourier alpha=4, k0=36; range [0.15,1.25] & 0.0299 & 0.0196 & 0.0199 & \textbf{0.0109} & 0.0195 & 0.0864 & 0.0192\\
36 & Power-law Fourier alpha=3, k0=24; range [-0.1,1.1] & 0.0328 & 0.0260 & 0.0264 & \textbf{0.0135} & 0.0255 & 0.1174 & 0.0240\\
37 & Power-law Fourier alpha=1.8, k0=12; range [-0.5,1.5] & 0.0846 & 0.0586 & 0.0596 & \textbf{0.0336} & 0.0548 & 0.2270 & 0.0599\\
38 & Power-law Fourier alpha=3, k0=24; range [-0.5,1.5] & 0.0765 & 0.0671 & 0.0675 & \textbf{0.0398} & 0.0635 & 0.2095 & 0.0660\\
39 & Sine mix f=[5, 9, 15, 23], decay=0.25; range [0,1.5] & 0.0512 & 0.0336 & 0.0343 & \textbf{0.0157} & 0.0302 & 0.1141 & 0.0327\\
40 & Sine mix f=[7, 19, 43, 89], decay=0.15; range [0,1.5] & 0.0371 & 0.0280 & 0.0281 & \textbf{0.0096} & 0.0286 & 0.0631 & 0.0285\\
41 & Sine mix f=[5, 9, 15, 23], decay=0.25; range [-0.3,1.3] & 0.0453 & 0.0461 & 0.0464 & \textbf{0.0199} & 0.0396 & 0.1616 & 0.0422\\
42 & Sine mix f=[7, 19, 43, 89], decay=0.15; range [-0.3,1.3] & 0.0458 & 0.0444 & 0.0447 & \textbf{0.0146} & 0.0446 & 0.0857 & 0.0378\\
43 & Sine mix f=[7, 19, 43, 89], decay=0.15; range [-0.2,1.2] & 0.0377 & 0.0359 & 0.0362 & \textbf{0.0118} & 0.0362 & 0.0790 & 0.0309\\
44 & Sine mix f=[5, 9, 15, 23], decay=0.25; range [-0.65,1.35] & 0.0959 & 0.1006 & 0.1023 & \textbf{0.0521} & 0.0918 & 0.2352 & 0.0952\\
45 & Sine mix f=[5, 9, 15, 23], decay=0.25; range [-0.5,1.5] & 0.0635 & 0.0713 & 0.0715 & \textbf{0.0305} & 0.0613 & 0.1940 & 0.0659\\
46 & Sine mix f=[5, 9, 15, 23], decay=0.25; range [-0.2,1.2] & 0.0389 & 0.0353 & 0.0356 & \textbf{0.0161} & 0.0307 & 0.1509 & 0.0321\\
47 & Sine mix f=[7, 19, 43, 89], decay=0.15; range [-0.5,1.5] & 0.0610 & 0.0602 & 0.0606 & \textbf{0.0213} & 0.0602 & 0.1030 & 0.0510\\
48 & Sine mix f=[5, 10, 20, 40, 80], decay=0.9; range [-0.1,1.1] & 0.0230 & 0.0184 & 0.0194 & \textbf{0.0058} & 0.0193 & 0.1406 & 0.0167\\
49 & Sine mix f=[4, 7, 11], decay=0.35; range [-0.3,1.3] & 0.0605 & 0.0413 & 0.0423 & \textbf{0.0164} & 0.0349 & 0.2108 & 0.0401\\
50 & Sine mix f=[7, 19, 43, 89], decay=0.15; range [-0.65,1.35] & 0.0843 & 0.0774 & 0.0779 & \textbf{0.0418} & 0.0776 & 0.1855 & 0.0762\\
51 & Sawtooth freq=2; range [0.15,1.25] & 0.0223 & 0.0147 & 0.0116 & 0.0101 & \textbf{0.0095} & 0.1640 & 0.0119\\
52 & Square wave freq=7, duty=0.35; range [0,1.2] & 0.0802 & 0.0571 & 0.0584 & \textbf{0.0134} & 0.0633 & 0.1245 & 0.0486\\
\end{longtable}
\endgroup

\begingroup\footnotesize
\setlength{\tabcolsep}{2pt}
\renewcommand{\arraystretch}{1.06}
\setlength{\LTcapwidth}{\textwidth}
\begin{longtable}{@{}C{0.034\textwidth} L{0.25\textwidth}@{\hspace{4pt}} *{8}{R{0.065\textwidth}}@{}}
\caption{Darcy Flow relative $L_2$ at the maximum common recorded training-core time 1468.919765 s (24.481996 min). All seven trained methods share this support; the baseline is the static initialization. Rows are two ID train/test references followed by 50 OOD coefficient-field datasets. Bold marks the best trained method per row. These are common-budget results, distinct from the accepted-endpoint attack and ablation records. Each method uses its last evaluated checkpoint at or before the time limit; the selected epochs differ, as recorded in the header.}\label{tab:appendix-darcy-52-relative-l2-time}\\
\toprule
No. & Dataset & Base & L1 & L2 & L3 Full & L3 Stop & Physics & Rand clean & Rand solver\\
\multicolumn{2}{l}{Selected checkpoint epoch} & 0 & 1620 & 1660 & 1855 & 3383 & 2228 & 1159 & 913\\
\multicolumn{2}{l}{Recorded core time (h)} & 0.00000 & 0.40788 & 0.40800 & 0.40784 & 0.40803 & 0.40791 & 0.40782 & 0.40764\\
\midrule
\endfirsthead
\toprule
No. & Dataset & Base & L1 & L2 & L3 Full & L3 Stop & Physics & Rand clean & Rand solver\\
\multicolumn{2}{l}{Selected checkpoint epoch} & 0 & 1620 & 1660 & 1855 & 3383 & 2228 & 1159 & 913\\
\multicolumn{2}{l}{Recorded core time (h)} & 0.00000 & 0.40788 & 0.40800 & 0.40784 & 0.40803 & 0.40791 & 0.40782 & 0.40764\\
\midrule
\endhead
\midrule
\multicolumn{10}{r}{continued on next page}\\
\endfoot
\bottomrule
\endlastfoot
1 & Train, binary GRF alpha=2, tau=3 & 0.0191 & 0.0197 & 0.0217 & 0.0336 & 0.0191 & \textbf{0.0162} & 0.0203 & 0.0163\\
2 & Test, binary GRF alpha=2, tau=3 & 0.0225 & 0.0269 & 0.0289 & 0.0397 & 0.0265 & 0.0281 & 0.0266 & \textbf{0.0253}\\
3 & matern smooth, frac=0.12, alpha=3.659, tau=2.056 & 0.1042 & 0.1065 & 0.1166 & \textbf{0.0541} & 0.0947 & 0.1011 & 0.0936 & 0.1091\\
4 & matern fine, frac=0.24, alpha=1.776, tau=11.864 & 0.0511 & 0.0538 & 0.0554 & 0.0484 & 0.0453 & 0.0521 & \textbf{0.0446} & 0.0553\\
5 & highpass grf, frac=0.20, alpha=1.399, tau=8.090 & 0.0664 & 0.0640 & 0.0721 & \textbf{0.0350} & 0.0536 & 0.0669 & 0.0504 & 0.0739\\
6 & bandpass grf, frac=0.16, alpha=2.165, tau=3.047 & 0.0835 & 0.0872 & 0.0954 & \textbf{0.0450} & 0.0792 & 0.0853 & 0.0764 & 0.0892\\
7 & wave mix, frac=0.12, alpha=2.343, tau=2.217 & 0.1318 & 0.1297 & 0.1383 & \textbf{0.0740} & 0.1166 & 0.1277 & 0.1167 & 0.1345\\
8 & blocky tiles, frac=0.24, alpha=3.854, tau=3.600 & 0.0640 & 0.0699 & 0.0689 & \textbf{0.0509} & 0.0572 & 0.0672 & 0.0591 & 0.0688\\
9 & rectangles, frac=0.20, alpha=1.441, tau=6.328 & 0.0745 & 0.0722 & 0.0797 & \textbf{0.0392} & 0.0607 & 0.0730 & 0.0595 & 0.0797\\
10 & cellular blobs, frac=0.16, alpha=1.974, tau=4.609 & 0.1200 & 0.1208 & 0.1264 & \textbf{0.0683} & 0.1064 & 0.1162 & 0.1080 & 0.1220\\
11 & matern smooth, frac=0.16, alpha=3.557, tau=2.390 & 0.0857 & 0.0830 & 0.0999 & \textbf{0.0422} & 0.0806 & 0.0792 & 0.0717 & 0.0862\\
12 & matern fine, frac=0.12, alpha=1.372, tau=13.359 & 0.1023 & 0.1016 & 0.1085 & \textbf{0.0519} & 0.0858 & 0.0977 & 0.0864 & 0.1059\\
13 & highpass grf, frac=0.24, alpha=1.110, tau=11.727 & 0.0372 & 0.0336 & 0.0410 & 0.0383 & 0.0259 & 0.0379 & \textbf{0.0230} & 0.0439\\
14 & bandpass grf, frac=0.20, alpha=1.805, tau=4.092 & 0.0610 & 0.0611 & 0.0688 & \textbf{0.0413} & 0.0530 & 0.0632 & 0.0488 & 0.0683\\
15 & wave mix, frac=0.16, alpha=2.262, tau=4.329 & 0.1166 & 0.1148 & 0.1236 & \textbf{0.0651} & 0.1034 & 0.1152 & 0.1017 & 0.1219\\
16 & blocky tiles, frac=0.12, alpha=2.005, tau=4.792 & 0.1233 & 0.1231 & 0.1294 & \textbf{0.0707} & 0.1077 & 0.1186 & 0.1093 & 0.1257\\
17 & rectangles, frac=0.24, alpha=1.677, tau=9.583 & 0.1035 & 0.1003 & 0.1095 & \textbf{0.0510} & 0.0889 & 0.0953 & 0.0902 & 0.1013\\
18 & cellular blobs, frac=0.20, alpha=3.272, tau=2.598 & 0.0930 & 0.0920 & 0.1021 & \textbf{0.0504} & 0.0823 & 0.0890 & 0.0805 & 0.0934\\
19 & matern smooth, frac=0.20, alpha=3.632, tau=2.957 & 0.0689 & 0.0765 & 0.0838 & \textbf{0.0430} & 0.0694 & 0.0741 & 0.0659 & 0.0766\\
20 & matern fine, frac=0.16, alpha=1.052, tau=9.110 & 0.0827 & 0.0810 & 0.0874 & \textbf{0.0447} & 0.0674 & 0.0811 & 0.0664 & 0.0888\\
21 & highpass grf, frac=0.12, alpha=1.265, tau=6.963 & 0.0978 & 0.0959 & 0.1037 & \textbf{0.0482} & 0.0813 & 0.0944 & 0.0809 & 0.1032\\
22 & bandpass grf, frac=0.24, alpha=2.182, tau=4.944 & 0.0659 & 0.0642 & 0.0761 & \textbf{0.0463} & 0.0591 & 0.0634 & 0.0570 & 0.0673\\
23 & wave mix, frac=0.20, alpha=2.694, tau=4.607 & 0.0972 & 0.0949 & 0.1034 & \textbf{0.0515} & 0.0850 & 0.0964 & 0.0820 & 0.1031\\
24 & blocky tiles, frac=0.16, alpha=3.295, tau=4.966 & 0.1089 & 0.1095 & 0.1152 & \textbf{0.0602} & 0.0964 & 0.1086 & 0.0966 & 0.1142\\
25 & rectangles, frac=0.12, alpha=1.432, tau=10.224 & 0.1410 & 0.1363 & 0.1466 & \textbf{0.0702} & 0.1210 & 0.1300 & 0.1220 & 0.1390\\
26 & cellular blobs, frac=0.24, alpha=2.775, tau=4.180 & 0.0860 & 0.0883 & 0.0931 & \textbf{0.0516} & 0.0758 & 0.0831 & 0.0757 & 0.0873\\
27 & matern smooth, frac=0.24, alpha=4.169, tau=1.663 & 0.0542 & 0.0515 & 0.0720 & \textbf{0.0336} & 0.0555 & 0.0530 & 0.0440 & 0.0560\\
28 & matern fine, frac=0.20, alpha=1.325, tau=12.825 & 0.0715 & 0.0720 & 0.0766 & \textbf{0.0402} & 0.0593 & 0.0717 & 0.0585 & 0.0777\\
29 & highpass grf, frac=0.16, alpha=1.592, tau=12.836 & 0.0922 & 0.0904 & 0.0986 & \textbf{0.0455} & 0.0781 & 0.0912 & 0.0762 & 0.0990\\
30 & bandpass grf, frac=0.12, alpha=1.897, tau=7.667 & 0.1032 & 0.1025 & 0.1116 & \textbf{0.0559} & 0.0895 & 0.0991 & 0.0878 & 0.1070\\
31 & wave mix, frac=0.24, alpha=1.718, tau=2.966 & 0.0907 & 0.0880 & 0.0976 & \textbf{0.0530} & 0.0817 & 0.0923 & 0.0761 & 0.0977\\
32 & blocky tiles, frac=0.20, alpha=2.062, tau=2.298 & 0.1036 & 0.1051 & 0.1093 & \textbf{0.0610} & 0.0912 & 0.1042 & 0.0924 & 0.1099\\
33 & rectangles, frac=0.16, alpha=1.242, tau=5.532 & 0.1241 & 0.1182 & 0.1299 & \textbf{0.0655} & 0.1066 & 0.1134 & 0.1054 & 0.1212\\
34 & cellular blobs, frac=0.12, alpha=2.193, tau=5.777 & 0.1227 & 0.1219 & 0.1303 & \textbf{0.0716} & 0.1088 & 0.1180 & 0.1094 & 0.1245\\
35 & matern smooth, frac=0.12, alpha=4.068, tau=2.803 & 0.1083 & 0.1119 & 0.1206 & \textbf{0.0561} & 0.1022 & 0.1084 & 0.0995 & 0.1131\\
36 & matern fine, frac=0.24, alpha=1.777, tau=7.674 & 0.0555 & 0.0589 & 0.0609 & \textbf{0.0383} & 0.0464 & 0.0569 & 0.0482 & 0.0607\\
37 & highpass grf, frac=0.20, alpha=1.385, tau=10.283 & 0.0566 & 0.0546 & 0.0622 & \textbf{0.0339} & 0.0439 & 0.0572 & 0.0407 & 0.0640\\
38 & bandpass grf, frac=0.16, alpha=1.272, tau=5.442 & 0.0922 & 0.0912 & 0.0985 & \textbf{0.0499} & 0.0787 & 0.0911 & 0.0773 & 0.0981\\
39 & wave mix, frac=0.12, alpha=1.723, tau=5.416 & 0.1276 & 0.1275 & 0.1316 & \textbf{0.0715} & 0.1081 & 0.1218 & 0.1134 & 0.1289\\
40 & blocky tiles, frac=0.24, alpha=2.238, tau=3.799 & 0.0807 & 0.0791 & 0.0851 & \textbf{0.0483} & 0.0696 & 0.0806 & 0.0676 & 0.0851\\
41 & rectangles, frac=0.20, alpha=1.498, tau=8.369 & 0.1146 & 0.1118 & 0.1193 & \textbf{0.0569} & 0.0986 & 0.1090 & 0.1006 & 0.1148\\
42 & cellular blobs, frac=0.16, alpha=3.472, tau=5.790 & 0.1134 & 0.1114 & 0.1168 & \textbf{0.0613} & 0.0959 & 0.1069 & 0.0993 & 0.1130\\
43 & matern smooth, frac=0.16, alpha=4.674, tau=1.274 & 0.0927 & 0.1007 & 0.1083 & \textbf{0.0454} & 0.0935 & 0.0997 & 0.0903 & 0.1024\\
44 & matern fine, frac=0.12, alpha=1.133, tau=8.361 & 0.1037 & 0.1034 & 0.1097 & \textbf{0.0538} & 0.0880 & 0.1013 & 0.0886 & 0.1095\\
45 & highpass grf, frac=0.24, alpha=1.182, tau=11.687 & 0.0387 & 0.0345 & 0.0425 & 0.0397 & 0.0276 & 0.0388 & \textbf{0.0247} & 0.0447\\
46 & bandpass grf, frac=0.20, alpha=1.226, tau=2.629 & 0.0767 & 0.0775 & 0.0846 & \textbf{0.0419} & 0.0664 & 0.0734 & 0.0651 & 0.0800\\
47 & wave mix, frac=0.16, alpha=1.826, tau=5.317 & 0.1119 & 0.1102 & 0.1186 & \textbf{0.0615} & 0.0982 & 0.1098 & 0.0977 & 0.1151\\
48 & blocky tiles, frac=0.12, alpha=2.727, tau=2.094 & 0.1208 & 0.1211 & 0.1264 & \textbf{0.0712} & 0.1049 & 0.1174 & 0.1085 & 0.1237\\
49 & rectangles, frac=0.24, alpha=1.572, tau=7.630 & 0.0848 & 0.0850 & 0.0899 & \textbf{0.0435} & 0.0715 & 0.0822 & 0.0726 & 0.0867\\
50 & cellular blobs, frac=0.20, alpha=3.355, tau=6.182 & 0.1094 & 0.1084 & 0.1158 & \textbf{0.0622} & 0.0969 & 0.1087 & 0.0990 & 0.1128\\
51 & matern smooth, frac=0.20, alpha=3.309, tau=2.228 & 0.0711 & 0.0793 & 0.0881 & \textbf{0.0419} & 0.0741 & 0.0780 & 0.0704 & 0.0803\\
52 & matern fine, frac=0.16, alpha=1.397, tau=9.918 & 0.0794 & 0.0790 & 0.0867 & \textbf{0.0422} & 0.0678 & 0.0781 & 0.0660 & 0.0840\\
\end{longtable}
\endgroup

\begingroup\footnotesize
\setlength{\tabcolsep}{2pt}
\renewcommand{\arraystretch}{1.06}
\setlength{\LTcapwidth}{\textwidth}
\begin{longtable}{@{}C{0.034\textwidth} L{0.25\textwidth}@{\hspace{4pt}} *{8}{R{0.065\textwidth}}@{}}
\caption{Darcy Flow relative $L_2$ at the maximum common epoch 3000. All seven trained methods share this support; the baseline is the static initialization. Rows are two ID train/test references followed by 50 OOD coefficient-field datasets. Bold marks the best trained method per row. These are common-budget results, distinct from the accepted-endpoint attack and ablation records. The trained columns all use epoch 3000, while their recorded training times differ.}\label{tab:appendix-darcy-52-relative-l2}\\
\toprule
No. & Dataset & Base & L1 & L2 & L3 Full & L3 Stop & Physics & Rand clean & Rand solver\\
\multicolumn{2}{l}{Selected checkpoint epoch} & 0 & 3000 & 3000 & 3000 & 3000 & 3000 & 3000 & 3000\\
\multicolumn{2}{l}{Recorded core time (h)} & 0.00000 & 0.80026 & 0.97706 & 0.65865 & 0.36181 & 0.54918 & 1.07504 & 1.25236\\
\midrule
\endfirsthead
\toprule
No. & Dataset & Base & L1 & L2 & L3 Full & L3 Stop & Physics & Rand clean & Rand solver\\
\multicolumn{2}{l}{Selected checkpoint epoch} & 0 & 3000 & 3000 & 3000 & 3000 & 3000 & 3000 & 3000\\
\multicolumn{2}{l}{Recorded core time (h)} & 0.00000 & 0.80026 & 0.97706 & 0.65865 & 0.36181 & 0.54918 & 1.07504 & 1.25236\\
\midrule
\endhead
\midrule
\multicolumn{10}{r}{continued on next page}\\
\endfoot
\bottomrule
\endlastfoot
1 & Train, binary GRF alpha=2, tau=3 & 0.0191 & 0.0192 & 0.0270 & 0.0358 & 0.0202 & 0.0162 & 0.0287 & \textbf{0.0145}\\
2 & Test, binary GRF alpha=2, tau=3 & 0.0225 & 0.0269 & 0.0291 & 0.0483 & \textbf{0.0244} & 0.0285 & 0.0347 & 0.0264\\
3 & matern smooth, frac=0.12, alpha=3.659, tau=2.056 & 0.1042 & 0.1029 & 0.0880 & \textbf{0.0629} & 0.0913 & 0.0981 & 0.0912 & 0.1045\\
4 & matern fine, frac=0.24, alpha=1.776, tau=11.864 & 0.0511 & 0.0523 & \textbf{0.0410} & 0.0512 & 0.0424 & 0.0516 & 0.0434 & 0.0561\\
5 & highpass grf, frac=0.20, alpha=1.399, tau=8.090 & 0.0664 & 0.0608 & 0.0438 & 0.0458 & 0.0486 & 0.0657 & \textbf{0.0411} & 0.0803\\
6 & bandpass grf, frac=0.16, alpha=2.165, tau=3.047 & 0.0835 & 0.0850 & 0.0710 & \textbf{0.0504} & 0.0748 & 0.0839 & 0.0746 & 0.0892\\
7 & wave mix, frac=0.12, alpha=2.343, tau=2.217 & 0.1318 & 0.1257 & 0.1093 & \textbf{0.0848} & 0.1131 & 0.1254 & 0.1105 & 0.1344\\
8 & blocky tiles, frac=0.24, alpha=3.854, tau=3.600 & 0.0640 & 0.0687 & \textbf{0.0511} & 0.0616 & 0.0538 & 0.0661 & 0.0578 & 0.0675\\
9 & rectangles, frac=0.20, alpha=1.441, tau=6.328 & 0.0745 & 0.0689 & 0.0513 & 0.0517 & 0.0560 & 0.0716 & \textbf{0.0508} & 0.0839\\
10 & cellular blobs, frac=0.16, alpha=1.974, tau=4.609 & 0.1200 & 0.1173 & 0.0984 & \textbf{0.0810} & 0.1027 & 0.1140 & 0.1025 & 0.1204\\
11 & matern smooth, frac=0.16, alpha=3.557, tau=2.390 & 0.0857 & 0.0796 & 0.0730 & \textbf{0.0511} & 0.0767 & 0.0768 & 0.0690 & 0.0809\\
12 & matern fine, frac=0.12, alpha=1.372, tau=13.359 & 0.1023 & 0.0973 & 0.0788 & \textbf{0.0608} & 0.0823 & 0.0950 & 0.0804 & 0.1066\\
13 & highpass grf, frac=0.24, alpha=1.110, tau=11.727 & 0.0372 & 0.0313 & \textbf{0.0215} & 0.0317 & 0.0224 & 0.0373 & 0.0225 & 0.0509\\
14 & bandpass grf, frac=0.20, alpha=1.805, tau=4.092 & 0.0610 & 0.0589 & 0.0440 & 0.0517 & 0.0482 & 0.0623 & \textbf{0.0427} & 0.0727\\
15 & wave mix, frac=0.16, alpha=2.262, tau=4.329 & 0.1166 & 0.1110 & 0.0948 & \textbf{0.0790} & 0.0992 & 0.1134 & 0.0937 & 0.1247\\
16 & blocky tiles, frac=0.12, alpha=2.005, tau=4.792 & 0.1233 & 0.1190 & 0.1008 & \textbf{0.0809} & 0.1041 & 0.1160 & 0.1043 & 0.1244\\
17 & rectangles, frac=0.24, alpha=1.677, tau=9.583 & 0.1035 & 0.0965 & 0.0816 & \textbf{0.0646} & 0.0857 & 0.0934 & 0.0837 & 0.0991\\
18 & cellular blobs, frac=0.20, alpha=3.272, tau=2.598 & 0.0930 & 0.0886 & 0.0744 & \textbf{0.0619} & 0.0787 & 0.0872 & 0.0757 & 0.0898\\
19 & matern smooth, frac=0.20, alpha=3.632, tau=2.957 & 0.0689 & 0.0756 & 0.0606 & \textbf{0.0460} & 0.0647 & 0.0730 & 0.0658 & 0.0766\\
20 & matern fine, frac=0.16, alpha=1.052, tau=9.110 & 0.0827 & 0.0774 & 0.0592 & \textbf{0.0549} & 0.0632 & 0.0793 & 0.0586 & 0.0928\\
21 & highpass grf, frac=0.12, alpha=1.265, tau=6.963 & 0.0978 & 0.0916 & 0.0736 & \textbf{0.0572} & 0.0774 & 0.0922 & 0.0731 & 0.1069\\
22 & bandpass grf, frac=0.24, alpha=2.182, tau=4.944 & 0.0659 & 0.0614 & 0.0520 & \textbf{0.0493} & 0.0556 & 0.0618 & 0.0530 & 0.0648\\
23 & wave mix, frac=0.20, alpha=2.694, tau=4.607 & 0.0972 & 0.0915 & 0.0753 & \textbf{0.0671} & 0.0801 & 0.0950 & 0.0732 & 0.1073\\
24 & blocky tiles, frac=0.16, alpha=3.295, tau=4.966 & 0.1089 & 0.1063 & 0.0875 & \textbf{0.0748} & 0.0920 & 0.1069 & 0.0905 & 0.1158\\
25 & rectangles, frac=0.12, alpha=1.432, tau=10.224 & 0.1410 & 0.1303 & 0.1154 & \textbf{0.0782} & 0.1184 & 0.1268 & 0.1143 & 0.1365\\
26 & cellular blobs, frac=0.24, alpha=2.775, tau=4.180 & 0.0860 & 0.0856 & 0.0682 & \textbf{0.0668} & 0.0722 & 0.0812 & 0.0724 & 0.0842\\
27 & matern smooth, frac=0.24, alpha=4.169, tau=1.663 & 0.0542 & 0.0492 & 0.0462 & \textbf{0.0350} & 0.0505 & 0.0519 & 0.0419 & 0.0517\\
28 & matern fine, frac=0.20, alpha=1.325, tau=12.825 & 0.0715 & 0.0689 & \textbf{0.0504} & 0.0522 & 0.0547 & 0.0703 & 0.0517 & 0.0803\\
29 & highpass grf, frac=0.16, alpha=1.592, tau=12.836 & 0.0922 & 0.0866 & 0.0689 & \textbf{0.0592} & 0.0735 & 0.0895 & 0.0672 & 0.1040\\
30 & bandpass grf, frac=0.12, alpha=1.897, tau=7.667 & 0.1032 & 0.0988 & 0.0828 & \textbf{0.0657} & 0.0864 & 0.0968 & 0.0822 & 0.1085\\
31 & wave mix, frac=0.24, alpha=1.718, tau=2.966 & 0.0907 & 0.0854 & 0.0711 & 0.0741 & 0.0766 & 0.0916 & \textbf{0.0659} & 0.1030\\
32 & blocky tiles, frac=0.20, alpha=2.062, tau=2.298 & 0.1036 & 0.1021 & 0.0818 & \textbf{0.0782} & 0.0867 & 0.1025 & 0.0856 & 0.1108\\
33 & rectangles, frac=0.16, alpha=1.242, tau=5.532 & 0.1241 & 0.1134 & 0.1005 & \textbf{0.0766} & 0.1041 & 0.1107 & 0.0981 & 0.1193\\
34 & cellular blobs, frac=0.12, alpha=2.193, tau=5.777 & 0.1227 & 0.1180 & 0.1020 & \textbf{0.0821} & 0.1054 & 0.1155 & 0.1047 & 0.1230\\
35 & matern smooth, frac=0.12, alpha=4.068, tau=2.803 & 0.1083 & 0.1094 & 0.0946 & \textbf{0.0631} & 0.0982 & 0.1063 & 0.0992 & 0.1124\\
36 & matern fine, frac=0.24, alpha=1.777, tau=7.674 & 0.0555 & 0.0568 & \textbf{0.0394} & 0.0473 & 0.0423 & 0.0555 & 0.0456 & 0.0596\\
37 & highpass grf, frac=0.20, alpha=1.385, tau=10.283 & 0.0566 & 0.0515 & 0.0346 & 0.0392 & 0.0390 & 0.0560 & \textbf{0.0324} & 0.0703\\
38 & bandpass grf, frac=0.16, alpha=1.272, tau=5.442 & 0.0922 & 0.0878 & 0.0703 & \textbf{0.0629} & 0.0745 & 0.0893 & 0.0703 & 0.1017\\
39 & wave mix, frac=0.12, alpha=1.723, tau=5.416 & 0.1276 & 0.1233 & 0.1019 & \textbf{0.0815} & 0.1052 & 0.1189 & 0.1090 & 0.1268\\
40 & blocky tiles, frac=0.24, alpha=2.238, tau=3.799 & 0.0807 & 0.0762 & \textbf{0.0601} & 0.0633 & 0.0647 & 0.0795 & 0.0603 & 0.0862\\
41 & rectangles, frac=0.20, alpha=1.498, tau=8.369 & 0.1146 & 0.1075 & 0.0896 & \textbf{0.0690} & 0.0942 & 0.1070 & 0.0925 & 0.1131\\
42 & cellular blobs, frac=0.16, alpha=3.472, tau=5.790 & 0.1134 & 0.1071 & 0.0883 & \textbf{0.0717} & 0.0921 & 0.1045 & 0.0933 & 0.1102\\
43 & matern smooth, frac=0.16, alpha=4.674, tau=1.274 & 0.0927 & 0.0991 & 0.0848 & \textbf{0.0521} & 0.0887 & 0.0983 & 0.0909 & 0.1040\\
44 & matern fine, frac=0.12, alpha=1.133, tau=8.361 & 0.1037 & 0.0993 & 0.0803 & \textbf{0.0645} & 0.0841 & 0.0990 & 0.0819 & 0.1123\\
45 & highpass grf, frac=0.24, alpha=1.182, tau=11.687 & 0.0387 & 0.0322 & \textbf{0.0232} & 0.0329 & 0.0243 & 0.0381 & 0.0239 & 0.0515\\
46 & bandpass grf, frac=0.20, alpha=1.226, tau=2.629 & 0.0767 & 0.0745 & 0.0582 & \textbf{0.0503} & 0.0621 & 0.0717 & 0.0606 & 0.0808\\
47 & wave mix, frac=0.16, alpha=1.826, tau=5.317 & 0.1119 & 0.1067 & 0.0900 & \textbf{0.0742} & 0.0943 & 0.1080 & 0.0908 & 0.1153\\
48 & blocky tiles, frac=0.12, alpha=2.727, tau=2.094 & 0.1208 & 0.1172 & 0.0984 & \textbf{0.0813} & 0.1016 & 0.1149 & 0.1045 & 0.1228\\
49 & rectangles, frac=0.24, alpha=1.572, tau=7.630 & 0.0848 & 0.0823 & 0.0627 & \textbf{0.0573} & 0.0675 & 0.0811 & 0.0660 & 0.0883\\
50 & cellular blobs, frac=0.20, alpha=3.355, tau=6.182 & 0.1094 & 0.1054 & 0.0879 & \textbf{0.0790} & 0.0927 & 0.1071 & 0.0930 & 0.1124\\
51 & matern smooth, frac=0.20, alpha=3.309, tau=2.228 & 0.0711 & 0.0783 & 0.0655 & \textbf{0.0412} & 0.0691 & 0.0769 & 0.0709 & 0.0811\\
52 & matern fine, frac=0.16, alpha=1.397, tau=9.918 & 0.0794 & 0.0759 & 0.0595 & \textbf{0.0512} & 0.0636 & 0.0765 & 0.0601 & 0.0860\\
\end{longtable}
\endgroup

\Needspace{7\baselineskip}
\subsection*{Matched Budgets, ID/OOD Tradeoffs, and NS Coverage}
\begin{multicols}{2}
The two ID rows and 50 OOD rows in each endpoint table answer different
questions. Full L3 improves shifted-input accuracy in these studies without
being the clean ID winner. At Burgers epoch 1000, its held-out ID relative
$L_2$ is 0.004643 versus 0.003364 for L1, whereas the OOD50 means are 0.023001
and 0.044703. The table below reports both epoch and runtime controls using
actual evaluated checkpoints at or before the common time limit. The historical
robustness table and the accepted Darcy attack/ablation records retain their own
checkpoint selections; their values are not substituted into these controls.

Random clean and random solver have different supervision. Random clean keeps
$\mathcal G(x)$ after changing the input and can create a mismatched label; its
Burgers epoch-1000 OOD mean is 0.162485. Random solver computes
$\mathcal G(x+\delta)$ for newly sampled perturbations in each batch. It avoids
the iterative attack search and completes more epochs at the same runtime,
but its sampling law does not adapt to the current model error. In the strict
Burgers time control, random solver reaches OOD50 error 0.037015 versus
0.023001 for Full L3. Correct relabeling and training throughput help the random
baseline, while the controlled outcome supports an additional benefit from
discrepancy-guided selection.

The finite network's allocation of approximation error across familiar and
shifted inputs is a plausible explanation of the ID/OOD tradeoff. The present
studies do not vary width, Fourier modes, depth, and data mixture independently,
so they do not establish an intrinsic capacity limit or a causal explanation.
Deployment should weigh ID accuracy, expected input shifts, robustness, and
time/memory together. The recorded trajectories show finite-budget learning
behavior, not a proof of asymptotic convergence or global optimality.
The quoted dispersion and paired tests concern inputs or timing repeats;
they do not quantify uncertainty over independent training seeds. Full external
training-pipeline reproductions and independent-seed quality studies remain
outside the completed comparisons.
\end{multicols}

\begin{table}[H]
\centering\footnotesize
\setlength{\tabcolsep}{3pt}
\renewcommand{\arraystretch}{1.1}
\caption{Burgers matched-budget summary. Time uses the last evaluated checkpoint at or before 7.427461557 h; Epoch uses the maximum common epoch 1000. ID is the held-out test split and OOD is the equal-weight mean of 50 shifted datasets. Full per-dataset values are listed in the preceding two Burgers tables.}\label{tab:revision-burgers-alignment}
\begin{tabular*}{\textwidth}{@{\extracolsep{\fill}}llrrrr@{}}
\toprule
Control & Method & Epoch & Core h & ID test & OOD50\\
\midrule
Time & L1 & 2756 & 7.424844 & 0.002830 & 0.043012\\
Time & L2 & 1499 & 7.426700 & 0.003053 & 0.044178\\
Time & L3 Full & 1000 & 7.427462 & 0.004643 & 0.023001\\
Time & L3 Stop & 2562 & 7.426865 & 0.002385 & 0.040487\\
Time & Random solver & 7859 & 7.426909 & 0.001648 & 0.037015\\
Epoch & L1 & 1000 & 2.696358 & 0.003364 & 0.044703\\
Epoch & L2 & 1000 & 4.942515 & 0.005542 & 0.045166\\
Epoch & L3 Full & 1000 & 7.427462 & 0.004643 & 0.023001\\
Epoch & L3 Stop & 1000 & 2.950285 & 0.005567 & 0.042295\\
Epoch & Random solver & 1000 & 0.939732 & 0.003786 & 0.043897\\
\bottomrule
\end{tabular*}
\end{table}

\begin{table}[H]
\centering\footnotesize
\setlength{\tabcolsep}{3pt}
\renewcommand{\arraystretch}{1.1}
\caption{Independent accepted Darcy Full/Stop ablation pairs (same seed). Equal updates holds 6,066 optimizer steps fixed; equal time uses approximately 0.408 core h. These ablation records do not replace the four common-budget 52-dataset tables.}\label{tab:revision-darcy-full-stop-alignment}
\begin{tabular*}{\textwidth}{@{\extracolsep{\fill}}llrrrrrr@{}}
\toprule
Control & Method & Epoch & Steps & Core h & ID test & OOD50 & Attack $\Delta$MSE\\
\midrule
Updates & L3 Full & 3033 & 6066 & 0.407926 & 0.036684 & 0.055236 & 2.101983e-7\\
Updates & L3 Stop & 3033 & 6066 & 0.365786 & 0.022761 & 0.079063 & 2.622299e-7\\
Time & L3 Full & 3033 & 6066 & 0.407926 & 0.036684 & 0.055236 & 2.101983e-7\\
Time & L3 Stop & 3383 & 6766 & 0.408033 & 0.022815 & 0.078479 & 2.914580e-7\\
\bottomrule
\end{tabular*}
\end{table}

\Needspace{12\baselineskip}
\begin{multicols}{2}
\subsection*{Navier--Stokes final-frame accuracy}
For input $j$, compare the predicted and numerical-solver final vorticity fields
at frame 19 on the $256\times256$ grid. Define
\begin{equation}
 \begin{aligned}
 \mathrm{RMSE}_m^{(j)}&=
 \left(256^{-2}\sum_{a,b}|e_{19,m}^{(j)}(a,b)|^2\right)^{1/2},\\
 e_{19,m}^{(j)}&=\hat\omega_{19,m}^{(j)}-\omega_{19}^{G,(j)}.
 \end{aligned}
\end{equation}
For each dataset $D$, first average this samplewise RMSE, then compute
\begin{equation}
 \Delta\mathrm{RMSE}(D)=
 |D|^{-1}\sum_{j\in D}
 \bigl(\mathrm{RMSE}_{6}^{(j)}-\mathrm{RMSE}_{\rm base}^{(j)}\bigr).
\end{equation}
Negative values mean improvement after six Full-L3 rounds. This is a
dataset-mean RMSE statistic, not the square root of a pooled MSE. The retained
402-dataset GRF summary omits 80 pre-existing configurations with input range
$[-1.5,-1.0]$. Those cases are unfavorable and are included explicitly in the
expanded summaries below. The GRF/LogGRF/NegLogGRF total counts the common
train/test references once. The data figure is
\figref{fig:app-g-ns-training-delta-rmse}.

Only the six-round quality comparison was completed. The multiple NS attack
objectives and short training-batch probes characterize attack quality and
computational cost, respectively; they do not supply a converged multi-method
NS training study. The prewarmed V100 cost extrapolations in
\tabref{tab:revision-ns-500epoch-budget} explain the remaining resource limit.
\end{multicols}
\begin{table}[H]
\centering\footnotesize
\setlength{\tabcolsep}{3pt}
\renewcommand{\arraystretch}{1.1}
\caption{NS round-six minus baseline dataset-mean final-frame RMSE, with favorable and unfavorable cases retained. The retained GRF subset excludes the 80 configurations with input range $[-1.5,-1.0]$. The broader pool counts the train/test references once.}\label{tab:revision-ns-quality-scope}
\begin{tabular*}{\textwidth}{@{\extracolsep{\fill}}lrrrrr@{}}
\toprule
Evaluation pool & Datasets & Improved & Worsened & Improved & Mean $\Delta$RMSE\\
\midrule
All GRF & 482 & 366 & 116 & 75.93\% & -0.014998\\
Retained GRF & 402 & 364 & 38 & 90.55\% & -0.046230\\
Excluded GRF range & 80 & 2 & 78 & 2.50\% & +0.141938\\
GRF + LogGRF + NegLogGRF & 1442 & 1015 & 427 & 70.39\% & -0.011762\\
\bottomrule
\end{tabular*}
\end{table}

\begingroup
\captionsetup[figure]{font=small,skip=2pt}
\setlength{\intextsep}{5pt}
\setlength{\textfloatsep}{5pt}
\setlength{\floatsep}{5pt}
\subsection*{Adversarial Training Pipelines}

\begin{figure}[H]
  \centering
  \includegraphics[width=0.92\textwidth]{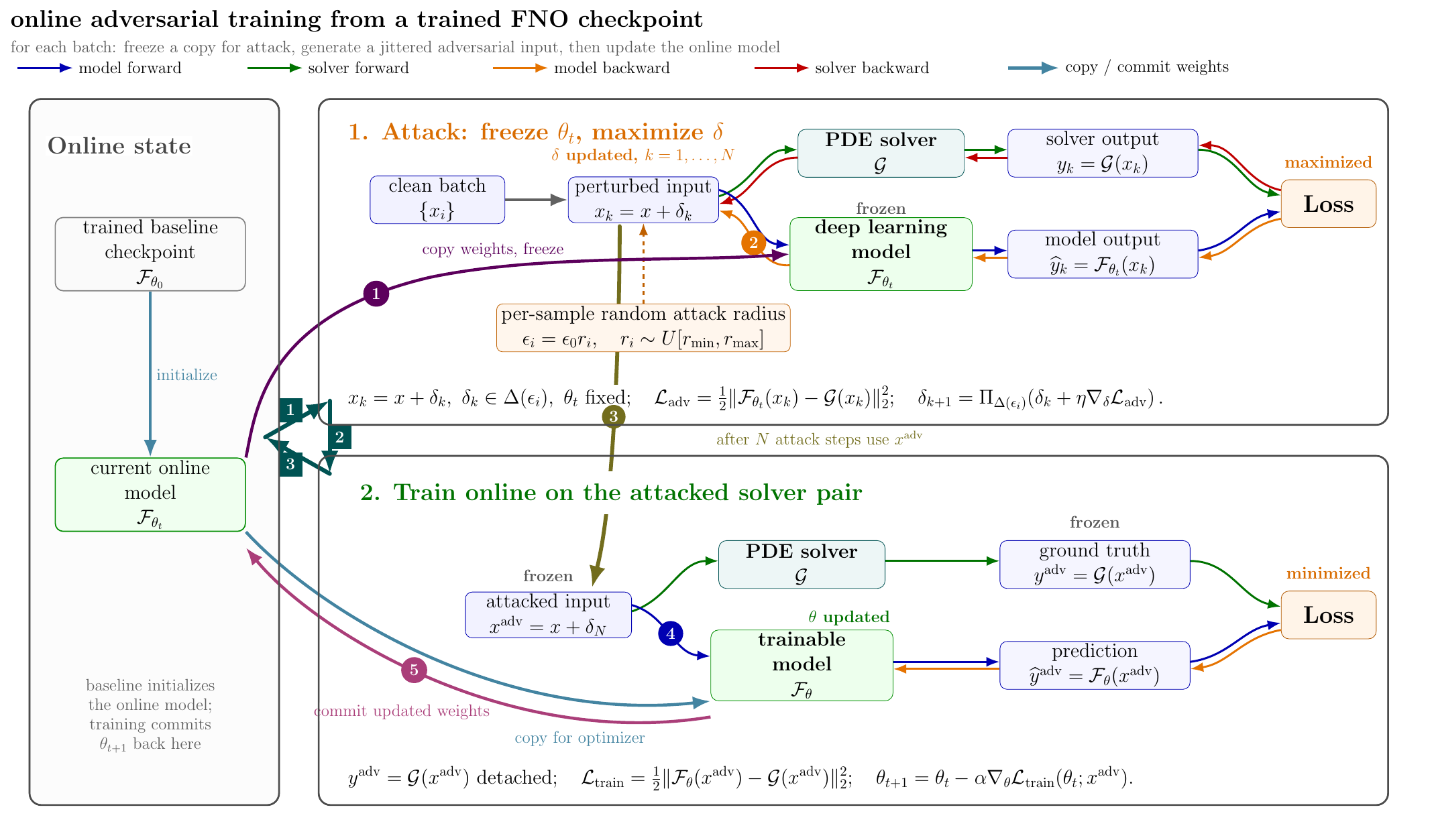}
  \caption{Solver-integrated adversarial training pipeline. The current model
  is copied into the attack stage, adversarial inputs are generated, and the
  trainable model is updated on those attacked samples before becoming the next
  current model.}
  \label{fig:adversarial-training-pipeline}
\end{figure}

\begin{figure}[H]
  \centering
  \includegraphics[width=0.92\textwidth]{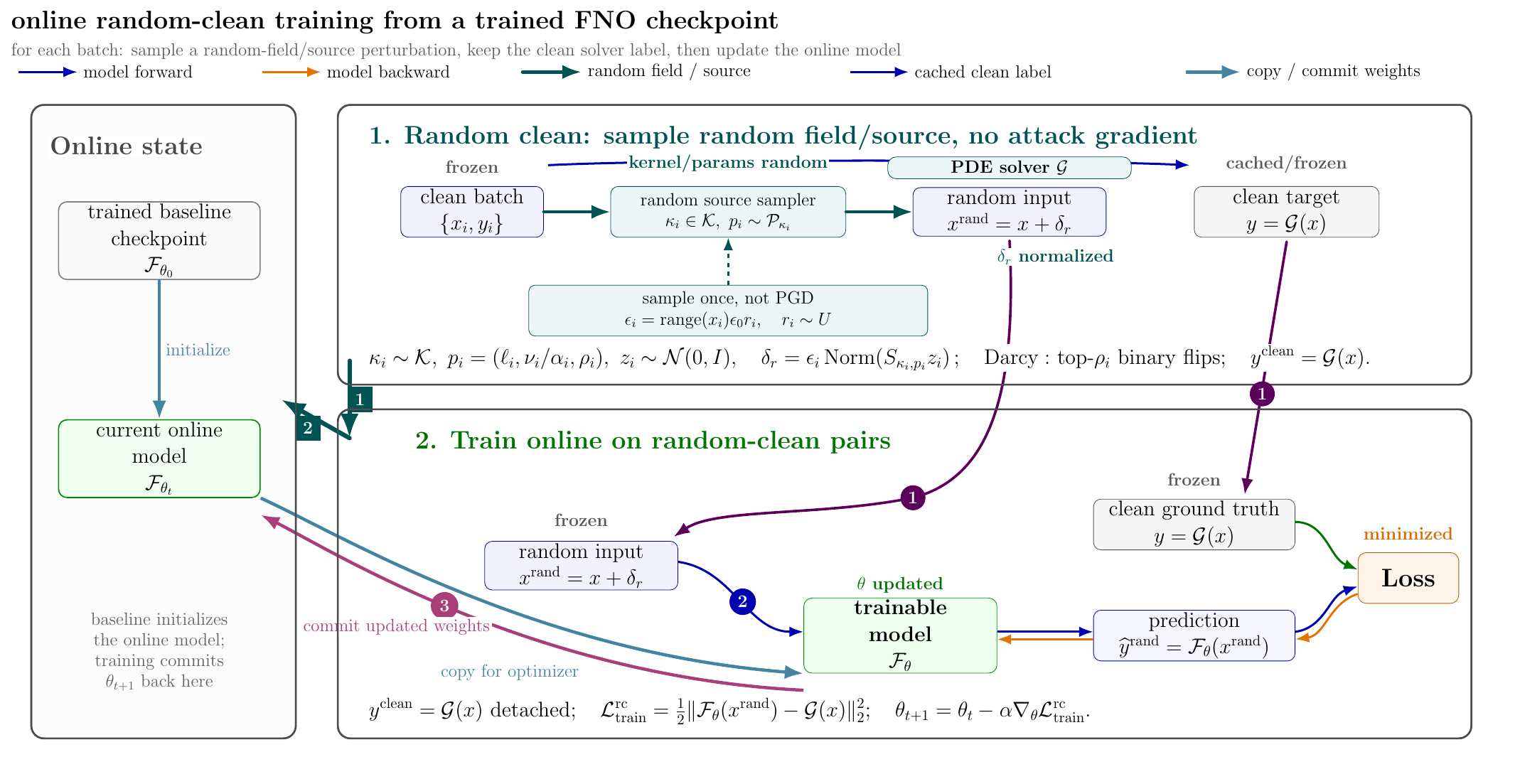}
  \caption{Random-clean training pipeline. Perturbations are randomly sampled,
  and the perturbed inputs are trained against the original clean solver target.}
  \label{fig:random-clean-training-pipeline}
\end{figure}

\begin{figure}[H]
  \centering
  \includegraphics[width=0.92\textwidth]{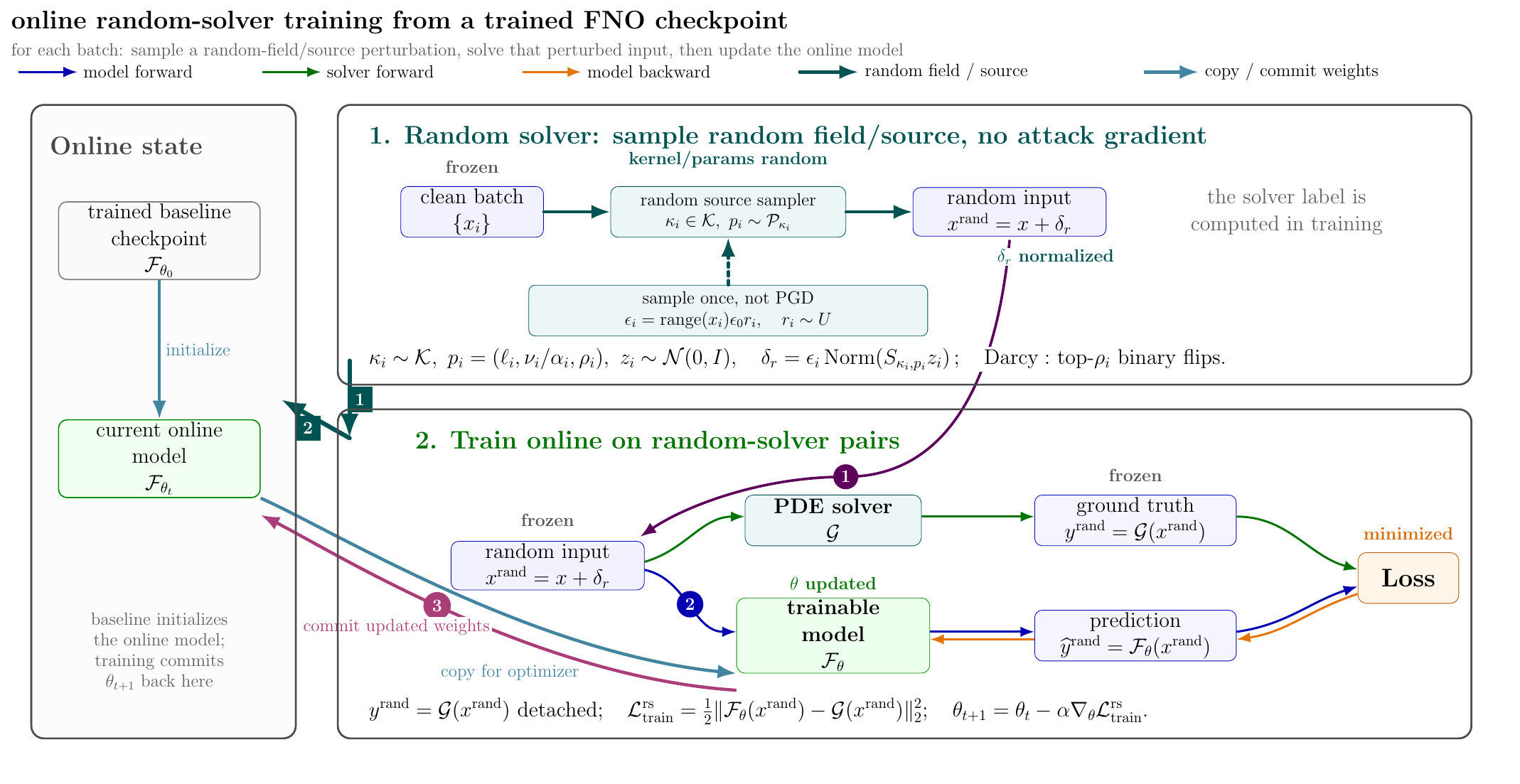}
  \caption{Random-solver training pipeline. Perturbations are randomly sampled,
  and each perturbed input is relabeled by the solver output at that perturbed
  input.}
  \label{fig:random-solver-training-pipeline}
\end{figure}
\endgroup

\begingroup
\captionsetup[figure]{font=small,skip=2pt}
\setlength{\intextsep}{5pt}
\setlength{\textfloatsep}{5pt}
\setlength{\floatsep}{5pt}
\nextappendixfigureheading{
\appendixsubsection{Solver-Integrated Adversarial Training}

\appendixdomainheading{Burgers}

\noindent\textbf{Common-time views: horizontal axis = measured hours.}
}

\begin{figure}[H]
  \useappendixfigureheading
  \centering
  \captionsetup{font=footnotesize,skip=1pt}
  \includegraphics[width=\textwidth,height=0.88\textheight,keepaspectratio]{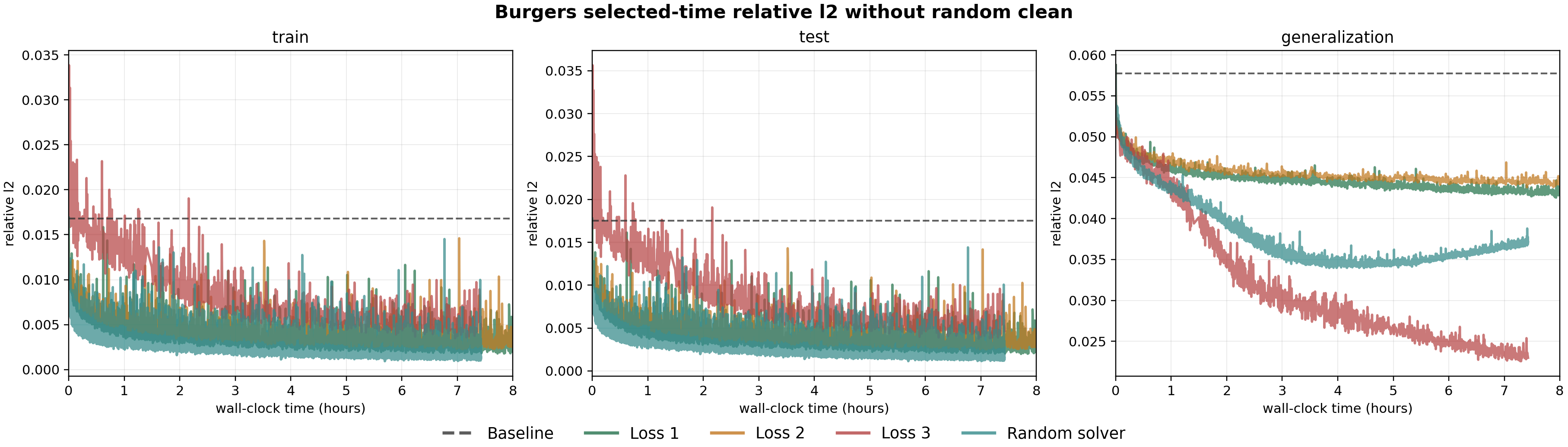}
  \caption{Burgers mean relative-$L_2$ trajectories against measured wall-clock time
  in hours: the supplied original train/test/OOD overview. Its original
  five-model legend contains the static baseline, L1, L2, Full L3 (named
  ``Loss 3'' in the image), and random solver. The curves retain their
  recorded time ranges. This historical overview does not include L3 Stop;
  the following updated panels include both Full and Stop on all 50 OOD
  datasets through the strict retained-method common time 7.427462 h.
  The common-epoch figures are a separate group below.}
  \label{fig:revision-burgers-common-time-mean-original}
\end{figure}

\begin{figure}[H]
  \useappendixfigureheading
  \centering
  \captionsetup{font=small,skip=1pt}
  \includegraphics[width=0.99\textwidth,height=0.80\textheight,keepaspectratio]{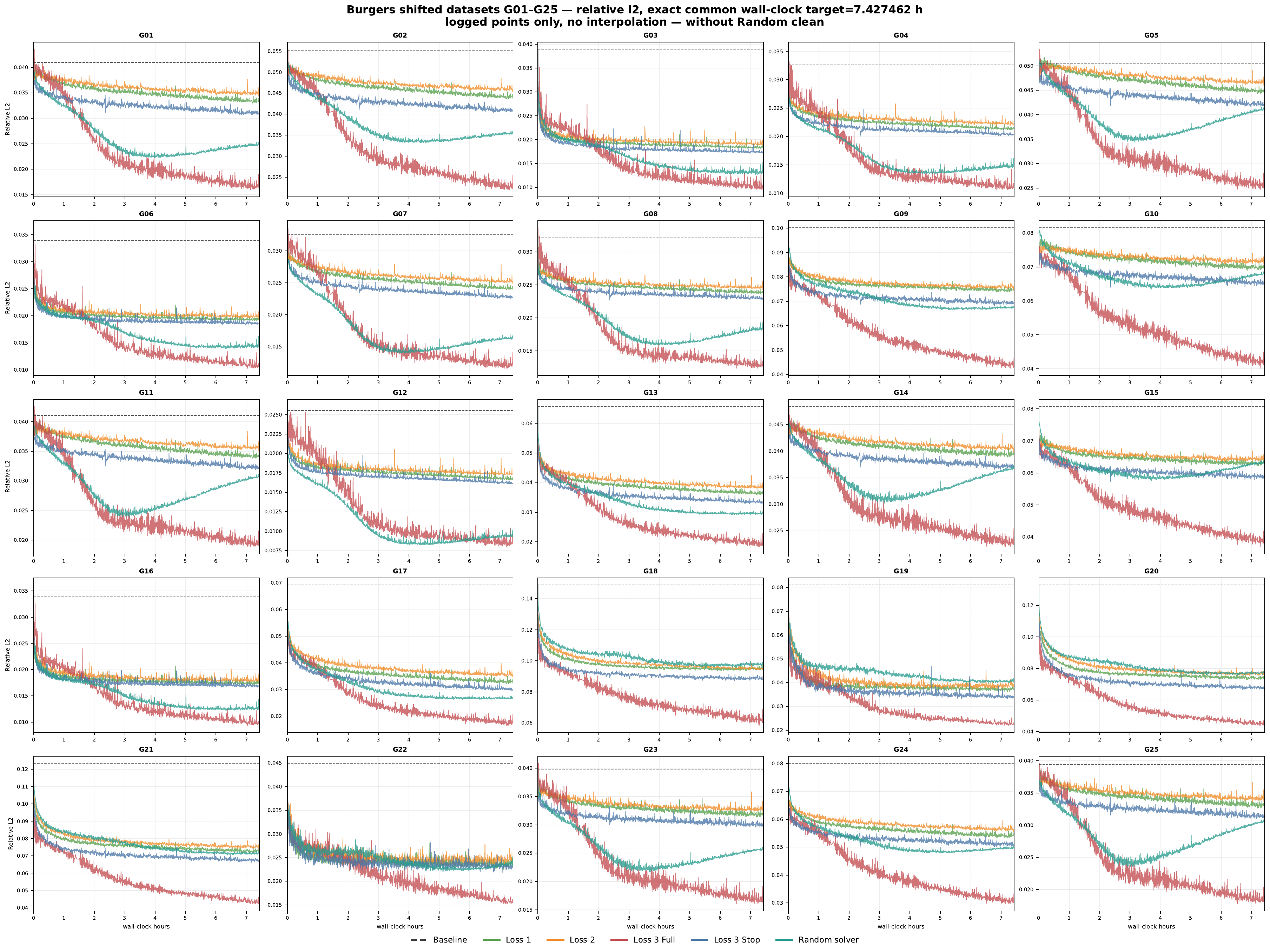}
  \caption{Burgers OOD datasets 1--25 on a wall-clock-hours horizontal axis through
  the retained-method common target 7.427462 h. This supplied time-axis view
  includes the baseline, L1, L2, L3 Full, L3 Stop, and random solver. The
  corresponding endpoint table is
  \tabref{tab:appendix-burgers-52-relative-l2-time}.}
  \label{fig:revision-burgers-common-time-1-25}
\end{figure}

\begin{figure}[H]
  \useappendixfigureheading
  \centering
  \captionsetup{font=small,skip=1pt}
  \includegraphics[width=0.99\textwidth,height=0.80\textheight,keepaspectratio]{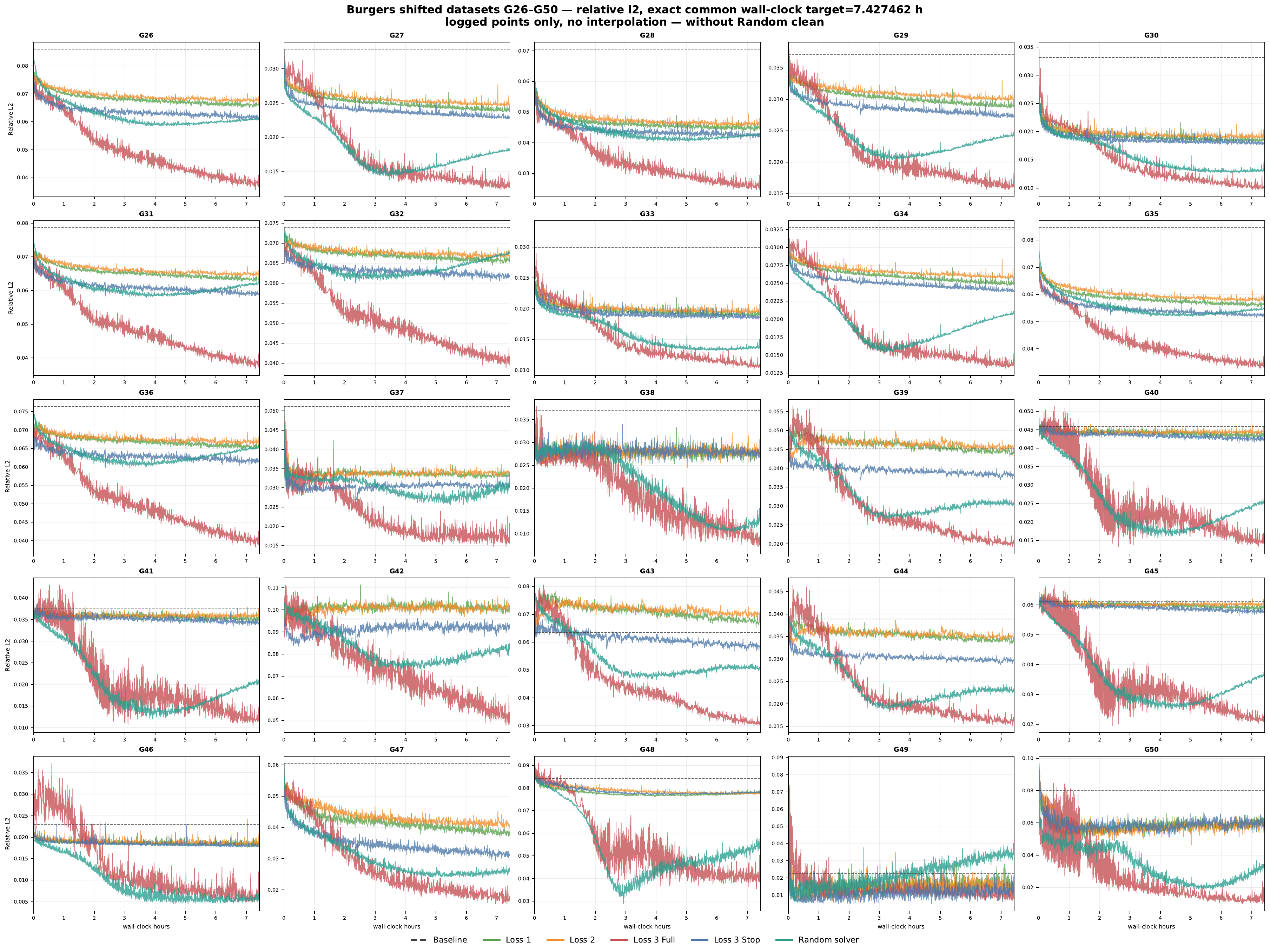}
  \caption{Burgers OOD datasets 26--50 using wall-clock hours and the same
  7.427462-h common target. Time is the independent axis in these panels;
  the following companion views instead use epoch.}
  \label{fig:revision-burgers-common-time-26-50}
\end{figure}

\begin{figure}[H]
  \useappendixfigureheading
  \centering
  \captionsetup{font=small,skip=1pt}
  \includegraphics[width=0.99\textwidth,height=0.80\textheight,keepaspectratio]{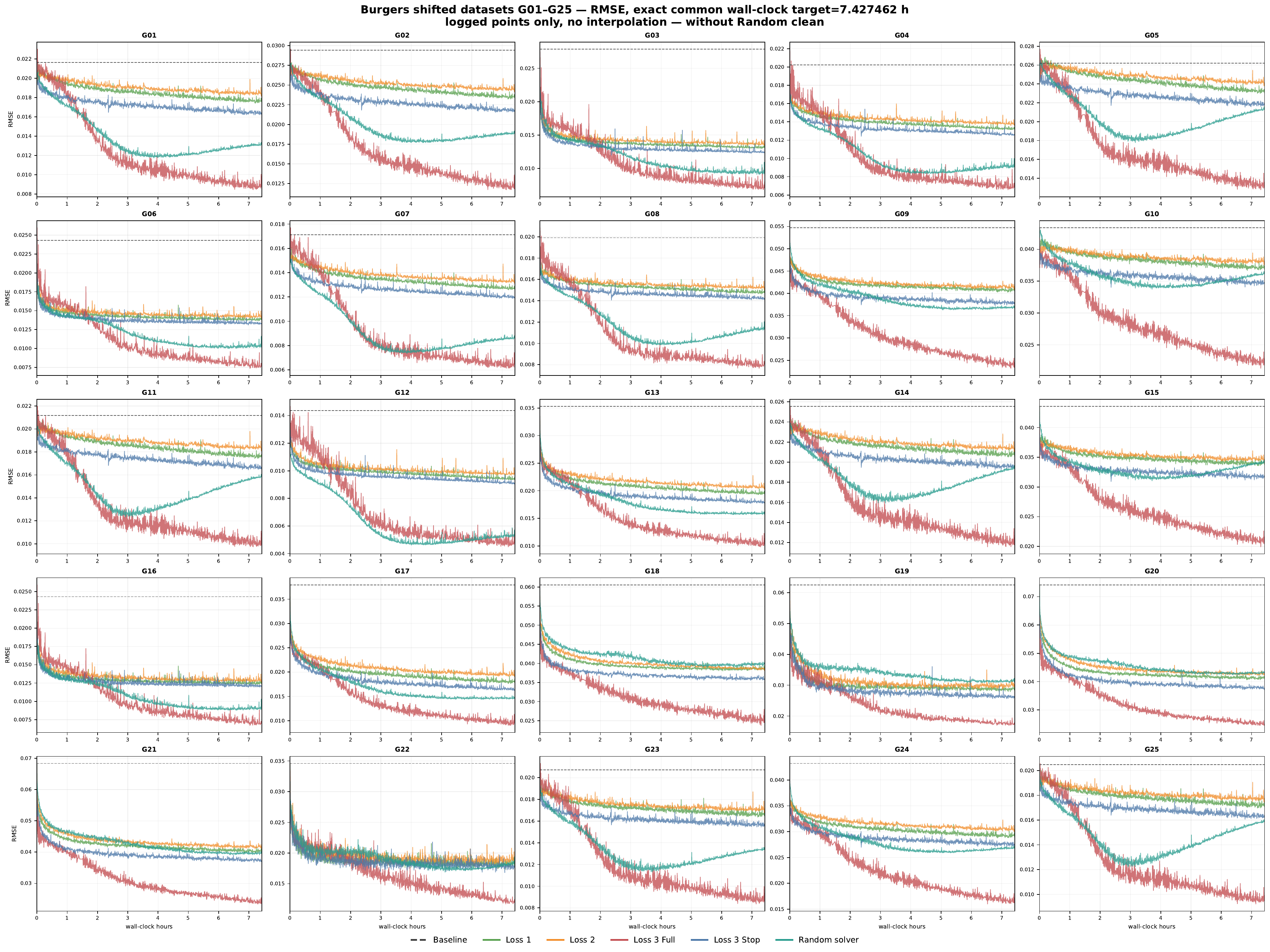}
  \caption{Burgers OOD datasets 1--25: the supplied RMSE companion on a wall-clock-hours
  axis through the strict common target 7.427462 h. The methods are baseline,
  L1, L2, L3 Full, L3 Stop, and random solver. It uses actual method-specific
  training times, rather than relabeling an epoch axis.}
  \label{fig:revision-burgers-common-time-rmse-1-25}
\end{figure}

\begin{figure}[H]
  \useappendixfigureheading
  \centering
  \captionsetup{font=small,skip=1pt}
  \includegraphics[width=0.99\textwidth,height=0.80\textheight,keepaspectratio]{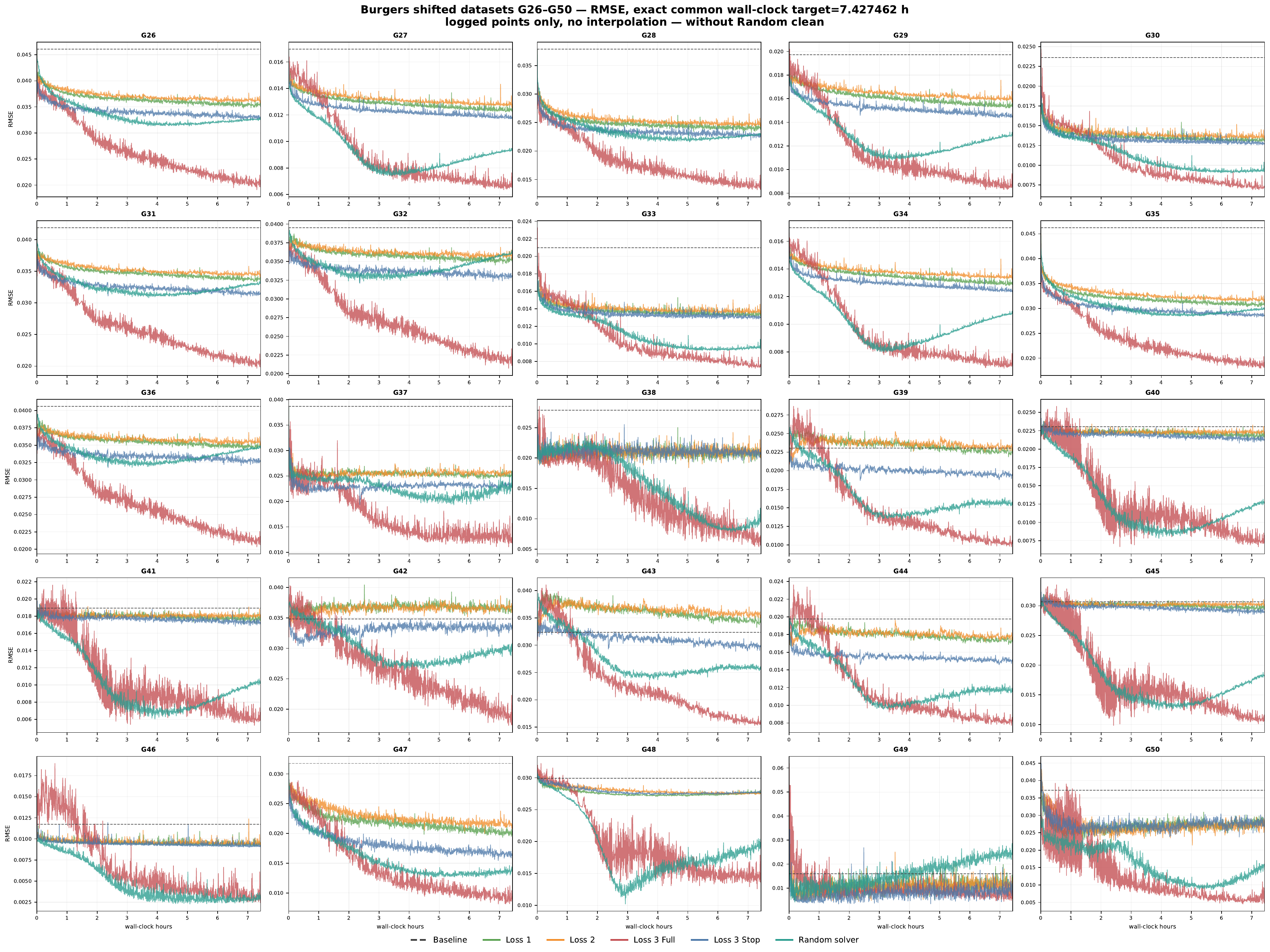}
  \caption{Burgers OOD datasets 26--50: RMSE against measured wall-clock hours with
  the same 7.427462-h common limit and Full/Stop comparison. The following
  figures switch the horizontal coordinate to epoch.}
  \label{fig:revision-burgers-common-time-rmse-26-50}
\end{figure}

\nextappendixfigureheading{
\noindent\textbf{Common-epoch views: horizontal axis = epoch.}
}

\begin{figure}[H]
  \useappendixfigureheading
  \centering
  \includegraphics[width=0.96\textwidth,height=0.82\textheight,keepaspectratio]{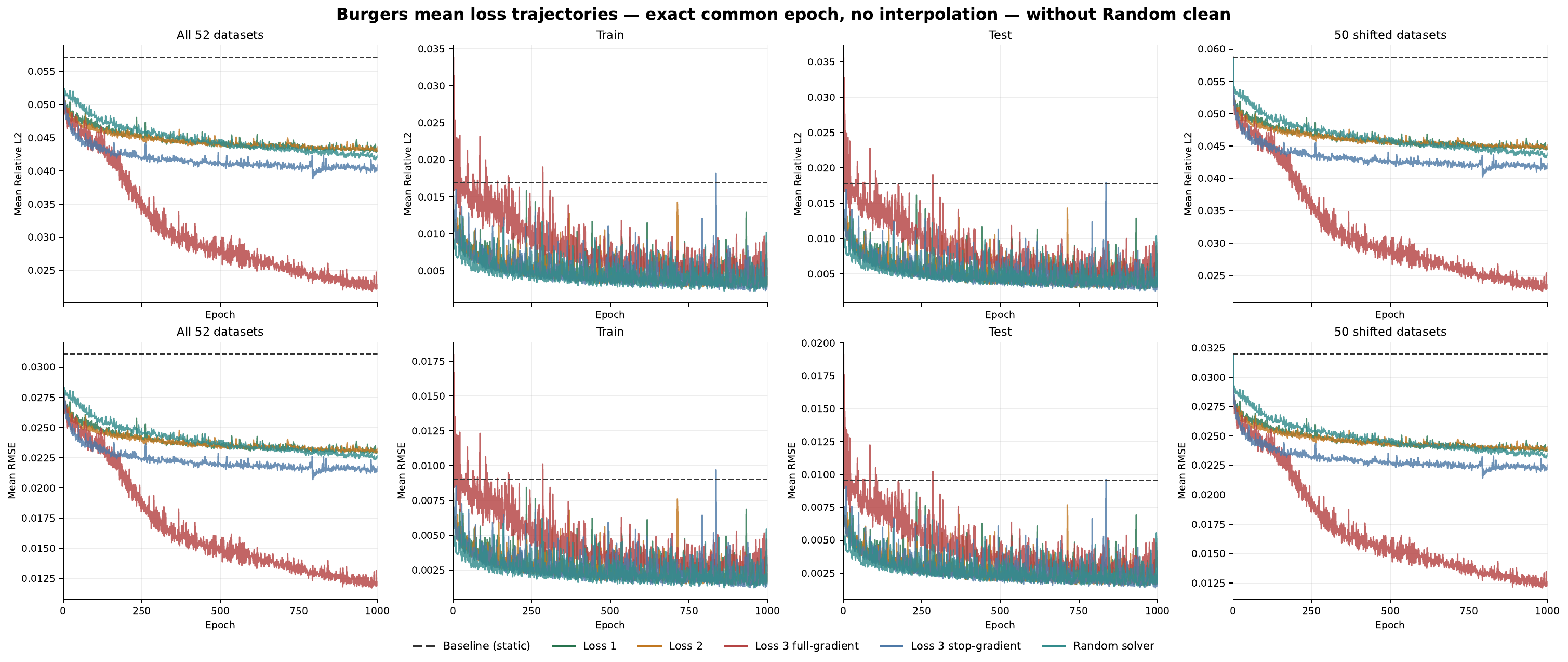}
  \caption{Burgers relative-$L_2$ mean curves through the maximum common epoch 1000.
  The panels compare the shared baseline, Loss 1, Loss 2, Loss 3 Full, Loss 3
  Stop, and random solver on the train split, test split, and 50 shifted OOD
  datasets. Random clean is omitted in this rescaled view so that the other
  curves remain readable.}
  \label{fig:app-g-burgers-training-selected-time}
\end{figure}

\begin{figure}[H]
  \useappendixfigureheading
  \centering
  \captionsetup{font=small,skip=1pt}
  \includegraphics[width=0.99\textwidth,height=0.80\textheight,keepaspectratio]{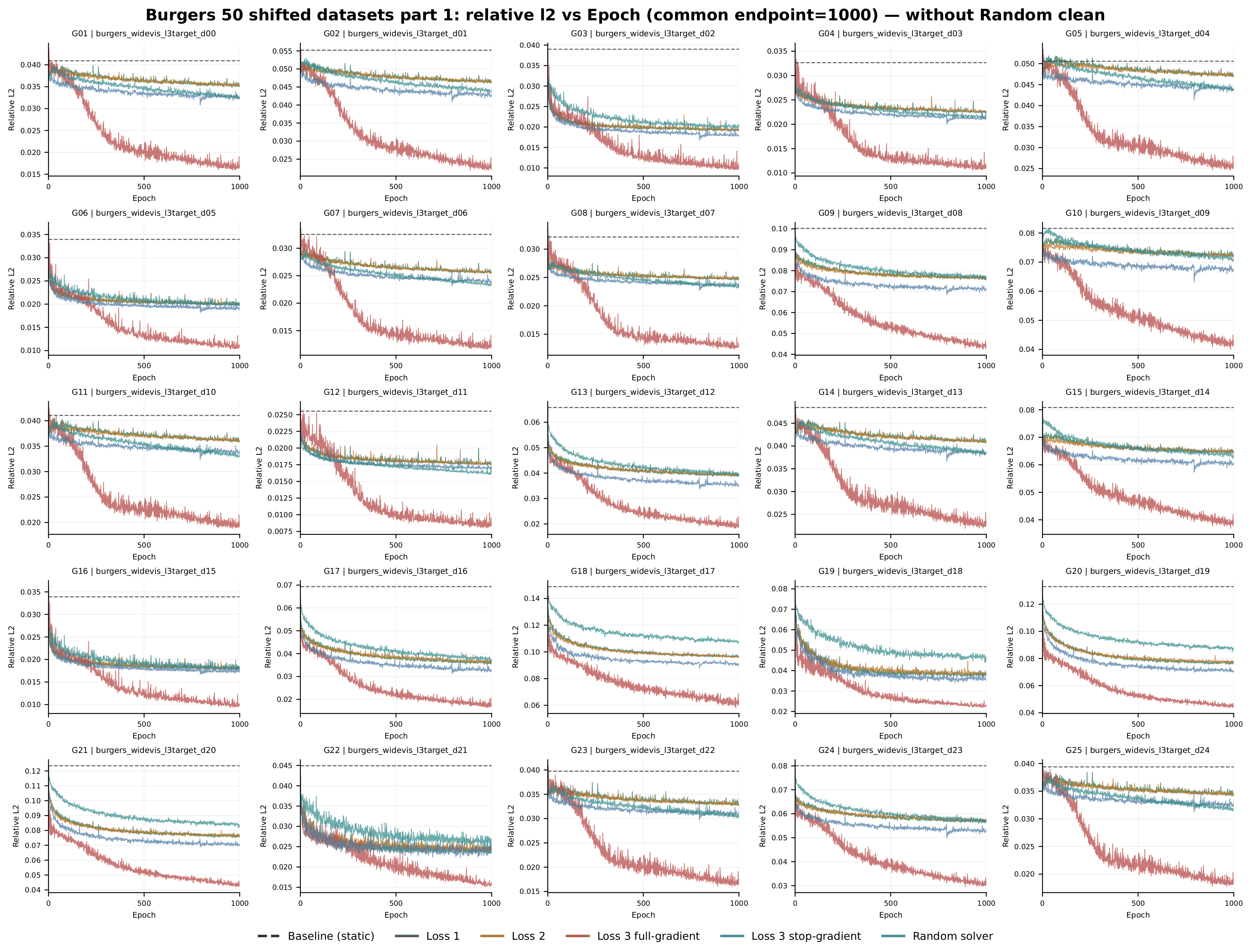}
  \caption{Burgers relative-$L_2$ curves for shifted datasets 1--25 through exact epoch
  1000. This rescaled view includes Loss 3 Full and Loss 3 Stop and omits random
  clean. The full-gradient objective gives the most consistent reductions over
  these held-out input distributions.}
  \label{fig:app-g-burgers-training-gen-1-25}
\end{figure}

\begin{figure}[H]
  \useappendixfigureheading
  \centering
  \captionsetup{font=small,skip=1pt}
  \includegraphics[width=0.99\textwidth,height=0.80\textheight,keepaspectratio]{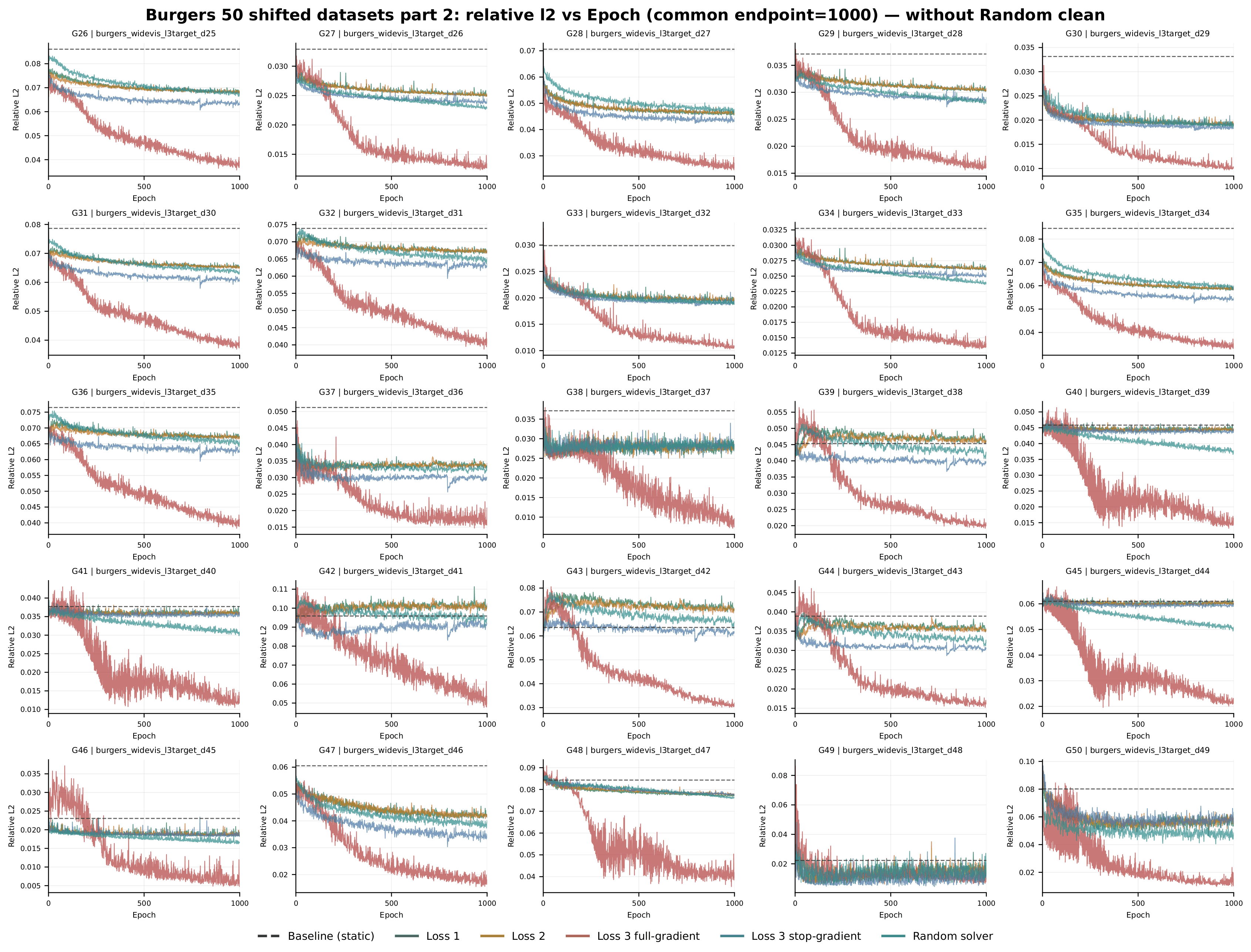}
  \caption{Burgers relative-$L_2$ curves for shifted datasets 26--50 through exact
  epoch 1000. This rescaled view includes Loss 3 Full and Loss 3 Stop and omits
  random clean.}
  \label{fig:app-g-burgers-training-gen-26-50}
\end{figure}

\begin{figure}[H]
  \useappendixfigureheading
  \centering
  \includegraphics[width=0.96\textwidth,height=0.82\textheight,keepaspectratio]{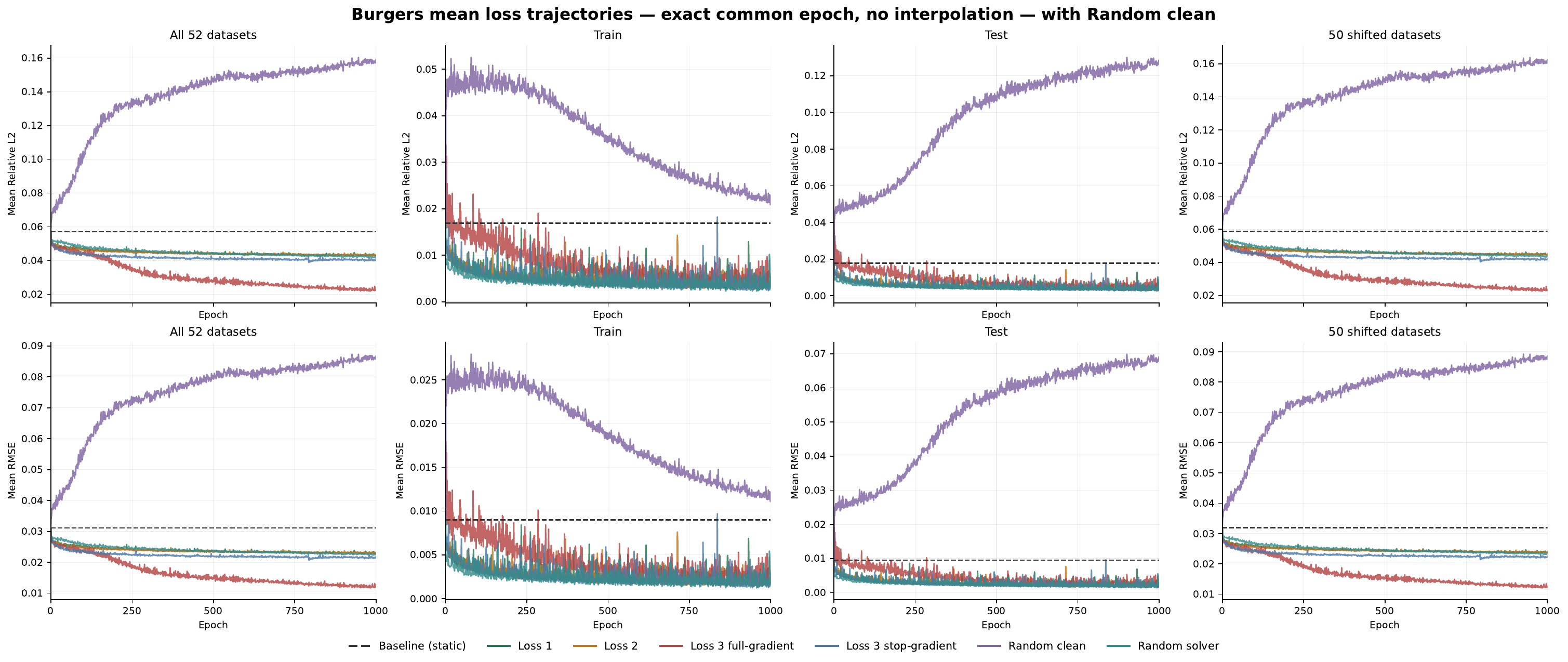}
  \caption{Burgers train, test, and 50-OOD mean relative-$L_2$ curves through exact
  epoch 1000 with random clean restored. This companion view uses the same
  checkpoints as \figref{fig:app-g-burgers-training-selected-time}; its wider
  vertical scale makes the poor random-clean trajectory visible without
  silently dropping that baseline.}
  \label{fig:revision-burgers-epoch1000-mean-with-random-clean}
\end{figure}

\begin{figure}[H]
  \useappendixfigureheading
  \centering
  \captionsetup{font=small,skip=1pt}
  \includegraphics[width=0.99\textwidth,height=0.80\textheight,keepaspectratio]{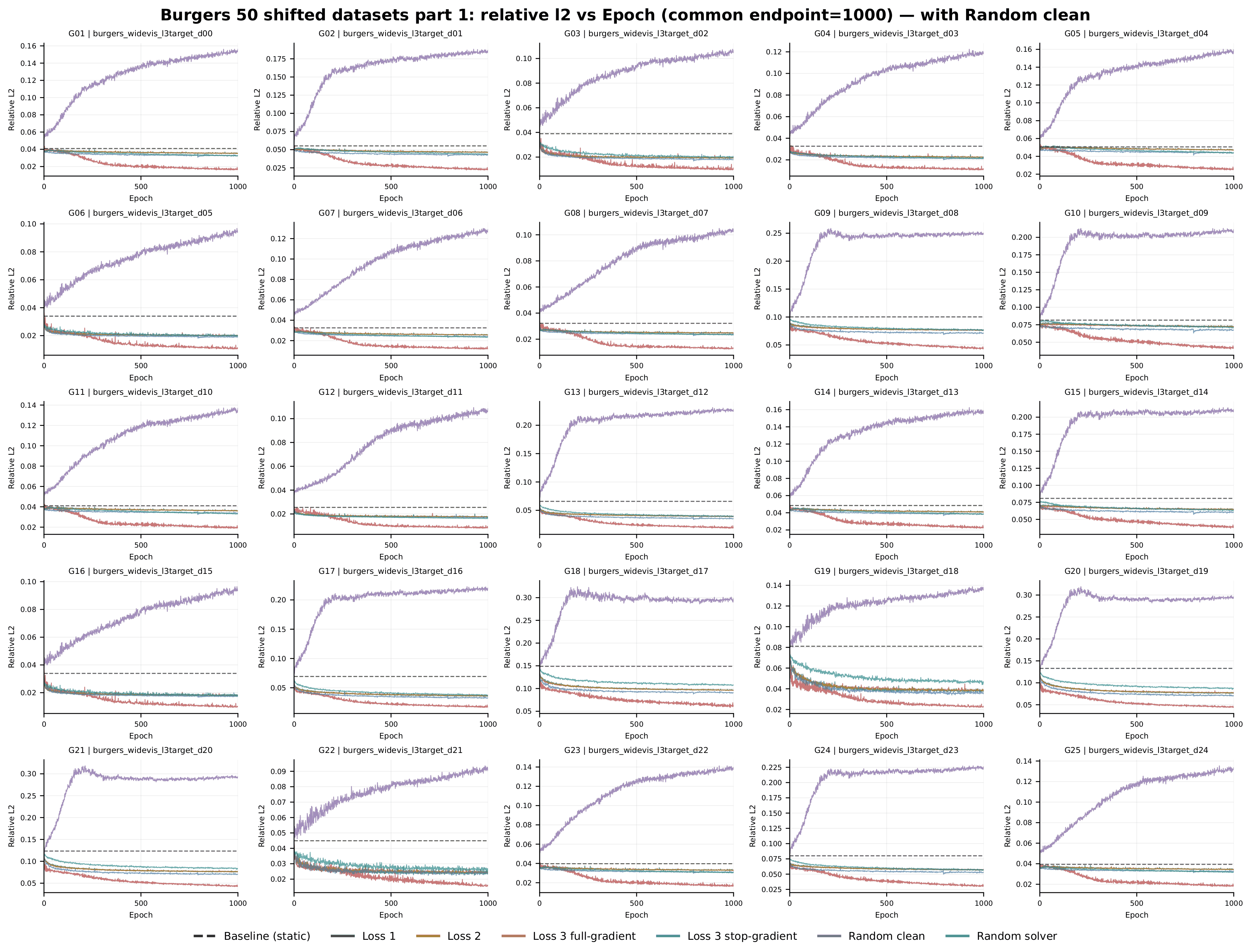}
  \caption{Burgers shifted datasets 1--25 through exact epoch 1000, including Loss 3
  Full, Loss 3 Stop, random clean, and random solver.}
  \label{fig:revision-burgers-epoch1000-ood-1-25-with-random-clean}
\end{figure}

\begin{figure}[H]
  \useappendixfigureheading
  \centering
  \captionsetup{font=small,skip=1pt}
  \includegraphics[width=0.99\textwidth,height=0.80\textheight,keepaspectratio]{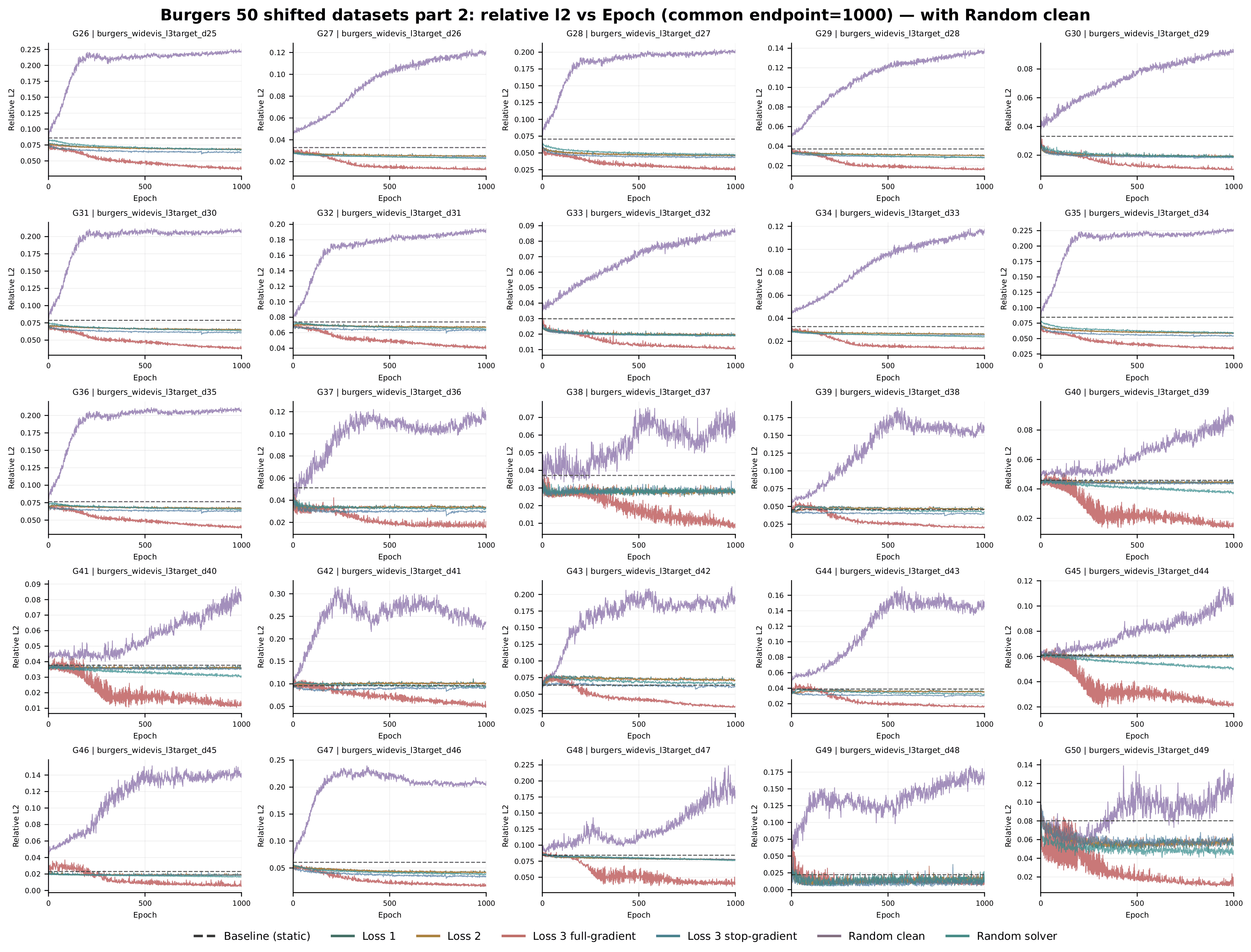}
  \caption{Burgers shifted datasets 26--50 through exact epoch 1000, including Loss 3
  Full, Loss 3 Stop, random clean, and random solver.}
  \label{fig:revision-burgers-epoch1000-ood-26-50-with-random-clean}
\end{figure}

\begin{figure}[H]
  \useappendixfigureheading
  \centering
  \captionsetup{font=small,skip=1pt}
  \includegraphics[width=0.99\textwidth,height=0.80\textheight,keepaspectratio]{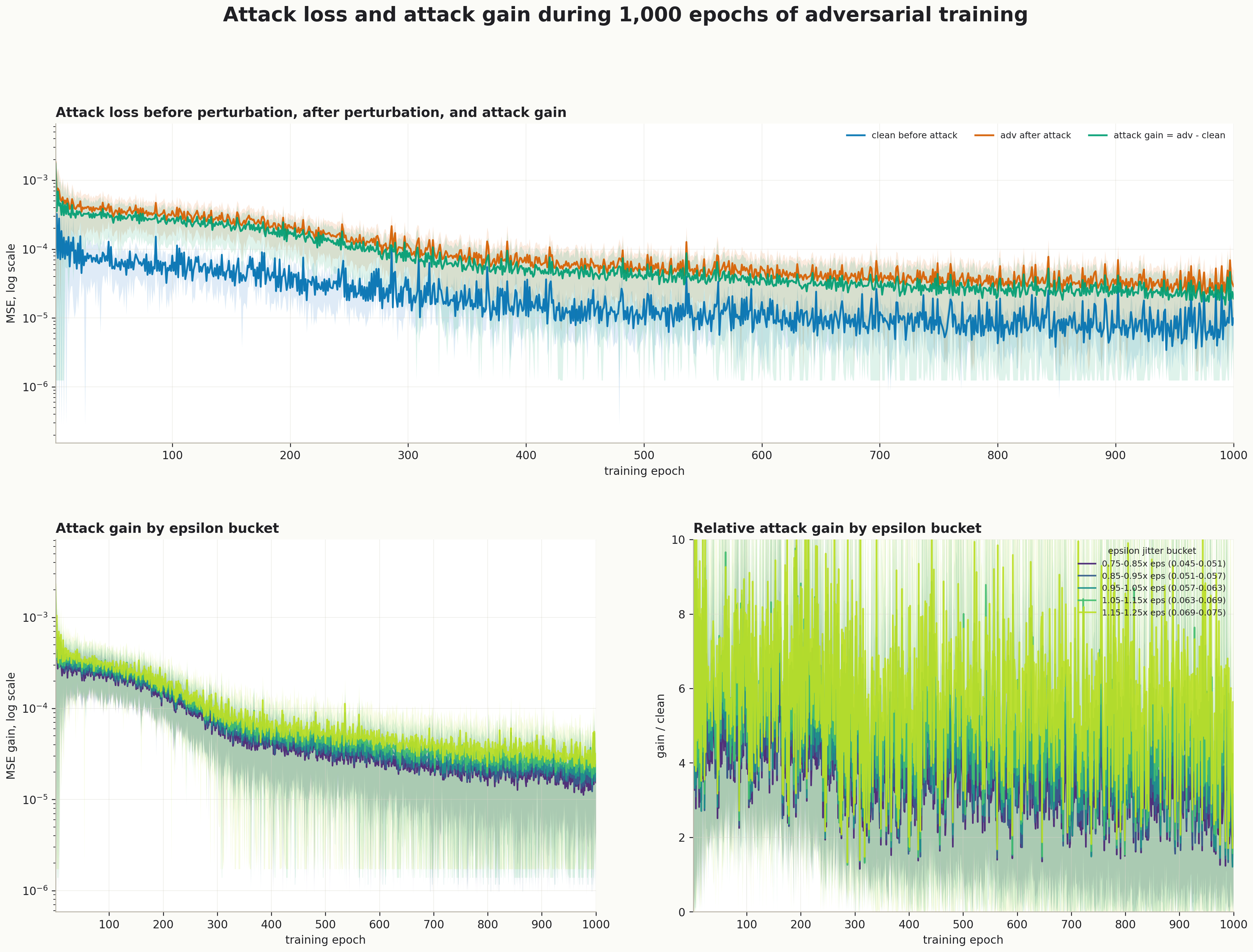}
  \caption{Burgers Loss 3 adversarial-training attack loss and gain over 1,000 epochs,
  including different epsilon budget buckets. Within these buckets, the Loss 3
  adversarial-attack loss increase decreases over training, indicating stronger
  robustness.}
  \label{fig:app-g-burgers-training-loss-gain}
\end{figure}

\begin{figure}[H]
  \useappendixfigureheading
  \centering
  \captionsetup{font=small,skip=1pt}
  \includegraphics[width=0.99\textwidth,height=0.80\textheight,keepaspectratio]{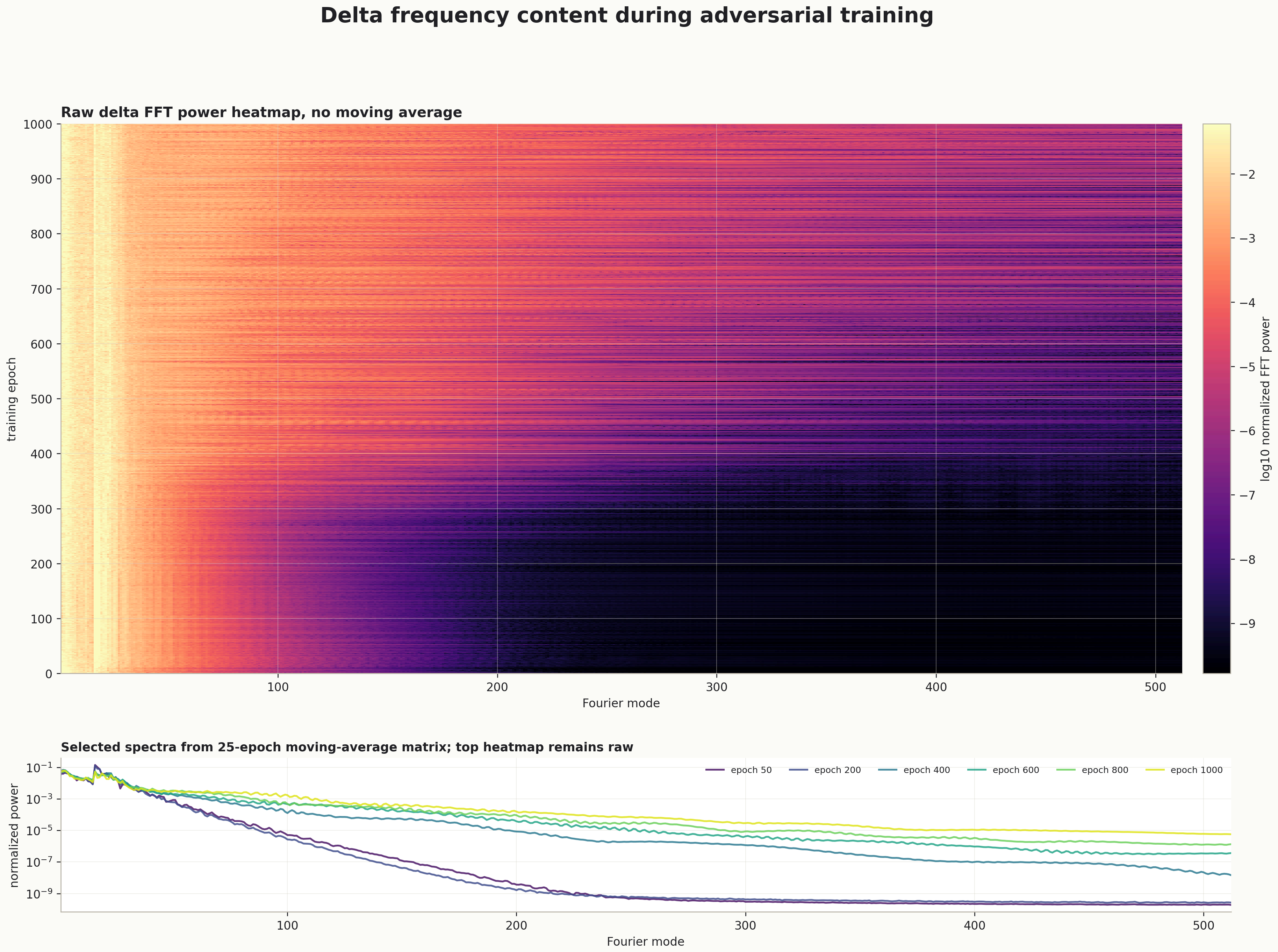}
  \caption{Burgers Loss 3 adversarial-training perturbation frequency content over
  epochs. As training progresses, Loss 3 shifts toward higher-frequency
  perturbations, unlike Loss 1 and Loss 2; this supports Loss 3 self-adaptivity,
  since a more robust model requires sharper perturbations to enlarge the
  model--solver error.}
  \label{fig:app-g-burgers-training-frequency}
\end{figure}

\begin{figure}[H]
  \useappendixfigureheading
  \centering
  \includegraphics[width=0.99\textwidth,height=0.74\textheight,keepaspectratio]{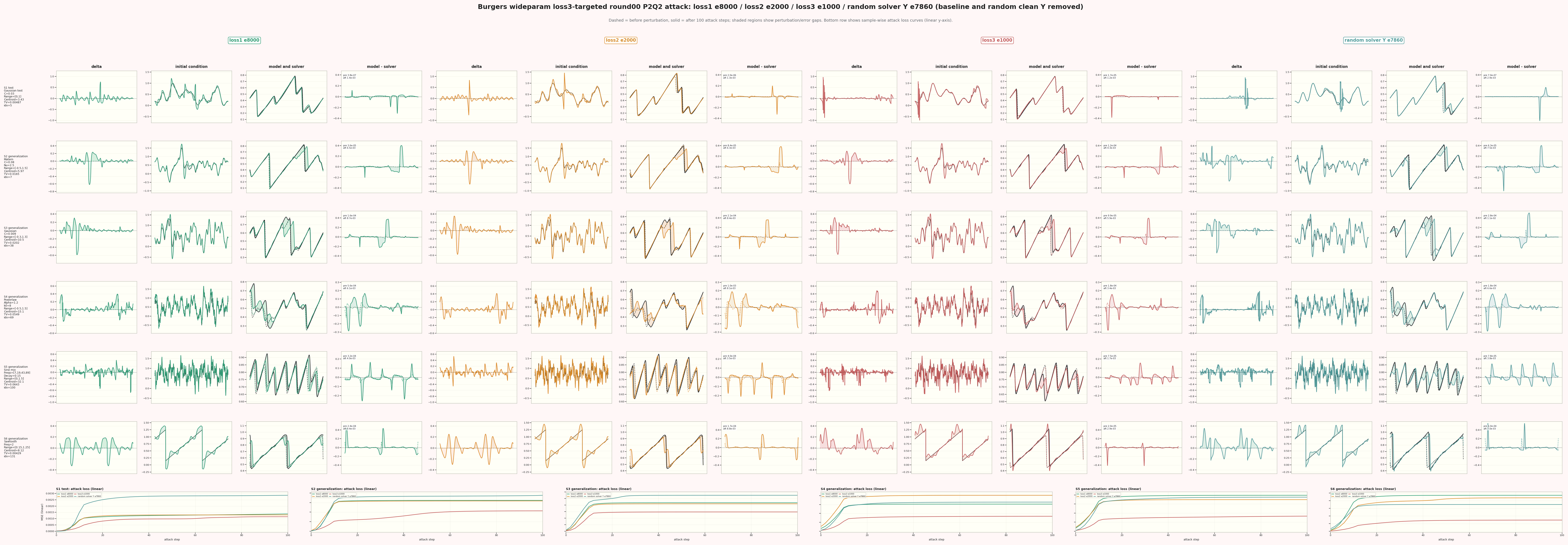}
  \caption{Burgers post-training perturbation sample grid comparing the models obtained
  from Loss 1, Loss 2, Loss 3, and random-solver adversarial training. Baseline
  and random clean are not plotted; the four displayed models are evaluated on
  the same selected initial conditions. After training, all four models are
  attacked with the same Loss 3 adversarial attack, so the labels denote the
  training objective rather than different evaluation objectives. The Loss 3-trained
  model has a smaller final loss increase, indicating the strongest robustness
  among these adversarial training variants.}
  \label{fig:app-g-burgers-training-delta-grid}
\end{figure}

\nextappendixfigureheading{
\noindent\textbf{Epsilon attack sweep.}
}

\begin{figure}[H]
  \useappendixfigureheading
  \centering
  \includegraphics[width=0.96\textwidth,height=0.82\textheight,keepaspectratio]{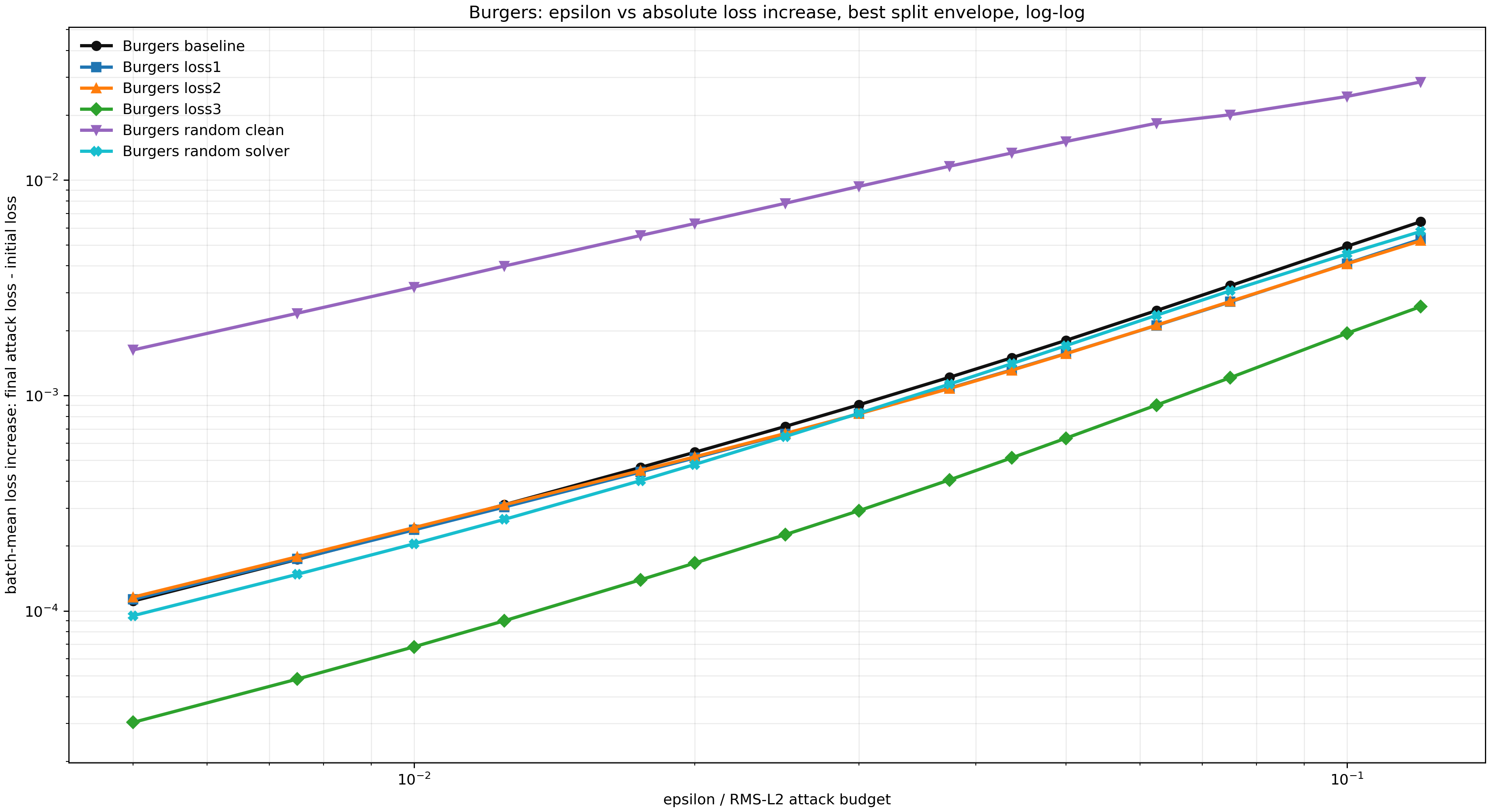}
  \caption{Burgers adversarial-training epsilon attack sweep under the same adversarial
  attack budget, showing the absolute final attack loss increase over the
  initial loss on log-log axes. Among the trained methods, Burgers loss 3 has
  the smallest loss increase across attack budgets, showing that it is the most
  robust model under the matched adversarial attack objective.}
  \label{fig:app-g-burgers-training-epsilon-loss-increase}
\end{figure}

\nextappendixfigureheading{
\appendixdomainheading{Darcy Flow}

\noindent\textbf{Training diagnostics and perturbation samples.}
}

\begin{figure}[H]
  \useappendixfigureheading
  \centering
  \includegraphics[width=0.96\textwidth,height=0.82\textheight,keepaspectratio]{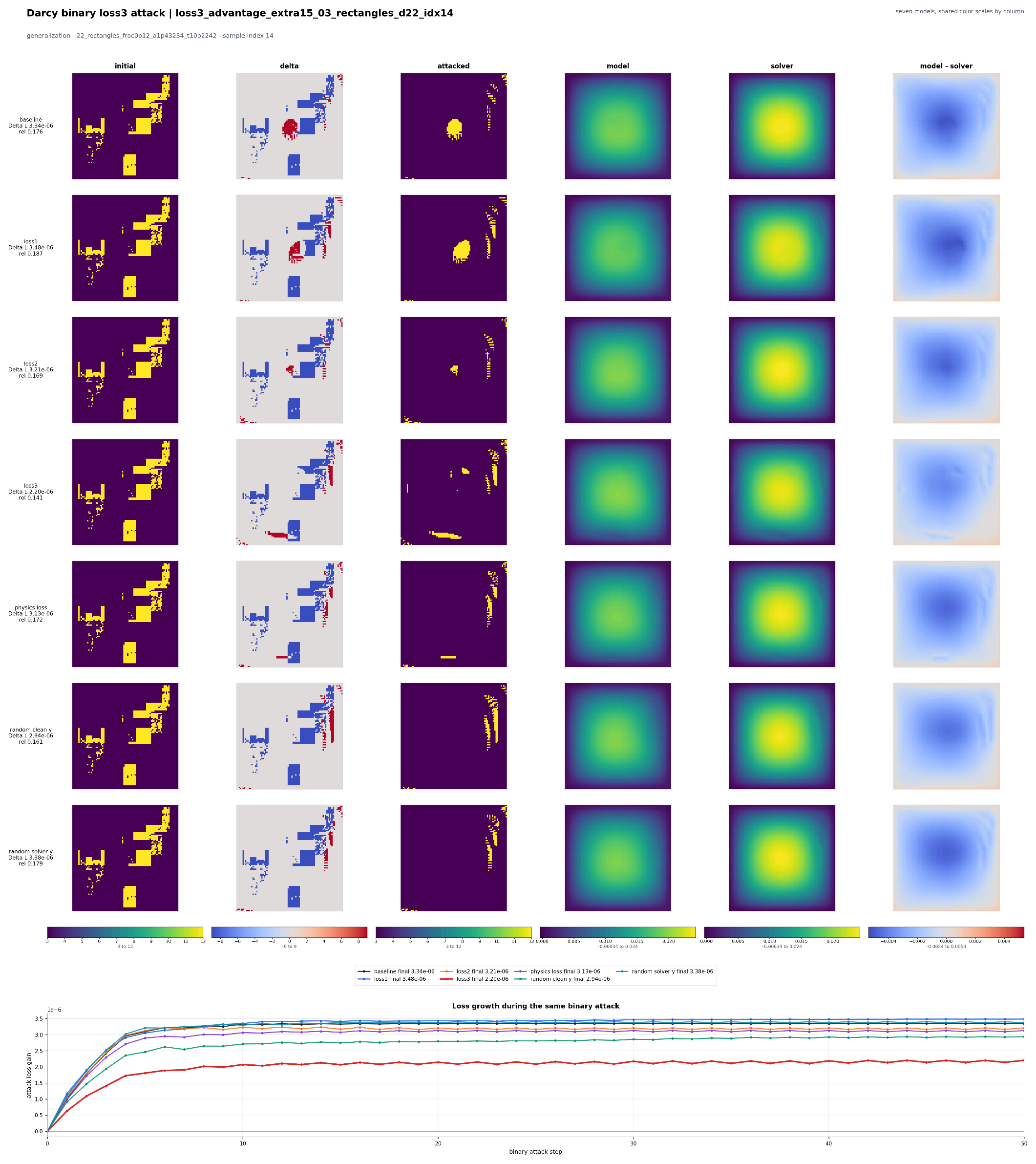}
  \caption{Darcy Flow binary loss-3 training sample grid, including coefficient,
  perturbation, model, solver, residual, and loss-growth panels.}
  \label{fig:app-g-darcy-training-binary-loss3-sample-grid}
\end{figure}

\nextappendixfigureheading{
\noindent\textbf{Generalization curves.}
}

\begin{figure}[H]
  \useappendixfigureheading
  \centering
  \captionsetup{font=footnotesize,skip=1pt}
  \includegraphics[width=\textwidth,height=0.88\textheight,keepaspectratio]{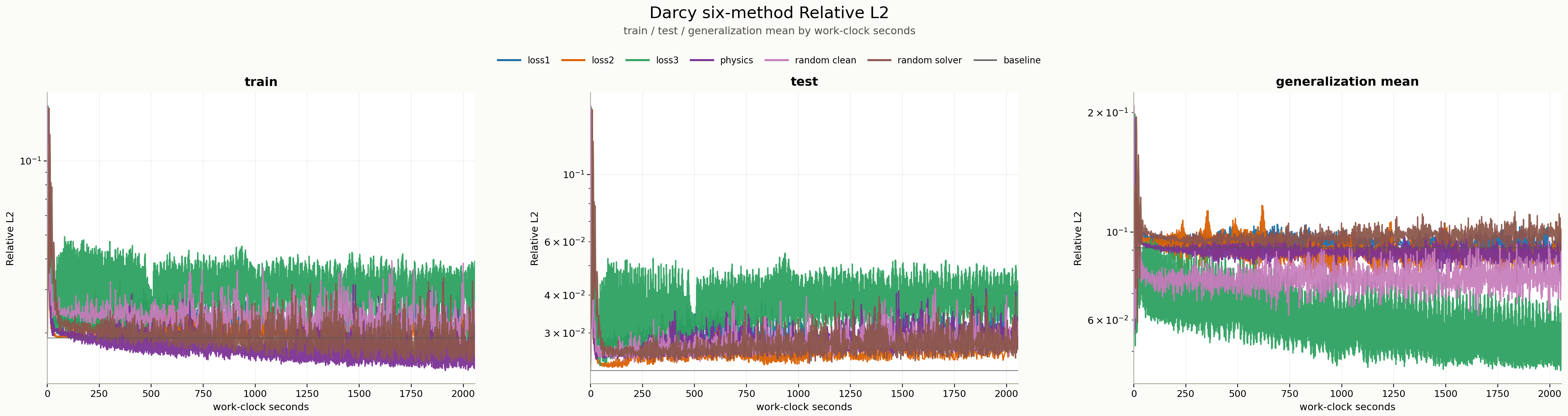}
  \caption{Darcy Flow six-method relative $L_2$ train, test, and generalization mean
  curves. Loss 3 gives the strongest improvement on the generalization data. On
  the training and test splits, Loss 3 performs worse than the other methods.}
  \label{fig:app-g-darcy-training-mean-curves}
\end{figure}

\begin{figure}[H]
  \useappendixfigureheading
  \centering
  \captionsetup{font=footnotesize,skip=1pt}
  \includegraphics[width=\textwidth,height=0.88\textheight,keepaspectratio]{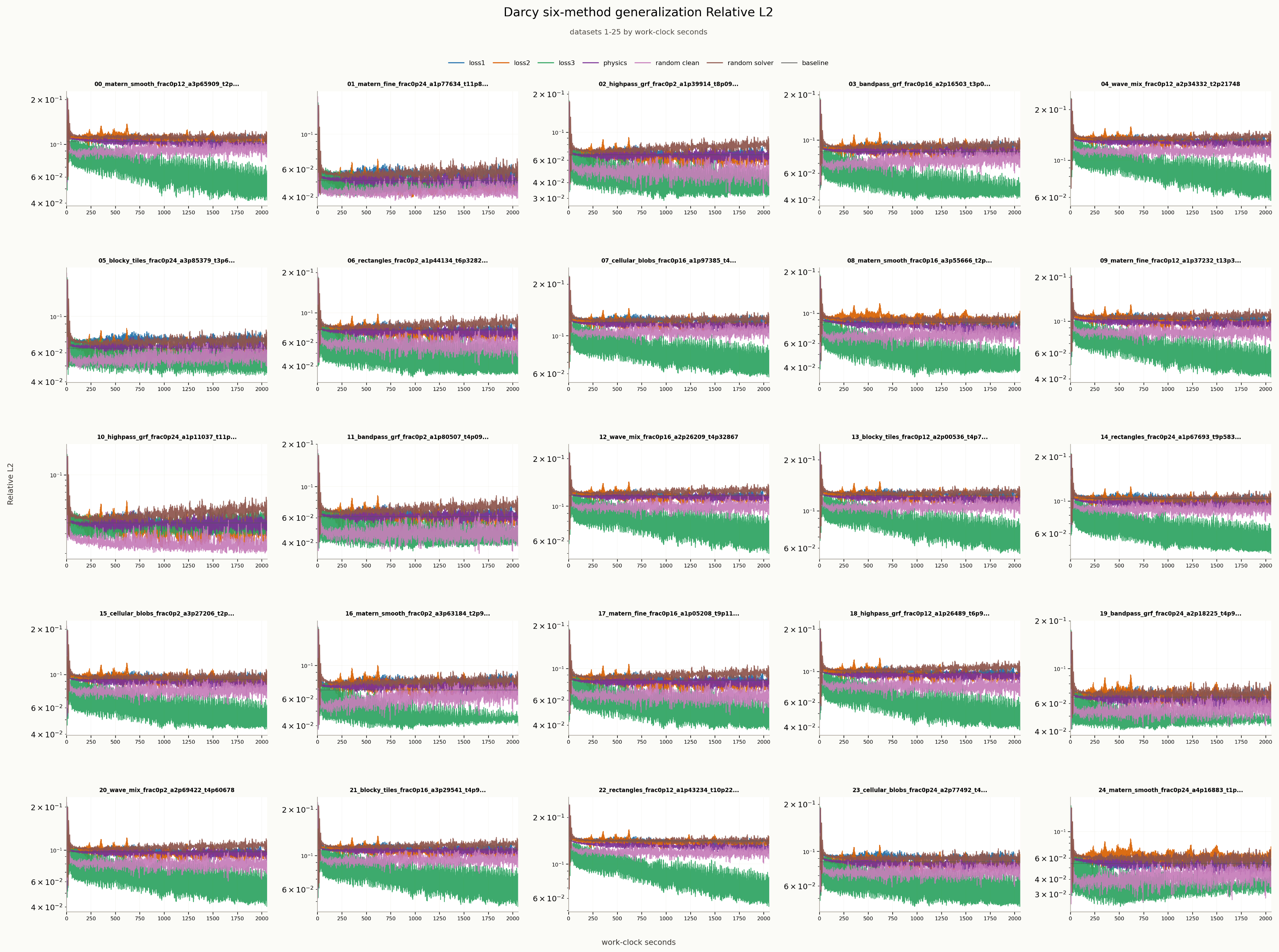}
  \caption{Darcy Flow generalization relative $L_2$ curves for datasets 1--25. Across
  these held-out cases, Loss 3 generally reaches lower generalization loss and
  decreases faster than the competing training strategies.}
  \label{fig:app-g-darcy-training-gen-1-25}
\end{figure}

\begin{figure}[H]
  \useappendixfigureheading
  \centering
  \captionsetup{font=footnotesize,skip=1pt}
  \includegraphics[width=\textwidth,height=0.88\textheight,keepaspectratio]{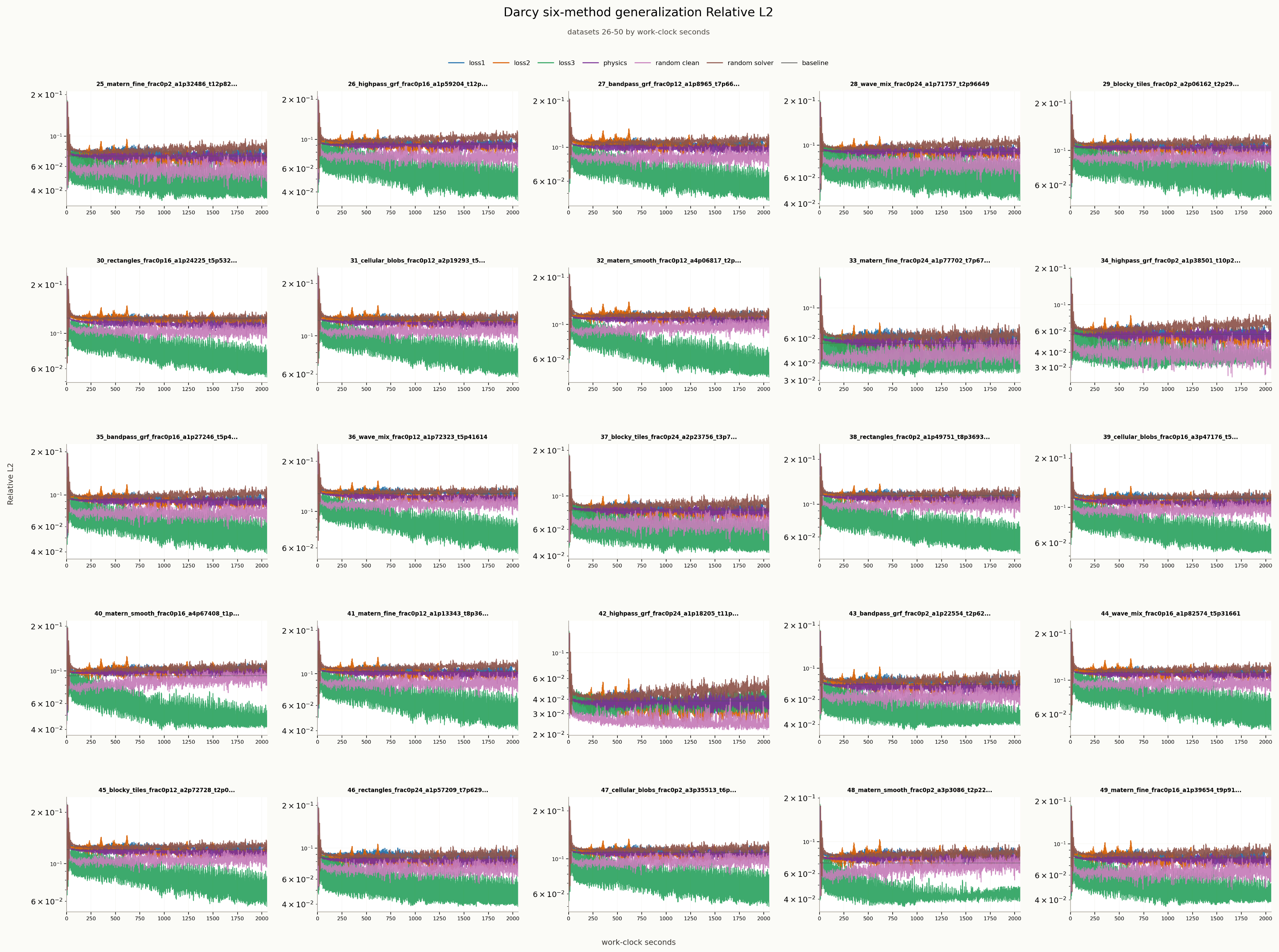}
  \caption{Darcy Flow generalization relative $L_2$ curves for datasets 26--50. These
  panels show the same pattern: the Loss 3 training objective tends to improve
  generalization more consistently and often reduces loss more quickly.}
  \label{fig:app-g-darcy-training-gen-26-50}
\end{figure}

\nextappendixfigureheading{
\noindent\textbf{Expanded eight-method wall-clock comparisons.}
}

\begin{figure}[H]
  \useappendixfigureheading
  \centering
  \captionsetup{font=footnotesize,skip=1pt}
  \includegraphics[width=\textwidth,height=0.88\textheight,keepaspectratio]{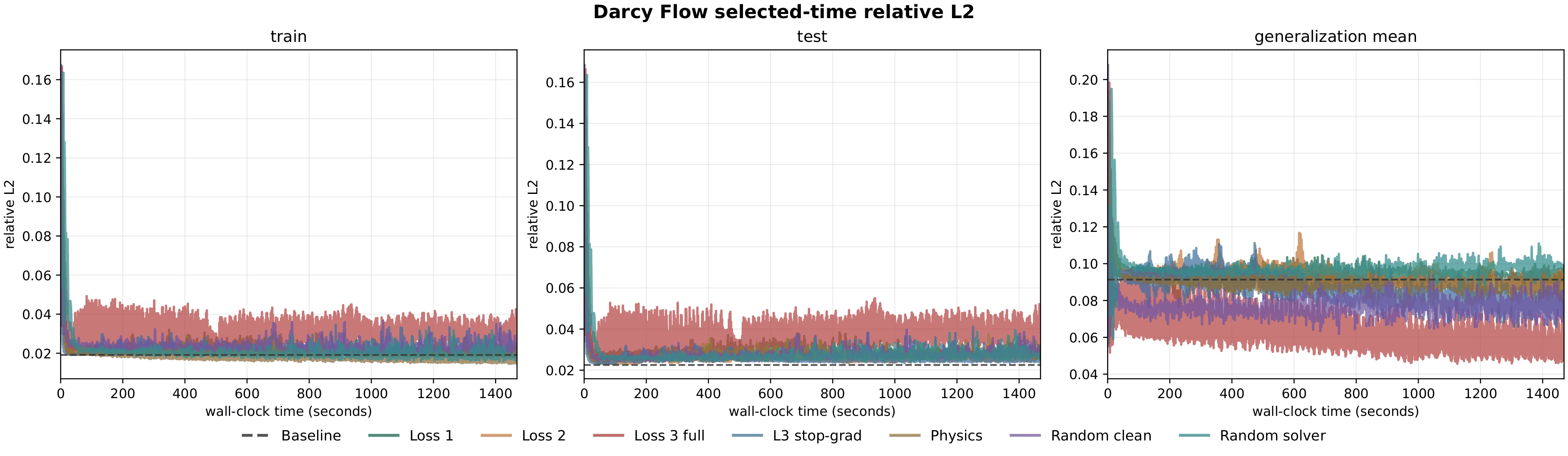}
  \caption{Darcy Flow wall-clock relative-$L_2$ summary for train, test, and the
  50-dataset OOD mean. The expanded comparison includes the shared baseline,
  Loss 1, Loss 2, Loss 3 Full, Loss 3 Stop, Physics, random clean, and random
  solver.}
  \label{fig:revision-darcy-wallclock-mean-l3-stop}
\end{figure}

\begin{figure}[H]
  \useappendixfigureheading
  \centering
  \captionsetup{font=footnotesize,skip=1pt}
  \includegraphics[width=\textwidth,height=0.88\textheight,keepaspectratio]{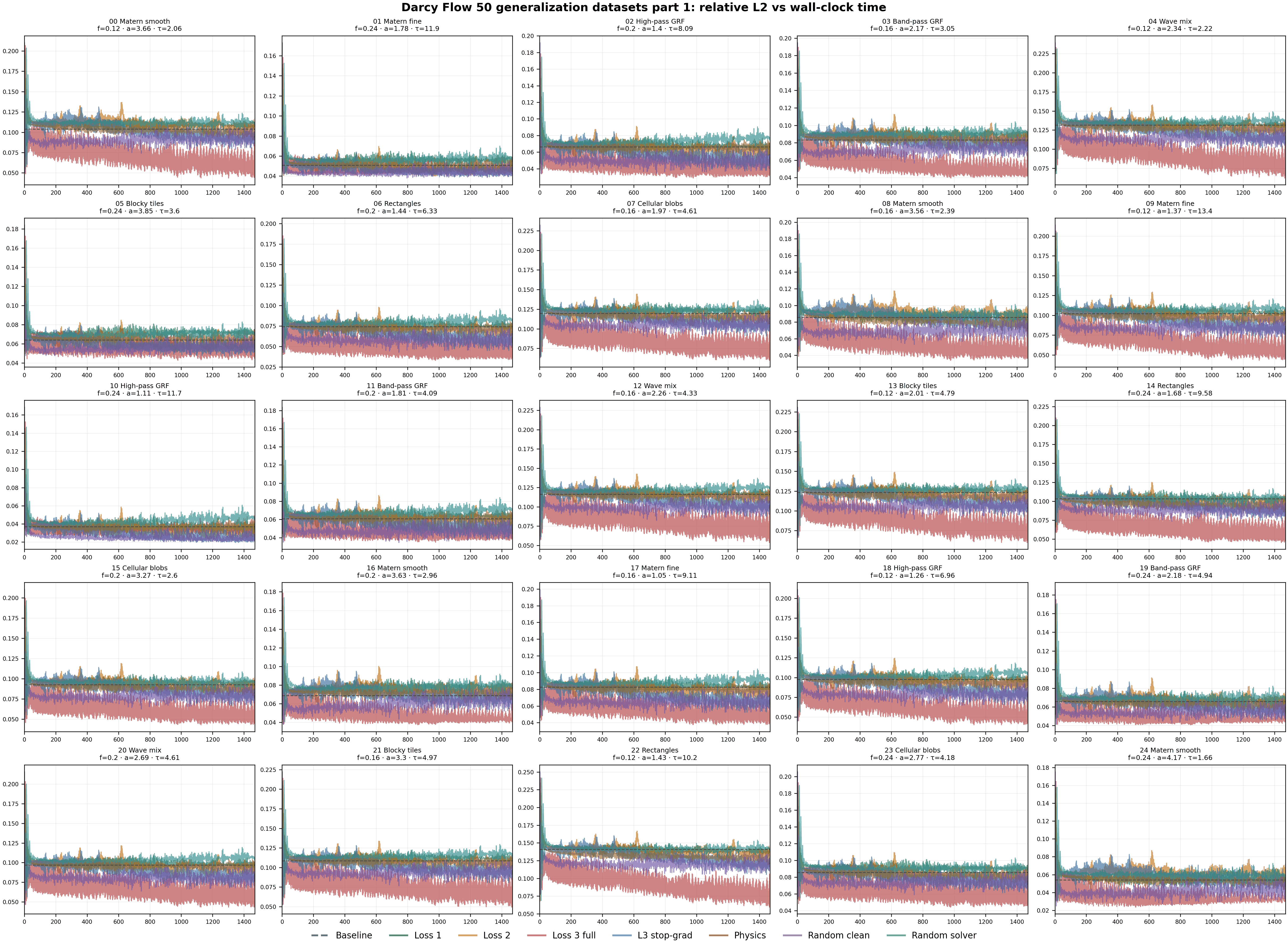}
  \caption{Darcy Flow wall-clock relative-$L_2$ curves for OOD datasets 1--25. These
  panels add Loss 3 Stop to the full eight-model comparison and use complete
  dataset titles without ellipses.}
  \label{fig:revision-darcy-wallclock-ood-1-25-l3-stop}
\end{figure}

\begin{figure}[H]
  \useappendixfigureheading
  \centering
  \captionsetup{font=footnotesize,skip=1pt}
  \includegraphics[width=\textwidth,height=0.88\textheight,keepaspectratio]{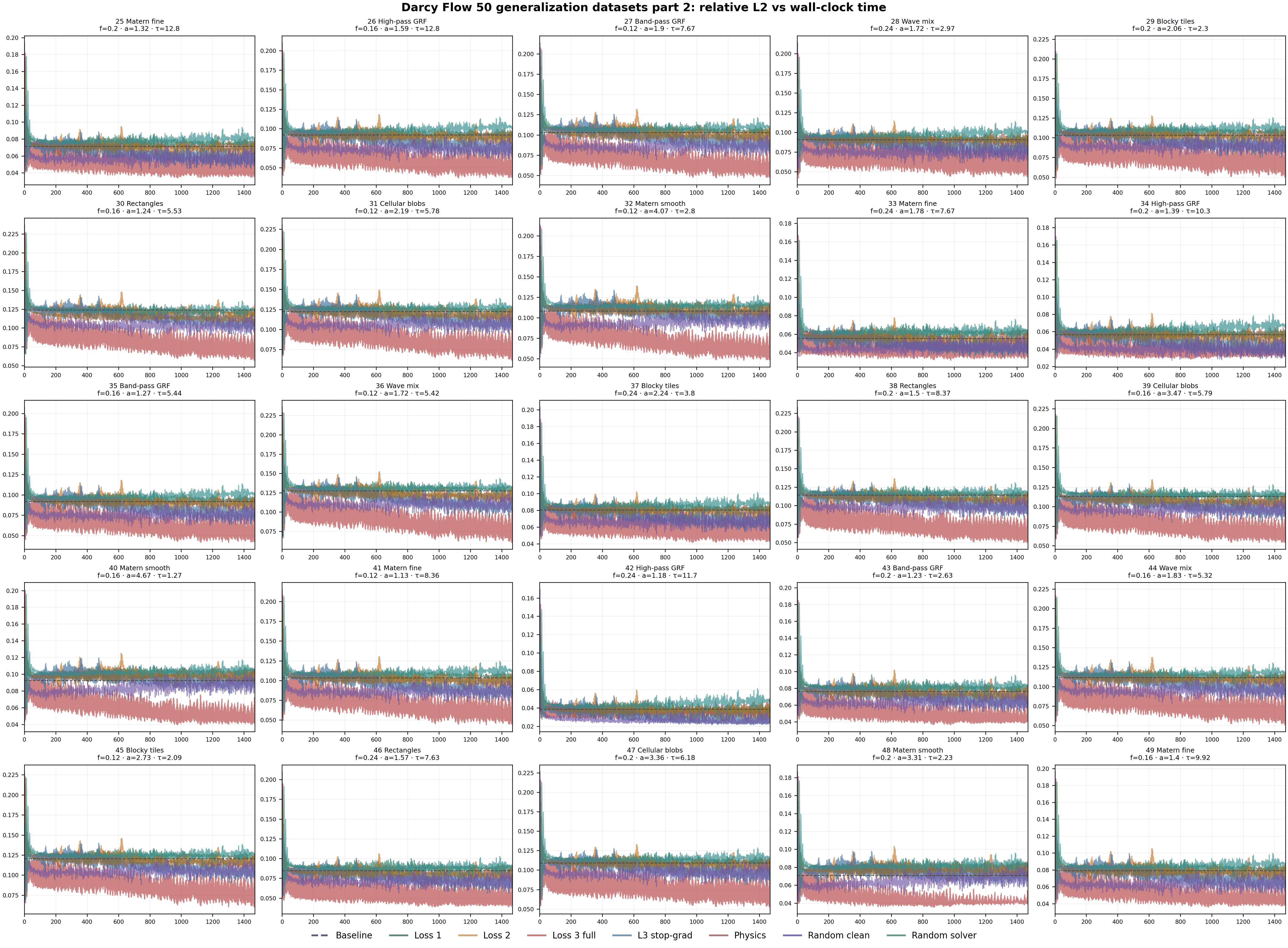}
  \caption{Darcy Flow wall-clock relative-$L_2$ curves for OOD datasets 26--50. These
  panels add Loss 3 Stop to the full eight-model comparison and use complete
  dataset titles without ellipses.}
  \label{fig:revision-darcy-wallclock-ood-26-50-l3-stop}
\end{figure}

\nextappendixfigureheading{
\noindent\textbf{Expanded eight-method common-epoch comparisons.}
}

\begin{figure}[H]
  \useappendixfigureheading
  \centering
  \captionsetup{font=footnotesize,skip=1pt}
  \includegraphics[width=\textwidth,height=0.88\textheight,keepaspectratio]{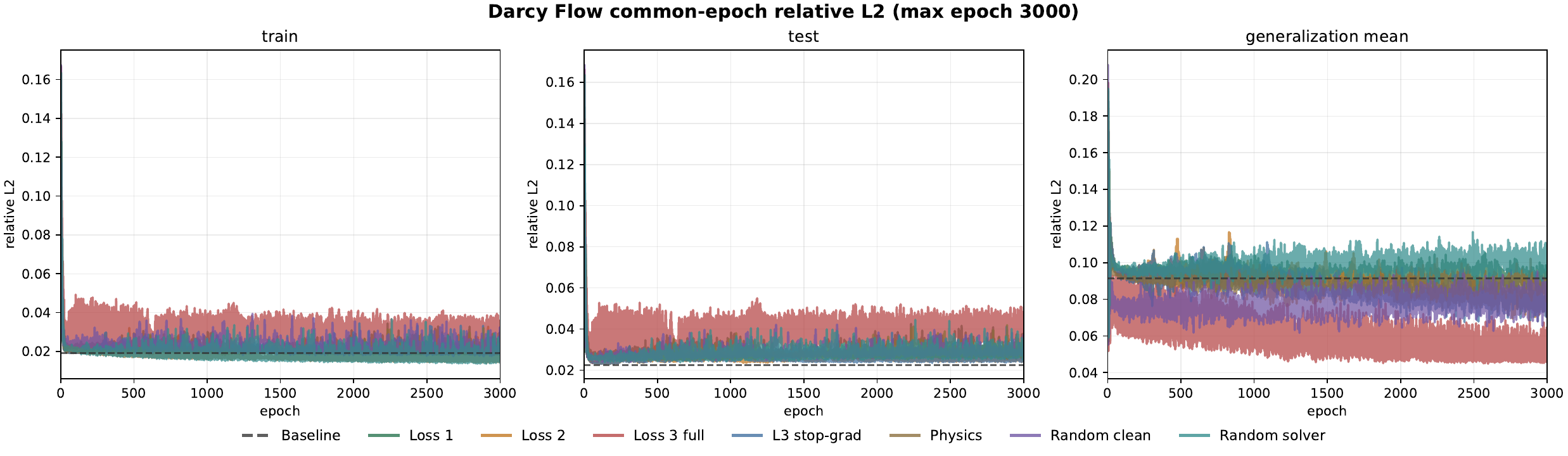}
  \caption{Darcy Flow train/test/OOD50 mean relative-$L_2$ trajectories on an epoch
  horizontal axis through the maximum common epoch 3000. Baseline, L1, L2,
  L3 Full, L3 Stop, Physics, random clean, and random solver are included.
  The common-epoch endpoint table is
  \tabref{tab:appendix-darcy-52-relative-l2}.}
  \label{fig:revision-darcy-common-epoch-mean}
\end{figure}

\begin{figure}[H]
  \useappendixfigureheading
  \centering
  \captionsetup{font=footnotesize,skip=1pt}
  \includegraphics[width=\textwidth,height=0.88\textheight,keepaspectratio]{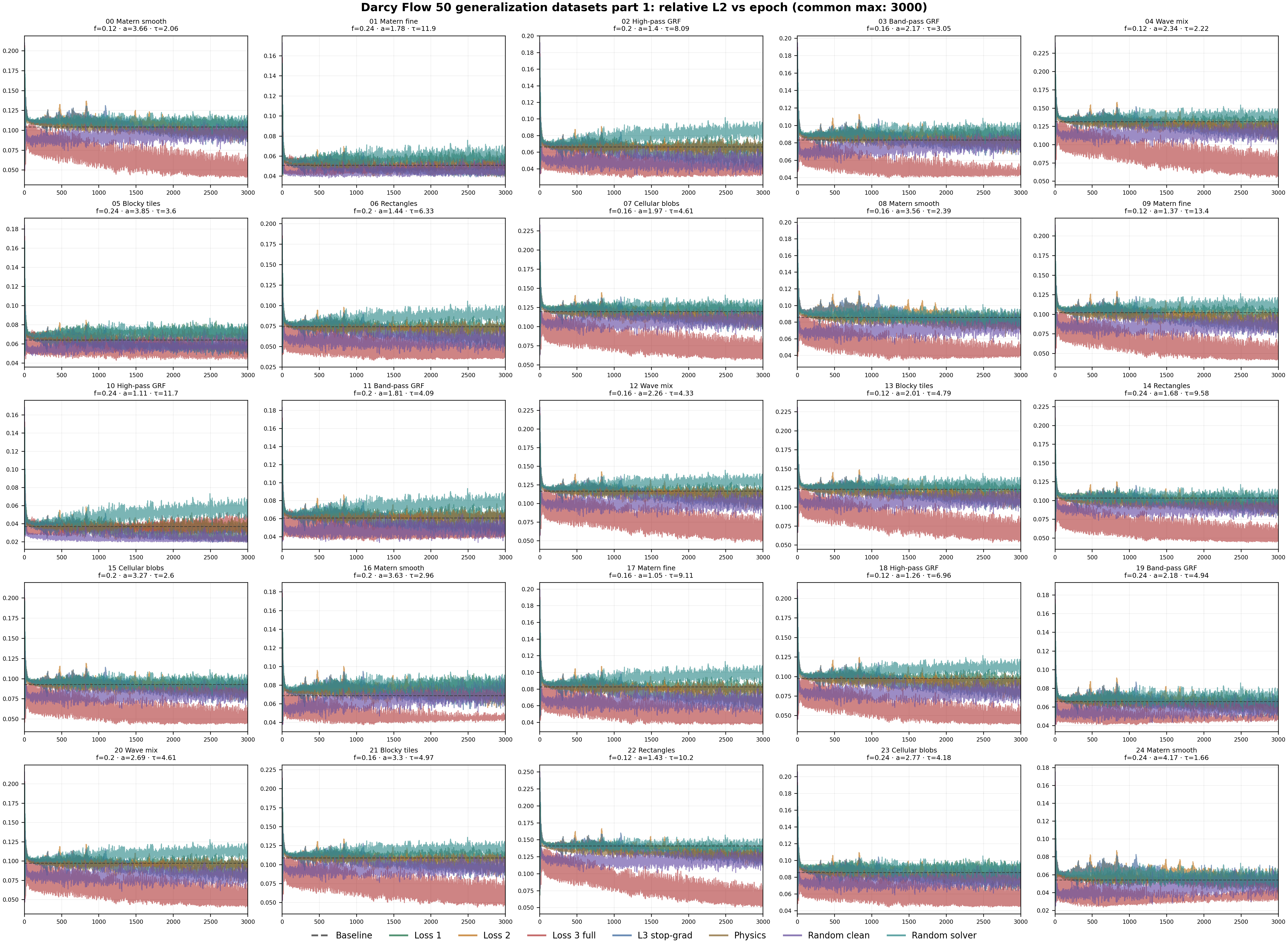}
  \caption{Darcy Flow OOD datasets 1--25 on an epoch axis through common epoch 3000,
  including both Full and Stop L3. This is separate from the preceding
  wall-clock-axis comparison.}
  \label{fig:revision-darcy-common-epoch-1-25}
\end{figure}

\begin{figure}[H]
  \useappendixfigureheading
  \centering
  \captionsetup{font=footnotesize,skip=1pt}
  \includegraphics[width=\textwidth,height=0.88\textheight,keepaspectratio]{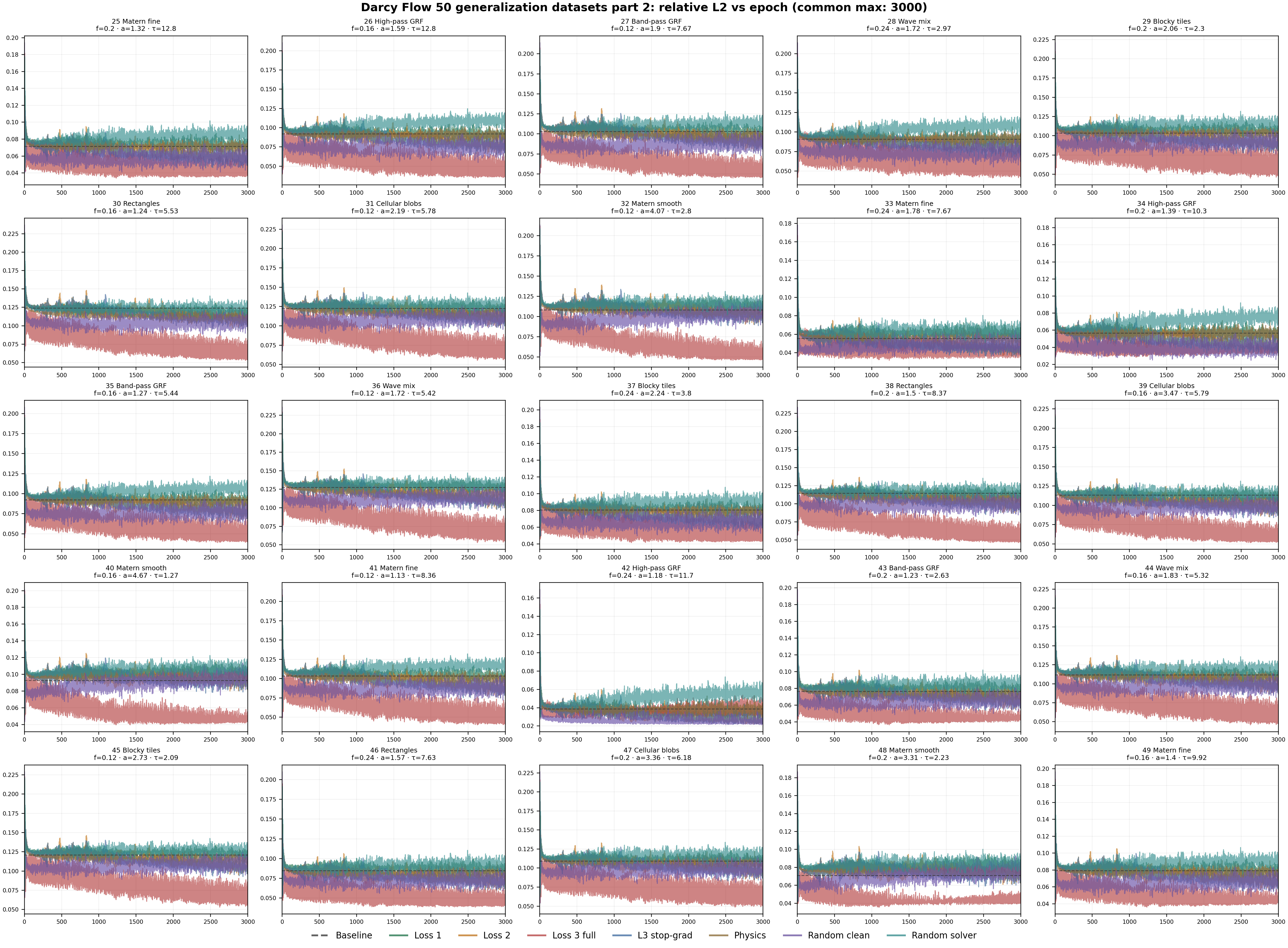}
  \caption{Darcy Flow OOD datasets 26--50 on an epoch axis through common epoch 3000,
  with the same eight-model legend as the mean and datasets 1--25 panels.}
  \label{fig:revision-darcy-common-epoch-26-50}
\end{figure}

\nextappendixfigureheading{
\noindent\textbf{Epsilon attack sweep.}
}

\begin{figure}[H]
  \useappendixfigureheading
  \centering
  \includegraphics[width=0.96\textwidth,height=0.82\textheight,keepaspectratio]{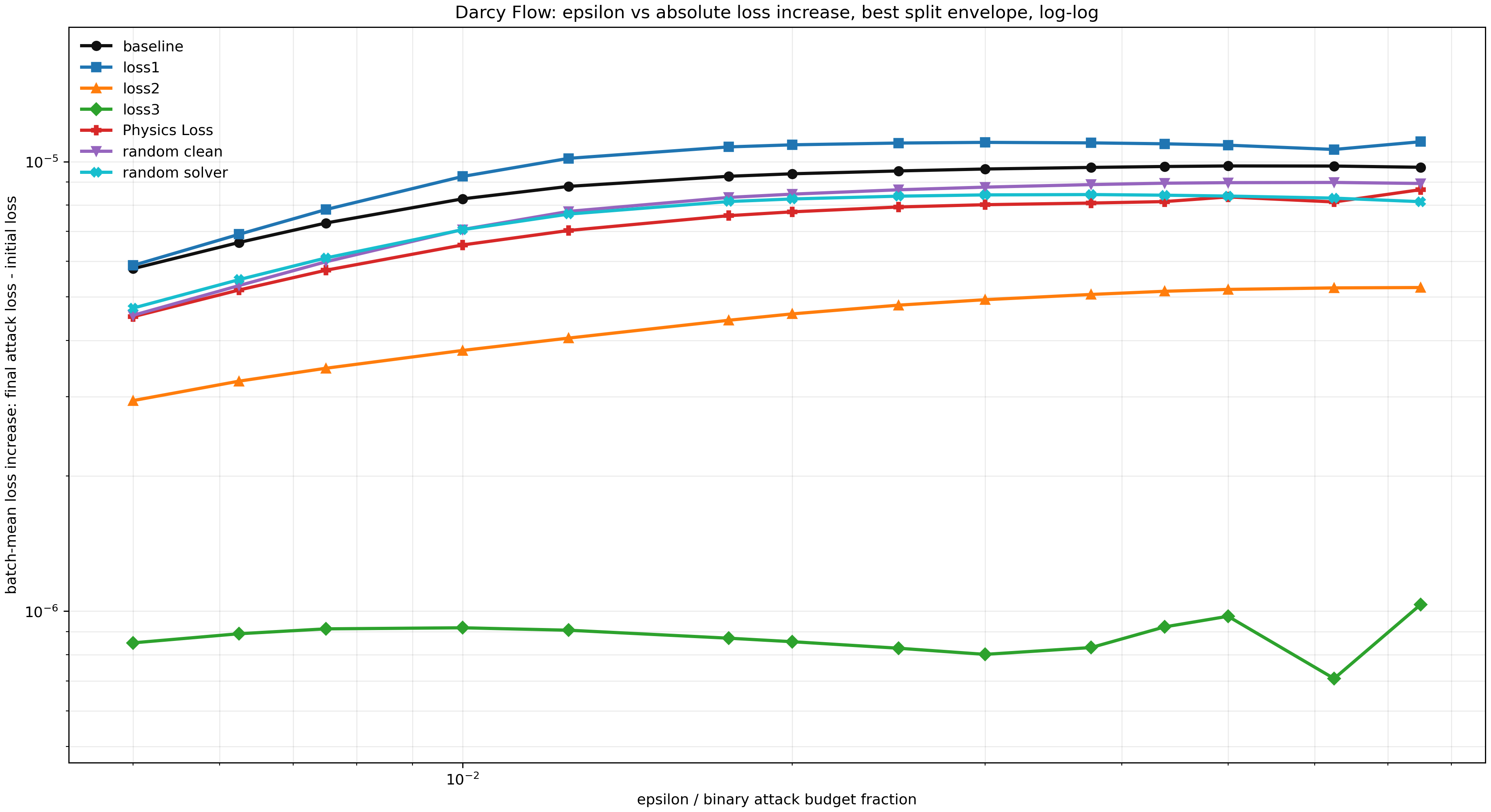}
  \caption{Darcy Flow adversarial-training epsilon attack sweep under the same attack
  budget. After training, the Loss 3 model has the smallest loss increase under
  the matched adversarial attack, indicating that it is the most robust among
  the compared Darcy Flow models.}
  \label{fig:app-g-darcy-training-epsilon-loss-increase}
\end{figure}

\FloatBarrier
\nextappendixfigureheading{
\appendixdomainheading{Navier--Stokes}
}

\begin{figure}[H]
  \useappendixfigureheading
  \centering
  \includegraphics[width=0.96\textwidth,height=0.82\textheight,keepaspectratio]{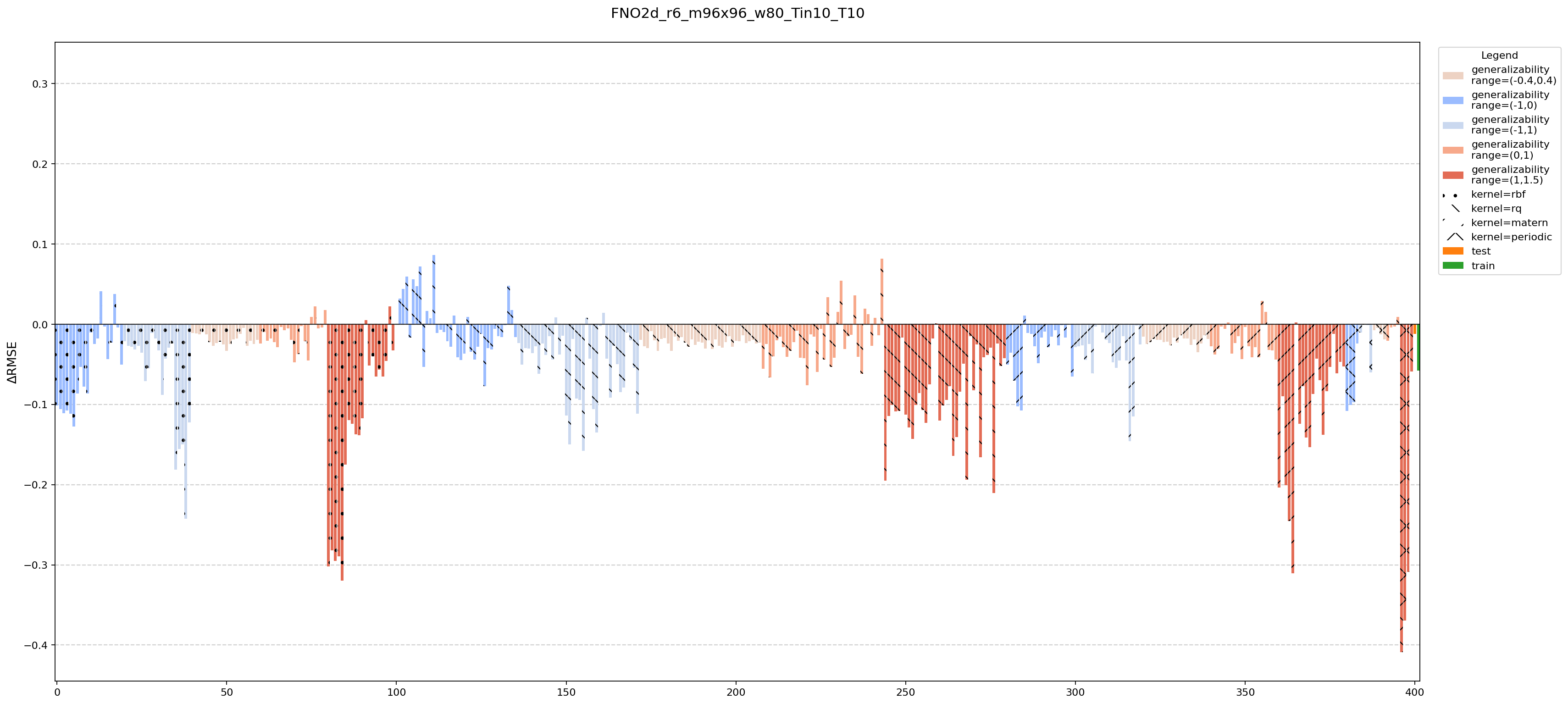}
  \caption{Navier--Stokes/FNO2d generalizability $\Delta$RMSE bar chart. Because
  Navier--Stokes adversarial training requires very large GPU memory and
  wall-clock time, we did not have enough compute for a complete systematic
  comparison. Instead, we ran six rounds of Loss 3 adversarial training on the
  FNO2d dataset and evaluated the trained model against the baseline on selected
  generalization datasets. Each bar is the round-six minus baseline mean of
  per-sample final-frame RMSE; negative values indicate improvement. The
  retained 402-dataset GRF subset has 364 improvements and 38 deteriorations.
  The excluded input range and the expanded 482/1,442-dataset results are
  reported in \tabref{tab:revision-ns-quality-scope}; this figure does not
  represent a multi-method long-training sweep.}
  \label{fig:app-g-ns-training-delta-rmse}
\end{figure}

\endgroup

\appendixsection{Robustness Metrics and Agreement Diagnostics}{app:robustness-metric-diagnostics}
\label{app:local-svd-diagnostics}
\begin{multicols}{2}

\subsection*{Metric Computation}

This appendix records how the robustness quantities in
\tabref{tab:robustness_metric_correlation_summary} and
\tabref{tab:robustness_vector_angle_summary} are computed.  The agreement
diagnostics are intentionally split by type: scalar quantities are compared
with correlations, while vector quantities are compared through directional
similarity or, equivalently, mean acute angle differences.  Let
\[
  e_\theta(x)=\mathcal{F}_\theta(x)-\mathcal{G}(x),
  \qquad
  J_{\mathcal E}(x)=D\mathcal{E}_\theta[x],
\]
where \(\mathcal{F}_\theta\) is the neural operator, \(\mathcal{G}\) is the
numerical solver, and \(\mathcal{E}_\theta=\mathcal{F}_\theta-\mathcal{G}\) is
the model-solver error operator.

\paragraph{Finite attack loss increase.}
For an evaluation input \(x\), the finite-budget attack runs projected-gradient
ascent inside the chosen perturbation ball and returns an attacked input
\(x+\delta_T\).  The reported increase is
\[
  \Delta\ell_{\rm atk}(x)
  =
  \ell_3(x+\delta_T)-\ell_3(x),
  \qquad
  \ell_3(z)=\frac{1}{2}\|e_\theta(z)\|_2^2 .
\]
Thus this number measures how much the solver-model discrepancy can be enlarged
by the actual finite attack used in evaluation.  Smaller values mean that the
trained model is harder to damage under the matched attack budget.

\paragraph{Jacobian-error norm.}
The Jacobian-error vector is
\[
  g_\theta(x)=J_{\mathcal E}(x)^\top e_\theta(x).
\]
It is the gradient of the squared solver-model error with respect to the input:
\[
  \left.\frac{d}{dt}\frac{1}{2}
  \|e_\theta(x+t h)\|_2^2\right|_{t=0}
  =
  \langle J_{\mathcal E}(x)^\top e_\theta(x),h\rangle .
\]
The reported quantity is \(\|J_{\mathcal E}^{\top}e_\theta\|_2\), averaged over
the recorded diagnostic samples.  It is a local proxy for the small-budget
attack loss increase, because the maximizing first-order direction is aligned
with \(J_{\mathcal E}^{\top}e_\theta\).
For a differentiable half-squared loss and a continuous $L_2$ ball,
\begin{equation}
 \max_{\|\delta\|_2\leq\epsilon}
 [\ell_3(x+\delta)-\ell_3(x)]
 =\epsilon\|J_{\mathcal E}(x)^\top e_\theta(x)\|_2+o(\epsilon).
 \label{eq:revision-local-attack-limit}
\end{equation}
Thus the attack increase per unit budget tends to J-error as the budget tends
to zero. One reverse-mode scalar-loss gradient computes it without forming a
dense Jacobian or running multi-step PGD. A spectral norm instead needs SVD or
iterated JVP/VJP products. The finite attack also depends on curvature, step
size, and projection. Darcy's binary flips form a discrete Hamming constraint;
the continuous local diagnostic is compared with those attacks empirically,
rather than using this continuous-ball expansion as a discrete guarantee.

\paragraph{Error-Jacobian spectral norm.}
The operator-norm diagnostic is the leading singular value
\[
  \sigma_1(J_{\mathcal E}(x))
  =
  \max_{\|h\|_2=1}\|J_{\mathcal E}(x)h\|_2 .
\]
This measures the largest local displacement of the error field, regardless of
whether that displacement is aligned with the current error vector
\(e_\theta(x)\).  It is therefore a useful sensitivity diagnostic, but it is not
as directly tied to final squared-error growth as
\(\|J_{\mathcal E}^{\top}e_\theta\|_2\).

\paragraph{Scalar correlation agreement.}
For each diagnostic pool, we correlate the per-sample finite attack loss
increase \(\Delta\ell_{\rm atk}\) with three local scalar metrics:
\(\sigma_1(D\mathcal F_\theta)\), \(\sigma_1(J_{\mathcal E})\), and
\(\|J_{\mathcal E}^{\top}e_\theta\|_2\). The updated table reports Pearson
\(r\) for every attack budget so that the coefficient matches the annotation in
the corresponding scatter panel. Burgers pools six exact-epoch-1000 checkpoints
and 25 inputs per checkpoint; Darcy Flow pools eight exact-epoch-3000
checkpoints and 21 OOD inputs per checkpoint. Navier--Stokes adds 20 matched
inputs from the retained baseline checkpoint. The Burgers and Darcy entries use
base-10 log coordinates; the Navier--Stokes entries use the recorded raw
coordinates. This is a scalar table only; it does not compare vector directions.

\paragraph{Model-only operator scale.}
\tabref{tab:model_only_jacobian_spectral_norms} separately reports
\(\sigma_1(D\mathcal{F}_\theta[x])\), with the solver column reporting
\(\sigma_1(D\mathcal{G}[x])\).  These model-only and solver-only operator
scales are not robustness metrics in our sense: a PDE solution operator can
legitimately move when its input is perturbed.  The relevant quantity for
solver-integrated robustness is whether the model-solver error remains stable,
not whether the neural operator itself has small local gain.
Tables~\ref{tab:model_solver_svd_subspace_similarity}
and~\ref{tab:darcy_model_solver_svd_subspace_similarity} add complementary
Burgers and Darcy Flow SVD-subspace diagnostics: the solver is the reference
subspace in every entry, so there is no separate solver column.

\end{multicols}

\begin{table}[H]
\centering
\footnotesize
\setlength{\tabcolsep}{3.2pt}
\renewcommand{\arraystretch}{1.25}
\caption{Model-only Jacobian spectral norms.  Entries are mean \(\pm\)
standard deviation of the largest singular value over the same \(25\) local
diagnostic samples.  The solver column reports \(\sigma_1(D\mathcal{G})\);
model columns report \(\sigma_1(D\mathcal{F}_\theta)\). Darcy Flow entries are
reported in \(10^{-3}\) units.  The robust
Loss 3 models in \tabref{tab:adversarial_training_robustness_summary} do not
show a corresponding decrease in this model-only operator scale.}
\label{tab:model_only_jacobian_spectral_norms}
\newcommand{\spectralentry}[2]{\begin{tabular}[c]{@{}c@{}}#1\\\(\pm\) #2\end{tabular}}
\begin{tabular}{@{}lcccccccc@{}}
\toprule
PDE & Solver & Baseline & Loss 1 & Loss 2 & Loss 3 &
\begin{tabular}[c]{@{}c@{}}Physics\\loss\end{tabular} &
\begin{tabular}[c]{@{}c@{}}Random\\clean\end{tabular} &
\begin{tabular}[c]{@{}c@{}}Random\\solver\end{tabular} \\
\midrule
Burgers &
\spectralentry{3.755}{0.823} &
\spectralentry{3.456}{0.805} &
\spectralentry{3.632}{0.889} &
\spectralentry{3.609}{0.897} &
\spectralentry{3.605}{0.884} &
-- &
\spectralentry{3.179}{1.613} &
\spectralentry{3.567}{0.978} \\
Darcy Flow &
\spectralentry{3.377}{0.7452} &
\spectralentry{1.250}{0.1375} &
\spectralentry{1.781}{0.2605} &
\spectralentry{1.712}{0.2495} &
\spectralentry{2.289}{0.3914} &
\spectralentry{1.817}{0.2742} &
\spectralentry{1.688}{0.2545} &
\spectralentry{1.860}{0.2840} \\
\bottomrule
\end{tabular}%
\end{table}

\begin{table}[H]
\centering
\footnotesize
\setlength{\tabcolsep}{3.2pt}
\renewcommand{\arraystretch}{1.25}
\caption{Model--solver SVD subspace similarity for Burgers. Entries are
mean \(\pm\) standard deviation over the same \(25\) local diagnostic samples.
For each input, \(s_k^R\) is the mean principal cosine between the leading
\(k\) right-singular subspaces of \(D\mathcal{F}_\theta[x]\) and
\(D\mathcal{G}[x]\). Higher values indicate stronger local alignment with the
solver input-response subspace. There is no solver column because the solver
subspace is the reference in every entry; these are local SVD diagnostics, not
standalone robustness metrics.}
\label{tab:model_solver_svd_subspace_similarity}
\newcommand{\subspaceentry}[2]{\begin{tabular}[c]{@{}c@{}}#1\\\(\pm\) #2\end{tabular}}
\begin{tabular}{@{}llcccccc@{}}
\toprule
PDE & Metric & Baseline & Loss 1 & Loss 2 & Loss 3 &
\begin{tabular}[c]{@{}c@{}}Random\\clean\end{tabular} &
\begin{tabular}[c]{@{}c@{}}Random\\solver\end{tabular} \\
\midrule
Burgers & \(s^R_1\) &
\subspaceentry{0.854}{0.301} &
\subspaceentry{0.938}{0.170} &
\subspaceentry{0.940}{0.143} &
\subspaceentry{0.960}{0.168} &
\subspaceentry{0.229}{0.198} &
\subspaceentry{0.921}{0.223} \\
Burgers & \(s^R_5\) &
\subspaceentry{0.902}{0.097} &
\subspaceentry{0.924}{0.083} &
\subspaceentry{0.924}{0.084} &
\subspaceentry{0.936}{0.085} &
\subspaceentry{0.533}{0.081} &
\subspaceentry{0.921}{0.104} \\
Burgers & \(s^R_{10}\) &
\subspaceentry{0.886}{0.059} &
\subspaceentry{0.911}{0.057} &
\subspaceentry{0.904}{0.059} &
\subspaceentry{0.958}{0.043} &
\subspaceentry{0.618}{0.057} &
\subspaceentry{0.904}{0.078} \\
Burgers & \(s^R_{20}\) &
\subspaceentry{0.732}{0.102} &
\subspaceentry{0.764}{0.100} &
\subspaceentry{0.761}{0.104} &
\subspaceentry{0.887}{0.069} &
\subspaceentry{0.646}{0.078} &
\subspaceentry{0.831}{0.099} \\
\bottomrule
\end{tabular}%
\end{table}

\begin{table}[H]
\centering
\footnotesize
\setlength{\tabcolsep}{3pt}
\renewcommand{\arraystretch}{1.18}
\caption{Model--solver block-2 SVD subspace similarity for Darcy Flow. Entries
are mean \(\pm\) standard deviation over the same \(25\) local diagnostic
samples. For each input, \(s_k^R\) is the mean principal cosine between the
leading \(k\) right-singular subspaces of the model and solver block-2
Jacobians. Higher values indicate stronger local alignment with the solver
input-response subspace. The Darcy Flow records contain \(k=1,3,5,10\); a
top-20 subspace was not recorded. These are local SVD diagnostics, not
standalone robustness metrics.}
\label{tab:darcy_model_solver_svd_subspace_similarity}
\newcommand{\darcysubspaceentry}[2]{\begin{tabular}[c]{@{}c@{}}#1\\\(\pm\) #2\end{tabular}}
\begin{tabular*}{\textwidth}{@{\extracolsep{\fill}}llccccccc@{}}
\toprule
PDE & Metric & Baseline & Loss 1 & Loss 2 & Loss 3 &
\begin{tabular}[c]{@{}c@{}}Physics\\loss\end{tabular} &
\begin{tabular}[c]{@{}c@{}}Random\\clean\end{tabular} &
\begin{tabular}[c]{@{}c@{}}Random\\solver\end{tabular} \\
\midrule
Darcy Flow & \(s^R_1\) &
\darcysubspaceentry{0.784}{0.051} &
\darcysubspaceentry{0.851}{0.027} &
\darcysubspaceentry{0.859}{0.028} &
\darcysubspaceentry{0.897}{0.025} &
\darcysubspaceentry{0.851}{0.026} &
\darcysubspaceentry{0.835}{0.032} &
\darcysubspaceentry{0.857}{0.026} \\
Darcy Flow & \(s^R_3\) &
\darcysubspaceentry{0.716}{0.054} &
\darcysubspaceentry{0.744}{0.044} &
\darcysubspaceentry{0.749}{0.045} &
\darcysubspaceentry{0.788}{0.035} &
\darcysubspaceentry{0.747}{0.041} &
\darcysubspaceentry{0.738}{0.051} &
\darcysubspaceentry{0.749}{0.041} \\
Darcy Flow & \(s^R_5\) &
\darcysubspaceentry{0.676}{0.070} &
\darcysubspaceentry{0.691}{0.055} &
\darcysubspaceentry{0.700}{0.056} &
\darcysubspaceentry{0.718}{0.058} &
\darcysubspaceentry{0.698}{0.055} &
\darcysubspaceentry{0.689}{0.062} &
\darcysubspaceentry{0.698}{0.056} \\
Darcy Flow & \(s^R_{10}\) &
\darcysubspaceentry{0.645}{0.058} &
\darcysubspaceentry{0.682}{0.048} &
\darcysubspaceentry{0.690}{0.052} &
\darcysubspaceentry{0.710}{0.042} &
\darcysubspaceentry{0.688}{0.048} &
\darcysubspaceentry{0.684}{0.056} &
\darcysubspaceentry{0.687}{0.049} \\
\bottomrule
\end{tabular*}%
\end{table}

\begin{table}[H]
\centering
\footnotesize
\setlength{\tabcolsep}{5pt}
\renewcommand{\arraystretch}{1.18}
\caption{Original-cohort scalar correlation diagnostics, retained separately
from the updated matched-checkpoint scatter studies. Burgers
uses pooled groups: all groups \(=25\) mixed inputs \(\times 6\) methods, and
generalization only \(=21\) shifted inputs \(\times 6\) methods.  Darcy Flow
uses epsilon-sweep rows, each with \(21\) generalization inputs \(\times 7\)
methods.  Here \(n\) is the number of records; \(r\) is Pearson correlation and
\(\rho\) is Spearman correlation with finite attack loss increase
\(\Delta\ell_{\rm atk}\).  Equivalently, \(R_m\) and \(\rho_m\) denote the
Pearson and Spearman correlations with scalar metric \(m\).}
\label{tab:robustness_metric_correlation_historical}
\begin{tabular*}{\textwidth}{@{\extracolsep{\fill}}llrcccc@{}}
\toprule
PDE & Diagnostic pool & \(n\) &
\(R_{\|J_{\mathcal E}^{\top}e_\theta\|_2}\) &
\(\rho_{\|J_{\mathcal E}^{\top}e_\theta\|_2}\) &
\(R_{\sigma_1(J_{\mathcal E})}\) &
\(\rho_{\sigma_1(J_{\mathcal E})}\) \\
\midrule
Burgers & all groups: 25 mixed \(\times 6\) & 150 & 0.8997 & 0.8976 & 0.6938 & 0.8312 \\
Burgers & gen. only: 21 shifted \(\times 6\) & 126 & 0.9254 & 0.8758 & 0.6377 & 0.7863 \\
\midrule
Darcy Flow & epsilon \(0.01{\times}\) & 147 & 0.5889 & 0.6204 & 0.4213 & 0.4254 \\
Darcy Flow & epsilon \(0.05{\times}\) & 147 & 0.5920 & 0.6481 & 0.4744 & 0.4777 \\
Darcy Flow & epsilon \(0.1{\times}\) & 147 & 0.5675 & 0.6548 & 0.5065 & 0.5060 \\
Darcy Flow & epsilon \(0.2{\times}\) & 147 & 0.4475 & 0.3780 & 0.4470 & 0.2972 \\
Darcy Flow & epsilon \(0.5{\times}\) & 147 & 0.3978 & 0.2982 & 0.3991 & 0.2093 \\
Darcy Flow & epsilon \(1{\times}\) & 147 & 0.3533 & 0.2096 & 0.3469 & 0.1145 \\
Darcy Flow & epsilon \(5{\times}\) & 147 & 0.3021 & 0.0497 & 0.3066 & -0.0135 \\
Darcy Flow & epsilon \(10{\times}\) & 147 & 0.2273 & -0.0960 & 0.2350 & -0.1534 \\
\bottomrule
\end{tabular*}
\end{table}
\begin{table}[H]
\centering
\footnotesize
\setlength{\tabcolsep}{3.4pt}
\renewcommand{\arraystretch}{1.12}
\caption{Pearson correlation between finite Loss-3 attack increase and three local Jacobian diagnostics. Burgers uses six checkpoints at exact epoch 1000 and 25 inputs per checkpoint ($n=150$). Darcy uses eight checkpoints at exact epoch 3000 and 21 OOD inputs per checkpoint; $n$ is smaller only when nonpositive attack increases are excluded from the log--log view. Navier--Stokes uses 20 matched inputs from one retained baseline checkpoint. The first two systems report correlations in base-10 log coordinates to match the scatter plots; Navier--Stokes reports the recorded raw-coordinate correlations.}
\label{tab:robustness_metric_correlation_summary}
\begin{tabular*}{\textwidth}{@{\extracolsep{\fill}}llcrrr@{}}
\toprule
PDE & Attack budget & $n$ & $r_{\sigma_1(D\mathcal F)}$ & $r_{\sigma_1(J_{\mathcal E})}$ & $r_{\|J_{\mathcal E}^{\top}e\|_2}$\\
\midrule
Burgers & $\epsilon=0.005$ (log--log) & 150 & -0.266 & 0.963 & 0.997\\
Burgers & $\epsilon=0.01$ (log--log) & 150 & -0.247 & 0.957 & 0.991\\
Burgers & $\epsilon=0.02$ (log--log) & 150 & -0.220 & 0.940 & 0.977\\
Burgers & $\epsilon=0.04$ (log--log) & 150 & -0.188 & 0.903 & 0.946\\
Burgers & $\epsilon=0.08$ (log--log) & 150 & -0.111 & 0.842 & 0.892\\
Burgers & $\epsilon=0.12$ (log--log) & 150 & -0.066 & 0.810 & 0.855\\
Burgers & $\epsilon=0.2$ (log--log) & 150 & -0.026 & 0.746 & 0.772\\
Burgers & $\epsilon=0.4$ (log--log) & 150 & 0.016 & 0.587 & 0.601\\
\midrule
Darcy Flow & $K=2$ flips (log--log) & 168 & -0.340 & 0.257 & 0.481\\
Darcy Flow & $K=9$ flips (log--log) & 164 & -0.327 & 0.261 & 0.489\\
Darcy Flow & $K=17$ flips (log--log) & 161 & -0.242 & 0.275 & 0.535\\
Darcy Flow & $K=34$ flips (log--log) & 160 & -0.272 & 0.302 & 0.557\\
Darcy Flow & $K=86$ flips (log--log) & 162 & -0.302 & 0.347 & 0.556\\
Darcy Flow & $K=172$ flips (log--log) & 165 & -0.393 & 0.382 & 0.604\\
Darcy Flow & $K=861$ flips (log--log) & 167 & -0.389 & 0.410 & 0.585\\
Darcy Flow & $K=1722$ flips (log--log) & 168 & -0.419 & 0.307 & 0.430\\
\midrule
Navier--Stokes & $\epsilon=8$ (raw) & 20 & 0.687 & 0.838 & 0.900\\
Navier--Stokes & $\epsilon=16$ (raw) & 20 & 0.721 & 0.832 & 0.859\\
Navier--Stokes & $\epsilon=32$ (raw) & 20 & 0.797 & 0.749 & 0.714\\
Navier--Stokes & $\epsilon=64$ (raw) & 20 & 0.800 & 0.695 & 0.627\\
\bottomrule
\end{tabular*}
\end{table}

\begingroup
\footnotesize
\setlength{\tabcolsep}{4.5pt}
\renewcommand{\arraystretch}{1.04}
\begin{longtable}{@{}llcrrr@{}}
\caption{Burgers panel-wise Pearson correlations reproduced from Figure~\ref{fig:revision-burgers-jacobian-scatter-loglog}. The three correlation columns follow the three scatter-plot columns. Each attack-budget block reports the pooled value and every method-specific value printed in the corresponding panel titles; correlations use base-10 log coordinates.}\label{tab:revision-burgers-scatter-panel-correlations}\\
\toprule
Budget & Group & $n$ & $r_{\sigma_1(D\mathcal F)}$ & $r_{\sigma_1(J_{\mathcal E})}$ & $r_{\|J_{\mathcal E}^{\top}e\|_2}$\\
\midrule
\endfirsthead
\toprule
Budget & Group & $n$ & $r_{\sigma_1(D\mathcal F)}$ & $r_{\sigma_1(J_{\mathcal E})}$ & $r_{\|J_{\mathcal E}^{\top}e\|_2}$\\
\midrule
\endhead
\midrule
\multicolumn{6}{r}{continued on next page}\\
\endfoot
\bottomrule
\endlastfoot
$\epsilon=0.005$ & Overall & 150 & -0.266 & +0.963 & +0.997\\
$\epsilon=0.005$ & Baseline & 25 & -0.153 & +0.916 & +0.997\\
$\epsilon=0.005$ & L1 & 25 & -0.298 & +0.967 & +0.998\\
$\epsilon=0.005$ & L2 & 25 & -0.269 & +0.963 & +0.999\\
$\epsilon=0.005$ & L3 Full & 25 & -0.361 & +0.983 & +1.000\\
$\epsilon=0.005$ & L3 Stop & 25 & -0.253 & +0.966 & +0.999\\
$\epsilon=0.005$ & Random solver & 25 & -0.299 & +0.977 & +0.998\\
\addlinespace[2pt]
$\epsilon=0.01$ & Overall & 150 & -0.247 & +0.957 & +0.991\\
$\epsilon=0.01$ & Baseline & 25 & -0.118 & +0.900 & +0.987\\
$\epsilon=0.01$ & L1 & 25 & -0.277 & +0.968 & +0.993\\
$\epsilon=0.01$ & L2 & 25 & -0.256 & +0.958 & +0.997\\
$\epsilon=0.01$ & L3 Full & 25 & -0.353 & +0.985 & +0.999\\
$\epsilon=0.01$ & L3 Stop & 25 & -0.229 & +0.962 & +0.996\\
$\epsilon=0.01$ & Random solver & 25 & -0.282 & +0.976 & +0.994\\
\addlinespace[2pt]
$\epsilon=0.02$ & Overall & 150 & -0.220 & +0.940 & +0.977\\
$\epsilon=0.02$ & Baseline & 25 & -0.063 & +0.852 & +0.951\\
$\epsilon=0.02$ & L1 & 25 & -0.253 & +0.964 & +0.983\\
$\epsilon=0.02$ & L2 & 25 & -0.245 & +0.947 & +0.991\\
$\epsilon=0.02$ & L3 Full & 25 & -0.338 & +0.985 & +0.996\\
$\epsilon=0.02$ & L3 Stop & 25 & -0.193 & +0.953 & +0.989\\
$\epsilon=0.02$ & Random solver & 25 & -0.248 & +0.972 & +0.986\\
\addlinespace[2pt]
$\epsilon=0.04$ & Overall & 150 & -0.188 & +0.903 & +0.946\\
$\epsilon=0.04$ & Baseline & 25 & -0.009 & +0.768 & +0.876\\
$\epsilon=0.04$ & L1 & 25 & -0.224 & +0.943 & +0.958\\
$\epsilon=0.04$ & L2 & 25 & -0.229 & +0.899 & +0.962\\
$\epsilon=0.04$ & L3 Full & 25 & -0.320 & +0.973 & +0.986\\
$\epsilon=0.04$ & L3 Stop & 25 & -0.155 & +0.922 & +0.967\\
$\epsilon=0.04$ & Random solver & 25 & -0.198 & +0.951 & +0.965\\
\addlinespace[2pt]
$\epsilon=0.08$ & Overall & 150 & -0.111 & +0.842 & +0.892\\
$\epsilon=0.08$ & Baseline & 25 & +0.071 & +0.661 & +0.774\\
$\epsilon=0.08$ & L1 & 25 & -0.156 & +0.885 & +0.903\\
$\epsilon=0.08$ & L2 & 25 & -0.127 & +0.808 & +0.885\\
$\epsilon=0.08$ & L3 Full & 25 & -0.226 & +0.917 & +0.930\\
$\epsilon=0.08$ & L3 Stop & 25 & -0.073 & +0.827 & +0.894\\
$\epsilon=0.08$ & Random solver & 25 & -0.109 & +0.879 & +0.906\\
\addlinespace[2pt]
$\epsilon=0.12$ & Overall & 150 & -0.066 & +0.810 & +0.855\\
$\epsilon=0.12$ & Baseline & 25 & +0.119 & +0.613 & +0.701\\
$\epsilon=0.12$ & L1 & 25 & -0.108 & +0.845 & +0.868\\
$\epsilon=0.12$ & L2 & 25 & -0.053 & +0.771 & +0.844\\
$\epsilon=0.12$ & L3 Full & 25 & -0.196 & +0.884 & +0.893\\
$\epsilon=0.12$ & L3 Stop & 25 & -0.025 & +0.760 & +0.834\\
$\epsilon=0.12$ & Random solver & 25 & -0.061 & +0.837 & +0.864\\
\addlinespace[2pt]
$\epsilon=0.2$ & Overall & 150 & -0.026 & +0.746 & +0.772\\
$\epsilon=0.2$ & Baseline & 25 & +0.134 & +0.473 & +0.528\\
$\epsilon=0.2$ & L1 & 25 & -0.032 & +0.840 & +0.847\\
$\epsilon=0.2$ & L2 & 25 & -0.045 & +0.777 & +0.839\\
$\epsilon=0.2$ & L3 Full & 25 & -0.135 & +0.772 & +0.764\\
$\epsilon=0.2$ & L3 Stop & 25 & +0.043 & +0.721 & +0.789\\
$\epsilon=0.2$ & Random solver & 25 & -0.028 & +0.770 & +0.790\\
\addlinespace[2pt]
$\epsilon=0.4$ & Overall & 150 & +0.016 & +0.587 & +0.601\\
$\epsilon=0.4$ & Baseline & 25 & +0.163 & +0.148 & +0.143\\
$\epsilon=0.4$ & L1 & 25 & +0.019 & +0.734 & +0.766\\
$\epsilon=0.4$ & L2 & 25 & +0.000 & +0.671 & +0.745\\
$\epsilon=0.4$ & L3 Full & 25 & -0.030 & +0.621 & +0.590\\
$\epsilon=0.4$ & L3 Stop & 25 & +0.009 & +0.687 & +0.758\\
$\epsilon=0.4$ & Random solver & 25 & +0.043 & +0.581 & +0.589\\
\addlinespace[2pt]
\end{longtable}
\endgroup

\begingroup
\footnotesize
\setlength{\tabcolsep}{4.5pt}
\renewcommand{\arraystretch}{1.04}
\begin{longtable}{@{}llcrrr@{}}
\caption{Darcy Flow panel-wise Pearson correlations reproduced from Figure~\ref{fig:revision-darcy-jacobian-scatter-loglog}. The three correlation columns follow the three scatter-plot columns. Each binary-flip-budget block reports the pooled value and every method-specific value printed in the corresponding panel titles; correlations use base-10 log coordinates after excluding nonpositive attack increases.}\label{tab:revision-darcy-scatter-panel-correlations}\\
\toprule
Budget & Group & $n$ & $r_{\sigma_1(D\mathcal F)}$ & $r_{\sigma_1(J_{\mathcal E})}$ & $r_{\|J_{\mathcal E}^{\top}e\|_2}$\\
\midrule
\endfirsthead
\toprule
Budget & Group & $n$ & $r_{\sigma_1(D\mathcal F)}$ & $r_{\sigma_1(J_{\mathcal E})}$ & $r_{\|J_{\mathcal E}^{\top}e\|_2}$\\
\midrule
\endhead
\midrule
\multicolumn{6}{r}{continued on next page}\\
\endfoot
\bottomrule
\endlastfoot
$K=2$ & Overall & 168 & -0.340 & +0.257 & +0.481\\
$K=2$ & Baseline & 21 & -0.631 & -0.131 & +0.265\\
$K=2$ & L1 & 21 & +0.041 & +0.054 & +0.284\\
$K=2$ & L2 & 21 & +0.238 & +0.134 & +0.296\\
$K=2$ & L3 Stop & 21 & +0.226 & +0.112 & +0.291\\
$K=2$ & L3 Full & 21 & -0.283 & +0.081 & +0.331\\
$K=2$ & Physics & 21 & +0.008 & +0.027 & +0.235\\
$K=2$ & Random clean & 21 & +0.294 & +0.337 & +0.533\\
$K=2$ & Random solver & 21 & -0.011 & +0.007 & +0.310\\
\addlinespace[2pt]
$K=9$ & Overall & 164 & -0.327 & +0.261 & +0.489\\
$K=9$ & Baseline & 21 & -0.564 & -0.066 & +0.338\\
$K=9$ & L1 & 21 & +0.199 & +0.063 & +0.289\\
$K=9$ & L2 & 20 & +0.363 & +0.172 & +0.389\\
$K=9$ & L3 Stop & 21 & +0.396 & +0.124 & +0.263\\
$K=9$ & L3 Full & 19 & -0.428 & +0.037 & +0.333\\
$K=9$ & Physics & 21 & +0.174 & +0.012 & +0.235\\
$K=9$ & Random clean & 20 & +0.255 & +0.383 & +0.620\\
$K=9$ & Random solver & 21 & +0.019 & -0.025 & +0.271\\
\addlinespace[2pt]
$K=17$ & Overall & 161 & -0.242 & +0.275 & +0.535\\
$K=17$ & Baseline & 21 & -0.510 & -0.021 & +0.379\\
$K=17$ & L1 & 21 & +0.171 & +0.057 & +0.281\\
$K=17$ & L2 & 20 & +0.345 & +0.262 & +0.496\\
$K=17$ & L3 Stop & 21 & +0.398 & +0.150 & +0.304\\
$K=17$ & L3 Full & 16 & +0.293 & +0.247 & +0.481\\
$K=17$ & Physics & 21 & +0.123 & -0.009 & +0.208\\
$K=17$ & Random clean & 20 & +0.158 & +0.357 & +0.605\\
$K=17$ & Random solver & 21 & +0.004 & -0.043 & +0.238\\
\addlinespace[2pt]
$K=34$ & Overall & 160 & -0.272 & +0.302 & +0.557\\
$K=34$ & Baseline & 21 & -0.447 & +0.037 & +0.423\\
$K=34$ & L1 & 21 & +0.171 & +0.005 & +0.229\\
$K=34$ & L2 & 20 & +0.360 & +0.307 & +0.551\\
$K=34$ & L3 Stop & 20 & +0.320 & +0.046 & +0.256\\
$K=34$ & L3 Full & 16 & +0.338 & +0.414 & +0.600\\
$K=34$ & Physics & 21 & +0.134 & -0.034 & +0.176\\
$K=34$ & Random clean & 20 & +0.127 & +0.370 & +0.620\\
$K=34$ & Random solver & 21 & +0.043 & -0.062 & +0.211\\
\addlinespace[2pt]
$K=86$ & Overall & 162 & -0.302 & +0.347 & +0.556\\
$K=86$ & Baseline & 21 & -0.359 & +0.111 & +0.489\\
$K=86$ & L1 & 21 & +0.190 & +0.026 & +0.263\\
$K=86$ & L2 & 20 & +0.349 & +0.265 & +0.491\\
$K=86$ & L3 Stop & 21 & +0.368 & +0.100 & +0.265\\
$K=86$ & L3 Full & 18 & +0.241 & +0.502 & +0.631\\
$K=86$ & Physics & 21 & +0.162 & -0.001 & +0.231\\
$K=86$ & Random clean & 19 & +0.077 & +0.330 & +0.610\\
$K=86$ & Random solver & 21 & -0.015 & -0.115 & +0.161\\
\addlinespace[2pt]
$K=172$ & Overall & 165 & -0.393 & +0.382 & +0.604\\
$K=172$ & Baseline & 21 & -0.292 & +0.114 & +0.475\\
$K=172$ & L1 & 21 & +0.254 & +0.041 & +0.276\\
$K=172$ & L2 & 20 & +0.367 & +0.338 & +0.568\\
$K=172$ & L3 Stop & 21 & +0.363 & +0.108 & +0.340\\
$K=172$ & L3 Full & 20 & -0.028 & +0.226 & +0.583\\
$K=172$ & Physics & 21 & +0.183 & +0.033 & +0.265\\
$K=172$ & Random clean & 20 & +0.131 & +0.385 & +0.650\\
$K=172$ & Random solver & 21 & +0.134 & -0.053 & +0.219\\
\addlinespace[2pt]
$K=861$ & Overall & 167 & -0.389 & +0.410 & +0.585\\
$K=861$ & Baseline & 21 & -0.269 & -0.091 & +0.254\\
$K=861$ & L1 & 21 & +0.234 & +0.174 & +0.432\\
$K=861$ & L2 & 21 & +0.274 & +0.457 & +0.602\\
$K=861$ & L3 Stop & 21 & +0.297 & +0.220 & +0.447\\
$K=861$ & L3 Full & 20 & -0.170 & -0.196 & +0.185\\
$K=861$ & Physics & 21 & +0.281 & +0.118 & +0.380\\
$K=861$ & Random clean & 21 & +0.318 & +0.446 & +0.541\\
$K=861$ & Random solver & 21 & +0.278 & +0.198 & +0.501\\
\addlinespace[2pt]
$K=1722$ & Overall & 168 & -0.419 & +0.307 & +0.430\\
$K=1722$ & Baseline & 21 & -0.178 & -0.268 & +0.070\\
$K=1722$ & L1 & 21 & -0.308 & -0.568 & -0.656\\
$K=1722$ & L2 & 21 & +0.144 & +0.385 & +0.504\\
$K=1722$ & L3 Stop & 21 & -0.518 & -0.522 & -0.498\\
$K=1722$ & L3 Full & 21 & -0.237 & -0.266 & -0.348\\
$K=1722$ & Physics & 21 & -0.295 & -0.578 & -0.659\\
$K=1722$ & Random clean & 21 & +0.164 & +0.471 & +0.491\\
$K=1722$ & Random solver & 21 & -0.304 & -0.603 & -0.796\\
\addlinespace[2pt]
\end{longtable}
\endgroup

\begin{multicols}{2}
\subsection*{Navier--Stokes: Matched-Input Correlations and Attack Data}
\begin{figure}[H]
  \centering
  \includegraphics[width=\textwidth,height=0.85\textheight,keepaspectratio]
    {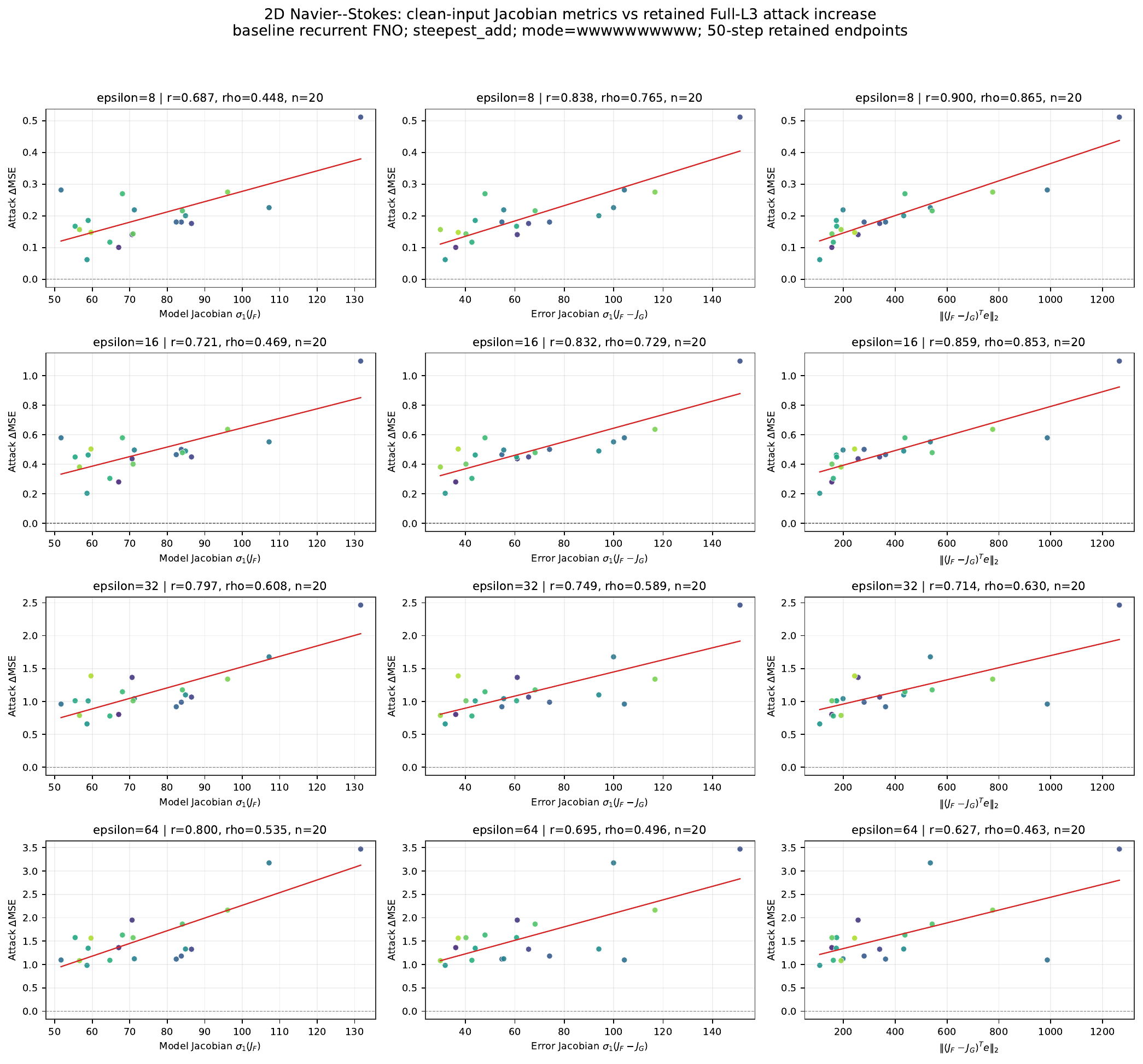}
  \caption{Navier--Stokes finite-attack $\Delta$MSE versus the three local
  Jacobian diagnostics, using the supplied raw-coordinate scatter plot.
  Rows correspond to $\epsilon=8,16,32,64$ and columns to
  $\sigma_1(D\mathcal F)$, $\sigma_1(J_{\mathcal E})$, and
  $\|J_{\mathcal E}^{\top}e\|_2$. Each panel uses the same 20 inputs from
  one retained recurrent-FNO checkpoint. The paired Pearson and Spearman
  coefficients are in \tabref{tab:revision-ns-scatter-correlations}.}
  \label{fig:revision-ns-jacobian-scatter}
\end{figure}

The NS study follows the same 20 initial conditions at four perturbation budgets,
giving 80 attack records from one retained recurrent-FNO checkpoint.
Figure~\ref{fig:revision-ns-jacobian-scatter} shows all three diagnostic columns;
Tables~\ref{tab:revision-ns-scatter-correlations} and
\ref{tab:revision-ns-scatter-input-data} report the paired correlations and the
input-level values used in those panels. These are across-input comparisons,
with the response and metric normalizations stated explicitly below.
\end{multicols}
\begin{table}[H]
\centering\footnotesize
\setlength{\tabcolsep}{3pt}
\renewcommand{\arraystretch}{1.1}
\caption{Navier--Stokes panel-wise correlations for Figure~\ref{fig:revision-ns-jacobian-scatter}. Each cell is Pearson $r$ / Spearman $\rho$ in raw coordinates; all recorded two-sided $p$ values are below 0.05. The same 20 inputs and one retained recurrent-FNO checkpoint are used at each budget.}\label{tab:revision-ns-scatter-correlations}
\begin{tabular*}{\textwidth}{@{\extracolsep{\fill}}lrrrr@{}}
\toprule
$\epsilon$ & $n$ & $\sigma_1(D\mathcal F)$ & $\sigma_1(J_{\mathcal E})$ & $\|J_{\mathcal E}^{\top}e\|_2$\\
\midrule
8 & 20 & 0.687 / 0.448 & 0.838 / 0.765 & 0.900 / 0.865\\
16 & 20 & 0.721 / 0.469 & 0.832 / 0.729 & 0.859 / 0.853\\
32 & 20 & 0.797 / 0.608 & 0.749 / 0.589 & 0.714 / 0.630\\
64 & 20 & 0.800 / 0.535 & 0.695 / 0.496 & 0.627 / 0.463\\
\bottomrule
\end{tabular*}
\end{table}

\begin{multicols}{2}
\paragraph{NS metric and response scope.}
The full-resolution $256^2$ Jacobian norms use matrix-free JVP/VJP power iteration, with relative-change tolerance $10^{-3}$ reached by all 20 inputs. J-error is the reverse-mode gradient norm of $\tfrac12\|e\|_2^2$. The retained 50-step Steepest Add full-L3 attacks use $\epsilon=8,16,32,64$. The plotted response is $\Delta\mathrm{MSE}=(\|e_{\rm adv}\|_2^2-\|e_{\rm clean}\|_2^2)/65536$, so its normalization differs from the half-squared local loss by a constant that does not change correlations.
The residual-aware metrics correlate more strongly at smaller budgets: J-error Pearson decreases from 0.900 to 0.627 across the sweep. At budgets 32 and 64 the model-Jacobian Pearson is larger, so there is no universal three-metric ranking at finite radius. These are across-input coefficients for one checkpoint, not across-trained-method correlations. A dense $65536\times65536$ float32 Jacobian alone occupies 16 GiB; matrix-free products avoid that allocation.
\end{multicols}

\begin{table}[H]
\centering\footnotesize
\setlength{\tabcolsep}{3pt}
\renewcommand{\arraystretch}{1.1}
\caption{The 20 matched NS inputs underlying the scatter panels. Columns give the three clean-input diagnostics and the measured attack-induced MSE increase at each budget; each row follows one input across all four attacks.}\label{tab:revision-ns-scatter-input-data}
\begin{tabular*}{\textwidth}{@{\extracolsep{\fill}}lrrrrrrr@{}}
\toprule
Input & $\sigma_1(D\mathcal F)$ & $\sigma_1(J_{\mathcal E})$ & $\|J_{\mathcal E}^{\top}e\|_2$ & $\Delta_8$ & $\Delta_{16}$ & $\Delta_{32}$ & $\Delta_{64}$\\
\midrule
0 & 67.045 & 36.081 & 155.2 & 0.10055 & 0.27984 & 0.80267 & 1.3607\\
1 & 70.615 & 60.976 & 256.65 & 0.1409 & 0.43672 & 1.3657 & 1.9495\\
2 & 86.463 & 65.564 & 339.75 & 0.17604 & 0.44926 & 1.0667 & 1.326\\
3 & 131.59 & 151.13 & 1264.6 & 0.5115 & 1.0982 & 2.4654 & 3.4692\\
4 & 83.756 & 74.048 & 279.92 & 0.18045 & 0.50029 & 0.98859 & 1.1805\\
5 & 82.4 & 54.759 & 362.66 & 0.18062 & 0.46472 & 0.91963 & 1.1153\\
6 & 51.665 & 104.33 & 986.6 & 0.28169 & 0.57872 & 0.96077 & 1.0961\\
7 & 107.16 & 99.952 & 535.39 & 0.22601 & 0.55117 & 1.6782 & 3.1732\\
8 & 71.216 & 55.511 & 198.57 & 0.21892 & 0.49615 & 1.0426 & 1.1216\\
9 & 84.864 & 93.998 & 432.63 & 0.20051 & 0.48932 & 1.0999 & 1.3307\\
10 & 58.59 & 31.824 & 108.59 & 0.061901 & 0.20296 & 0.6588 & 0.98103\\
11 & 58.893 & 43.966 & 172.68 & 0.18556 & 0.46226 & 1.0088 & 1.3476\\
12 & 55.424 & 60.7 & 173.91 & 0.16724 & 0.44867 & 1.0109 & 1.5755\\
13 & 64.683 & 42.626 & 160.99 & 0.117 & 0.30383 & 0.77922 & 1.0898\\
14 & 68.038 & 47.921 & 437.37 & 0.26978 & 0.57873 & 1.1473 & 1.6299\\
15 & 84.035 & 68.204 & 542.49 & 0.21601 & 0.47824 & 1.1764 & 1.8637\\
16 & 70.876 & 40.215 & 155.69 & 0.14336 & 0.40056 & 1.0103 & 1.5727\\
17 & 96.107 & 116.72 & 776.01 & 0.27505 & 0.63596 & 1.3392 & 2.163\\
18 & 56.588 & 29.869 & 190.99 & 0.15679 & 0.38097 & 0.7876 & 1.0823\\
19 & 59.651 & 37.088 & 243.2 & 0.14792 & 0.50296 & 1.3884 & 1.5642\\
\bottomrule
\end{tabular*}
\end{table}

\begin{multicols}{2}
\paragraph{Vector-direction similarity.}
The vector diagnostic compares three input-space directions: the finite attack
direction \(\delta_T\), the Jacobian-error direction
\(g_\theta=J_{\mathcal E}^{\top}e_\theta\), and the leading right singular
direction \(v_1\) of \(J_{\mathcal E}\).  The available Burgers summaries store
absolute-cosine direction statistics, which we report as
\[
  s(u,v)=\left|\frac{\langle u,v\rangle}{\|u\|_2\|v\|_2}\right|,
  \qquad \theta_s(u,v)=\arccos s(u,v).
\]

\end{multicols}

\begin{table}[H]
\centering
\footnotesize
\setlength{\tabcolsep}{4pt}
\renewcommand{\arraystretch}{1.16}
\caption{Vector-direction similarity diagnostics for Burgers. Entries are
similarity score / angle in degrees from the available absolute-cosine
summaries.}
\label{tab:robustness_vector_angle_summary}
\begin{tabular*}{\textwidth}{@{\extracolsep{\fill}}lrrrr@{}}
\toprule
Diagnostic pool & \(n\) &
\((\delta_T,v_1)\) &
\((\delta_T,g_\theta)\) &
\((v_1,g_\theta)\) \\
\midrule
all six methods: 25 mixed \(\times 6\) & 150 & 0.246/74.9 & 0.760/39.3 & 0.288/72.1 \\
base/Loss 1--3 only: 25 mixed \(\times 4\) & 100 & 0.200/77.6 & 0.780/37.7 & 0.240/75.2 \\
\bottomrule
\end{tabular*}
\end{table}

\begingroup
\captionsetup[figure]{font=small,skip=2pt}
\setlength{\intextsep}{5pt}
\setlength{\textfloatsep}{5pt}
\setlength{\floatsep}{5pt}
\begin{figure}[H]
  \centering
  \includegraphics[width=0.86\textwidth]{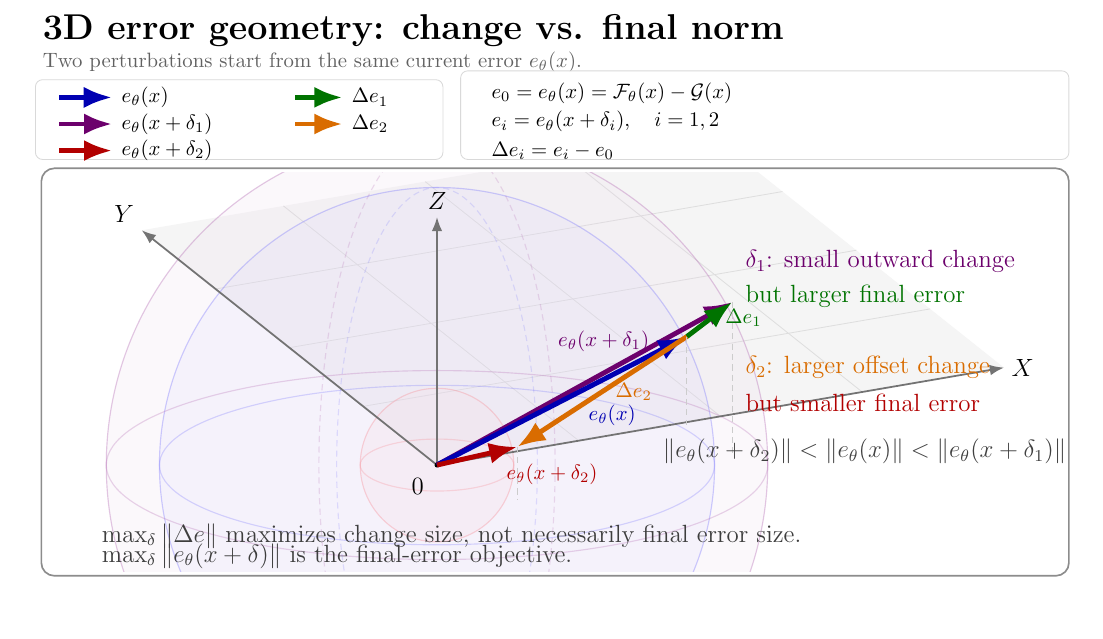}
  \caption{Error-vector geometry for adversarial loss increase. The figure
  contrasts final error norm growth with the error-difference direction used by
  local Jacobian-error diagnostics.}
  \label{fig:error-difference-vs-final-norm}
\end{figure}
\endgroup

\begin{multicols}{2}
\subsection*{Singular-Function Geometry of the Jacobian-Error Direction}

Let $T:=D\mathcal{E}_\theta[x]:\mathcal{X}\to\mathcal{Y}$ be the local error
Jacobian. The empirical Jacobians in this paper are finite-dimensional
matrices and therefore admit the usual matrix SVD. The continuum formula below
is the compact-operator analogue: assume that $T$ is compact. Without this
compactness assumption, a general bounded operator may have continuous
spectrum and need not admit a countable singular-function expansion. Let
$e:=e_\theta[x]$ denote the current error field. For compact $T$, the singular
system is constructed by applying the spectral theorem to the positive
self-adjoint compact operator $T^*T$:
\begin{equation}
  \begin{aligned}
    T^*T v_i &= \sigma_i^2 v_i,\\
    u_i &= \sigma_i^{-1}T v_i,\\
    T v_i &= \sigma_i u_i,\\
    T^*u_i &= \sigma_i v_i .
  \end{aligned}
\end{equation}
Here $v_i\in\mathcal{X}$ are right singular functions,
$u_i\in\mathcal{Y}$ are left singular functions, and the positive singular
values are indexed as a finite or countable nonincreasing sequence
$\sigma_1\ge\sigma_2\ge\cdots>0$, repeated with multiplicity. If there are
infinitely many positive singular values, then $\sigma_i\downarrow0$. Zero
singular directions, if present, lie in $\ker(T)$ and do not contribute to the
expansions below.

For simplicity, suppose the leading singular value is simple. If it has
multiplicity greater than one, replace $u_1$ and $v_1$ by the corresponding
leading left and right singular subspaces.

Decomposing the current error field into its range-relevant left singular
components,
\begin{equation}
  \begin{aligned}
    e
    &=
    \sum_i
    \langle e,u_i\rangle_{\mathcal{Y}} u_i
    +
    e_\perp,\\
    e_\perp
    &\in
    \ker(T^*)=(\overline{\operatorname{range}(T)})^\perp,
  \end{aligned}
\end{equation}
gives
\begin{equation}
  D\mathcal{E}_\theta[x]^*e_\theta[x]
  =
  T^*e
  =
  \sum_i
  \sigma_i
  \langle e,u_i\rangle_{\mathcal{Y}} v_i .
\end{equation}
Thus the coefficient of the Jacobian-error direction along the leading
right singular direction $v_1$ is
\begin{equation}
  \sigma_1
  \langle e,u_1\rangle_{\mathcal{Y}},
\end{equation}
so the weight is controlled by both the singular value and the projection of
the current error field onto the corresponding left singular function. In
particular,
\begin{equation}
  \frac{
    |\langle T^*e,v_1\rangle_{\mathcal{X}}|
  }{
    \|T^*e\|_{\mathcal{X}}
  }
  =
  \frac{
    \sigma_1|\langle e,u_1\rangle_{\mathcal{Y}}|
  }{
    \left(
      \sum_i
      \sigma_i^2
      |\langle e,u_i\rangle_{\mathcal{Y}}|^2
    \right)^{1/2}
  } .
\end{equation}
Therefore the Jacobian-error direction is close to the leading operator-norm
input direction $v_1$ only when the leading weighted left-singular projection
dominates:
\begin{equation}
  \sigma_1|\langle e,u_1\rangle_{\mathcal{Y}}|
  \gg
  \left(
    \sum_{i\ge2}
    \sigma_i^2
    |\langle e,u_i\rangle_{\mathcal{Y}}|^2
  \right)^{1/2}.
\end{equation}
If the current error field instead projects mainly onto other left singular
functions, then $T^*e$ points toward the corresponding right singular
directions, and the Jacobian-error direction need not be close to the
operator-norm direction.
\end{multicols}

\begin{multicols}{2}
\subsection*{Local SVD Diagnostics of Model, Solver, and Error Jacobians}
The local SVD diagnostics below interpret the solver-integrated
attack results. At selected Burgers inputs, we linearize the trained neural
operator $\mathcal{F}_\theta$, the solver $\mathcal{G}$, and the error operator
$\mathcal{E}_\theta=\mathcal{F}_\theta-\mathcal{G}$ through the Jacobians
$D\mathcal{F}_\theta[x]$, $D\mathcal{G}[x]$, and
$D\mathcal{E}_\theta[x]$. The right singular vectors live in the input
perturbation space, so they are the local directions most directly related to
attack perturbations.

For two Jacobians $A$ and $B$, let
$V_A^{(k)}=[v_{A,1},\ldots,v_{A,k}]$ and
$V_B^{(k)}=[v_{B,1},\ldots,v_{B,k}]$ be their leading $k$ right singular
vectors. The top-$k$ subspace similarity uses the principal cosines
\begin{equation*}
  \rho_i^{(k)}
  =
  \sigma_i\!\left((V_A^{(k)})^\top V_B^{(k)}\right),
  \quad
  s_k(A,B)=\frac{1}{k}\sum_{i=1}^k \rho_i^{(k)} .
\end{equation*}
This score compares the two spans as subspaces, allowing the best rotation
within the leading $k$-dimensional directions rather than forcing rank-$i$ to
match rank-$i$. The rank-matched vector angle is instead
\begin{equation*}
  \alpha_r(A,B)
  =
  \arccos |\langle v_{A,r},v_{B,r}\rangle| .
\end{equation*}

This is only a local linear probe. A pure Jacobian operator-norm problem would
follow the top right singular vector, but $\mathcal{L}_3$ loss growth also
depends on the current error field through
$D\mathcal{E}_\theta[x]^\top e_\theta(x)$, and finite PGD includes nonlinear
and projection effects. The probe is therefore approximate, but it explains
the same pattern seen in the attacks: model-only objectives follow common
model--solver modes, while $\mathcal{L}_3$ exposes mismatch modes that are more
effective for increasing model--solver discrepancy and are often more
oscillatory.

For FNO, \tabref{tab:local-svd-subspace-similarity} and
\tabref{tab:local-svd-rank-angles} show the key mechanism. The leading right
singular directions of the model and solver are very close, so perturbations
that follow the model response also move the solver in nearly the same local
direction. The leading right singular directions of the error operator are
much farther from the model directions. Thus $\mathcal{L}_1$ or $\mathcal{L}_2$ directions
can change both outputs together without strongly increasing their difference;
large model--solver discrepancy requires the residual directions exposed by
$D\mathcal{E}_\theta[x]$.
\end{multicols}

\begin{table}[H]
\centering
\footnotesize
\caption{Top-$k$ right-singular subspace similarity. Entries are the mean
principal cosine $s_k$; higher value indicates stronger local alignment. For FNO,
the model and solver subspaces are highly aligned, while the error subspace
is separated; this is especially clear at $\nu=10^{-2}$. DeepONet behaves
differently: its model and error subspaces are close to each other and both
are far from the solver subspace.}
\label{tab:local-svd-subspace-similarity}
\setlength{\tabcolsep}{3pt}
\begin{tabular}{llrrrr}
\toprule
Family & Operators & $k=1$ & $k=2$ & $k=4$ & $k=8$ \\
\midrule
FNO, $\nu=10^{-3}$ & $\mathcal{F}_\theta,\mathcal{G}$ & $0.996$ & $0.997$ & $0.947$ & $0.981$ \\
FNO, $\nu=10^{-3}$ & $\mathcal{F}_\theta,\mathcal{E}_\theta$ & $0.218$ & $0.784$ & $0.816$ & $0.602$ \\
FNO, $\nu=10^{-3}$ & $\mathcal{G},\mathcal{E}_\theta$ & $0.206$ & $0.789$ & $0.799$ & $0.609$ \\
FNO, $\nu=10^{-2}$ & $\mathcal{F}_\theta,\mathcal{G}$ & $0.999$ & $1.000$ & $1.000$ & $0.999$ \\
FNO, $\nu=10^{-2}$ & $\mathcal{F}_\theta,\mathcal{E}_\theta$ & $0.427$ & $0.290$ & $0.156$ & $0.088$ \\
FNO, $\nu=10^{-2}$ & $\mathcal{G},\mathcal{E}_\theta$ & $0.438$ & $0.294$ & $0.157$ & $0.088$ \\
DeepONet, $\nu=10^{-2}$ & $\mathcal{F}_\theta,\mathcal{G}$ & $0.125$ & $0.164$ & $0.182$ & $0.170$ \\
DeepONet, $\nu=10^{-2}$ & $\mathcal{F}_\theta,\mathcal{E}_\theta$ & $0.985$ & $0.979$ & $0.977$ & $0.981$ \\
DeepONet, $\nu=10^{-2}$ & $\mathcal{G},\mathcal{E}_\theta$ & $0.010$ & $0.011$ & $0.015$ & $0.024$ \\
\bottomrule
\end{tabular}
\end{table}

\begin{table}[H]
\centering
\footnotesize
\caption{Rank-matched right singular vector angles, in degrees. Unlike the
top-$k$ subspace similarity in \tabref{tab:local-svd-subspace-similarity}, this
table forces the $r$th right singular vector of one Jacobian to match the $r$th
right singular vector of the other Jacobian. The FNO rows show small
model--solver angles but large model--error angles, so model- or solver-driven
directions mostly move both outputs together; discrepancy growth needs the
error directions.}
\label{tab:local-svd-rank-angles}
\setlength{\tabcolsep}{3pt}
\begin{tabular}{llrrrrrrrr}
\toprule
Family & Operators & $r=1$ & $r=2$ & $r=3$ & $r=4$ & $r=5$ & $r=6$ & $r=7$ & $r=8$ \\
\midrule
FNO, $\nu=10^{-3}$ & $\mathcal{F}_\theta,\mathcal{G}$ & $4.86$ & $5.18$ & $23.71$ & $39.07$ & $53.83$ & $41.17$ & $27.04$ & $40.44$ \\
FNO, $\nu=10^{-3}$ & $\mathcal{F}_\theta,\mathcal{E}_\theta$ & $74.32$ & $55.8$ & $61.36$ & $81.66$ & $72.68$ & $79.56$ & $87.88$ & $88.62$ \\
FNO, $\nu=10^{-2}$ & $\mathcal{F}_\theta,\mathcal{G}$ & $2.32$ & $2.15$ & $1.76$ & $1.5$ & $1.79$ & $2.85$ & $3.65$ & $6.77$ \\
FNO, $\nu=10^{-2}$ & $\mathcal{F}_\theta,\mathcal{E}_\theta$ & $62.53$ & $89.9$ & $89.96$ & $89.87$ & $89.95$ & $89.96$ & $89.91$ & $89.73$ \\
\bottomrule
\end{tabular}
\end{table}

\begin{table}[H]
\centering
\footnotesize
\caption{FNO leading singular values. Values are averaged over the five local
inputs used in the SVD diagnostic.}
\label{tab:local-svd-singular-values}
\setlength{\tabcolsep}{3pt}
\begin{tabular*}{\textwidth}{@{\extracolsep{\fill}}llrrrrrrrr@{}}
\toprule
Model & Jacobian & $\sigma_1$ & $\sigma_2$ & $\sigma_3$ & $\sigma_4$ & $\sigma_5$ & $\sigma_6$ & $\sigma_7$ & $\sigma_8$ \\
\midrule
FNO, $\nu=10^{-3}$ & $D\mathcal{F}_\theta$ & $3.895$ & $3.125$ & $2.114$ & $1.784$ & $1.048$ & $0.645$ & $0.542$ & $0.47$ \\
FNO, $\nu=10^{-3}$ & $D\mathcal{G}$ & $4.095$ & $3.277$ & $2.207$ & $1.854$ & $1.084$ & $0.676$ & $0.538$ & $0.483$ \\
FNO, $\nu=10^{-3}$ & $D\mathcal{E}_\theta$ & $0.868$ & $0.727$ & $0.46$ & $0.311$ & $0.263$ & $0.246$ & $0.224$ & $0.199$ \\
FNO, $\nu=10^{-2}$ & $D\mathcal{F}_\theta$ & $1.379$ & $1.24$ & $0.88$ & $0.657$ & $0.473$ & $0.323$ & $0.289$ & $0.16$ \\
FNO, $\nu=10^{-2}$ & $D\mathcal{G}$ & $1.373$ & $1.234$ & $0.88$ & $0.654$ & $0.471$ & $0.32$ & $0.286$ & $0.156$ \\
FNO, $\nu=10^{-2}$ & $D\mathcal{E}_\theta$ & $0.037$ & $0.031$ & $0.031$ & $0.031$ & $0.031$ & $0.031$ & $0.031$ & $0.031$ \\
\bottomrule
\end{tabular*}
\end{table}

\begin{table}[H]
\centering
\footnotesize
\caption{FNO Fourier-energy quantiles of individual leading right singular
vectors. Each row reports one right singular vector $v_r$; values are averaged
over the five local inputs, but not averaged across ranks. The quantity
$q_p(v_r)$ is reported as a Fourier mode index: for length-$1024$ vectors, the
non-DC one-sided rFFT modes are $1,\ldots,512$. For each local input, $q_p$ is
the first mode where the cumulative non-DC raw rFFT energy reaches $p\%$; the
table reports the mean of these five integer modes, hence non-integer entries.
Larger values indicate higher-frequency directions.}
\label{tab:local-svd-frequency}
\setlength{\tabcolsep}{2pt}
\begin{tabular*}{\textwidth}{@{\extracolsep{\fill}}lllrrrrrrrrr@{}}
\toprule
Model & Rank & Jacobian & $q_1$ & $q_5$ & $q_{10}$ & $q_{25}$ & $q_{50}$ & $q_{70}$ & $q_{90}$ & $q_{95}$ & $q_{99}$ \\
\midrule
FNO, $\nu=10^{-3}$ & $1$ & $D\mathcal{F}_\theta$ & $1$ & $1$ & $1$ & $1$ & $1.2$ & $1.8$ & $2.8$ & $5.4$ & $8.6$ \\
FNO, $\nu=10^{-3}$ & $1$ & $D\mathcal{G}$ & $1$ & $1$ & $1$ & $1$ & $1.2$ & $1.8$ & $2.8$ & $5.8$ & $10.6$ \\
FNO, $\nu=10^{-3}$ & $1$ & $D\mathcal{E}_\theta$ & $1$ & $1$ & $1$ & $1$ & $1.6$ & $2.4$ & $4.2$ & $6.8$ & $15.4$ \\
\midrule
FNO, $\nu=10^{-3}$ & $2$ & $D\mathcal{F}_\theta$ & $1$ & $1$ & $1$ & $1$ & $1.6$ & $2$ & $3.6$ & $4.8$ & $9.4$ \\
FNO, $\nu=10^{-3}$ & $2$ & $D\mathcal{G}$ & $1$ & $1$ & $1$ & $1$ & $1.6$ & $2.2$ & $3.6$ & $5$ & $10.2$ \\
FNO, $\nu=10^{-3}$ & $2$ & $D\mathcal{E}_\theta$ & $1$ & $1$ & $1$ & $1$ & $1.6$ & $2.6$ & $4.2$ & $8.2$ & $17$ \\
\midrule
FNO, $\nu=10^{-3}$ & $3$ & $D\mathcal{F}_\theta$ & $1$ & $1$ & $1$ & $1.2$ & $2.2$ & $3$ & $4.2$ & $5.2$ & $9.2$ \\
FNO, $\nu=10^{-3}$ & $3$ & $D\mathcal{G}$ & $1$ & $1$ & $1$ & $1.2$ & $2$ & $3$ & $4.2$ & $5.2$ & $11$ \\
FNO, $\nu=10^{-3}$ & $3$ & $D\mathcal{E}_\theta$ & $1$ & $1$ & $1$ & $1.4$ & $2.4$ & $3.6$ & $10.8$ & $14.8$ & $41$ \\
\midrule
FNO, $\nu=10^{-3}$ & $4$ & $D\mathcal{F}_\theta$ & $1$ & $1$ & $1$ & $1.2$ & $2$ & $3$ & $4$ & $5$ & $9$ \\
FNO, $\nu=10^{-3}$ & $4$ & $D\mathcal{G}$ & $1$ & $1$ & $1$ & $1.4$ & $2.2$ & $3.4$ & $4.8$ & $5.6$ & $10.2$ \\
FNO, $\nu=10^{-3}$ & $4$ & $D\mathcal{E}_\theta$ & $1.4$ & $2.4$ & $3.8$ & $6.8$ & $11.2$ & $13.8$ & $16.4$ & $19.6$ & $34.2$ \\
\midrule
FNO, $\nu=10^{-2}$ & $1$ & $D\mathcal{F}_\theta$ & $1$ & $1$ & $1$ & $1$ & $1$ & $1$ & $2$ & $2.4$ & $4.2$ \\
FNO, $\nu=10^{-2}$ & $1$ & $D\mathcal{G}$ & $1$ & $1$ & $1$ & $1$ & $1$ & $1$ & $2$ & $2.4$ & $4.2$ \\
FNO, $\nu=10^{-2}$ & $1$ & $D\mathcal{E}_\theta$ & $20.8$ & $34.4$ & $43.2$ & $59.2$ & $80.2$ & $93.6$ & $109$ & $114.2$ & $121.4$ \\
\midrule
FNO, $\nu=10^{-2}$ & $2$ & $D\mathcal{F}_\theta$ & $1$ & $1$ & $1$ & $1$ & $1$ & $1$ & $1.4$ & $2$ & $3.8$ \\
FNO, $\nu=10^{-2}$ & $2$ & $D\mathcal{G}$ & $1$ & $1$ & $1$ & $1$ & $1$ & $1$ & $1.4$ & $2$ & $3.8$ \\
FNO, $\nu=10^{-2}$ & $2$ & $D\mathcal{E}_\theta$ & $94.8$ & $157.2$ & $197.6$ & $272.2$ & $355$ & $410$ & $465.6$ & $481.4$ & $498.4$ \\
\midrule
FNO, $\nu=10^{-2}$ & $3$ & $D\mathcal{F}_\theta$ & $1$ & $1$ & $1$ & $1$ & $1$ & $1.6$ & $2$ & $2.4$ & $3.8$ \\
FNO, $\nu=10^{-2}$ & $3$ & $D\mathcal{G}$ & $1$ & $1$ & $1$ & $1$ & $1$ & $1.6$ & $1.8$ & $2.4$ & $3.8$ \\
FNO, $\nu=10^{-2}$ & $3$ & $D\mathcal{E}_\theta$ & $66.6$ & $107.6$ & $135$ & $188$ & $254$ & $333.6$ & $403.6$ & $428.8$ & $465.4$ \\
\midrule
FNO, $\nu=10^{-2}$ & $4$ & $D\mathcal{F}_\theta$ & $1$ & $1.2$ & $1.2$ & $1.4$ & $2$ & $2$ & $2.2$ & $3.2$ & $4.2$ \\
FNO, $\nu=10^{-2}$ & $4$ & $D\mathcal{G}$ & $1$ & $1.2$ & $1.2$ & $1.4$ & $2$ & $2$ & $2.2$ & $3.2$ & $4.2$ \\
FNO, $\nu=10^{-2}$ & $4$ & $D\mathcal{E}_\theta$ & $52.8$ & $85$ & $107.2$ & $151.6$ & $232.4$ & $404.6$ & $474.8$ & $489$ & $502.8$ \\
\midrule
\bottomrule
\end{tabular*}
\end{table}

\begin{multicols}{2}
For FNO, especially at $\nu=10^{-2}$, $D\mathcal{F}_\theta[x]$ and
$D\mathcal{G}[x]$ share leading input subspaces, so model-only objectives excite
common modes; $D\mathcal{E}_\theta[x]$ cancels that shared response and exposes
weaker but more discrepancy-relevant, higher-frequency mismatch directions.
\tabref{tab:local-svd-singular-values} shows that these error directions have
smaller singular values, while \tabref{tab:local-svd-frequency} shows that they
are much higher frequency. At $\nu=10^{-2}$, the model and solver directions have
$q_{50}(v_r)\le 2$ for ranks $1$--$4$, while the error directions already have
$q_{50}=80.2$ at rank $1$ and $q_{50}\ge 232.4$ for ranks $2$--$4$; this is a
rank-wise effect, not an artifact of averaging $v_1,\ldots,v_4$ together. The
tail quantiles give the same message: the model and solver modes remain below
$q_{99}\le4.2$, whereas the error modes reach $q_{99}=121.4$--$502.8$.
\end{multicols}

\begin{figure}[H]
  \centering
  \includegraphics[height=0.88\textheight,keepaspectratio]
    {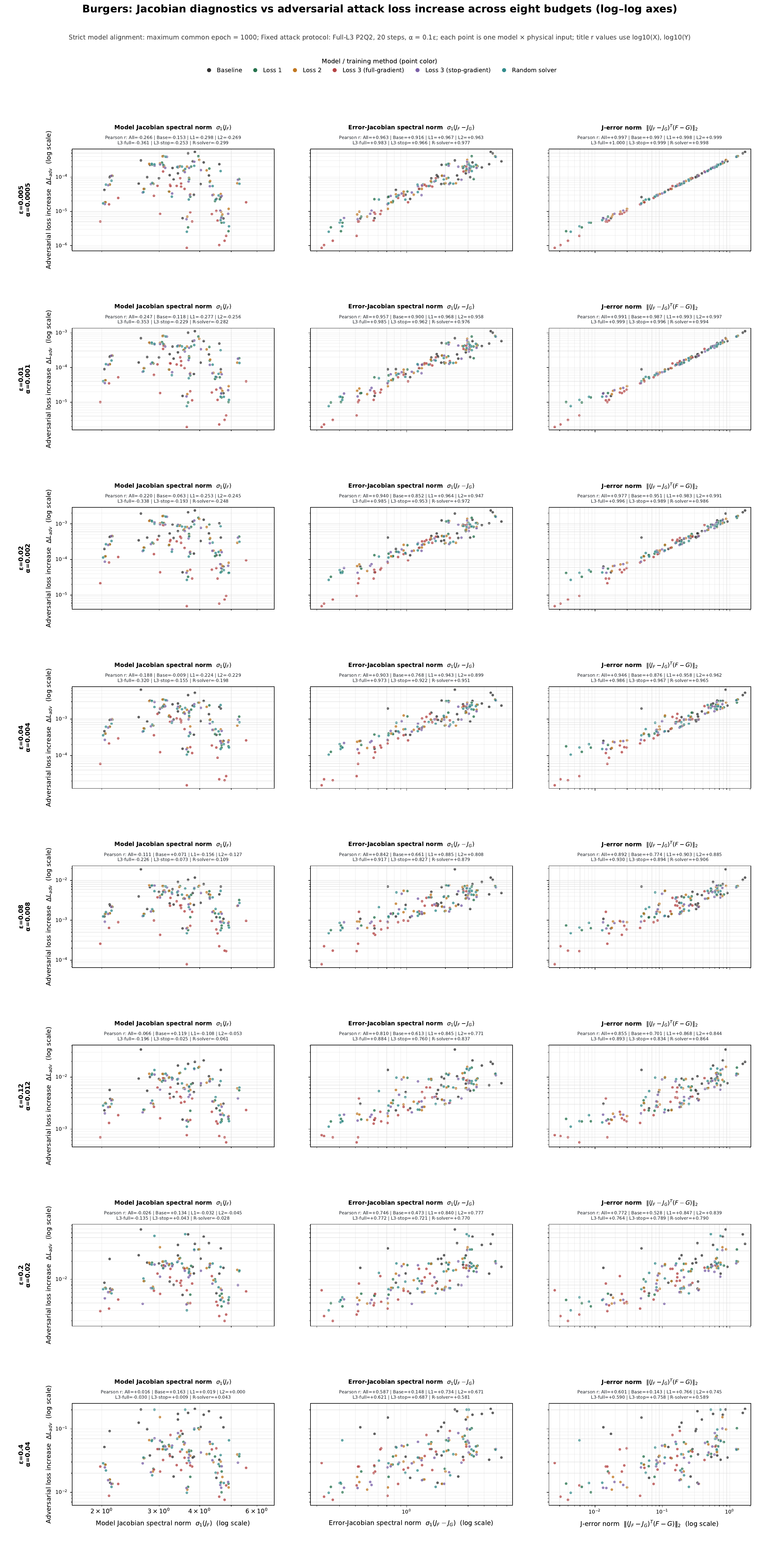}
  \caption{Burgers finite-attack increase versus three Jacobian diagnostics on
  base-10 log--log axes. Rows are the eight attack budgets and columns are the
  model Jacobian spectral norm, error-Jacobian spectral norm, and
  Jacobian-error norm. Each panel pools six exact-epoch-1000 checkpoints and 25
  inputs per checkpoint ($n=150$).}
  \label{fig:revision-burgers-jacobian-scatter-loglog}
\end{figure}

\begin{figure}[H]
  \centering
  \includegraphics[height=0.88\textheight,keepaspectratio]
    {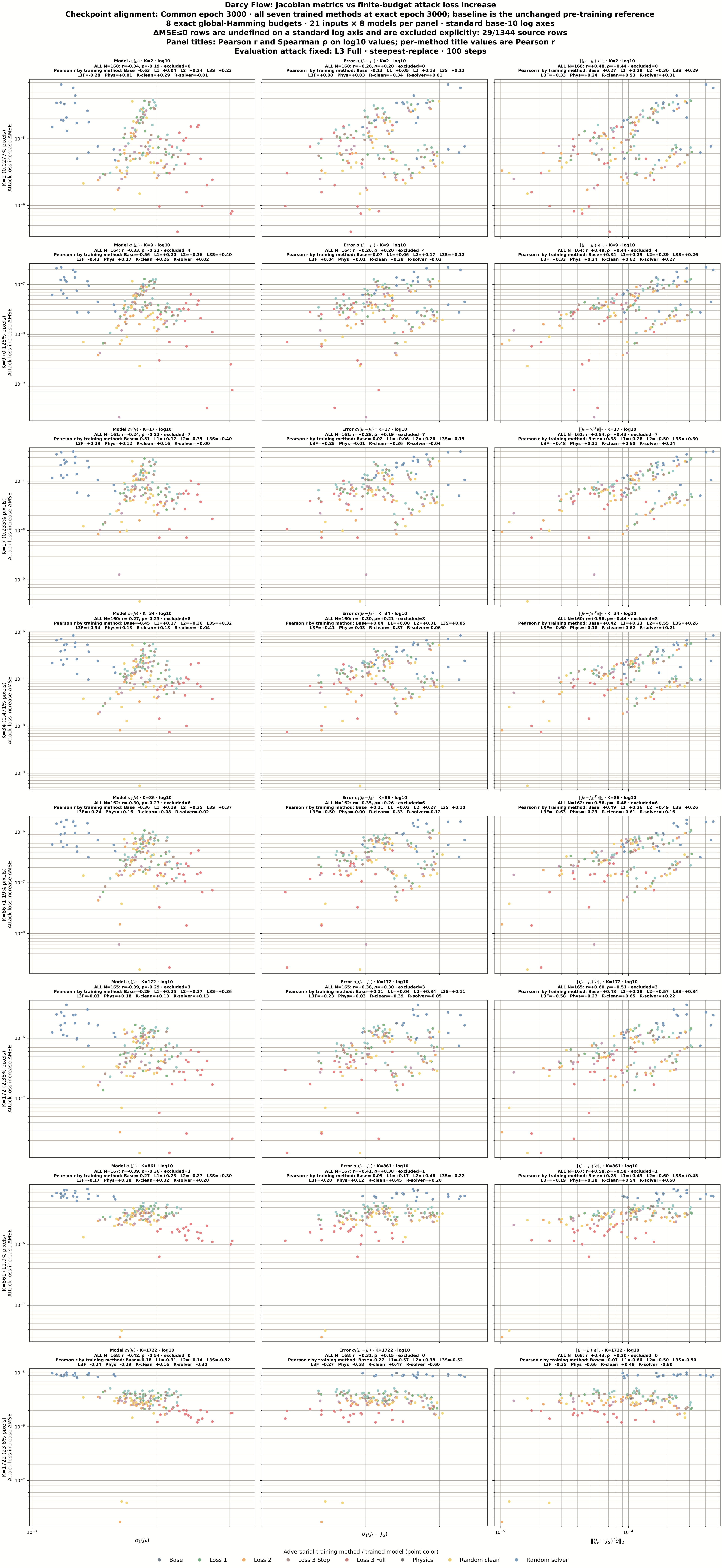}
  \caption{Darcy Flow finite-attack increase versus three Jacobian diagnostics
  on base-10 log--log axes. Rows are eight binary-flip budgets and columns are
  the model Jacobian spectral norm, error-Jacobian spectral norm, and
  Jacobian-error norm. Each panel starts from eight exact-epoch-3000
  checkpoints and 21 fixed OOD inputs per checkpoint; nonpositive attack
  increases are explicitly excluded from the log view and the panel-specific
  sample count is reported in
  \tabref{tab:robustness_metric_correlation_summary}.}
  \label{fig:revision-darcy-jacobian-scatter-loglog}
\end{figure}


\begin{thebibliography}{36}
\providecommand{\natexlab}[1]{#1}

\bibitem[{Adesoji and Chen(2022)}]{adesoji2022evaluating}
Adesoji, A.~D.; and Chen, P.-Y. 2022.
\newblock Evaluating the Adversarial Robustness for Fourier Neural Operators.
\newblock In \emph{ICLR 2022 Workshop on Socially Responsible Machine
  Learning}.

\bibitem[{Andriushchenko et~al.(2020)Andriushchenko, Croce, Flammarion, and
  Hein}]{andriushchenko2019square}
Andriushchenko, M.; Croce, F.; Flammarion, N.; and Hein, M. 2020.
\newblock Square Attack: A Query-Efficient Black-Box Adversarial Attack via
  Random Search.
\newblock In \emph{Computer Vision -- ECCV 2020}, Lecture Notes in Computer
  Science, 484--501. Springer.

\bibitem[{Bousquet and Elisseeff(2002)}]{bousquet2002stability}
Bousquet, O.; and Elisseeff, A. 2002.
\newblock Stability and Generalization.
\newblock \emph{Journal of Machine Learning Research}, 2: 499--526.

\bibitem[{Brendel, Rauber, and Bethge(2018)}]{brendel2017decision}
Brendel, W.; Rauber, J.; and Bethge, M. 2018.
\newblock Decision-Based Adversarial Attacks: Reliable Attacks Against
  Black-Box Machine Learning Models.
\newblock In \emph{International Conference on Learning Representations}.

\bibitem[{Brown et~al.(2017)Brown, Man{\'e}, Roy, Abadi, and
  Gilmer}]{brown2017adversarialpatch}
Brown, T.~B.; Man{\'e}, D.; Roy, A.; Abadi, M.; and Gilmer, J. 2017.
\newblock Adversarial Patch.
\newblock In \emph{NIPS 2017 Workshop on Machine Learning and Computer
  Security}.

\bibitem[{Carlini and Wagner(2017)}]{carlini2017towards}
Carlini, N.; and Wagner, D. 2017.
\newblock Towards Evaluating the Robustness of Neural Networks.
\newblock In \emph{IEEE Symposium on Security and Privacy}, 39--57.

\bibitem[{Chen, Jordan, and Wainwright(2020)}]{chen2019hopskipjump}
Chen, J.; Jordan, M.~I.; and Wainwright, M.~J. 2020.
\newblock {HopSkipJumpAttack}: A Query-Efficient Decision-Based Attack.
\newblock In \emph{IEEE Symposium on Security and Privacy}, 1277--1294.

\bibitem[{Ciss{\'e} et~al.(2017)Ciss{\'e}, Bojanowski, Grave, Dauphin, and
  Usunier}]{cisse2017parseval}
Ciss{\'e}, M.; Bojanowski, P.; Grave, E.; Dauphin, Y.; and Usunier, N. 2017.
\newblock Parseval Networks: Improving Robustness to Adversarial Examples.
\newblock In \emph{Proceedings of the 34th International Conference on Machine
  Learning}, volume~70 of \emph{Proceedings of Machine Learning Research},
  854--863. PMLR.

\bibitem[{Croce and Hein(2020)}]{croce2020autoattack}
Croce, F.; and Hein, M. 2020.
\newblock Reliable Evaluation of Adversarial Robustness with an Ensemble of
  Diverse Parameter-Free Attacks.
\newblock In \emph{Proceedings of the 37th International Conference on Machine
  Learning}, volume 119 of \emph{Proceedings of Machine Learning Research},
  2206--2216. PMLR.

\bibitem[{Enyeart and Lin(2024)}]{enyeart2024adversarial}
Enyeart, D.; and Lin, G. 2024.
\newblock Adversarial Autoencoders in Operator Learning.
\newblock \emph{arXiv preprint arXiv:2412.07811}.

\bibitem[{Goodfellow, Shlens, and Szegedy(2015)}]{goodfellow2015explaining}
Goodfellow, I.~J.; Shlens, J.; and Szegedy, C. 2015.
\newblock Explaining and Harnessing Adversarial Examples.
\newblock In \emph{International Conference on Learning Representations}.

\bibitem[{Huang et~al.(2026)Huang, Ma, Wang, and Xiang}]{huang2026stablepdenet}
Huang, C.; Ma, C.; Wang, K.; and Xiang, Y. 2026.
\newblock {StablePDENet}: Enhancing Stability of Operator Learning for Solving
  Differential Equations.
\newblock \emph{arXiv preprint arXiv:2601.06472}.

\bibitem[{Kurakin, Goodfellow, and Bengio(2017)}]{kurakin2016physical}
Kurakin, A.; Goodfellow, I.; and Bengio, S. 2017.
\newblock Adversarial Examples in the Physical World.
\newblock In \emph{ICLR Workshop}.

\bibitem[{Li et~al.(2021)Li, Kovachki, Azizzadenesheli, Liu, Bhattacharya,
  Stuart, and Anandkumar}]{li2021fourier}
Li, Z.; Kovachki, N.; Azizzadenesheli, K.; Liu, B.; Bhattacharya, K.; Stuart,
  A.; and Anandkumar, A. 2021.
\newblock Fourier Neural Operator for Parametric Partial Differential
  Equations.
\newblock In \emph{International Conference on Learning Representations}.

\bibitem[{Liu et~al.(2023)Liu, Park, Hoang, Hasson, and
  Huan}]{liu2022forecastingattacks}
Liu, L.; Park, Y.; Hoang, T.~N.; Hasson, H.; and Huan, J. 2023.
\newblock Robust Multivariate Time-Series Forecasting: Adversarial Attacks and
  Defense Mechanisms.
\newblock In \emph{International Conference on Learning Representations}.

\bibitem[{Lu et~al.(2021)Lu, Jin, Pang, Zhang, and
  Karniadakis}]{lu2021deeponet}
Lu, L.; Jin, P.; Pang, G.; Zhang, Z.; and Karniadakis, G.~E. 2021.
\newblock Learning Nonlinear Operators via {DeepONet} Based on the Universal
  Approximation Theorem of Operators.
\newblock \emph{Nature Machine Intelligence}, 3(3): 218--229.

\bibitem[{Madry et~al.(2018)Madry, Makelov, Schmidt, Tsipras, and
  Vladu}]{madry2018towards}
Madry, A.; Makelov, A.; Schmidt, L.; Tsipras, D.; and Vladu, A. 2018.
\newblock Towards Deep Learning Models Resistant to Adversarial Attacks.
\newblock In \emph{International Conference on Learning Representations}.

\bibitem[{Mode and Hoque(2020)}]{mode2020mtsregression}
Mode, G.~R.; and Hoque, K.~A. 2020.
\newblock Adversarial Examples in Deep Learning for Multivariate Time Series
  Regression.
\newblock In \emph{IEEE Applied Imagery Pattern Recognition Workshop}, 1--10.

\bibitem[{Moosavi-Dezfooli, Fawzi, and Frossard(2016)}]{moosavi2016deepfool}
Moosavi-Dezfooli, S.-M.; Fawzi, A.; and Frossard, P. 2016.
\newblock {DeepFool}: A Simple and Accurate Method to Fool Deep Neural
  Networks.
\newblock In \emph{IEEE Conference on Computer Vision and Pattern Recognition},
  2574--2582.

\bibitem[{Nguyen and Raff(2019)}]{nguyen2018regression}
Nguyen, A.~T.; and Raff, E. 2019.
\newblock Adversarial Attacks, Regression, and Numerical Stability
  Regularization.
\newblock In \emph{AAAI-19 Workshop on Engineering Dependable and Secure
  Machine Learning Systems}.

\bibitem[{Oommen et~al.(2025)Oommen, Khodakarami, Bora, Wang, and
  Karniadakis}]{oommen2025turbulent}
Oommen, V.; Khodakarami, S.; Bora, A.; Wang, Z.; and Karniadakis, G.~E. 2025.
\newblock Learning Turbulent Flows with Generative Models: Super-resolution,
  Forecasting, and Sparse Flow Reconstruction.
\newblock \emph{arXiv preprint arXiv:2509.08752}.

\bibitem[{Ouyang et~al.(2025)Ouyang, Zhu, Xiong, Liu, and Lu}]{ouyang2025rams}
Ouyang, W.; Zhu, M.; Xiong, W.; Liu, S.-W.; and Lu, L. 2025.
\newblock {RAMS}: Residual-Based Adversarial-Gradient Moving Sample Method for
  Scientific Machine Learning in Solving Partial Differential Equations.
\newblock \emph{arXiv preprint arXiv:2509.01234}.

\bibitem[{Papernot et~al.(2016)Papernot, McDaniel, Jha, Fredrikson, Celik, and
  Swami}]{papernot2016limitations}
Papernot, N.; McDaniel, P.; Jha, S.; Fredrikson, M.; Celik, Z.~B.; and Swami,
  A. 2016.
\newblock The Limitations of Deep Learning in Adversarial Settings.
\newblock In \emph{IEEE European Symposium on Security and Privacy}, 372--387.

\bibitem[{Rahman et~al.(2022)Rahman, Florez, Anandkumar, Ross, and
  Azizzadenesheli}]{rahman2022gano}
Rahman, M.~A.; Florez, M.~A.; Anandkumar, A.; Ross, Z.~E.; and Azizzadenesheli,
  K. 2022.
\newblock Generative Adversarial Neural Operators.
\newblock \emph{Transactions on Machine Learning Research}.

\bibitem[{Ribeiro et~al.(2023)Ribeiro, Zachariah, Bach, and
  Sch{\"o}n}]{ribeiro2023linearregression}
Ribeiro, A.~H.; Zachariah, D.; Bach, F.; and Sch{\"o}n, T.~B. 2023.
\newblock Regularization Properties of Adversarially-Trained Linear Regression.
\newblock In \emph{Advances in Neural Information Processing Systems},
  volume~36, 23658--23670.

\bibitem[{Roy, Chakraborty, and Alam(2026)}]{roy2026beyond}
Roy, S.; Chakraborty, S.; and Alam, S.~B. 2026.
\newblock Beyond Uniform Sampling: Synergistic Active Learning and Input
  Denoising for Robust Neural Operators.
\newblock \emph{arXiv preprint arXiv:2604.13316}.

\bibitem[{Roy et~al.(2026)Roy, Kobayashi, Chakraborty, {Rizwan-uddin}, and
  Alam}]{roy2026vulnerabilities}
Roy, S.; Kobayashi, K.; Chakraborty, S.; {Rizwan-uddin}; and Alam, S.~B. 2026.
\newblock Adversarial Vulnerabilities in Neural Operator Digital Twins:
  Gradient-Free Attacks on Nuclear Thermal-Hydraulic Surrogates.
\newblock \emph{arXiv preprint arXiv:2603.22525}.

\bibitem[{Schwalbe-Koda, Tan, and
  G{\'o}mez-Bombarelli(2021)}]{schwalbe2021differentiable}
Schwalbe-Koda, D.; Tan, A.~R.; and G{\'o}mez-Bombarelli, R. 2021.
\newblock Differentiable Sampling of Molecular Geometries with
  Uncertainty-Based Adversarial Attacks.
\newblock \emph{Nature Communications}, 12: 5104.

\bibitem[{Shi et~al.(2025)Shi, Li, Guo, Wu, and Zhao}]{shi2025wbar}
Shi, S.; Li, Y.; Guo, Z.; Wu, B.; and Zhao, Y. 2025.
\newblock {PINNs} Failure Region Localization and Refinement through White-box
  Adversarial Attack.
\newblock \emph{Neurocomputing}, 132055.

\bibitem[{Su, Vargas, and Sakurai(2019)}]{su2017onepixel}
Su, J.; Vargas, D.~V.; and Sakurai, K. 2019.
\newblock One Pixel Attack for Fooling Deep Neural Networks.
\newblock \emph{IEEE Transactions on Evolutionary Computation}, 23(5):
  828--841.

\bibitem[{Tong et~al.(2018)Tong, Yu, Alfeld, and
  Vorobeychik}]{tong2018adversarialregression}
Tong, L.; Yu, S.; Alfeld, S.; and Vorobeychik, Y. 2018.
\newblock Adversarial Regression with Multiple Learners.
\newblock In \emph{Proceedings of the 35th International Conference on Machine
  Learning}, volume~80 of \emph{Proceedings of Machine Learning Research},
  4946--4954. PMLR.

\bibitem[{Wang, Wang, and Perdikaris(2021)}]{wang2021physicsdeeponet}
Wang, S.; Wang, H.; and Perdikaris, P. 2021.
\newblock Learning the Solution Operator of Parametric Partial Differential
  Equations with Physics-Informed {DeepONets}.
\newblock \emph{Science Advances}, 7(40): eabi8605.

\bibitem[{Wang et~al.(2019)Wang, Ma, Bailey, Yi, Zhou, and
  Gu}]{wang2019convergence}
Wang, Y.; Ma, X.; Bailey, J.; Yi, J.; Zhou, B.; and Gu, Q. 2019.
\newblock On the Convergence and Robustness of Adversarial Training.
\newblock In \emph{Proceedings of the 36th International Conference on Machine
  Learning}, volume~97 of \emph{Proceedings of Machine Learning Research},
  6586--6595. PMLR.

\bibitem[{Xu and Mannor(2012)}]{xu2012robustness}
Xu, H.; and Mannor, S. 2012.
\newblock Robustness and Generalization.
\newblock \emph{Machine Learning}, 86(3): 391--423.

\bibitem[{Zhang et~al.(2019)Zhang, Yu, Jiao, Xing, El~Ghaoui, and
  Jordan}]{zhang2019trades}
Zhang, H.; Yu, Y.; Jiao, J.; Xing, E.~P.; El~Ghaoui, L.; and Jordan, M.~I.
  2019.
\newblock Theoretically Principled Trade-off between Robustness and Accuracy.
\newblock In \emph{Proceedings of the 36th International Conference on Machine
  Learning}, volume~97 of \emph{Proceedings of Machine Learning Research},
  7472--7482. PMLR.

\bibitem[{Zhu et~al.(2023)Zhu, Zhang, Jiao, Karniadakis, and
  Lu}]{zhu2023reliable}
Zhu, M.; Zhang, H.; Jiao, A.; Karniadakis, G.~E.; and Lu, L. 2023.
\newblock Reliable Extrapolation of Deep Neural Operators Informed by Physics
  or Sparse Observations.
\newblock \emph{Computer Methods in Applied Mechanics and Engineering}, 412:
  116064.

\end{thebibliography}
\end{document}